\documentclass[11pt,onecolumn]{article}

\usepackage[T1]{fontenc}
\usepackage[utf8]{inputenc}
\usepackage{lmodern}
\usepackage{textcomp}

\usepackage[a4paper,margin=1in]{geometry}

\usepackage{amsmath,amssymb,amsthm,mathtools}
\usepackage{bm}

\usepackage{booktabs}
\usepackage{longtable}
\usepackage{tabularx}
\usepackage{multirow}
\usepackage{makecell}
\usepackage{array}
\newcolumntype{Y}{>{\centering\arraybackslash}X}
\usepackage{siunitx}
\usepackage{xcolor}
\usepackage{tikz}
\usetikzlibrary{arrows.meta,positioning,calc,fit,backgrounds,
  shapes.geometric,shapes.misc,decorations.pathreplacing,
  patterns,patterns.meta,matrix,shadows.blur,3d}
\usepackage{pgfplots}
\pgfplotsset{compat=1.18}
\pgfplotsset{/pgf/number format/1000 sep={}}
\usepgfplotslibrary{fillbetween,groupplots,statistics,polar}

\usepackage[ruled,vlined,linesnumbered]{algorithm2e}

\usepackage{enumitem}
\usepackage{microtype}
\usepackage{caption}
\usepackage{subcaption}
\usepackage{float}
\usepackage{placeins}

\usepackage[numbers,sort&compress]{natbib}

\usepackage{hyperref}
\usepackage{fancyhdr}
\usepackage{cleveref}

\crefname{algocf}{Algorithm}{Algorithms}
\Crefname{algocf}{Algorithm}{Algorithms}

\definecolor{gnnink}{HTML}{1A1A1A}
\definecolor{gnnblue}{HTML}{1F4E79}
\definecolor{gnnteal}{HTML}{2C7FB8}
\definecolor{gnngreen}{HTML}{41AB5D}
\definecolor{gnnamber}{HTML}{E8A317}
\definecolor{gnnred}{HTML}{C0392B}
\definecolor{gnngray}{HTML}{6E6E6E}
\definecolor{gnnlight}{HTML}{EAF1F7}
\definecolor{linknavy}{HTML}{1A3D6D}
\definecolor{citegreen}{HTML}{2C6E49}

\hypersetup{
  colorlinks=true,
  linkcolor=linknavy,
  citecolor=citegreen,
  urlcolor=linknavy,
  pdftitle={Graph Neural Networks Applications Across Domains: All Insights You Need},
  pdfauthor={},
  bookmarksnumbered=true
}

\theoremstyle{definition}
\newtheorem{definition}{Definition}
\newtheorem{problem}{Problem}
\theoremstyle{plain}

\newtheorem{proposition}{Proposition}

\theoremstyle{remark}

\newcommand{\graph}{\mathcal{G}}
\newcommand{\vset}{\mathcal{V}}
\newcommand{\eset}{\mathcal{E}}
\newcommand{\Adj}{\mathbf{A}}
\newcommand{\Deg}{\mathbf{D}}
\newcommand{\Lap}{\mathbf{L}}
\newcommand{\Feat}{\mathbf{X}}
\newcommand{\Hid}{\mathbf{H}}
\newcommand{\Wgt}{\mathbf{W}}
\newcommand{\msg}{\mathbf{m}}
\newcommand{\nbr}{\mathcal{N}}
\newcommand{\real}{\mathbb{R}}
\newcommand{\softmax}{\operatorname{softmax}}
\newcommand{\relu}{\operatorname{ReLU}}

\DeclareMathOperator*{\AGG}{AGGREGATE}
\DeclareMathOperator*{\UPD}{UPDATE}
\DeclareMathOperator*{\READOUT}{READOUT}

\newcommand{\hv}{\mathbf{h}}                 % node embedding vector
\newcommand{\xv}{\mathbf{x}}                 % node feature vector
\newcommand{\ev}{\mathbf{e}}                 % edge feature vector
\newcommand{\bias}{\mathbf{b}}               % bias vector
\newcommand{\Iden}{\mathbf{I}}               % identity matrix
\newcommand{\Atil}{\widetilde{\mathbf{A}}}   % self-loop augmented adjacency
\newcommand{\Dtil}{\widetilde{\mathbf{D}}}   % self-loop augmented degree
\newcommand{\Lnorm}{\widehat{\mathbf{L}}}    % symmetric normalised Laplacian
\newcommand{\trans}{^{\!\top}}               % transpose
\newcommand{\concat}{\,\Vert\,}              % concatenation

\tikzset{
  gnode/.style={circle, draw=gnnblue, fill=gnnlight, line width=0.6pt,
                minimum size=7mm, font=\small, inner sep=1pt},
  gbox/.style={rectangle, rounded corners=2pt, draw=gnnblue, fill=gnnlight,
               line width=0.6pt, minimum height=8mm, align=center, font=\small},
  gedge/.style={draw=gnngray, line width=0.7pt},
  garrow/.style={-{Stealth[length=2mm]}, draw=gnnink, line width=0.7pt},
  glabel/.style={font=\footnotesize\itshape, text=gnngray},
  node distance=10mm and 14mm
}

\newcommand{\hbfull}{\tikz[baseline=-0.6ex]{\filldraw[gnnink] (0,0) circle (0.45ex);}}
\newcommand{\hbhalf}{\tikz[baseline=-0.6ex]{\draw[gnnink,line width=0.3pt] (0,0) circle (0.45ex);
  \fill[gnnink] (0,0) -- (0,0.45ex) arc (90:270:0.45ex) -- cycle;}}
\newcommand{\hbempty}{\tikz[baseline=-0.6ex]{\draw[gnnink,line width=0.3pt] (0,0) circle (0.45ex);}}

\definecolor{gnnpurple}{HTML}{6A4C93}
\definecolor{gnnpink}{HTML}{D6336C}
\definecolor{gnnocean}{HTML}{0B7285}
\definecolor{gnnslate}{HTML}{34495E}
\definecolor{gnnlime}{HTML}{5C940D}
\definecolor{gnnrust}{HTML}{B5651D}
\definecolor{tintblue}{HTML}{DBE7F3}
\definecolor{tintteal}{HTML}{D4ECF6}
\definecolor{tintgreen}{HTML}{DCF1E3}
\definecolor{tintamber}{HTML}{FBE9CB}
\definecolor{tintred}{HTML}{F8DBD7}
\definecolor{tintpurple}{HTML}{E7DDF1}
\definecolor{tintgray}{HTML}{EAEDF0}
\definecolor{tintpink}{HTML}{F9DAE5}
\definecolor{tintocean}{HTML}{D3EAEF}
\definecolor{tintlime}{HTML}{E6EFD1}

\tikzset{
  cnode/.style={circle, draw=#1!75!black, fill=#1, text=white, line width=0.7pt,
                minimum size=8.5mm, font=\small\bfseries, inner sep=1pt},
  cnode/.default=gnnblue,
  sbox/.style 2 args={rectangle, rounded corners=3pt, draw=#1!80!black, fill=#2,
                line width=0.9pt, minimum height=9mm, align=center, font=\small,
                inner sep=4.5pt, text=gnnink},
  chip/.style={rectangle, rounded corners=2pt, fill=#1, text=white,
               font=\footnotesize\bfseries, inner sep=3.5pt, minimum height=5.5mm, align=center},
  chip/.default=gnnblue,
  carrow/.style={-{Stealth[length=2.8mm,width=2.4mm]}, draw=#1, line width=1.2pt},
  carrow/.default=gnnslate,
  cedge/.style={draw=#1, line width=1pt},
  cedge/.default=gnnslate,
  panel/.style={rectangle, rounded corners=4pt, draw=#1!55!white, fill=#1!7!white, line width=0.7pt},
  panel/.default=gnnslate,
}

\tikzset{
  pics/ic person/.style n args={1}{code={\filldraw[#1,draw=#1!70!black,line width=0.4pt]
      (0,0.14) circle (0.12);
      \filldraw[#1,draw=#1!70!black,line width=0.4pt]
      (-0.20,-0.30) to[out=80,in=200] (-0.05,-0.04) -- (0.05,-0.04)
      to[out=-20,in=100] (0.20,-0.30) -- cycle;}},
  pics/ic person/.default={gnnblue},
  pics/ic mol/.style n args={1}{code={\draw[#1!60!black,line width=0.7pt]
      (-0.18,-0.12)--(0,0.16) (0,0.16)--(0.18,-0.12);
      \filldraw[#1,draw=#1!70!black,line width=0.4pt] (0,0.16) circle (0.085);
      \filldraw[#1!75!white,draw=#1!70!black,line width=0.4pt] (-0.18,-0.12) circle (0.075);
      \filldraw[#1!75!white,draw=#1!70!black,line width=0.4pt] (0.18,-0.12) circle (0.075);}},
  pics/ic mol/.default={gnngreen},
  pics/ic brain/.style n args={1}{code={\filldraw[#1!30!white,draw=#1!75!black,line width=0.6pt]
      (0,0) ellipse (0.22 and 0.18);
      \draw[#1!75!black,line width=0.5pt] (0,0.16) -- (0,-0.16);
      \draw[#1!75!black,line width=0.45pt] (-0.11,0.05) to[out=-30,in=210] (-0.02,0.04);
      \draw[#1!75!black,line width=0.45pt] (0.04,0.02) to[out=-30,in=210] (0.13,0.0);}},
  pics/ic brain/.default={gnnpink},
  pics/ic sensor/.style n args={1}{code={\filldraw[#1,draw=#1!70!black,line width=0.4pt]
      (-0.1,-0.12) rectangle (0.1,0.1);
      \draw[#1!80!black,line width=0.6pt] (0.16,-0.04) arc (-40:40:0.12);
      \draw[#1!80!black,line width=0.6pt] (0.22,-0.08) arc (-40:40:0.20);}},
  pics/ic sensor/.default={gnnocean},
  pics/ic bolt/.style n args={1}{code={\filldraw[#1,draw=#1!70!black,line width=0.4pt]
      (0.04,0.22)--(-0.12,0.0)--(0.0,0.0)--(-0.04,-0.22)--(0.13,0.04)--(0.02,0.04)--cycle;}},
  pics/ic bolt/.default={gnnamber},
  pics/ic doc/.style n args={1}{code={\filldraw[white,draw=#1!80!black,line width=0.55pt]
      (-0.13,-0.18) rectangle (0.13,0.2);
      \draw[#1!80!black,line width=0.45pt] (-0.08,0.1)--(0.08,0.1) (-0.08,0.02)--(0.08,0.02)
      (-0.08,-0.06)--(0.08,-0.06) (-0.08,-0.12)--(0.03,-0.12);}},
  pics/ic doc/.default={gnnslate},
  pics/ic factory/.style n args={1}{code={\filldraw[#1,draw=#1!70!black,line width=0.4pt]
      (-0.2,-0.16) -- (-0.2,0.02) -- (-0.02,-0.08) -- (-0.02,0.02) -- (0.16,-0.08)
      -- (0.16,-0.16) -- cycle;
      \fill[#1!70!black] (-0.16,0.02) rectangle (-0.12,0.14);}},
  pics/ic factory/.default={gnnrust},
  pics/ic car/.style n args={1}{code={\filldraw[#1,draw=#1!70!black,line width=0.4pt]
      (-0.2,-0.06) -- (-0.14,-0.06) -- (-0.08,0.06) -- (0.08,0.06) -- (0.13,-0.06)
      -- (0.2,-0.06) -- (0.2,-0.12) -- (-0.2,-0.12) -- cycle;
      \fill[gnnink] (-0.11,-0.14) circle (0.045); \fill[gnnink] (0.11,-0.14) circle (0.045);}},
  pics/ic car/.default={gnnred},
  pics/ic atom/.style n args={1}{code={\draw[#1!70!black,line width=0.5pt]
      (-0.16,-0.16)--(0.16,-0.16)--(0.16,0.16)--(-0.16,0.16)--cycle (-0.16,0)--(0.16,0) (0,-0.16)--(0,0.16);
      \foreach \x in {-0.16,0,0.16}{\foreach \y in {-0.16,0,0.16}{
        \filldraw[#1,draw=#1!70!black,line width=0.3pt] (\x,\y) circle (0.035);}}}},
  pics/ic atom/.default={gnnpurple},
  pics/ic shield/.style n args={1}{code={\filldraw[#1!25!white,draw=#1!80!black,line width=0.6pt]
      (0,0.2) -- (0.16,0.12) -- (0.16,-0.06) -- (0,-0.2) -- (-0.16,-0.06) -- (-0.16,0.12) -- cycle;
      \draw[#1!80!black,line width=0.7pt] (-0.07,0.02) -- (-0.01,-0.07) -- (0.09,0.08);}},
  pics/ic shield/.default={gnnteal},
}

\definecolor{orcidgreen}{HTML}{A6CE39}
\definecolor{scholarblue}{HTML}{4285F4}
\definecolor{linkedinblue}{HTML}{0A66C2}
\newcommand{\orcidID}[1]{\href{#1}{\begin{tikzpicture}[baseline=-0.55ex]
  \node[circle,fill=orcidgreen,inner sep=0pt,minimum size=2.0ex]{};
  \node[text=white,font=\bfseries,inner sep=0pt]{\fontsize{4.6}{4.6}\selectfont iD};\end{tikzpicture}}}
\newcommand{\scholarID}[1]{\href{#1}{\begin{tikzpicture}[baseline=-0.55ex,y=1ex,x=1ex]
  \fill[scholarblue] (0,0.62) -- (1.0,1.16) -- (2.0,0.62) -- (1.0,0.08) -- cycle;
  \fill[scholarblue] (0.5,0.46) .. controls (1.0,0.22) .. (1.5,0.46) -- (1.5,0.16)
        .. controls (1.0,-0.06) .. (0.5,0.16) -- cycle;\end{tikzpicture}}}
\newcommand{\linkedinID}[1]{\href{#1}{\begin{tikzpicture}[baseline=-0.55ex]
  \node[rounded corners=0.35ex,fill=linkedinblue,inner sep=0pt,minimum size=2.0ex]{};
  \node[text=white,font=\bfseries,inner sep=0pt]{\fontsize{5}{5}\selectfont in};\end{tikzpicture}}}
\newcommand{\AuthorOrcidURL}{https://orcid.org/0009-0003-7116-5080}
\newcommand{\AuthorScholarURL}{https://scholar.google.com/citations?user=N2il6NQAAAAJ\&hl=en\&oi=ao}
\newcommand{\AuthorLinkedinURL}{https://www.linkedin.com/in/abderaouf-bahi-phd-7b934a195/}
\newcommand{\authoricons}{\orcidID{\AuthorOrcidURL}\,\scholarID{\AuthorScholarURL}\,\linkedinID{\AuthorLinkedinURL}}
\fancypagestyle{plain}{\fancyhf{}\fancyfoot[C]{\footnotesize\thepage}%
  }

\begin{document}

%================================ TITLE / FRONT MATTER =========================
\title{\bfseries Graph Neural Networks Applications Across Domains:\\[2pt]
\large All Insights You Need}

\author{%
  {\large Abderaouf Bahi}\thanks{Computer Science and Applied Mathematics Laboratory (LIMA),
Faculty of Science and Technology, Chadli Bendjedid University, 
P.O. Box 73, El Tarf 36000, Algeria }~\authoricons\\[3pt]
  \normalsize \texttt{a.bahi@univ-eltarf.dz}%
}

\maketitle

%--------------------------------- Abstract -----------------------------------
\begin{abstract}
\noindent
Graph neural networks have moved from a niche representation-learning technique
to the default model class wherever data carry relational structure. The
interesting question is no longer whether message passing helps on a given
dataset, but where graph structure earns its computational cost and where it
does not. This survey organises the field around a single design space, derives
the spectral and spatial formulations from shared first principles, and connects
expressive power to the Weisfeiler-Leman hierarchy with explicit statements of
what current architectures can and cannot separate. Against that methodological
backbone we examine twelve application domains, among them recommendation and
social networks, knowledge graphs and language-model integration, drug discovery
and molecular property learning, healthcare and neuroscience, computer vision,
traffic and urban computing, power and renewable-energy systems, wireless and
sixth-generation networks, fraud and cybersecurity, industrial prognostics, materials
science, and climate modelling. For each domain we specify the graph-construction
choices and their costs, identify which architecture families dominate and why,
and separate reported gains from artefacts of weak baselines or favourable splits.
A cross-domain comparison exposes recurring patterns: heterophily and scale
undercut the same models almost everywhere, temporal graphs remain harder than
their static counterparts, and the architectures that top public leaderboards are
seldom the ones that reach deployment. We treat over-smoothing, over-squashing,
robustness, distribution shift, fairness, and explainability not as a closing
checklist but as the constraints that decide adoption. Finally, we assess the
claim that graph foundation models and language-model integration constitute a
genuine break from task-specific networks, and argue that the evidence is
suggestive rather than settled. Qualitative judgements are the authors' own and
are marked as such; every reported number is attributed to its source.
\end{abstract}


\newpage

%--------------------------------- Contents -----------------------------------
{\hypersetup{linkcolor=gnnink}
\tableofcontents
\listoffigures
\listoftables
\listofalgorithms}
\newpage

%============================ ABBREVIATIONS TABLE ==============================
\section*{Abbreviations}
\addcontentsline{toc}{section}{Abbreviations}
\label{sec:abbrev}
\renewcommand{\arraystretch}{1.15}
\begin{longtable}{@{}p{0.16\textwidth}p{0.78\textwidth}@{}}
\toprule
\textbf{Acronym} & \textbf{Expansion} \\
\midrule
\endfirsthead
\toprule
\textbf{Acronym} & \textbf{Expansion} \\
\midrule
\endhead
\bottomrule
\endlastfoot
6G        & Sixth-generation mobile network \\
ACC       & Anomaly correlation coefficient \\
AUC       & Area under the curve \\
AUROC     & Area under the receiver operating characteristic curve \\
BERT      & Bidirectional encoder representations from transformers \\
BS        & Base station \\
CNN       & Convolutional neural network \\
DER       & Distributed energy resource \\
DG        & Distributed generation \\
DTI       & Drug-target interaction \\
EEG       & Electroencephalography \\
EHR       & Electronic health record \\
ETA       & Estimated time of arrival \\
F1        & F-measure (harmonic mean of precision and recall) \\
fMRI      & Functional magnetic resonance imaging \\
GAN       & Generative adversarial network \\
GAT       & Graph attention network \\
GCL       & Graph contrastive learning \\
GCN       & Graph convolutional network \\
GFM       & Graph foundation model \\
GGNN      & Gated graph neural network \\
GIN       & Graph isomorphism network \\
GNN       & Graph neural network \\
GPU       & Graphics processing unit \\
GraphGPS  & General, powerful, scalable graph transformer framework \\
GraphRAG  & Graph-based retrieval-augmented generation \\
GraphSAGE & Graph sample and aggregate \\
GRU       & Gated recurrent unit \\
IIoT      & Industrial internet of things \\
IoT       & Internet of things \\
KG        & Knowledge graph \\
KGE       & Knowledge graph embedding \\
LLM       & Large language model \\
MAE       & Mean absolute error \\
MAPE      & Mean absolute percentage error \\
MLP       & Multilayer perceptron \\
MPNN      & Message passing neural network \\
MRR       & Mean reciprocal rank \\
NDCG      & Normalized discounted cumulative gain \\
NLP       & Natural language processing \\
ODE       & Ordinary differential equation \\
PMU       & Phasor measurement unit \\
PV        & Photovoltaic \\
QSAR      & Quantitative structure-activity relationship \\
RAG       & Retrieval-augmented generation \\
RL        & Reinforcement learning \\
RMSE      & Root mean square error \\
ROC       & Receiver operating characteristic \\
RUL       & Remaining useful life \\
SCADA     & Supervisory control and data acquisition \\
SSL       & Self-supervised learning \\
ST-GNN    & Spatio-temporal graph neural network \\
VAE       & Variational autoencoder \\
VLM       & Vision-language model \\
WL        & Weisfeiler-Leman \\
\end{longtable}

%============================== NOTATION TABLE ================================
\section*{Notation}
\addcontentsline{toc}{section}{Notation}
\label{sec:notation}
\renewcommand{\arraystretch}{1.25}
\begin{longtable}{@{}p{0.22\textwidth}p{0.72\textwidth}@{}}
\toprule
\textbf{Symbol} & \textbf{Meaning} \\
\midrule
\endfirsthead
\toprule
\textbf{Symbol} & \textbf{Meaning} \\
\midrule
\endhead
\bottomrule
\endlastfoot
$\graph=(\vset,\eset)$        & Graph with node set $\vset$ and edge set $\eset$ \\
$n=\lvert\vset\rvert$         & Number of nodes \\
$m=\lvert\eset\rvert$         & Number of edges \\
$v,u$                         & Individual nodes; $(u,v)\in\eset$ an edge \\
$\nbr(v)$                     & Neighbourhood of node $v$ \\
$\Adj\in\real^{n\times n}$    & Adjacency matrix \\
$\Atil=\Adj+\Iden$            & Adjacency matrix with self-loops \\
$\Deg$                        & Diagonal degree matrix, $\Deg_{ii}=\sum_j \Adj_{ij}$ \\
$\Iden$                       & Identity matrix \\
$\Lap=\Deg-\Adj$              & Combinatorial graph Laplacian \\
$\Lnorm=\Iden-\Deg^{-1/2}\Adj\Deg^{-1/2}$ & Symmetric normalised Laplacian \\
$\Feat\in\real^{n\times d}$   & Node feature matrix; row $\xv_v$ is the feature of $v$ \\
$\ev_{uv}$                    & Feature vector of edge $(u,v)$ \\
$\Hid^{(l)}\in\real^{n\times d_l}$ & Node representations at layer $l$, with $\Hid^{(0)}=\Feat$ \\
$\hv_v^{(l)}$                 & Representation of node $v$ at layer $l$ \\
$d_l$                         & Hidden width at layer $l$ \\
$L$                           & Number of message-passing layers \\
$\Wgt^{(l)},\bias^{(l)}$      & Learnable weight matrix and bias at layer $l$ \\
$\msg_{u\to v}^{(l)}$         & Message sent from $u$ to $v$ at layer $l$ \\
$\alpha_{uv}$                 & Attention coefficient on edge $(u,v)$ \\
$\sigma(\cdot)$               & Element-wise nonlinearity \\
$\relu,\softmax$              & Rectified linear unit and softmax operators \\
$\odot$                       & Hadamard (element-wise) product \\
$\concat$                     & Vector concatenation \\
$\mathcal{R}$                 & Relation set in heterogeneous or knowledge graphs \\
$T$                           & Number of timestamps in a temporal graph \\
$\theta$                      & Collection of all learnable parameters \\
$\AGG$                        & Permutation-invariant neighbourhood aggregation operator \\
$\UPD$                        & Update function combining a node with its aggregated neighbourhood \\
$\READOUT$                    & Permutation-invariant graph-level readout (pooling) \\
$\Dtil=\Deg+\Iden$            & Degree matrix with self-loops, paired with $\Atil$ \\
$(\cdot)^{\trans}$            & Matrix or vector transpose \\
$\hbfull\;\hbhalf\;\hbempty$  & Strong / moderate / weak on the qualitative scale used in the comparison tables \\
\end{longtable}

\FloatBarrier
\newpage

%==============================================================================
%  PHASE 1 BODY SECTIONS ARE APPENDED BELOW THIS LINE, IN OUTLINE ORDER.
%  (Phase 0 deliverable ends here; the skeleton already compiles.)
%==============================================================================

% <<< BODY:START >>>
% (sections 1..20 inserted here during Phase 1)
\setcounter{table}{0}% front-matter longtables (abbreviations, notation) must not count
\setcounter{figure}{0}
%================================ SECTION 1 ===================================
\section{Introduction}
\label{sec:intro}

Relational structure is the rule rather than the exception in the data that modern machine learning is asked to model. Molecules are atoms joined by bonds, social platforms are users joined by interactions, power systems are buses joined by transmission lines, and a knowledge base is a set of entities joined by typed relations. Treating such data as independent feature vectors discards the very signal that separates one instance from another. Graph neural networks (GNNs) answer this by computing representations that depend explicitly on a node's connectivity, propagating information along edges so that the embedding of an entity reflects the entities around it. The construction is simple to state and has proven unusually general: the same neighbourhood-aggregation recursion that classifies a paper from its citations also predicts whether a molecule inhibits bacterial growth \cite{stokes2020antibiotic} and forecasts global weather at skill competitive with operational numerical systems \cite{lam2023graphcast}.

The mechanism underlying this generality is neighbourhood aggregation. Each node carries a vector representation that is updated by combining its own state with a summary of its neighbours' states, so that after $k$ rounds a node has incorporated information from its $k$-hop neighbourhood. Two properties make the construction what it is. The aggregation is invariant to the ordering of neighbours, which encodes the assumption that a node is defined by the set of entities it connects to rather than by any arrangement imposed on them. The update is shared across all nodes, which lets a single trained model apply to graphs of arbitrary size and shape. Invariance to neighbour order and parameter sharing across positions are what separate a graph network from a multilayer perceptron applied to concatenated features, and they are the source of both its strengths and its characteristic failures.

The literature has grown in step with the method's reach, and several broad surveys map its methodological core \cite{wu2021survey,zhou2020review,zhang2022dlgsurvey,chami2022taxonomy}. Earlier reviews of graph convolutional networks \cite{zhang2019gcnreview} and of graph representation learning more broadly \cite{hamilton2017repsurvey} cover the foundations, and the field is sometimes framed within the wider programme of geometric deep learning, which unifies learning on grids, groups, graphs, and manifolds under the principle of invariance to a symmetry group \cite{bronstein2017gdl,bronstein2021gdl}; a compact synthesis casts the area as the study of connection itself \cite{velickovic2023everything}. What these references establish, and what we take as settled, is that graph learning has left the experimental stage. It is the default model class wherever data carry relational structure, with mature subfields in recommendation \cite{wu2023recsurvey}, knowledge graphs \cite{ji2021kgsurvey}, molecular modelling \cite{wieder2020molsurvey}, traffic forecasting \cite{jiang2022trafficsurvey}, and power systems \cite{liao2021powersurvey}, among others. The question that remains open, and that motivates this survey, is sharper than whether graphs help. It is where the inductive bias of message passing earns its computational cost, and where a simpler model would do as well or better. A practitioner choosing a model for a new problem is poorly served by a catalogue of architectures; what they need is a sense of which structural assumptions hold in their domain and which fail.

\subsection{What distinguishes this survey}
\label{subsec:distinguish}

Most existing surveys are organized as catalogues. They proceed method by method or application by application, summarizing each contribution in turn. This format has value as a reference, yet it cannot answer the comparative question, because the answer lives in the gaps between papers rather than within any one of them. Whether attention helps depends on the graph; whether a deeper network helps depends on homophily; whether a reported gain is real depends on the baseline it was measured against. This survey is built around those comparisons. We fix one notation and one design space, derive the spectral and spatial formulations from shared first principles, and then read every application domain through the same analytical schema, asking in each case what the nodes and edges are, which architecture family dominates and why, and how much of the reported improvement survives contact with a strong non-graph baseline.

The existing reference works divide cleanly into two groups, and neither occupies the position this survey takes. The first group is methodological. Comprehensive treatments of architectures and theory \cite{wu2021survey,zhou2020review,chami2022taxonomy}, the textbook account of representation learning \cite{hamilton2020book}, and focused reviews of expressiveness \cite{sato2020expressivesurvey,morris2023wlstory}, self-supervision \cite{wu2021sslsurvey,xie2022sslreview}, dynamic graphs \cite{kazemi2020dynamicsurvey}, and heterogeneous graphs \cite{wang2022hgsurvey} explain the machinery in depth but say little about how it behaves once a particular domain imposes its constraints. The second group is applied. Surveys of recommendation \cite{wu2023recsurvey}, knowledge graphs \cite{ye2022kgsurvey}, drug discovery \cite{sun2020drugsurvey}, traffic \cite{jiang2022trafficsurvey}, time series \cite{jin2024gnn4ts}, power systems \cite{liao2021powersurvey}, wireless communication \cite{shen2023theory2practice}, and trustworthiness \cite{dai2024trustworthy} go deep on a single area but cannot compare across areas, since comparison sits outside their remit. The position this survey occupies is the one between them: a treatment methodological enough to explain why a model behaves as it does, and applied enough to span the domains where that behaviour is tested, with the explicit aim of holding them up against one another.

A second commitment is to criticism. Graph learning carries a measurement problem that its rate of publication tends to obscure. Standard node-classification benchmarks are small, their splits differ across papers, and the construction of larger and more carefully controlled benchmarks has repeatedly shown that the distance between elaborate architectures and well-tuned simple ones is narrower than headline numbers suggest \cite{dwivedi2023benchmarking,hu2020ogb}. We treat such findings as central rather than incidental. Where the evidence for a method is strong we say so; where a claimed advantage rests on a weak baseline, a favourable split, or a metric that rewards the wrong behaviour, we say that too. The intent is not to disparage but to calibrate, because a survey that reports every claimed improvement at face value is less useful than one that separates the durable results from the transient ones.

The contribution of graph structure deserves a concrete statement, because the benefit is real but bounded, and being precise about it prevents two opposite errors. On graphs where neighbouring nodes tend to share labels, a property called homophily, message passing behaves as a learned form of label propagation, and even an untrained smoothing of features is a strong baseline \cite{wu2019sgc}. The graph supplies a prior that adjacent nodes resemble one another, and a model that exploits this prior needs less labelled data than one that treats nodes in isolation. The same mechanism turns into a liability when the prior is wrong. On a fraud graph, where fraudulent accounts deliberately attach themselves to legitimate ones for cover, averaging over neighbours erases the signal it was meant to extract, and a careless graph model can fall behind a tabular classifier that ignores the edges entirely. The recurring lesson of the application chapters is that the value of a graph model is contingent on the alignment between the edges and the target, and that this alignment must be checked rather than assumed.

\subsection{Patterns that only appear across domains}
\label{subsec:patterns}

Reading domains side by side pays off because the same failure modes recur, and naming them once clarifies all of them. Heterophily, the situation in which connected nodes differ rather than agree, degrades the standard smoothing-based models in fraud graphs, in web graphs, and in several molecular tasks alike \cite{zhu2020h2gcn,pei2020geomgcn}. Scale imposes the same family of compromises, whether through sampling, clustering, or algebraic simplification, on anyone working with industrial graphs, no matter whether the nodes are users, sensors, or proteins \cite{hamilton2017graphsage,chiang2019clustergcn}. Temporal graphs are harder than static ones in every domain that has them, and the field's progress on evolving topologies trails its progress on fixed ones \cite{kazemi2020dynamicsurvey,skarding2021dynamicsurvey}. There is also a divergence, visible only when domains are compared, between the architectures that top public leaderboards and the ones that reach production, where data freshness, inference latency, and the cost of retraining outweigh a point of accuracy. Surfacing observations of this kind is the central purpose of the cross-domain analysis.

We depart from convention as well in how the field's open problems are handled. Over-smoothing, over-squashing, limited expressive power, brittleness under adversarial perturbation, distribution shift, opacity, and unfairness are usually relegated to a closing section that reads as a checklist. We instead treat them as the constraints that decide whether a method can be deployed, and we connect each to the domains where it bites hardest. Over-smoothing is an architectural curiosity on a citation benchmark and a genuine obstacle in deep molecular networks; the expressiveness ceiling set by the Weisfeiler-Leman test \cite{xu2019gin,morris2019wlneural} matters little for recommendation and a great deal for tasks that hinge on counting substructures; an explanation method that satisfies a vision audience may fail a clinician's evidentiary standard \cite{yuan2023explainsurvey}. The constraints are general, but their severity is domain-specific.

The most consequential recent development, and the one most in need of sober assessment, is the convergence of graph learning with large language models (LLMs) and the associated ambition of building graph foundation models (GFMs) \cite{liu2023gfmsurvey,mao2024gfm}. Two distinct claims travel under this banner and deserve to be separated. The first holds that language models can absorb part of the work GNNs do, reasoning over graph-structured inputs rendered as text, and that retrieval over graphs improves their factual reliability \cite{edge2024graphrag,jin2024llmongraphs}. The second holds that a single pretrained model can transfer across graphs and tasks the way a language model transfers across text, removing the need to train a bespoke network for each dataset \cite{liu2024ofa,galkin2024ultra}. The first claim rests on a growing and credible body of work. The second is closer to aspiration than to achievement, because graphs lack the shared vocabulary that makes text transferable, and we argue that the present evidence for genuine cross-graph generalization is suggestive rather than settled.

\subsection{Scope, conventions, and contributions}
\label{subsec:scope}

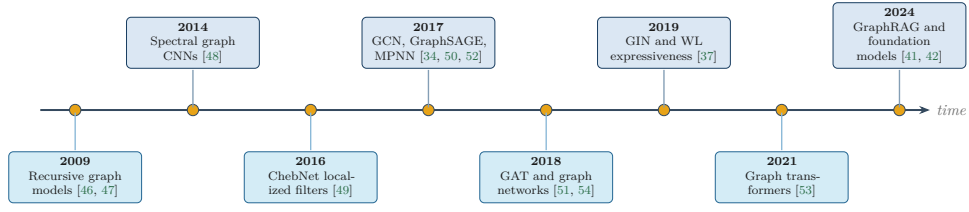
\begin{figure}[t]
\centering
\resizebox{0.8\textwidth}{!}{%
\begin{tikzpicture}[
   tlA/.style={rounded corners=3pt, draw=gnnblue!80!black, fill=tintblue, text width=28mm, align=center, font=\scriptsize, inner sep=3.5pt, text=gnnink},
   tlB/.style={rounded corners=3pt, draw=gnnteal!80!black, fill=tintteal, text width=28mm, align=center, font=\scriptsize, inner sep=3.5pt, text=gnnink},
   dot/.style={circle, draw=gnnslate, fill=gnnamber, minimum size=2.6mm, inner sep=0pt}]
  \draw[-{Stealth[length=2.5mm]}, line width=1.3pt, draw=gnnslate] (-0.8,0) -- (19.6,0)
        node[right, font=\small\itshape, text=gnngray] {time};
  \foreach \x in {0,2.7,5.4,8.1,10.8,13.5,16.2,18.9} { \node[dot] at (\x,0) {}; }
  \node[tlA, anchor=south] at (2.7,0.95)  {\textbf{2014}\\Spectral graph CNNs \cite{bruna2014spectral}};
  \node[tlA, anchor=south] at (8.1,0.95)  {\textbf{2017}\\GCN, GraphSAGE, MPNN \cite{kipf2017gcn,hamilton2017graphsage,gilmer2017mpnn}};
  \node[tlA, anchor=south] at (13.5,0.95) {\textbf{2019}\\GIN and WL expressiveness \cite{xu2019gin}};
  \node[tlA, anchor=south] at (18.9,0.95) {\textbf{2024}\\GraphRAG and foundation models \cite{edge2024graphrag,mao2024gfm}};
  \node[tlB, anchor=north] at (0,-0.95)    {\textbf{2009}\\Recursive graph models \cite{gori2005newmodel,scarselli2009gnn}};
  \node[tlB, anchor=north] at (5.4,-0.95)  {\textbf{2016}\\ChebNet localized filters \cite{defferrard2016chebnet}};
  \node[tlB, anchor=north] at (10.8,-0.95) {\textbf{2018}\\GAT and graph networks \cite{velickovic2018gat,battaglia2018relational}};
  \node[tlB, anchor=north] at (16.2,-0.95) {\textbf{2021}\\Graph transformers \cite{ying2021graphormer}};
  \foreach \x in {2.7,8.1,13.5,18.9} { \draw[gnnblue!55, line width=0.9pt] (\x,0.06)  -- (\x,0.92);  }
  \foreach \x in {0,5.4,10.8,16.2}   { \draw[gnnteal!55, line width=0.9pt] (\x,-0.06) -- (\x,-0.92); }
\end{tikzpicture}}
\caption[Milestones in the development of GNN architectures]{Milestones in the development of graph neural network architectures, from recursive models through message passing, attention, transformers, and foundation-model efforts.}
\label{fig:timeline}
\end{figure}

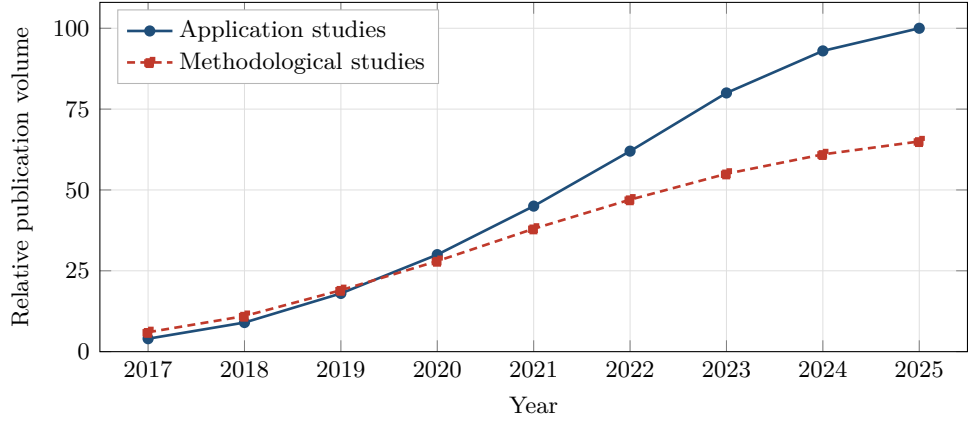
\begin{figure}[t]
\centering
\begin{tikzpicture}
\begin{axis}[
  width=0.82\textwidth, height=6.2cm,
  xlabel={Year}, ylabel={Relative publication volume},
  xmin=2016.5, xmax=2025.5, ymin=0, ymax=108,
  xtick={2017,2018,2019,2020,2021,2022,2023,2024,2025},
  ytick={0,25,50,75,100},
  grid=both, grid style={gnngray!22, line width=0.3pt},
  tick label style={font=\footnotesize}, label style={font=\footnotesize},
  legend style={font=\footnotesize, at={(0.02,0.98)}, anchor=north west, draw=gnngray!50},
  legend cell align=left]
\addplot[gnnblue, mark=*, line width=1pt, mark size=1.6pt] coordinates
  {(2017,4)(2018,9)(2019,18)(2020,30)(2021,45)(2022,62)(2023,80)(2024,93)(2025,100)};
\addlegendentry{Application studies}
\addplot[gnnred, mark=square*, line width=1pt, mark size=1.6pt, densely dashed] coordinates
  {(2017,6)(2018,11)(2019,19)(2020,28)(2021,38)(2022,47)(2023,55)(2024,61)(2025,65)};
\addlegendentry{Methodological studies}
\end{axis}
\end{tikzpicture}
\caption[Illustrative trajectory of GNN research]{Illustrative trajectory of graph neural network research, normalised to the most recent year of application work.}
\label{fig:pubtrend}
\end{figure}

The scope of this survey is the application of GNNs across domains, read through a shared methodological lens. \Cref{fig:timeline} sketches the lineage that produced the current toolkit, from the recursive models of the mid-2000s \cite{gori2005newmodel,scarselli2009gnn} through the spectral constructions that tied graph learning to signal processing \cite{bruna2014spectral,defferrard2016chebnet}, the convolutional and attentional designs that made it practical \cite{kipf2017gcn,hamilton2017graphsage,velickovic2018gat}, the message-passing abstraction that unified them \cite{gilmer2017mpnn}\footnote{Reference implementation: \url{https://github.com/brain-research/mpnn}}, and the transformers and foundation-model efforts that define the present frontier \cite{ying2021graphormer,mao2024gfm}. The volume of application research has risen sharply over the same period, as \Cref{fig:pubtrend} indicates. We do not attempt to be exhaustive at the level of individual papers, an impossible goal given the publication rate, but we do aim to be complete at the level of domains, mechanisms, and the failure modes that connect them.

A word on boundaries. The models examined here are graph neural networks in the message-passing sense together with their close relatives, including the spectral constructions that preceded them and the graph transformers that generalize them. Classical network science, graph kernels, and probabilistic graphical models enter only as points of comparison, not as subjects in their own right. On the application side the organizing principle is the domain rather than the task, so a method that performs link prediction appears wherever its links live, in a knowledge graph, a recommender, or a molecular interaction network, rather than collected into a single chapter on link prediction. This choice follows from the survey's thesis that the domain, through the meaning it assigns to nodes and edges, determines which methods work, so that grouping by domain reveals more than grouping by abstract task.

This survey is written for a reader who already knows what a neural network is and wants a defensible map of where graph methods stand. We assume familiarity with gradient-based learning and basic linear algebra and build the graph-specific machinery from there. Two conventions govern the comparative material. Quantitative claims appear only when a cited source supplies the number, and that source is named in place; where no such number exists we use a three-level qualitative scale, drawn as filled, half, and open circles, that records the authors' own assessment and is labelled as such. Charts that present trends not tied to a single benchmark are marked illustrative. These conventions are deliberately conservative, since the alternative, presenting estimated or half-remembered numbers as if they were measured, would defeat the survey's purpose.

The contributions of this work are the following.
\begin{enumerate}[leftmargin=*,itemsep=2pt,topsep=3pt]
\item A unified treatment of GNN mechanisms under one notation and one design space, derived from the message-passing framework rather than presented as a list of named models.
\item A cross-domain comparative analysis that reads twelve application areas through a common schema and surfaces the structural assumptions, dominant architectures, and failure modes they share.
\item A critical appraisal that distinguishes durable empirical results from artefacts of weak baselines and favourable evaluation, applied uniformly across domains.
\item An account of the constraints that govern deployment, connected to the domains where each is most severe rather than gathered into a closing list.
\item A measured evaluation of graph foundation models and language-model integration that separates the claims the evidence supports from those it does not.
\end{enumerate}

The remainder of this paper is organized to move from mechanism to application to synthesis. \Cref{sec:foundations} fixes the graph-theoretic preliminaries and the learning settings, after which \Cref{sec:arch} develops the architectures from the message-passing recursion and \Cref{sec:taxonomy} organizes them into a design space. The application chapters follow, opening with recommendation and social networks in \Cref{sec:social} and continuing through knowledge graphs and language-model integration in \Cref{sec:kg}, the molecular and biological sciences in \Cref{sec:drug} and \Cref{sec:health}, perception in \Cref{sec:vision}, the mobility and infrastructure settings of \Cref{sec:traffic,sec:power,sec:iot}, the security and industrial domains of \Cref{sec:fraud,sec:industrial}, and the physical sciences in \Cref{sec:science}. \Cref{sec:cross} draws the cross-domain comparison together, \Cref{sec:challenges} examines the constraints on deployment, and \Cref{sec:gfm,sec:future} turn to foundation models and the open road before \Cref{sec:conclusion} concludes.

% ---- Figure F1: historical evolution timeline ----

% ---- Figure F21: illustrative publication trend ----

\Cref{tab:surveycompare} positions the present survey against a representative set of both, and the contrast it records, full breadth across the application domains rather than depth in one, is the gap this survey is written to fill.

\begin{table}[t]
\centering
\caption[Positioning against representative prior graph-neural-network surveys.]{Positioning against representative prior graph neural network surveys. For each survey, \hbfull{} marks an application domain or cross-cutting theme treated in depth, \hbhalf{} one touched in passing, and \hbempty{} one not addressed; the marks reflect each survey's primary scope rather than a measured quantity. The breadth that distinguishes the present survey is the full bottom row.}
\label{tab:surveycompare}
\footnotesize
\setlength{\tabcolsep}{3.2pt}
\resizebox{\textwidth}{!}{%
\begin{tabular}{@{}p{2.05cm}p{2.75cm}c*{13}{c}@{}}
\toprule
\textbf{Survey} & \textbf{Primary focus} & \textbf{Yr} & \rotatebox[origin=l]{90}{Recommendation} & \rotatebox[origin=l]{90}{Knowledge \& LMs} & \rotatebox[origin=l]{90}{Molecules} & \rotatebox[origin=l]{90}{Healthcare} & \rotatebox[origin=l]{90}{Vision} & \rotatebox[origin=l]{90}{Traffic} & \rotatebox[origin=l]{90}{Power} & \rotatebox[origin=l]{90}{Wireless/6G} & \rotatebox[origin=l]{90}{Security} & \rotatebox[origin=l]{90}{Industrial} & \rotatebox[origin=l]{90}{Materials} & \rotatebox[origin=l]{90}{Theory} & \rotatebox[origin=l]{90}{Trustworthy} \\
\midrule
Wu et al.~\cite{wu2021survey} & General methods \& taxonomy & 2021 & \hbhalf & \hbhalf & \hbhalf & \hbempty & \hbhalf & \hbhalf & \hbempty & \hbempty & \hbempty & \hbempty & \hbempty & \hbhalf & \hbempty \\
Zhou et al.~\cite{zhou2020review} & Methods \& applications & 2020 & \hbhalf & \hbhalf & \hbhalf & \hbempty & \hbhalf & \hbempty & \hbempty & \hbempty & \hbempty & \hbempty & \hbempty & \hbhalf & \hbempty \\
Zhang et al.~\cite{zhang2022dlgsurvey} & Deep learning on graphs & 2022 & \hbhalf & \hbhalf & \hbhalf & \hbempty & \hbhalf & \hbempty & \hbempty & \hbempty & \hbempty & \hbempty & \hbempty & \hbhalf & \hbempty \\
Wu et al.~\cite{wu2023recsurvey} & Recommender systems & 2023 & \hbfull & \hbhalf & \hbempty & \hbempty & \hbempty & \hbempty & \hbempty & \hbempty & \hbempty & \hbempty & \hbempty & \hbempty & \hbempty \\
Wang et al.~\cite{wang2022hgsurvey} & Heterogeneous embedding & 2022 & \hbhalf & \hbhalf & \hbempty & \hbempty & \hbempty & \hbempty & \hbempty & \hbempty & \hbhalf & \hbempty & \hbempty & \hbempty & \hbempty \\
Ye et al.~\cite{ye2022kgsurvey} & Knowledge graphs & 2022 & \hbempty & \hbfull & \hbempty & \hbempty & \hbempty & \hbempty & \hbempty & \hbempty & \hbempty & \hbempty & \hbempty & \hbempty & \hbempty \\
Jin et al.~\cite{jin2024llmongraphs} & LLMs on graphs & 2024 & \hbhalf & \hbfull & \hbhalf & \hbempty & \hbempty & \hbempty & \hbempty & \hbempty & \hbempty & \hbempty & \hbempty & \hbempty & \hbempty \\
Ren et al.~\cite{ren2024llmgraphsurvey} & LLMs for graphs & 2024 & \hbhalf & \hbfull & \hbempty & \hbempty & \hbempty & \hbempty & \hbempty & \hbempty & \hbempty & \hbempty & \hbempty & \hbempty & \hbempty \\
Peng et al.~\cite{peng2024graphragsurvey} & Graph retrieval-augmented gen. & 2024 & \hbempty & \hbfull & \hbempty & \hbempty & \hbempty & \hbempty & \hbempty & \hbempty & \hbempty & \hbempty & \hbempty & \hbempty & \hbempty \\
Wieder et al.~\cite{wieder2020molsurvey} & Molecular property prediction & 2020 & \hbempty & \hbempty & \hbfull & \hbempty & \hbempty & \hbempty & \hbempty & \hbempty & \hbempty & \hbempty & \hbempty & \hbempty & \hbempty \\
Liao et al.~\cite{liao2021powersurvey} & Power systems & 2022 & \hbempty & \hbempty & \hbempty & \hbempty & \hbempty & \hbempty & \hbfull & \hbempty & \hbempty & \hbempty & \hbempty & \hbempty & \hbempty \\
Jiang et al.~\cite{jiang2022trafficsurvey} & Traffic forecasting & 2022 & \hbempty & \hbempty & \hbempty & \hbempty & \hbempty & \hbfull & \hbempty & \hbempty & \hbempty & \hbempty & \hbempty & \hbempty & \hbempty \\
Jin et al.~\cite{jin2024gnn4ts} & Time-series forecasting & 2024 & \hbempty & \hbempty & \hbempty & \hbempty & \hbempty & \hbhalf & \hbhalf & \hbempty & \hbempty & \hbhalf & \hbempty & \hbempty & \hbempty \\
Sun et al.~\cite{sun2022advsurvey} & Adversarial robustness & 2022 & \hbempty & \hbempty & \hbempty & \hbempty & \hbempty & \hbempty & \hbempty & \hbempty & \hbhalf & \hbempty & \hbempty & \hbempty & \hbhalf \\
Dai et al.~\cite{dai2024trustworthy} & Trustworthy GNNs & 2024 & \hbempty & \hbempty & \hbempty & \hbempty & \hbempty & \hbempty & \hbempty & \hbempty & \hbhalf & \hbempty & \hbempty & \hbempty & \hbfull \\
Yuan et al.~\cite{yuan2023explainsurvey} & Explainability & 2023 & \hbempty & \hbempty & \hbempty & \hbempty & \hbempty & \hbempty & \hbempty & \hbempty & \hbempty & \hbempty & \hbempty & \hbempty & \hbhalf \\
Sato~\cite{sato2020expressivesurvey} & Expressive power & 2020 & \hbempty & \hbempty & \hbempty & \hbempty & \hbempty & \hbempty & \hbempty & \hbempty & \hbempty & \hbempty & \hbempty & \hbfull & \hbempty \\
Liu et al.~\cite{liu2023gfmsurvey} & Graph foundation models & 2023 & \hbhalf & \hbhalf & \hbhalf & \hbempty & \hbempty & \hbempty & \hbempty & \hbempty & \hbempty & \hbempty & \hbempty & \hbhalf & \hbempty \\
\textbf{This survey} & \textbf{Cross-domain, unified} & \textbf{2026} & \hbfull & \hbfull & \hbfull & \hbfull & \hbfull & \hbfull & \hbfull & \hbfull & \hbfull & \hbfull & \hbfull & \hbfull & \hbfull \\
\bottomrule
\end{tabular}}
\end{table}

\FloatBarrier

%================================ SECTION 2 ===================================
\section{Mathematical foundations}
\label{sec:foundations}

The machinery of graph learning rests on a small amount of linear algebra and a precise statement of what a model is asked to predict. This section fixes both. We introduce the matrices that describe a graph and the spectral objects derived from them, define the learning settings that recur throughout the survey, and state the invariance properties that any sound architecture must respect. The notation established here is used without further comment in every later section.

\subsection{Graphs and their matrices}
\label{subsec:graphmatrices}

A graph records which entities are related and, in the attributed case that concerns machine learning, attaches a description to each entity.

\begin{definition}[Attributed graph]
\label{def:graph}
An attributed graph $\graph=(\vset,\eset,\Feat)$ consists of a set of $n=\lvert\vset\rvert$ nodes, a set of edges $\eset\subseteq\vset\times\vset$ with $m=\lvert\eset\rvert$, and a feature matrix $\Feat\in\real^{n\times d}$ whose $v$-th row $\xv_v$ describes node $v$. Edges may carry features $\ev_{uv}$ and, in the relational setting, a type drawn from a relation set $\mathcal{R}$.
\end{definition}

Connectivity is captured by the adjacency matrix,
\begin{equation}\label{eq:adj}
\Adj_{uv}=
\begin{cases}
1 & (u,v)\in\eset,\\
0 & \text{otherwise,}
\end{cases}
\end{equation}
which becomes symmetric for undirected graphs and admits real weights $\Adj_{uv}\in\real_{\ge 0}$ when edges differ in strength. The number of edges incident to a node is its degree, collected on the diagonal of
\begin{equation}\label{eq:deg}
\Deg=\operatorname{diag}(d_1,\dots,d_n),\qquad d_v=\sum_{u\in\vset}\Adj_{vu}.
\end{equation}
The interplay of $\Adj$ and $\Deg$ produces the Laplacian, the single most consequential operator in graph signal processing,
\begin{equation}\label{eq:laplacian}
\Lap=\Deg-\Adj .
\end{equation}
For an undirected graph $\Lap$ is symmetric and positive semidefinite, its smallest eigenvalue is zero with the constant vector as eigenvector, and the multiplicity of the zero eigenvalue counts the connected components. Two normalized variants appear repeatedly, the symmetric form
\begin{equation}\label{eq:normlap}
\Lnorm=\Iden-\Deg^{-1/2}\Adj\,\Deg^{-1/2}
\end{equation}
and the random-walk form
\begin{equation}\label{eq:rwlap}
\mathbf{L}_{\mathrm{rw}}=\Iden-\Deg^{-1}\Adj .
\end{equation}
Normalization bounds the spectrum, so that the eigenvalues of $\Lnorm$ lie in $[0,2]$, which matters because it lets spectral filters be expressed as stable polynomials of a bounded operator.

The Laplacian carries a spectral structure that connects graphs to classical signal processing. Writing its eigendecomposition as
\begin{equation}\label{eq:eig}
\Lnorm=\mathbf{U}\,\bm{\Lambda}\,\mathbf{U}\trans,\qquad
\bm{\Lambda}=\operatorname{diag}(\lambda_1,\dots,\lambda_n),\quad
0=\lambda_1\le\cdots\le\lambda_n\le 2,
\end{equation}
the orthonormal eigenvectors in $\mathbf{U}$ play the role of Fourier modes on the graph, with small eigenvalues indexing smooth modes that vary slowly across edges and large eigenvalues indexing oscillatory ones. The graph Fourier transform of a signal $\xv\in\real^{n}$ and its inverse follow immediately,
\begin{equation}\label{eq:gft}
\hat{\xv}=\mathbf{U}\trans\xv,\qquad \xv=\mathbf{U}\,\hat{\xv}.
\end{equation}
The notion of smoothness that these modes formalize is made quantitative by the Dirichlet energy, the quadratic form
\begin{equation}\label{eq:dirichlet}
\xv\trans\Lap\,\xv=\tfrac{1}{2}\sum_{(u,v)\in\eset}\Adj_{uv}\,(x_u-x_v)^2 ,
\end{equation}
which is small when the signal agrees across edges and large when it disagrees. This single expression returns in the analysis of over-smoothing, where repeated message passing drives the Dirichlet energy of node representations toward zero and erases the distinctions a classifier needs.

One renormalized operator deserves separate mention because it underlies the most widely used architecture in the field. Adding self-loops and symmetrically normalizing gives
\begin{equation}\label{eq:normadj}
\hat{\Adj}=\Dtil^{-1/2}\,\Atil\,\Dtil^{-1/2},\qquad
\Atil=\Adj+\Iden,\quad \Dtil=\Deg+\Iden ,
\end{equation}
an operator whose eigenvalues are shifted into a range that stabilizes repeated multiplication. The derivation that turns \cref{eq:normadj} into a graph convolution is taken up in the next section; here it is enough to record that the matrix exists and why its spectrum is well behaved. The Laplacian's properties are what make the spectral viewpoint usable. Because it is symmetric and positive semidefinite, it has a full set of real, orthogonal eigenvectors and nonnegative eigenvalues, which lets any graph signal be decomposed into a sum of these eigenvectors much as a time signal decomposes into sinusoids. The eigenvalues order the eigenvectors from smooth to oscillatory, the smallest belonging to signals that vary little across edges and the largest to signals that flip sign between neighbours, and this ordering is the bridge between the algebra of the matrix and the intuition of frequency on a graph. Everything the spectral methods do, and much of the analysis of over-smoothing, rests on this decomposition existing and behaving well.

\subsection{Variants of the basic graph}
\label{subsec:variants}

The simple undirected graph is a starting point rather than the only object of interest, and several variants recur across the application chapters with direct consequences for how a model is built. Directed graphs break the symmetry of $\Adj$, so that aggregation must distinguish incoming from outgoing edges and may maintain separate transformations for each direction; citation, traffic, and provenance graphs all carry this asymmetry, and discarding it throws away information about who influences whom. Weighted graphs replace the binary entries of \cref{eq:adj} with real strengths, which lets the data express that some neighbours matter more than others before any learning takes place. Heterogeneous and relational graphs attach a type to every node and edge and are described by a family of adjacency matrices, one per relation,
\begin{equation}\label{eq:reladj}
\{\Adj_r\}_{r\in\mathcal{R}},\qquad \Adj_r\in\real^{n\times n},
\end{equation}
a representation that knowledge graphs and many industrial systems require and that the relational architectures of the next section consume directly. Bipartite graphs split the nodes into two disjoint sets with edges only between them, the natural shape of a user-item recommendation graph. Signed graphs admit negative edges that encode distrust or opposition, which inverts the usual smoothing assumption on the edges that carry them. Hypergraphs let a single edge join more than two nodes at once, capturing group interactions that a pairwise graph can only approximate. Temporal graphs add a time index, either as an ordered sequence of snapshots
\begin{equation}\label{eq:tempseq}
\{\graph^{(t)}\}_{t=1}^{T},\qquad \graph^{(t)}=\big(\vset^{(t)},\eset^{(t)},\Feat^{(t)}\big),
\end{equation}
or as a stream of timestamped events, and the gap between these two encodings shapes the dynamic architectures examined later. None of these variants overturns the principle of neighbourhood aggregation; each changes only what counts as a neighbour and how a message is formed. These variants matter because real data rarely arrives as the simple undirected graph that introductory treatments assume. A directed graph distinguishes the source and target of a relation, which a citation or a transaction requires; a weighted graph records the strength of a connection, which a correlation or a distance supplies; a graph with typed nodes and edges captures the several kinds of entity and relation that knowledge graphs and many real systems contain; and a graph that changes over time describes the many domains where structure evolves. Each variant adjusts what a neighbour is and how its message is formed, and the survey's domains draw on all of them, which is why the abstract framework is stated generally enough to specialize to each rather than fixed to the simplest case.

\subsection{Spectral filters and random walks}
\label{subsec:spectral}

Two views of propagation organize much of what follows, and both descend from the Laplacian. The spectral view treats learning as filtering. A spectral filter modulates each Fourier mode by a learned function of its frequency,
\begin{equation}\label{eq:specfilter}
g_{\bm{\theta}}(\Lnorm)=\mathbf{U}\,g_{\bm{\theta}}(\bm{\Lambda})\,\mathbf{U}\trans ,
\end{equation}
so that a low-pass filter suppressing large eigenvalues smooths a signal across edges, while a high-pass filter preserves the differences between neighbours that heterophilous tasks depend on. The eigenvalues carry structural meaning of their own, since the smallest nonzero one, the algebraic connectivity, measures how well connected the graph is and how quickly information mixes across it. The spatial view treats propagation as a random walk. The row-normalized transition matrix
\begin{equation}\label{eq:rwtrans}
\mathbf{P}=\Deg^{-1}\Adj
\end{equation}
moves probability mass from a node to its neighbours, and its powers describe multi-step diffusion. Personalized PageRank tempers that diffusion with a restart that holds mass near a source,
\begin{equation}\label{eq:ppr}
\bm{\pi}=\alpha\big(\Iden-(1-\alpha)\,\hat{\Adj}\big)^{-1}\mathbf{e},
\end{equation}
with teleport probability $\alpha$ and a one-hot source $\mathbf{e}$, and this exact expression resurfaces as the propagation rule of architectures that decouple feature transformation from neighbourhood mixing. The spectral and spatial views describe the same operator from two directions, and the architectures of the next section amount to choices about which of these descriptions to make learnable. The progression from early spectral methods to the convolutions used today is a progression toward locality and efficiency. Defining a filter directly in the spectral domain requires the eigendecomposition of the Laplacian, which is expensive and ties the filter to a single graph, and it produces filters that are not localized in space. Approximating the filter by a polynomial of the Laplacian removes the need for the eigendecomposition and makes the filter act within a bounded number of hops, and truncating the polynomial to its first term yields the simple neighbourhood average that the most widely used convolution performs. Each step trades some of the generality of an arbitrary spectral filter for the locality, efficiency, and transferability that practical use demands, which is why the spatial view came to dominate even though the spectral view explains where it comes from.

\subsection{Tasks and learning settings}
\label{subsec:settings}

Graph learning problems are organized by the level at which a prediction is made. At the node level the model assigns a label or value to each vertex; at the edge level it scores pairs of vertices; at the graph level it summarizes an entire structure into a single output. The three settings share machinery but differ in their readouts and in the way evaluation is set up, and conflating them is a frequent source of confusion when results are compared.

\begin{problem}[Node classification]
\label{prob:node}
Given a graph $\graph$ and labels $y_v$ for nodes $v$ in a training subset $\vset_{\mathrm{train}}\subset\vset$, predict the labels of the remaining nodes $\vset\setminus\vset_{\mathrm{train}}$.
\end{problem}

\begin{problem}[Link prediction]
\label{prob:link}
Given a graph in which only a subset of edges is observed, assign to each candidate pair $(u,v)$ a score reflecting the probability that the edge exists, and rank unobserved pairs accordingly.
\end{problem}

\begin{problem}[Graph classification]
\label{prob:graph}
Given a collection of graphs with associated labels $\{(\graph_i,y_i)\}_{i=1}^{N}$, learn a function that maps a previously unseen graph to its label.
\end{problem}

Cutting across these levels is the distinction between transductive and inductive learning, which is often left implicit and which changes what a reported number means. In the transductive setting the test nodes are present in the graph during training, so the model sees their features and their position in the topology and need only infer their labels; classic semi-supervised node classification on a single citation network is transductive. In the inductive setting the model must generalize to nodes or whole graphs unseen at training time, which is the regime that matters for deployment, since production systems encounter new users, new molecules, and new sensors continuously. An architecture that performs well transductively can fail inductively when it has implicitly memorized positions rather than learning a transferable function, and the sampling-based methods discussed later were motivated in large part by the need to train inductively at scale.

A second cross-cutting axis is the amount and kind of supervision. Fully supervised training assumes a label for every training example, semi-supervised training exploits a graph in which only a fraction of nodes are labelled, and self-supervised training dispenses with task labels during pretraining and instead constructs a learning signal from the data itself. Graphs are unusually well suited to the semi-supervised regime because the structure links labelled to unlabelled nodes and lets information flow between them, which is precisely why a small labelled set can go a long way on a homophilous graph and why the same approach disappoints when the homophily assumption fails.

The objectives that instantiate these settings are standard but worth writing down so that later variations are legible against them. Node classification minimizes a cross-entropy over the labelled set,
\begin{equation}\label{eq:nodece}
\mathcal{L}_{\mathrm{node}}=-\sum_{v\in\vset_{\mathrm{train}}}\sum_{c=1}^{C} y_{vc}\,\log \hat{y}_{vc},
\qquad
\hat{\mathbf{y}}_v=\softmax\!\big(\Wgt_o\,\hv_v^{(L)}\big),
\end{equation}
where $\hv_v^{(L)}$ is the representation produced after $L$ rounds of message passing and $\Wgt_o$ is an output projection. Link prediction scores a pair by comparing the endpoints' representations, a common choice being
\begin{equation}\label{eq:linkscore}
s_{uv}=\sigma\!\big(\hv_u^{(L)\top}\hv_v^{(L)}\big),
\end{equation}
trained against observed edges and sampled negatives $\eset^{-}$ through a binary cross-entropy,
\begin{equation}\label{eq:linkbce}
\mathcal{L}_{\mathrm{link}}
=-\!\!\sum_{(u,v)\in\eset}\!\log s_{uv}
\;-\!\!\sum_{(u,v)\in\eset^{-}}\!\log\big(1-s_{uv}\big).
\end{equation}
Graph-level prediction first collapses the node representations into a single vector through a permutation-invariant readout,
\begin{equation}\label{eq:readout}
\hv_{\graph}=\READOUT\big(\{\hv_v^{(L)}:v\in\vset\}\big),
\end{equation}
and then applies a standard loss to the resulting graph embedding,
\begin{equation}\label{eq:graphce}
\mathcal{L}_{\mathrm{graph}}=\sum_{i=1}^{N}\ell\!\big(g(\hv_{\graph_i}),\,y_i\big),
\end{equation}
with $\ell$ a cross-entropy for classification or a squared error for regression and $g$ a small prediction head.

\subsection{Invariance, equivariance, and the constraint they impose}
\label{subsec:invariance}

A graph has no canonical ordering of its nodes, and any function that pretends otherwise is modelling an artefact of how the data happened to be stored. The correct behaviour is captured by two symmetry requirements. Let $\mathbf{P}$ be an $n\times n$ permutation matrix.

\begin{definition}[Permutation equivariance and invariance]
\label{def:perm}
A node-level function $f$ is permutation equivariant if relabelling the nodes relabels its output in the same way, and a graph-level function $\phi$ is permutation invariant if relabelling the nodes leaves its output unchanged.
\end{definition}

Stated as equations, equivariance reads
\begin{equation}\label{eq:perm-equiv}
f\big(\mathbf{P}\Adj\mathbf{P}\trans,\ \mathbf{P}\Feat\big)=\mathbf{P}\,f(\Adj,\Feat),
\end{equation}
and invariance reads
\begin{equation}\label{eq:perm-inv}
\phi\big(\mathbf{P}\Adj\mathbf{P}\trans,\ \mathbf{P}\Feat\big)=\phi(\Adj,\Feat).
\end{equation}
These conditions are not decorative. They rule out treating the adjacency matrix as an ordinary image-like array on which position-dependent filters slide, which is why convolution on graphs cannot be defined by analogy to a regular grid and must instead be built from operations that commute with permutation. Equivariance is the property a message-passing layer must have, since the per-node update is applied identically everywhere; invariance is the property a graph-level readout must have, which forces the readout to be a symmetric function of its inputs such as a sum, a mean, or a maximum, and explains why these unglamorous aggregators are so hard to improve upon.

The self-supervised objectives that appear in several domains share a common shape that can be stated before any specific instantiation. A pretext task transforms the graph into one or more views and asks the model to recover a target generated from the data,
\begin{equation}\label{eq:ssl-generic}
\mathcal{L}_{\mathrm{ssl}}=\mathbb{E}\big[\ell_{\mathrm{pretext}}\big(f_\theta(\tilde{\graph}),\,t(\graph)\big)\big],
\end{equation}
where $\tilde{\graph}$ is an augmented view, $t(\graph)$ a self-generated target such as a masked feature, a removed edge, or a representation of a second view, and $\ell_{\mathrm{pretext}}$ the matching loss. The contrastive and generative realizations of \cref{eq:ssl-generic} are developed alongside the architectures that use them.

A final preliminary concerns expressive power, which the next section treats in depth and which can be framed here. Because a message-passing layer is permutation equivariant and aggregates a multiset of neighbour states, two nodes whose neighbourhoods are identical up to relabelling receive identical representations regardless of how the network is parameterized. The natural yardstick for what such a model can separate is therefore the graph isomorphism problem, the question of whether two graphs are the same up to relabelling, together with the Weisfeiler-Leman colour-refinement procedure that supplies a practical if incomplete test for it. This lens explains a strength and a limit that recur across domains at once: a graph network distinguishes structures exactly as well as colour refinement does, which suffices for most node-level prediction and falls short on tasks that hinge on counting specific substructures such as rings or cliques. The distinction between invariance and equivariance is worth stating precisely, because the two serve different needs. A graph-level prediction should be invariant, returning the same value however the nodes are numbered, since a molecule's solubility does not depend on the order in which its atoms are listed. A node-level prediction should be equivariant, so that permuting the nodes permutes the outputs correspondingly rather than scrambling them, since relabelling the nodes should relabel their predictions in step. Message passing is built to be equivariant at every layer and invariant after a symmetric readout, which is exactly the pairing the two kinds of task require, and recognizing which symmetry a problem demands is the first check on whether an architecture is appropriate to it.

\subsection{Metrics and benchmarks}
\label{subsec:metrics}

Comparisons across this survey rest on a small set of metrics whose definitions are fixed here so that later tables read unambiguously. Classification at the node and graph levels is reported through accuracy, the fraction of correct predictions,
\begin{equation}\label{eq:accuracy}
\mathrm{Accuracy}=\frac{\mathrm{TP}+\mathrm{TN}}{\mathrm{TP}+\mathrm{TN}+\mathrm{FP}+\mathrm{FN}},
\end{equation}
a quantity that becomes misleading under class imbalance and is then supplemented by precision and recall,
\begin{equation}\label{eq:precrec}
\mathrm{Precision}=\frac{\mathrm{TP}}{\mathrm{TP}+\mathrm{FP}},\qquad
\mathrm{Recall}=\frac{\mathrm{TP}}{\mathrm{TP}+\mathrm{FN}},
\end{equation}
and by their harmonic mean, the F1 score,
\begin{equation}\label{eq:f1}
\mathrm{F1}=\frac{2\,\mathrm{Precision}\cdot\mathrm{Recall}}{\mathrm{Precision}+\mathrm{Recall}} .
\end{equation}
Ranking and detection tasks favour AUROC, which integrates the true-positive rate against the false-positive rate over all thresholds,
\begin{equation}\label{eq:auroc}
\mathrm{AUROC}=\int_{0}^{1}\mathrm{TPR}\big(\mathrm{FPR}^{-1}(x)\big)\,\mathrm{d}x ,
\end{equation}
and is preferred in fraud detection and link prediction because it does not commit to a single operating point. Regression tasks, common in molecular property prediction and forecasting, report error rather than agreement, most often the root mean square error,
\begin{equation}\label{eq:rmse}
\mathrm{RMSE}=\sqrt{\tfrac{1}{n}\textstyle\sum_{i=1}^{n}(y_i-\hat{y}_i)^2},
\end{equation}
the mean absolute error,
\begin{equation}\label{eq:mae}
\mathrm{MAE}=\tfrac{1}{n}\textstyle\sum_{i=1}^{n}\lvert y_i-\hat{y}_i\rvert ,
\end{equation}
and, where relative error is the quantity of interest, the mean absolute percentage error,
\begin{equation}\label{eq:mape}
\mathrm{MAPE}=\frac{100}{n}\textstyle\sum_{i=1}^{n}\left\lvert\frac{y_i-\hat{y}_i}{y_i}\right\rvert .
\end{equation} The graph descriptors, spectral objects, learning objectives, invariance properties, and evaluation metrics introduced in this section, \cref{eq:deg,eq:laplacian,eq:normlap,eq:rwlap,eq:gft,eq:specfilter,eq:rwtrans,eq:linkscore,eq:linkbce,eq:graphce,eq:perm-equiv,eq:perm-inv,eq:accuracy,eq:precrec,eq:f1,eq:auroc,eq:rmse,eq:mae,eq:mape}, are used without further comment in the sections that follow.
The choice among these is not neutral, and several disagreements in the application literature turn out on inspection to be disputes about which metric to optimize rather than about which model is stronger.

Benchmarks give these settings their empirical content, and a small number of suites account for most reported comparisons. \Cref{tab:benchmarksummary} lists the ones that recur across this survey, with the level they target and the property that made each notable. The trajectory they trace is itself informative: the early citation networks are small enough that careful tuning of a simple model rivals elaborate architectures, the later large-scale and controlled suites were built precisely to expose that fact, and the molecular and scientific benchmarks introduced split protocols, such as scaffold splitting, that test generalization rather than memorization.

\begin{table}[t]
\centering
\caption[Representative benchmark suites for graph learning]{Representative benchmark suites for graph learning. Sources cite the works that established each suite's common use in graph neural network evaluation.}
\label{tab:benchmarksummary}
\renewcommand{\arraystretch}{1.25}
\begin{tabularx}{\textwidth}{@{}l l X l@{}}
\toprule
\textbf{Suite} & \textbf{Levels} & \textbf{Notable property} & \textbf{Source} \\
\midrule
Citation networks &
\makecell[l]{node} &
Long-standing semi-supervised default; now considered too small to discriminate models &
\cite{kipf2017gcn} \\
Open Graph Benchmark &
\makecell[l]{node, link,\\ graph} &
Standardized large-scale datasets and splits introduced to curb small-data overfitting &
\cite{hu2020ogb} \\
MoleculeNet &
\makecell[l]{graph} &
Broad molecular property collection with scaffold splits that test out-of-distribution generalization &
\cite{wu2018moleculenet} \\
Benchmarking GNNs &
\makecell[l]{node, edge,\\ graph} &
Controlled medium-scale suite showing that many architectures perform comparably under matched budgets &
\cite{dwivedi2023benchmarking} \\
\bottomrule
\end{tabularx}
\end{table}

The objects assembled here, the Laplacian and its spectrum, the three task levels with their objectives, and the permutation symmetries, are the fixed points against which the next section reads the proliferation of architectures as variations on one idea.

\FloatBarrier

%================================ SECTION 3 ===================================
\section{Architectures and mechanisms}
\label{sec:arch}

\begin{figure}[tb]
\centering
\resizebox{\textwidth}{!}{%
\begin{tikzpicture}[
  font=\sffamily,
  >=Stealth,
  stage/.style={rounded corners=4pt, draw=#1!75!black, fill=#1!12, line width=0.8pt,
                text=gnnink, align=center, inner sep=6pt, minimum height=20mm},
  layer/.style={rounded corners=3pt, draw=gnnteal!80!black, fill=tintteal, line width=0.7pt,
                minimum width=11mm, minimum height=15mm, align=center, font=\scriptsize, text=gnnink},
  head/.style={rounded corners=3pt, draw=#1!80!black, fill=#1!14, line width=0.8pt,
               text=gnnink, align=center, font=\scriptsize, inner sep=4pt, minimum width=26mm, minimum height=12mm},
  gnode/.style={circle, draw=#1!80!black, fill=#1!70, inner sep=0pt, minimum size=4.2mm},
  flow/.style={-{Stealth[length=3mm]}, line width=1.1pt, draw=gnnslate},
  clab/.style={font=\scriptsize\itshape, text=gnngray, align=center}]

% ---- Stage 1: input graph ----
\node[stage=gnnblue, minimum width=30mm] (inp) at (0,0) {};
\node[clab, anchor=south] at (inp.north) {Input graph $\graph=(\vset,\eset)$};
\begin{scope}[shift={(-1.05,-0.05)}, scale=0.95]
  \node[gnode=gnnblue]  (a) at (0,0.5)   {};
  \node[gnode=gnnteal]  (b) at (0.9,0.85){};
  \node[gnode=gnngreen] (c) at (1.05,-0.1){};
  \node[gnode=gnnamber] (d) at (0.15,-0.55){};
  \node[gnode=gnnpurple](e) at (1.5,0.45) {};
  \draw[gnnslate,line width=0.7pt] (a)--(b) (a)--(d) (b)--(c) (c)--(d) (b)--(e) (c)--(e);
\end{scope}
\node[clab, anchor=north, text width=30mm] at (inp.south) {node features $\Feat$, edges $\Adj$};

% ---- Stage 2: feature init ----
\node[stage=gnnocean, minimum width=22mm, right=10mm of inp] (init) {$\Hid^{(0)}=\Feat$\\[2pt]\scriptsize embed nodes\\\scriptsize and edges};
\node[clab, anchor=south] at (init.north) {Initialisation};

% ---- Stage 3: L message-passing layers ----
\node[stage=gnnteal, minimum width=52mm, right=10mm of init, minimum height=24mm] (mp) {};
\node[clab, anchor=south] at (mp.north) {$L$ message-passing layers};
\node[layer] (l1) at ($(mp.center)+(-1.6,0)$) {Layer 1\\[3pt]\scriptsize\textbf{AGG}\\\scriptsize neighbours\\[2pt]\scriptsize\textbf{UPD}\\\scriptsize combine};
\node[layer] (l2) at ($(mp.center)+(0,0)$)    {Layer 2};
\node[layer] (lL) at ($(mp.center)+(1.6,0)$)  {Layer $L$};
\draw[-{Stealth[length=2mm]}, line width=0.9pt, draw=gnnteal!70!black] (l1)--(l2);
\draw[-{Stealth[length=2mm]}, line width=0.9pt, draw=gnnteal!70!black, densely dotted] (l2)--(lL);
\node[clab, anchor=north, text width=52mm] at (mp.south)
  {each node mixes its $k$-hop neighbourhood; receptive field grows with depth};

% ---- Stage 4: representations + readout ----
\node[stage=gnngreen, minimum width=24mm, right=10mm of mp, minimum height=24mm] (out) {$\Hid^{(L)}$\\[2pt]\scriptsize node embeddings\\[5pt]\scriptsize\textbf{READOUT}\\\scriptsize $\to \hv_\graph$};
\node[clab, anchor=south] at (out.north) {Representations};

% ---- Stage 5: task heads ----
\node[head=gnnblue]  (hn) at ($(out.east)+(2.7,1.35)$)  {Node task\\\scriptsize classify / regress $v$};
\node[head=gnnamber] (he) at ($(out.east)+(2.7,0)$)     {Edge task\\\scriptsize link prediction $(u,v)$};
\node[head=gnnpurple](hg) at ($(out.east)+(2.7,-1.35)$) {Graph task\\\scriptsize classify / regress $\graph$};

% ---- flows ----
\draw[flow] (inp)--(init);
\draw[flow] (init)--(mp);
\draw[flow] (mp)--(out);
\draw[flow] (out.east) to[out=20,in=180]  (hn.west);
\draw[flow] (out.east) to[out=0,in=180]   (he.west);
\draw[flow] (out.east) to[out=-20,in=180] (hg.west);
\end{tikzpicture}}
\caption[End-to-end graph neural network pipeline.]{The end-to-end pipeline shared by most graph neural networks. A graph with node features and edges is embedded into initial states, refined by $L$ message-passing layers that each aggregate a node's neighbours and update its state, and read out into node embeddings and, through a permutation-invariant readout, a graph embedding. A task head then turns these representations into node-level, edge-level, or graph-level predictions. The same backbone serves every domain in this survey; only the graph construction and the task head change.}
\label{fig:e2epipeline}
\end{figure}
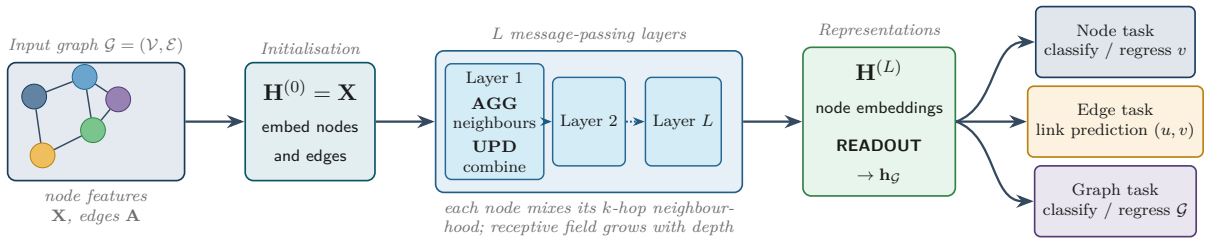

The number of named graph architectures is large and grows monthly, but the number of distinct ideas behind them is small. This section develops the mechanisms from a single abstraction, message passing, and treats the well-known models as points in the space that abstraction defines. \Cref{fig:e2epipeline} shows the end-to-end pipeline this abstraction produces, from an input graph through the message-passing layers to a task head, which the rest of the section fills in. Reading them this way makes their relationships legible: spectral and spatial convolutions are two derivations of the same averaging operator, attention is a learned reweighting of that average, and the expressiveness limits of all of them trace to a single combinatorial fact about colour refinement. The presentation moves from the framework to its instances, then to the variations that handle depth, global context, time, heterogeneity, and the absence of labels.

\subsection{The message-passing framework}
\label{subsec:mpnn}

\begin{figure}[t]
\centering
\resizebox{0.92\textwidth}{!}{%
\begin{tikzpicture}
  % neighbour nodes (teal)
  \node[cnode=gnnteal] (u1) at (0,1.6) {$u_1$};
  \node[cnode=gnnteal] (u2) at (0,0) {$u_2$};
  \node[cnode=gnnteal] (u3) at (0,-1.6) {$u_3$};
  % grouping panel behind neighbours
  \begin{scope}[on background layer]
    \node[panel=gnnteal, fit=(u1)(u3), inner xsep=5mm, inner ysep=4mm] (nb) {};
  \end{scope}
  \node[chip=gnnteal] at (nb.north) {neighbours $\nbr(v)$};
  % aggregate box (amber)
  \node[sbox={gnnamber}{tintamber}, text width=24mm] (agg) at (4.2,0)
    {\textbf{AGGREGATE}\\[1pt]\footnotesize permutation\\\footnotesize invariant};
  % target node (blue, larger)
  \node[cnode=gnnblue, minimum size=13mm] (v) at (8.4,0) {$v$};
  % messages from neighbours
  \draw[carrow=gnnteal] (u1) to[out=0,in=158] (agg.west);
  \draw[carrow=gnnteal] (u2) -- (agg.west);
  \draw[carrow=gnnteal] (u3) to[out=0,in=202] (agg.west);
  % aggregated message to v
  \draw[carrow=gnnblue] (agg) -- node[above,font=\footnotesize]{$\msg_v^{(l)}$} (v);
  % own-state contribution (self loop)
  \draw[carrow=gnnslate] (v.north) to[out=55,in=125,looseness=5]
        node[above=0.5mm,font=\scriptsize]{$\hv_v^{(l)}$} (v.north west);
  % update equation in its own separated panel (no overlap)
  \node[sbox={gnnslate}{tintgray}, text width=92mm] (eq) at (4.2,-3.6)
    {$\displaystyle \hv_v^{(l+1)}=\UPD\!\big(\hv_v^{(l)},\ \AGG_{u\in\nbr(v)}\,\msg_{u\to v}^{(l)}\big)$};
\end{tikzpicture}}
\caption[One round of message passing at a node]{One round of message passing at node $v$: messages from the neighbours $u_1,u_2,u_3$ are combined by a permutation-invariant aggregator and used together with the node's own state $\hv_v^{(l)}$ to update $v$. The update rule appears below.}
\label{fig:mpnn}
\end{figure}
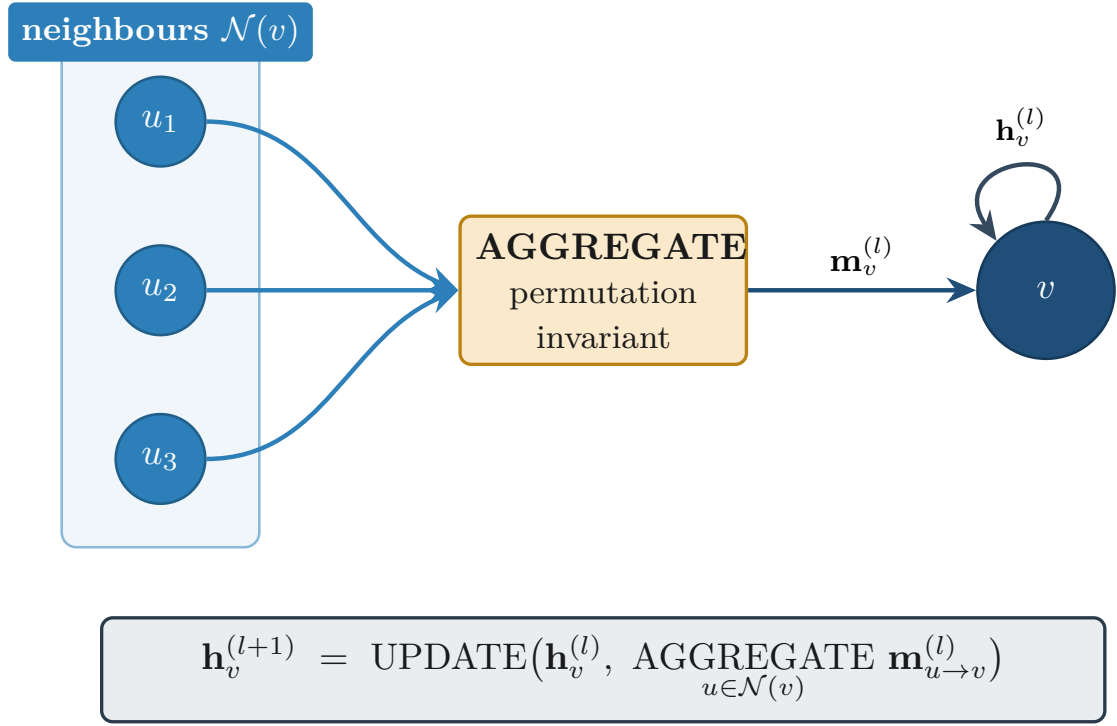

Almost every graph neural network in use computes node representations by alternating two operations, the gathering of information from a node's neighbours and the updating of the node's own state with what was gathered \cite{gilmer2017mpnn,battaglia2018relational}. Writing $\hv_v^{(l)}$ for the representation of node $v$ after $l$ rounds, a message from a neighbour $u$ is formed as
\begin{equation}\label{eq:msg}
\msg_{u\to v}^{(l)}=\textsc{msg}\big(\hv_u^{(l)},\,\hv_v^{(l)},\,\ev_{uv}\big),
\end{equation}
the messages from the neighbourhood are combined by a permutation-invariant aggregator,
\begin{equation}\label{eq:agg}
\msg_v^{(l)}=\AGG\big(\{\msg_{u\to v}^{(l)}:u\in\nbr(v)\}\big),
\end{equation}
and the node state advances through
\begin{equation}\label{eq:upd}
\hv_v^{(l+1)}=\UPD\big(\hv_v^{(l)},\,\msg_v^{(l)}\big).
\end{equation}
After $L$ rounds a graph-level prediction applies the readout of \cref{eq:readout}. The three functions $\textsc{msg}$, $\AGG$, and $\UPD$ are where architectures differ, and the constraint from \cref{subsec:invariance} fixes what they may be: $\AGG$ must be invariant to the order of its inputs, which is why sums, means, and maxima dominate, while $\textsc{msg}$ and $\UPD$ are shared across nodes so that the layer is equivariant. \Cref{alg:mpnn} states the recursion in full, and \Cref{fig:mpnn} shows one round on a small neighbourhood.

\begin{algorithm}[t]
\caption{Generic message passing}
\label{alg:mpnn}
\KwIn{Graph $\graph=(\vset,\eset)$, node features $\{\xv_v\}$, edge features $\{\ev_{uv}\}$, layers $L$}
\KwOut{Node representations $\{\hv_v^{(L)}\}$}
$\hv_v^{(0)} \gets \xv_v$ \textbf{for all} $v\in\vset$\;
\For{$l \gets 0$ \KwTo $L-1$}{
  \ForEach{$v \in \vset$}{
    \ForEach{$u \in \nbr(v)$}{
      $\msg_{u\to v}^{(l)} \gets \textsc{msg}\big(\hv_u^{(l)},\hv_v^{(l)},\ev_{uv}\big)$\;
    }
    $\msg_v^{(l)} \gets \AGG\big(\{\msg_{u\to v}^{(l)} : u \in \nbr(v)\}\big)$\;
    $\hv_v^{(l+1)} \gets \UPD\big(\hv_v^{(l)}, \msg_v^{(l)}\big)$\;
  }
}
\Return $\{\hv_v^{(L)}\}$\;
\end{algorithm}

This abstraction is more than bookkeeping. It tells us that a graph network of depth $L$ sees exactly the $L$-hop rooted subtree around each node and nothing beyond it, which bounds what any such model can compute and frames the over-squashing problem taken up later as the difficulty of routing information through narrow bottlenecks in that subtree. It also makes clear why so much architectural effort concentrates on the aggregator: the expressive power of the whole network is limited by how much the aggregator preserves about the multiset of neighbour states. This multiset view clarifies what aggregation must accomplish and where simple choices fall short. Summation preserves the full multiset up to the injectivity of the functions around it, which is why it underlies the most expressive message-passing networks, while averaging discards the count of neighbours and the maximum discards all but the dominant one, each losing information some tasks need. The practical consequence is that the aggregator is not an interchangeable detail but a determinant of what the network can represent, and the right choice depends on whether the task cares about the presence of a feature, its prevalence, or its extreme, which is a question worth asking explicitly rather than settling by default.

\subsection{From spectral convolution to the graph convolutional network}
\label{subsec:gcn}

The first principled graph convolutions were defined in the spectral domain by analogy to filtering. Using the eigenbasis of \cref{eq:eig}, a filter $g_{\bm{\theta}}$ acts on a signal $\xv$ by
\begin{equation}\label{eq:specconv}
g_{\bm{\theta}}\star\xv=\mathbf{U}\,g_{\bm{\theta}}(\bm{\Lambda})\,\mathbf{U}\trans\xv ,
\end{equation}
which is exact but impractical, since it requires the full eigendecomposition and produces filters that are global and not localized on the graph \cite{bruna2014spectral}. Localizing the filter and removing the eigendecomposition are achieved together by writing $g_{\bm{\theta}}$ as a truncated expansion in Chebyshev polynomials of the scaled Laplacian \cite{defferrard2016chebnet},
\begin{equation}\label{eq:cheb}
g_{\bm{\theta}}\star\xv\approx\sum_{k=0}^{K}\theta_k\,T_k(\tilde{\Lap})\,\xv,
\qquad \tilde{\Lap}=\tfrac{2}{\lambda_{\max}}\Lnorm-\Iden,
\end{equation}
with the polynomials defined by the recurrence
\begin{equation}\label{eq:chebrecur}
T_0(x)=1,\quad T_1(x)=x,\quad T_k(x)=2x\,T_{k-1}(x)-T_{k-2}(x).
\end{equation}
Because $T_k(\tilde{\Lap})$ touches only the $k$-hop neighbourhood, the filter is now strictly localized, and because everything is expressed through sparse matrix products the cost scales with the number of edges rather than with $n^3$. The order $K$ of the expansion sets the trade-off that recurs whenever locality meets reach: a small $K$ produces a tightly localized filter that is cheap and stable but blind to structure more than a few hops away, while a large $K$ widens the receptive field at the cost of more parameters and a greater risk of the numerical instability that motivated rescaling the Laplacian in the first place. The Chebyshev basis is preferred over the naive monomial basis $\{\Lnorm^k\}$ for exactly this reason, since its near-orthogonality on the spectral interval keeps the conditioning of the fit under control as $K$ grows \cite{defferrard2016chebnet}. The graph convolutional network follows from a deliberate oversimplification of \cref{eq:cheb}: keep only $K=1$, tie the two coefficients, and apply the renormalization of \cref{eq:normadj} to keep the spectrum in check \cite{kipf2017gcn}\footnote{Reference implementation: \url{https://github.com/tkipf/gcn}}. The result is the layer
\begin{equation}\label{eq:gcnlayer}
\Hid^{(l+1)}=\sigma\big(\hat{\Adj}\,\Hid^{(l)}\,\Wgt^{(l)}\big),
\end{equation}
which is the single most used equation in the field. Its appeal is its economy: one sparse multiplication mixes neighbours, one dense multiplication transforms features, and one nonlinearity completes the layer. Its weakness is the same economy, since the fixed degree-normalized averaging in $\hat{\Adj}$ cannot express that some neighbours should count more than others, cannot avoid mixing across class boundaries on a heterophilous graph, and degrades when stacked deeply. \Cref{alg:gcn} gives the forward pass and training loop.

\begin{algorithm}[t]
\caption{GCN forward pass and training}
\label{alg:gcn}
\KwIn{Graph $\graph$, features $\Feat$, labels $\{y_v\}_{v\in\vset_{\mathrm{train}}}$, layers $L$, learning rate $\eta$}
\KwOut{Trained weights $\{\Wgt^{(l)}\}$ and output projection $\Wgt_o$}
$\hat{\Adj} \gets \Dtil^{-1/2}\,\Atil\,\Dtil^{-1/2}$ \tcp*{precompute once}
\While{not converged}{
  $\Hid^{(0)} \gets \Feat$\;
  \For{$l \gets 0$ \KwTo $L-1$}{
    $\Hid^{(l+1)} \gets \relu\big(\hat{\Adj}\,\Hid^{(l)}\,\Wgt^{(l)}\big)$\;
  }
  $\widehat{\mathbf{Y}} \gets \softmax\big(\Hid^{(L)}\,\Wgt_o\big)$\;
  $\mathcal{L} \gets$ cross-entropy of $\widehat{\mathbf{Y}}$ against labels on $\vset_{\mathrm{train}}$ \tcp*{\cref{eq:nodece}}
  update $\{\Wgt^{(l)}\},\Wgt_o$ by descent with rate $\eta$\;
}
\Return $\{\Wgt^{(l)}\},\Wgt_o$\;
\end{algorithm}

\subsection{Spatial convolution and neighbourhood sampling}
\label{subsec:sage}

The spatial view dispenses with the spectrum and defines convolution directly as aggregation over neighbours, which is both more intuitive and more flexible, since the aggregator can be any learnable symmetric function rather than a polynomial of the Laplacian \cite{monti2017monet,atwood2016dcnn,niepert2016patchysan}. The architecture that made this view practical at scale is GraphSAGE, whose contribution is less a new aggregator than a training scheme: instead of operating on the whole graph it samples a fixed number of neighbours at each hop and aggregates over the sample \cite{hamilton2017graphsage}\footnote{Reference implementation: \url{https://github.com/williamleif/GraphSAGE}}. A layer transforms a node by combining its previous state with an aggregate of sampled neighbour states,
\begin{equation}\label{eq:sage}
\hv_v^{(k)}=\sigma\Big(\Wgt^{(k)}\big[\hv_v^{(k-1)}\concat \textsc{agg}_k\big(\{\hv_u^{(k-1)}:u\in\mathcal{S}_k(v)\}\big)\big]\Big),
\end{equation}
where $\mathcal{S}_k(v)$ is the sampled neighbourhood and $\textsc{agg}_k$ a mean, pooling, or recurrent aggregator. Two consequences matter for deployment. Sampling bounds the receptive field so that the memory and time of a minibatch no longer depend on the size of the whole graph, which is what lets the method train on graphs with hundreds of millions of edges. The concatenation of a node's own state with the neighbour aggregate, rather than their summation, preserves a distinction between self and context that pure averaging discards. The same sampling idea was later refined by importance sampling and by clustering the graph into subgraphs for each minibatch \cite{chen2018fastgcn,chiang2019clustergcn,zeng2020graphsaint}, techniques returned to when scalability is examined directly. \Cref{alg:sage} gives the sampling procedure.

\begin{algorithm}[t]
\caption{GraphSAGE minibatch with neighbour sampling}
\label{alg:sage}
\KwIn{Graph $\graph$, features $\{\xv_v\}$, depth $K$, sample sizes $\{S_k\}$, aggregators $\{\textsc{agg}_k\}$, minibatch $\mathcal{B}$}
\KwOut{Embeddings $\{\hv_v^{(K)}:v\in\mathcal{B}\}$}
$\hv_v^{(0)} \gets \xv_v$\;
\For{$k \gets 1$ \KwTo $K$}{
  \ForEach{$v$ in the $k$-hop support of $\mathcal{B}$}{
    $\mathcal{S}_k(v) \gets$ sample $S_k$ neighbours uniformly from $\nbr(v)$\;
    $\mathbf{a}_v \gets \textsc{agg}_k\big(\{\hv_u^{(k-1)} : u \in \mathcal{S}_k(v)\}\big)$\;
    $\hv_v^{(k)} \gets \sigma\big(\Wgt^{(k)}\,[\hv_v^{(k-1)} \concat \mathbf{a}_v]\big)$\;
    $\hv_v^{(k)} \gets \hv_v^{(k)} / \lVert \hv_v^{(k)}\rVert_2$\;
  }
}
\Return $\{\hv_v^{(K)} : v \in \mathcal{B}\}$\;
\end{algorithm}

\subsection{Attention on graphs}
\label{subsec:gat}

Fixed averaging weights every neighbour by degree alone, which is rarely what the task wants. The graph attention network replaces the fixed coefficients with learned ones computed from the endpoints' features \cite{velickovic2018gat}\footnote{Reference implementation: \url{https://github.com/PetarV-/GAT}}. An unnormalized score for the edge $(u,v)$ is
\begin{equation}\label{eq:gat-score}
e_{uv}=\mathrm{LeakyReLU}\big(\mathbf{a}\trans[\Wgt\hv_u\concat \Wgt\hv_v]\big),
\end{equation}
these scores are normalized over the neighbourhood by a softmax,
\begin{equation}\label{eq:gat-alpha}
\alpha_{uv}=\frac{\exp(e_{uv})}{\sum_{w\in\nbr(v)}\exp(e_{wv})},
\end{equation}
and the node is updated as an attention-weighted combination, typically with $K$ heads concatenated for stability,
\begin{equation}\label{eq:gat-update}
\hv_v'=\big\Vert_{k=1}^{K}\ \sigma\Big(\sum_{u\in\nbr(v)}\alpha_{uv}^{k}\,\Wgt^{k}\hv_u\Big).
\end{equation}
Attention buys two things. It lets the model down-weight neighbours that would otherwise corrupt a representation, which is the mechanism behind much of its advantage on noisy graphs, and the coefficients $\alpha_{uv}$ offer a first, if unreliable, handle on interpretability. What attention does not buy is a change in expressive power: because it still aggregates a multiset of neighbour states, it sits under the same Weisfeiler-Leman ceiling as the unweighted models, a point developed next. \Cref{alg:gat} lists the computation.

\begin{algorithm}[t]
\caption{GAT layer with multi-head attention}
\label{alg:gat}
\KwIn{Node features $\{\hv_v\}$, per-head weights $\{\Wgt^k\}$, attention vectors $\{\mathbf{a}^k\}$, heads $K$}
\KwOut{Updated features $\{\hv_v'\}$}
\For{$k \gets 1$ \KwTo $K$}{
  \ForEach{$v\in\vset$ and $u\in\nbr(v)\cup\{v\}$}{
    $e_{uv}^{k} \gets \mathrm{LeakyReLU}\big((\mathbf{a}^k)\trans[\Wgt^k\hv_u \concat \Wgt^k\hv_v]\big)$\;
  }
  \ForEach{$v\in\vset$}{
    $\alpha_{uv}^{k} \gets \mathrm{softmax}_u\big(e_{uv}^{k}\big)$ over $u\in\nbr(v)\cup\{v\}$\;
  }
}
\ForEach{$v\in\vset$}{
  $\hv_v' \gets \big\Vert_{k=1}^{K}\ \sigma\big(\sum_{u}\alpha_{uv}^{k}\,\Wgt^k\hv_u\big)$\;
}
\Return $\{\hv_v'\}$\;
\end{algorithm}

\subsection{Edge features and message functions}
\label{subsec:edge}

The message function of \cref{eq:msg} was written to depend on the edge feature $\ev_{uv}$, and several architectures take that dependence seriously rather than discarding it. The simplest route concatenates the edge feature into the message, but a more expressive one lets the edge feature parameterize the transformation applied to the neighbour, as in edge-conditioned convolution, where a small network maps each edge label to a weight matrix,
\begin{equation}\label{eq:ecc}
\msg_{u\to v}=\Theta(\ev_{uv})\,\hv_u,\qquad \Theta:\real^{d_e}\to\real^{d\times d},
\end{equation}
so that a bond type in a molecule or a relation in a scene graph selects how its endpoint is read \cite{simonovsky2017ecc}. The general message-passing formulation makes the edge network a first-class component and was introduced to unify the molecular models that need it \cite{gilmer2017mpnn}. Gating offers a complementary refinement, replacing the fixed combination of self and neighbour terms with learned gates that control how much of each message survives, which stabilizes training on deeper stacks \cite{bresson2017gatedgcn}. The practical lesson is that edges are not decoration: on graphs where the edge type carries the chemistry or the semantics, a model that folds the edge into the message function, rather than into the topology alone, is consistently the stronger choice. The mechanisms for using edge features range from the simple to the elaborate. The plainest approach concatenates an edge's features onto the neighbour message before aggregation, letting the network modulate what each neighbour contributes by the nature of the connection. A more expressive approach conditions the message transformation itself on the edge, so that an edge generates the weight matrix applied to its neighbour, which lets the relationship determine not merely what is sent but how it is transformed. Some architectures maintain edge representations updated in parallel with node representations, passing information from nodes to their incident edges and back, which is natural when edges carry as much meaning as nodes, as bonds do in molecules or transactions do in payment networks. The recurring finding is that the more of a problem's signal lives on its edges, the more it matters to give edges first-class treatment rather than collapsing them into the adjacency structure.

\subsection{Geometric and equivariant message passing}
\label{subsec:equivariant}

A different constraint arises when nodes carry coordinates in space rather than only abstract features, as atoms, particles, and meshes do. The relevant symmetry is then not only permutation but also rotation, translation, and reflection, and a model that ignores it wastes capacity learning what geometry already guarantees. Equivariant message passing builds the symmetry in, keeping a scalar feature that is invariant to rigid motion alongside a vector feature that rotates with the input. The $E(n)$-equivariant network is the clearest instance, forming messages from relative distances and moving coordinates along relative directions,
\begin{equation}\label{eq:egnn-m}
\msg_{u\to v}=\phi_m\big(\hv_u,\hv_v,\lVert \mathbf{x}_u-\mathbf{x}_v\rVert^2\big),
\end{equation}
\begin{equation}\label{eq:egnn-x}
\mathbf{x}_v'=\mathbf{x}_v+\sum_{u\in\nbr(v)}(\mathbf{x}_v-\mathbf{x}_u)\,\phi_x\big(\msg_{u\to v}\big),
\end{equation}
so that a rotation of the input produces the same rotation of the output while the scalar predictions do not change at all \cite{satorras2021egnn}. The payoff is sample efficiency: a force field or a binding affinity that respects these symmetries by construction needs far fewer examples than one that must rediscover them, which is why geometric variants dominate the molecular and physical-simulation chapters and are revisited there in detail. The geometric models form a hierarchy of increasing structure. The simplest respect only the distances between points, which already suffices to make a model invariant to rotation and translation while discarding directional information. More expressive models retain directional information by passing vector-valued messages that rotate with the system, which is necessary when the quantity of interest, such as a force, is itself directional. The most structured incorporate angular and higher-order geometric information, capturing the arrangement of neighbours around a point rather than only their distances. Each step up the hierarchy adds expressiveness at a cost in computation, and the appropriate level depends on whether the target property depends on distances alone, on directions, or on the full local geometry, a choice the molecular and materials chapters return to because it determines both accuracy and cost in those domains.

\subsection{Expressive power and the Weisfeiler-Leman connection}
\label{subsec:gin}

A precise question underlies the comparison of all these models: which graphs can a message-passing network tell apart? The answer is given by an old combinatorial algorithm. Colour refinement, the one-dimensional Weisfeiler-Leman test, repeatedly recolours each node by hashing its current colour together with the multiset of its neighbours' colours,
\begin{equation}\label{eq:wl}
c_v^{(t+1)}=\textsc{hash}\big(c_v^{(t)},\,\{\!\{c_u^{(t)}:u\in\nbr(v)\}\!\}\big),
\end{equation}
and two graphs that receive different colour histograms are certainly non-isomorphic. The structural parallel to \cref{eq:agg,eq:upd} is exact, and it has a sharp consequence.

\begin{proposition}[Weisfeiler-Leman ceiling]
\label{prop:wl}
Any message-passing network of the form \cref{eq:msg,eq:agg,eq:upd} maps two nodes to different representations only if colour refinement assigns them different colours. The bound is attained: with an injective aggregator and update, a network is as discriminative as the one-dimensional Weisfeiler-Leman test, and the graph isomorphism network realizes such a model \cite{xu2019gin,morris2019wlneural}.
\end{proposition}

The construction that attains the bound is GIN, which sums neighbour states, scales the self term, and passes the result through a multilayer perceptron capable of approximating the injective function the proof requires \cite{xu2019gin}\footnote{Reference implementation: \url{https://github.com/weihua916/powerful-gnns}},
\begin{equation}\label{eq:gin}
\hv_v^{(l+1)}=\mathrm{MLP}^{(l)}\Big((1+\epsilon^{(l)})\,\hv_v^{(l)}+\sum_{u\in\nbr(v)}\hv_u^{(l)}\Big).
\end{equation}
The summation is the load-bearing choice. A mean aggregator loses the size of the neighbourhood and a max aggregator loses its multiplicity, so both are strictly weaker than the sum at separating structures, a distinction that principal-neighbourhood aggregation later exploited by combining several aggregators at once \cite{corso2020pna}. The ceiling is not merely theoretical. Tasks that require counting triangles, distinguishing regular graphs, or detecting specific motifs sit beyond one-dimensional colour refinement, and higher-order networks that operate on tuples of nodes were proposed precisely to climb the hierarchy, at a cost in computation that has so far kept them from wide use \cite{morris2019wlneural,sato2020expressivesurvey,morris2024futuretheory}.

The hierarchy is worth stating plainly, since it organizes a large theoretical literature. The $k$-dimensional Weisfeiler-Leman test colours tuples of $k$ nodes rather than single nodes, and its power grows strictly with $k$, so that the two-dimensional test separates graphs the one-dimensional test cannot and the three-dimensional test separates graphs that defeat the two-dimensional one. Networks have been designed to match each level, but the cost of operating on $k$-tuples grows as $n^k$, which confines all but the first level to small graphs. A more practical line raises expressive power without climbing the full hierarchy, by injecting random or positional features that break the symmetry colour refinement is blind to, or by running a base network on many subgraphs and aggregating the results, which detects substructures a single pass misses \cite{sato2020expressivesurvey}. None of these has displaced the plain sum aggregator in routine use, and the gap between what the theory shows is possible and what the benchmarks actually reward remains one of the more honest tensions in the field. \Cref{alg:gin} gives the forward pass with a jumping-knowledge readout.

\begin{algorithm}[t]
\caption{GIN forward pass with graph-level readout}
\label{alg:gin}
\KwIn{Features $\{\xv_v\}$, layers $L$, MLPs $\{\mathrm{MLP}^{(l)}\}$, scalars $\{\epsilon^{(l)}\}$}
\KwOut{Graph embedding $\hv_{\graph}$}
$\hv_v^{(0)} \gets \xv_v$\;
\For{$l \gets 0$ \KwTo $L-1$}{
  \ForEach{$v\in\vset$}{
    $\hv_v^{(l+1)} \gets \mathrm{MLP}^{(l)}\big((1+\epsilon^{(l)})\hv_v^{(l)} + \sum_{u\in\nbr(v)}\hv_u^{(l)}\big)$\;
  }
}
$\hv_{\graph} \gets \big\Vert_{l=0}^{L}\ \sum_{v\in\vset}\hv_v^{(l)}$ \tcp*{concatenate per-layer sums}
\Return $\hv_{\graph}$\;
\end{algorithm}

\subsection{Depth, residual connections, and decoupled propagation}
\label{subsec:depth}

Convolutional networks for images grow more capable as they deepen, and the early expectation that graph networks would behave likewise proved wrong. Stacking many message-passing layers drives node representations toward a common value, the over-smoothing effect analyzed through the Dirichlet energy of \cref{eq:dirichlet}, so that beyond a few layers accuracy falls rather than rises \cite{oono2020expressive}. Several architectural responses recover depth without collapse. Jumping-knowledge networks let the final representation draw on every intermediate layer rather than only the last, which preserves the sharper, less smoothed features of early layers \cite{xu2018jknet}. Decoupling propagation from transformation removes the problem at its source: personalized propagation applies the closed-form diffusion of \cref{eq:ppr} to features that were transformed once, so that depth of propagation no longer entails depth of nonlinear mixing \cite{klicpera2019appnp}. Identity mappings and initial-residual connections, as in GCNII, let a layer default to passing its input through and so train stably at many layers \cite{chen2020gcnii}, and edge dropping during training regularizes the smoothing \cite{rong2020dropedge}. A more radical reading is that the nonlinearities were never doing much work: simplified graph convolution removes them entirely, collapsing $L$ layers into a single fixed propagation $\hat{\Adj}^{L}$ followed by a linear classifier,
\begin{equation}\label{eq:sgc}
\widehat{\mathbf{Y}}=\softmax\big(\hat{\Adj}^{L}\,\Feat\,\Wgt\big),
\end{equation}
and matches the accuracy of the full model on standard benchmarks while training far faster \cite{wu2019sgc}. The success of \cref{eq:sgc} is among the strongest pieces of evidence for the survey's recurring caution: on the datasets where graph networks are usually compared, most of the measured performance comes from neighbourhood averaging, not from the depth or nonlinearity that more elaborate models add. This observation has a constructive side, since if much of the benefit comes from propagation rather than from deep nonlinear transformation, then decoupling the two, propagating features across the graph and learning a shallow predictor on the result, recovers most of the performance at a fraction of the cost. Several influential models take exactly this route, separating the smoothing the graph provides from the transformation the labels require, and their success is itself evidence for the claim that the graph's contribution is the propagation, with the deep machinery around it adding less than its prominence suggests.

\subsection{Normalization, regularization, and stable training}
\label{subsec:training}

The mechanisms that make graph networks train reliably are easy to overlook and decisive in practice. Normalization comes first. Applying standard batch normalization across nodes is complicated by the fact that a minibatch on a graph is a set of interdependent neighbourhoods rather than independent samples, and graph-specific schemes instead normalize within a node's feature vector or rescale node representations so that their pairwise distances do not collapse as depth grows \cite{zhao2020pairnorm}, countering the smoothing measured by \cref{eq:dirichlet}. Regularization comes second. Dropping edges at random during training, the graph analogue of dropout, both regularizes and slows over-smoothing by thinning the paths along which information mixes \cite{rong2020dropedge}, while feature masking and weight decay play their usual roles. A quieter third factor is the optimization itself: the sparse, irregular structure of graph computation produces gradients with high variance across nodes of different degree, so learning rates and initializations tuned for dense networks often transfer poorly. These details rarely headline a paper, yet the difference between a method that reproduces and one that does not frequently lies here rather than in the layer equation, which is part of why the carefully tuned baselines mentioned throughout this survey are so hard to beat. The stabilizing techniques deserve emphasis because they are easy to overlook and disproportionately consequential. Normalization that keeps representations at a consistent scale across layers, dropout adapted to graphs by removing edges rather than only features, and careful initialization together determine whether a model trains stably and generalizes, often more than the architectural choices that receive more attention. A recurring lesson of the empirical literature is that a simple architecture with these elements tuned carefully outperforms an elaborate one without them, which is part of why reported comparisons are so sensitive to implementation details and why reproducing a result can require matching them closely.

\subsection{Graph transformers}
\label{subsec:transformer}

Message passing restricts every layer to a one-hop exchange, which makes long-range dependencies expensive to capture and is one origin of over-squashing. Graph transformers remove the restriction by letting every node attend to every other, using the self-attention of the sequence transformer,
\begin{equation}\label{eq:gt-attn}
\mathrm{Attn}(\mathbf{Q},\mathbf{K},\mathbf{V})=\softmax\Big(\tfrac{\mathbf{Q}\mathbf{K}\trans}{\sqrt{d}}\Big)\mathbf{V},
\end{equation}
with queries, keys, and values projected from node features \cite{dwivedi2021gt,ying2021graphormer}. Global attention discards the graph, so structure has to be reinjected, and how to do so is the central design question. One family adds positional encodings derived from the Laplacian eigenvectors,
\begin{equation}\label{eq:lap-pe}
\mathbf{P}_{\mathrm{pe}}=[\,\mathbf{u}_1,\dots,\mathbf{u}_k\,],
\end{equation}
appending the leading nonconstant eigenvectors to the node features as a learned sense of position \cite{kreuzer2021san,dwivedi2021gt}. Graphormer instead biases the attention scores by encodings of degree, spatial distance, and the edges on the shortest path between two nodes \cite{ying2021graphormer}. A hybrid that has held up well combines a local message-passing branch with a global attention branch in every layer, on the argument that the two capture complementary scales \cite{rampasek2022gps}, and other work shows that even a pure transformer over node and edge tokens, given the right encodings, is competitive \cite{chen2022sat,kim2022tokengt}.

Scalability is the standing obstacle, since the dense attention of \cref{eq:gt-attn} costs $\mathcal{O}(n^2)$ in time and memory and does not survive contact with a graph of millions of nodes. The responses mirror those developed for long sequences: restrict attention to a sampled or local neighbourhood, approximate the softmax with a low-rank or kernel factorization that brings the cost down to near-linear, or coarsen the graph so that global attention runs over a manageable number of supernodes. Each buys scale by returning some of the global reach that motivated the transformer in the first place, and which compromise is right depends on whether the long-range dependencies the task needs are dense or sparse.

The honest assessment is mixed. Graph transformers win clearly on small graphs where global structure matters, such as molecules, and their quadratic attention cost makes them awkward on the large sparse graphs where message passing is cheapest, so the choice between the two families is still governed by graph size more than by any decisive accuracy advantage. The central design question for a graph transformer is how to inject structure into an architecture that by default treats its input as an unordered set. Two devices recur. Positional encodings give each node a signature of its location in the graph, often derived from the eigenvectors of the Laplacian, which play the role that sequence position plays in a language transformer. Structural encodings instead summarize the local topology around a node, its degree or the pattern of short walks from it, so that attention can condition on structure even without explicit edges. A common compromise keeps message passing for local structure and adds a transformer layer for global interaction, which captures long-range dependencies that message passing handles poorly while retaining the inductive bias that makes message passing data-efficient, and this hybrid is where much of the practical value of graph transformers has been found.

\subsection{Temporal and dynamic networks}
\label{subsec:temporal}

Static message passing assumes a fixed topology, an assumption that fails for traffic networks, financial transaction graphs, and social interactions, all of which evolve \cite{kazemi2020dynamicsurvey,skarding2021dynamicsurvey}. Two encodings of time, introduced in \cref{eq:tempseq}, lead to two families of model. Discrete-time methods treat the history as a sequence of snapshots and compose a spatial encoder with a temporal one, most often by feeding the per-snapshot node embeddings of a GNN into a recurrent cell so that each node carries a state summarizing its past \cite{li2018dcrnn}\footnote{Reference implementation: \url{https://github.com/liyaguang/DCRNN}}. Continuous-time methods instead consume a stream of timestamped events and update the representations of the nodes involved as each event arrives, which avoids the information loss of snapshotting at the cost of a more intricate training procedure \cite{zhang2024tgnntrends}. A persistent difficulty, and a reason progress here trails the static case, is that the structure and the signal change together, so a model must separate genuine temporal pattern from drift in the graph itself; recent work that tunes a node's state from its evolving structure rather than from snapshots alone is one attempt to address this \cite{li2026est}. \Cref{alg:tgnn} gives the discrete-time template.

\begin{algorithm}[t]
\caption{Discrete-time dynamic GNN update}
\label{alg:tgnn}
\KwIn{Snapshot sequence $\{\graph^{(t)}\}_{t=1}^{T}$, spatial encoder $f$, recurrent cell $\textsc{rnn}$}
\KwOut{Temporal node states $\{\mathbf{s}_v^{(T)}\}$}
initialize $\mathbf{s}_v^{(0)}$ for all $v$\;
\For{$t \gets 1$ \KwTo $T$}{
  $\{\hv_v^{(t)}\} \gets f\big(\graph^{(t)}\big)$ \tcp*{spatial encoding of snapshot $t$}
  \ForEach{$v\in\vset^{(t)}$}{
    $\mathbf{s}_v^{(t)} \gets \textsc{rnn}\big(\mathbf{s}_v^{(t-1)},\,\hv_v^{(t)}\big)$ \tcp*{temporal update}
  }
}
\Return $\{\mathbf{s}_v^{(T)}\}$\;
\end{algorithm}

\subsection{Heterogeneous and relational message passing}
\label{subsec:hetero}

Many graphs carry typed nodes and edges, and collapsing the types throws away the information that the type encodes. Relational graph convolution keeps a separate transformation per relation and sums the per-relation aggregates, using the family of adjacency matrices from \cref{eq:reladj},
\begin{equation}\label{eq:rgcn}
\hv_v^{(l+1)}=\sigma\Big(\Wgt_0^{(l)}\hv_v^{(l)}+\sum_{r\in\mathcal{R}}\sum_{u\in\nbr_r(v)}\tfrac{1}{c_{v,r}}\Wgt_r^{(l)}\hv_u^{(l)}\Big),
\end{equation}
with a per-relation normalizer $c_{v,r}$ \cite{schlichtkrull2018rgcn}. The obvious cost is parameters, one weight matrix per relation, which on knowledge graphs with thousands of relation types forces basis or block decompositions of the $\Wgt_r$. Heterogeneous attention takes a different route, learning attention at two levels, within a relation and then across relations, so that the model can decide which relation types matter for a given node \cite{wang2019han}. A third approach borrows composition operators from knowledge-graph embedding to combine a neighbour with the relation connecting it, which keeps the parameter count flat in the number of relations \cite{vashishth2020compgcn}. These designs recur throughout the application chapters, since knowledge graphs, recommendation graphs, and industrial systems are heterogeneous by nature \cite{wang2022hgsurvey}. The methods for heterogeneous graphs divide along a clear line. One approach maintains separate transformations for each type of node and edge, so that a relation has its own way of passing messages, which is faithful but grows expensive as the number of relations rises, prompting parameter-sharing schemes that decompose each relation's transformation into a combination of shared bases. Another approach reasons along metapaths, sequences of relation types that define meaningful composite connections, such as the author-paper-author path that links collaborators in a citation network, aggregating along these paths rather than over raw edges. The trade-off mirrors one seen throughout the survey, between faithfully modelling every distinction the data presents and controlling the parameters and computation that fidelity costs, and the right balance depends on how many relation types a graph has and how much each matters to the task.

\subsection{Pooling and graph-level readout}
\label{subsec:pooling}

Graph-level tasks need the readout of \cref{eq:readout} to compress a set of node states into one vector, and the choice ranges from a simple sum or mean to learned hierarchical coarsening. Differentiable pooling learns a soft assignment of nodes to clusters at each level and coarsens both the features and the adjacency accordingly,
\begin{equation}\label{eq:diffpool}
\mathbf{S}^{(l)}=\softmax\big(\mathrm{GNN}_{\mathrm{pool}}^{(l)}(\Adj^{(l)},\Hid^{(l)})\big),\quad
\Adj^{(l+1)}=\mathbf{S}^{(l)\top}\Adj^{(l)}\mathbf{S}^{(l)},
\end{equation} The layer definitions collected above, from the spectral convolution and its Chebyshev recurrence through the convolutional, sampling, attention, edge-conditioned, equivariant, isomorphism, positional-encoding, and pooling updates, \cref{eq:specconv,eq:chebrecur,eq:gcnlayer,eq:sage,eq:gat-score,eq:gat-alpha,eq:gat-update,eq:ecc,eq:egnn-m,eq:egnn-x,eq:wl,eq:gin,eq:lap-pe,eq:diffpool}, recur throughout the application chapters.
so that a coarse graph is fed to the next block \cite{ying2018diffpool}. Cheaper alternatives score nodes and keep the top fraction, dropping the rest, which preserves a sparse structure and runs faster though it discards information more bluntly \cite{lee2019sagpool,gao2019graphunets}. A spectral variant pools by approximately minimizing a normalized-cut objective, tying the learned clusters to graph connectivity \cite{bianchi2020mincutpool}. The empirical record is sobering: across many graph-classification benchmarks a plain sum or mean readout is hard to beat, and the gains from elaborate pooling are smaller and less consistent than their sophistication suggests, which is another instance of the pattern that simple symmetric aggregation is a strong and stubborn baseline. The pooling methods themselves span a spectrum from flat to hierarchical. Flat readout summarizes all node representations in a single step, by summing, averaging, or taking a maximum, or by a small attention mechanism that weights nodes by learned importance, and it is simple, permutation-invariant, and surprisingly hard to beat. Hierarchical pooling instead coarsens the graph in stages, repeatedly merging nodes into clusters and pooling within them, which mirrors the way image networks downsample and can capture structure at multiple scales. The hierarchical methods are more expressive in principle and sometimes in practice, but they introduce choices, how to cluster and how many stages to use, that add complexity and can be unstable, which is why a well-tuned flat readout remains the default that more elaborate pooling must justify itself against.

\subsection{Learning without labels}
\label{subsec:ssl}

Labels are scarce on large graphs, and self-supervised pretraining fills the gap by manufacturing a learning signal from structure, following the template of \cref{eq:ssl-generic}. The dominant family is contrastive. Deep graph infomax trains a node encoder to distinguish true node-graph pairs from corrupted ones by maximizing a mutual-information estimate \cite{velickovic2019dgi}, and the more general contrastive recipe builds two augmented views of a graph, encodes both, and pulls the representations of the same node together while pushing different nodes apart through an InfoNCE objective,
\begin{equation}\label{eq:infonce}
\mathcal{L}_{\mathrm{NCE}}=-\sum_{i}\log\frac{\exp\!\big(\mathrm{sim}(\mathbf{z}_{1,i},\mathbf{z}_{2,i})/\tau\big)}{\sum_{j}\exp\!\big(\mathrm{sim}(\mathbf{z}_{1,i},\mathbf{z}_{2,j})/\tau\big)},
\end{equation}
with augmentations such as edge dropping, feature masking, and subgraph sampling \cite{you2020graphcl,zhu2020grace,hassani2020mvgrl}. \Cref{alg:gcl} gives the procedure. Two observations temper the enthusiasm. The choice of augmentation matters more than the choice of loss, and augmentations that work on one graph can destroy the signal on another, which is why some methods drop negatives and augmentation entirely in favour of a bootstrapped target \cite{thakoor2022bgrl}. The generative alternative, masking node features or edges and reconstructing them, has become competitive and is conceptually simpler \cite{hou2022graphmae}, and graph-level pretraining by predicting masked attributes or context is the route most relevant to the foundation-model ambitions discussed later \cite{hu2020pretrain,hu2020gptgnn,qiu2020gcc,sun2020infograph}. Whether any of this pretraining transfers across genuinely different graphs, as opposed to improving sample efficiency on a fixed one, is the open question that the chapter on foundation models returns to \cite{wu2021sslsurvey,xie2022sslreview}.

\begin{algorithm}[t]
\caption{Graph contrastive pretraining}
\label{alg:gcl}
\KwIn{Graph $\graph$, encoder $f_\theta$, augmentation distributions $t_1,t_2$, temperature $\tau$, rate $\eta$}
\KwOut{Pretrained parameters $\theta$}
\While{not converged}{
  $\tilde{\graph}_1 \gets t_1(\graph)$,\quad $\tilde{\graph}_2 \gets t_2(\graph)$ \tcp*{two augmented views}
  $\mathbf{Z}_1 \gets f_\theta(\tilde{\graph}_1)$,\quad $\mathbf{Z}_2 \gets f_\theta(\tilde{\graph}_2)$\;
  $\mathcal{L} \gets$ InfoNCE over positive pairs $(\mathbf{z}_{1,i},\mathbf{z}_{2,i})$ \tcp*{\cref{eq:infonce}}
  update $\theta \gets \theta - \eta\,\nabla_\theta \mathcal{L}$\;
}
\Return $\theta$\;
\end{algorithm}

\subsection{Reading the families against one another}
\label{subsec:archcompare}

\begin{figure}[t]
\centering
\resizebox{0.9\textwidth}{!}{%
\begin{tikzpicture}[card/.style 2 args={rounded corners=3pt, draw=#1!80!black, fill=#2, text width=30mm, minimum height=28mm, align=center, font=\small, text=gnnink, inner sep=4pt}]
  \node[card={gnnblue}{tintblue}] (a) {\textbf{GCN}\\[5pt]$\sigma\!\big(\hat{\Adj}\,\Hid\,\Wgt\big)$\\[7pt]\scriptsize fixed degree-normalized mean};
  \node[card={gnnteal}{tintteal}, right=8mm of a] (b) {\textbf{GraphSAGE}\\[5pt]$\sigma\!\big(\Wgt[\hv_v\concat\bar{\hv}_{\nbr}]\big)$\\[7pt]\scriptsize sample then aggregate; inductive};
  \node[card={gnngreen}{tintgreen}, right=8mm of b] (c) {\textbf{GAT}\\[5pt]$\sigma\!\big(\sum_u\alpha_{uv}\Wgt\hv_u\big)$\\[7pt]\scriptsize learned attention weights};
  \node[card={gnnamber}{tintamber}, right=8mm of c] (d) {\textbf{GIN}\\[5pt]$\mathrm{MLP}\big((1{+}\epsilon)\hv_v{+}\!\sum_u\hv_u\big)$\\[7pt]\scriptsize injective sum; 1-WL power};
  \node[card={gnnpurple}{tintpurple}, right=8mm of d] (e) {\textbf{Graph transformer}\\[5pt]$\softmax\!\big(\tfrac{\mathbf{Q}\mathbf{K}\trans}{\sqrt{d}}\big)\mathbf{V}$\\[7pt]\scriptsize global attention with positional encoding};
\end{tikzpicture}}
\caption[Defining update of five architecture families]{The defining update of five architecture families.}
\label{fig:archcompare}
\end{figure}
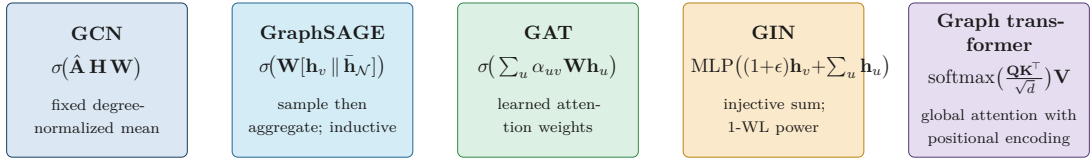

The mechanisms above are easier to choose between when set side by side. \Cref{fig:archcompare} contrasts the defining update of five families, \Cref{tab:archfamilies} records their qualitative properties, and \Cref{tab:complexity} states their asymptotic costs. Across every symbolic table in this survey, \hbfull{} marks a strong capability, \hbhalf{} a moderate or partial one, and \hbempty{} a weak or absent one, and these marks record the authors' qualitative judgement rather than a measured quantity. The comparison rewards a few summary judgments. Degree-normalized averaging is cheap and surprisingly strong but inflexible; attention adds useful selectivity at a modest cost and a weak interpretability bonus; sum aggregation is the right default when structural discrimination matters; sampling is the lever that makes any of them scale; and global attention is worth its quadratic price only when the graph is small enough to pay it. None of the families dominates the others across all of scalability, expressiveness, heterophily tolerance, and interpretability, which is why the design space, rather than any single model, is the useful object.

A few practical defaults follow from the comparison and survive contact with real datasets. On a large, sparsely labelled, broadly homophilous graph, a sampled spatial model such as GraphSAGE or a carefully tuned GCN is the right first attempt, cheap to train and hard to beat. When the task turns on fine structural distinctions, as in the graph classification of molecules, sum aggregation in the style of GIN or an equivariant model is the better starting point. When neighbours of different classes mix, the homophily assumption fails and a model with learned or signed attention, or an explicit high-pass component, is needed. When the graph is small but long-range structure matters, a graph transformer earns its quadratic cost. The recurring mistake is to reach for the most elaborate architecture first; the evidence assembled here favours starting simple, tuning honestly, and adding mechanism only when a measured failure demands it.

\begin{table}[t]
\centering
\caption[Qualitative comparison of architecture families]{Qualitative comparison of architecture families. Symbols follow the strong/moderate/weak scale defined in the text and record the authors' assessments, not measured quantities.}
\label{tab:archfamilies}
\renewcommand{\arraystretch}{1.2}
\small
\begin{tabularx}{\textwidth}{@{}l Y Y Y Y Y@{}}
\toprule
\textbf{Family} & \textbf{Inductive} & \makecell{\textbf{Expressive}\\\textbf{power}} & \textbf{Scalability} & \textbf{Heterophily} & \makecell{\textbf{Interpret-}\\\textbf{ability}} \\
\midrule
GCN               & \hbhalf  & \hbhalf  & \hbhalf  & \hbempty & \hbhalf  \\
GraphSAGE         & \hbfull  & \hbhalf  & \hbfull  & \hbempty & \hbhalf  \\
GAT               & \hbfull  & \hbhalf  & \hbhalf  & \hbhalf  & \hbfull  \\
GIN               & \hbfull  & \hbfull  & \hbempty & \hbempty & \hbhalf  \\
Graph transformer & \hbfull  & \hbfull  & \hbempty & \hbfull  & \hbhalf  \\
\bottomrule
\end{tabularx}
\end{table}

\begin{table}[t]
\centering
\caption[Asymptotic per-layer cost of architecture families]{Asymptotic cost of one layer, with $n$ nodes, $m$ edges, hidden width $d$, and $S$ sampled neighbours per node. Complexities follow the analyses in the cited works; constants and lower-order terms are omitted.}
\label{tab:complexity}
\renewcommand{\arraystretch}{1.3}
\begin{tabularx}{\textwidth}{@{}l l l X@{}}
\toprule
\textbf{Family} & \textbf{Time} & \textbf{Memory} & \textbf{Comment} \\
\midrule
GCN &
$\mathcal{O}(m d + n d^2)$ &
$\mathcal{O}(n d + m)$ &
Sparse propagation plus dense transform \cite{kipf2017gcn} \\
GraphSAGE &
$\mathcal{O}(S^{K} n d^2)$ &
$\mathcal{O}(S^{K} d)$ per node &
Cost set by sample size, not graph size \cite{hamilton2017graphsage} \\
GAT &
$\mathcal{O}(m d + n d^2)$ &
$\mathcal{O}(n d + m)$ &
Adds per-edge attention scoring \cite{velickovic2018gat} \\
GIN &
$\mathcal{O}(m d + n d^2)$ &
$\mathcal{O}(n d + m)$ &
Sum aggregation with an MLP update \cite{xu2019gin} \\
Graph transformer &
$\mathcal{O}(n^2 d)$ &
$\mathcal{O}(n^2)$ &
Dense attention over all node pairs \cite{ying2021graphormer} \\
\bottomrule
\end{tabularx}
\end{table}

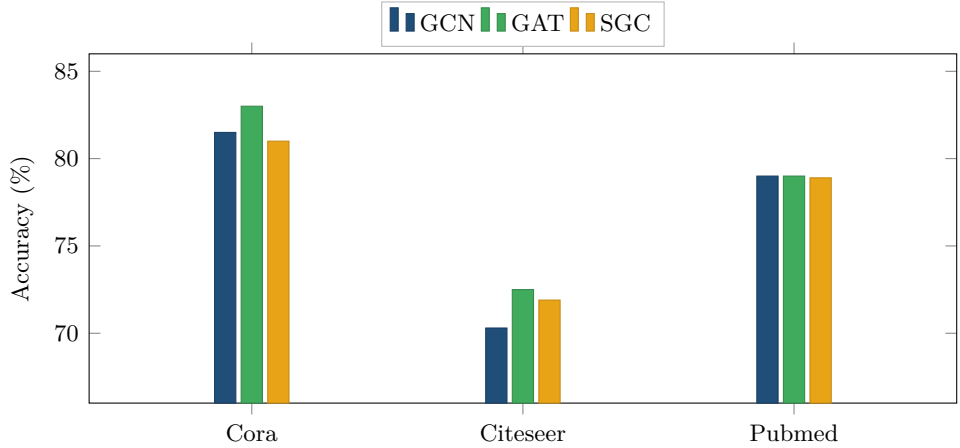
\begin{figure}[t]
\centering
\begin{tikzpicture}
\begin{axis}[
  ybar, bar width=8pt,
  width=0.82\textwidth, height=6.2cm,
  symbolic x coords={Cora,Citeseer,Pubmed}, xtick=data,
  ymin=66, ymax=86, ylabel={Accuracy (\%)},
  enlarge x limits=0.3,
  tick label style={font=\footnotesize}, label style={font=\footnotesize},
  legend style={font=\footnotesize, at={(0.5,1.02)}, anchor=south, legend columns=3, draw=gnngray!50},
  legend cell align=left]
\addplot[draw=gnnblue!80!black, fill=gnnblue] coordinates {(Cora,81.5)(Citeseer,70.3)(Pubmed,79.0)};
\addlegendentry{GCN}
\addplot[draw=gnngreen!80!black, fill=gnngreen] coordinates {(Cora,83.0)(Citeseer,72.5)(Pubmed,79.0)};
\addlegendentry{GAT}
\addplot[draw=gnnamber!85!black, fill=gnnamber] coordinates {(Cora,81.0)(Citeseer,71.9)(Pubmed,78.9)};
\addlegendentry{SGC}
\end{axis}
\end{tikzpicture}
\caption[Reported node-classification accuracy on citation benchmarks]{Reported node-classification accuracy on standard citation benchmarks, with values drawn from \Cref{tab:nodeaccnum} and the sources cited there \cite{kipf2017gcn,velickovic2018gat,wu2019sgc}. The narrow spread across very different models is the point.}
\label{fig:nodeacc}
\end{figure}

Where numbers are available from primary sources, they reinforce rather than overturn these judgments. \Cref{tab:nodeaccnum} collects reported node-classification accuracy for three representative models on the standard citation benchmarks, and \Cref{fig:nodeacc} shows the same figures. The differences are real but small, a few points separate the strongest from the simplest, and the simplest of the three is a linear model. This is the empirical texture the survey asks readers to keep in view: graph structure helps, the architectural refinements help less than their proliferation implies, and the gap narrows further once baselines are tuned with the same care as the methods they are meant to lose to. The honest conclusion of comparing the families is that the differences among them are smaller than the volume of work on each would suggest, and that the choice among a graph convolution, an attention-based model, and a general message-passing network is often less consequential than the choices of how the graph is built, how the model is regularized, and how the baselines are tuned. This is not a counsel of indifference, since the families do differ in the inductive biases they bring and the structure they can exploit, but it is a reminder that architectural novelty is one lever among several and frequently not the most powerful, a theme the application chapters bear out repeatedly.

\begin{table}[t]
\centering
\caption[Reported node-classification accuracy (\%)]{Node-classification accuracy (\%) on standard citation benchmarks, as reported in the cited works. Splits and training protocols differ across papers, so the values are indicative rather than strictly comparable.}
\label{tab:nodeaccnum}
\begin{tabular}{@{}l S[table-format=2.1] S[table-format=2.1] S[table-format=2.1] l@{}}
\toprule
\textbf{Method} & {\textbf{Cora}} & {\textbf{Citeseer}} & {\textbf{Pubmed}} & \textbf{Source} \\
\midrule
GCN & 81.5 & 70.3 & 79.0 & \cite{kipf2017gcn} \\
GAT & 83.0 & 72.5 & 79.0 & \cite{velickovic2018gat} \\
SGC & 81.0 & 71.9 & 78.9 & \cite{wu2019sgc} \\
\bottomrule
\end{tabular}
\end{table}

\FloatBarrier

\section{Taxonomy and design space}
\label{sec:taxonomy}

The previous section read the major architectures as variations on a single mechanism. The number of those variations is now large enough that a reader needs more than a list to hold them in mind, and this section supplies the organizing structure: a taxonomy that sorts architectures by what they do, a set of design axes along which any model can be located, and a comparison of what the main families are good and bad at. The aim is a vocabulary precise enough that each application can be described by the region of this space its methods occupy rather than by name alone.

\subsection{A taxonomy of architectures}
\label{subsec:taxonomy}

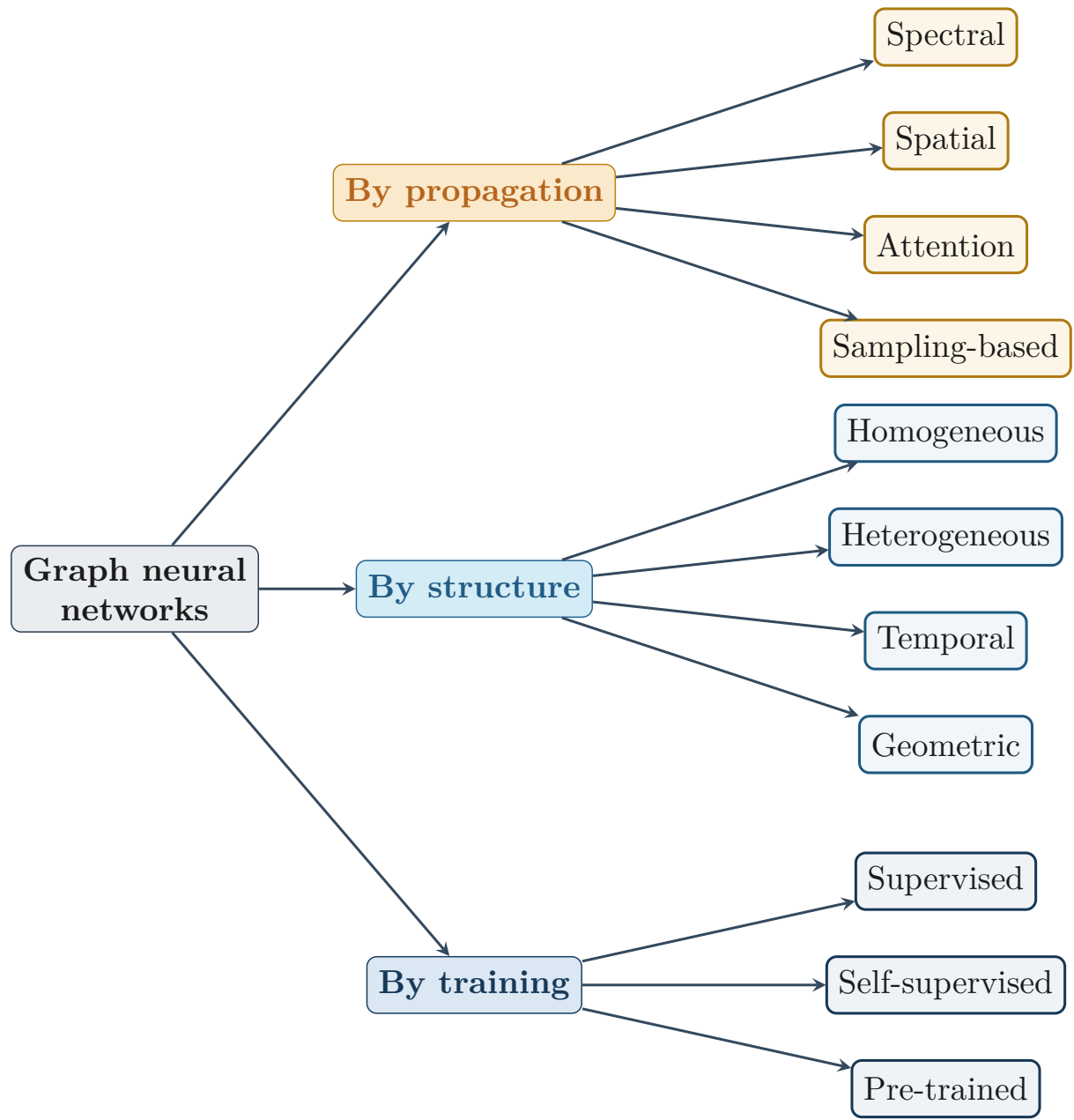
\begin{figure}[t]
\centering
\resizebox{\textwidth}{!}{%
\begin{tikzpicture}[
  grow=right, edge from parent/.style={draw=gnnslate,->,>=stealth, line width=0.8pt},
  level 1/.style={sibling distance=42mm, level distance=36mm},
  level 2/.style={sibling distance=11mm, level distance=50mm},
  every node/.style={rounded corners=3pt, align=center, font=\small, inner sep=3.5pt, minimum height=6mm}]
\node[draw=gnnslate!85!black, fill=tintgray, font=\small\bfseries, text=gnnink] {Graph neural\\networks}
  child { node[draw=gnnblue!80!black, fill=tintblue, text=gnnblue!72!black, font=\small\bfseries] {By training}
    child { node[draw=gnnblue!70!black, fill=gnnblue!7, text=gnnink] {Pre-trained} }
    child { node[draw=gnnblue!70!black, fill=gnnblue!7, text=gnnink] {Self-supervised} }
    child { node[draw=gnnblue!70!black, fill=gnnblue!7, text=gnnink] {Supervised} } }
  child { node[draw=gnnteal!80!black, fill=tintteal, text=gnnteal!72!black, font=\small\bfseries] {By structure}
    child { node[draw=gnnteal!70!black, fill=gnnteal!7, text=gnnink] {Geometric} }
    child { node[draw=gnnteal!70!black, fill=gnnteal!7, text=gnnink] {Temporal} }
    child { node[draw=gnnteal!70!black, fill=gnnteal!7, text=gnnink] {Heterogeneous} }
    child { node[draw=gnnteal!70!black, fill=gnnteal!7, text=gnnink] {Homogeneous} } }
  child { node[draw=gnnamber!80!black, fill=tintamber, text=gnnrust, font=\small\bfseries] {By propagation}
    child { node[draw=gnnamber!75!black, fill=gnnamber!10, text=gnnink] {Sampling-based} }
    child { node[draw=gnnamber!75!black, fill=gnnamber!10, text=gnnink] {Attention} }
    child { node[draw=gnnamber!75!black, fill=gnnamber!10, text=gnnink] {Spatial} }
    child { node[draw=gnnamber!75!black, fill=gnnamber!10, text=gnnink] {Spectral} } };
\end{tikzpicture}}
\caption[Taxonomy of graph neural network architectures.]{Taxonomy of graph neural network architectures along three independent axes: how a layer propagates information, what kind of graph it is built for, and how it is trained. A concrete model makes a choice on each axis, so the branches are not mutually exclusive.}
\label{fig:taxonomy}
\end{figure}

Three questions separate one graph network from another, and they form the top level of the taxonomy in \cref{fig:taxonomy}. The first asks how a layer propagates information: whether it filters in the spectral domain, passes messages in the spatial domain, weights neighbours by attention, or restricts computation to a sampled subset of the graph. The second asks what kind of graph the model is built for: a single homogeneous graph, a heterogeneous or relational one with typed nodes and edges, a temporal graph that changes over time, or a geometric structure in which node coordinates carry meaning. The third asks how the model is trained: under direct supervision, under a self-supervised objective that manufactures its own targets, or as a pre-trained model intended for transfer. These axes are not mutually exclusive, and most concrete systems make a choice on each. A recommendation model might pass messages on a bipartite heterogeneous graph trained with a contrastive loss, which places it at once in three branches of \cref{fig:taxonomy}. The taxonomy is therefore a description of independent decisions rather than a partition into disjoint classes, and its value is that it makes those decisions explicit instead of burying them in a model name.

\subsection{The design space as a set of axes}
\label{subsec:designspace}

Underneath the taxonomy sits a finer design space whose coordinates are the choices a layer actually instantiates. Every message-passing layer can be written as a composition of three operations, a function that forms a message from a neighbour, a permutation-invariant aggregator that pools those messages, and an update that combines the pooled message with the node's current state,
\begin{equation}\label{eq:designtemplate}
\hv_v^{(l+1)}=\UPD\!\Big(\hv_v^{(l)},\ \AGG_{u\in\nbr(v)}\,\msg\big(\hv_v^{(l)},\hv_u^{(l)},\ev_{uv}\big)\Big),
\end{equation}
and the named architectures of the previous section are recovered by filling in these three slots in particular ways. \Cref{tab:designspace} lists the axes this template exposes together with the options each commonly takes. Reading the table as a product space is instructive, because even the modest counts shown multiply into thousands of combinations, of which the literature has named and tested only a small fraction. Most papers vary a single axis while holding the others fixed, so the apparent diversity of architectures overstates the diversity of ideas, and large regions of the design space remain unexamined not because they were tried and failed but because no one has reported them. The observation tempers any claim that the field has converged on an optimal design. What it has converged on is a small, well-trodden path through a space it has mostly not mapped, and several of the cross-domain comparisons later in this survey are best read as reports from the few points along that path where careful measurement exists.

\begin{table}[t]
\centering
\caption[The graph neural network design space.]{The design space exposed by the message-passing template of \cref{eq:designtemplate}. Each axis is chosen largely independently, so the entries multiply into a far larger space than the named architectures occupy.}
\label{tab:designspace}
\small
\begin{tabularx}{\textwidth}{@{}lXX@{}}
\toprule
Axis & Common options & Principal effect \\
\midrule
Message function & identity, linear map, edge-conditioned map, attention weight & sets what a neighbour contributes and whether edge features enter \\
Aggregator & sum, mean, max, attention-weighted sum & fixes expressive power and sensitivity to neighbourhood size \\
Update & linear with nonlinearity, gated recurrence, residual sum & controls how new and old state combine and whether deep training stays stable \\
Propagation depth & shallow stack, deep with residuals, decoupled propagation & trades receptive field against over-smoothing \\
Normalization & symmetric degree scaling, pairwise-distance rescaling, edge dropping & stabilizes training and slows feature collapse \\
Readout & sum, mean, max, hierarchical pooling, virtual node & determines how node states become a graph representation \\
Training objective & supervised loss, contrastive loss, generative reconstruction & sets what the representations are optimized to capture \\
\bottomrule
\end{tabularx}
\end{table}

\subsection{What the families are good at}
\label{subsec:capability}

A design choice buys some capabilities at the expense of others, and \cref{tab:capability} records the trade-offs that the main families make. The entries use the symbolic scale defined in the previous section, and they encode the authors' reading of the evidence rather than any single measured quantity. The pattern across rows carries more meaning than any individual cell. Expressive power and scalability pull in opposite directions, since the isomorphism-targeted design that maximizes what a model can distinguish relies on an aggregator that resists the sampling tricks on which large-graph training depends, while the sampling-based design that scales to billions of edges gives up some of that distinguishing power. Attention buys interpretability of a limited kind, because the learned weights can be inspected, though the inspection is not always faithful to what actually drives a prediction. Handling heterophily, the setting in which neighbours tend to differ rather than agree, remains a weakness shared by most families that were built on an assumption of smoothness, and it is the capability on which the standard benchmarks are least representative of the wider population of graphs. No row dominates the others, which is the central fact a capability taxonomy is meant to convey: choosing an architecture is choosing which weaknesses to accept.

\begin{table}[t]
\centering
\caption[Capability comparison across architecture families.]{Capabilities of the main architecture families, rated on the symbolic scale defined in the text. Entries record the authors' synthesis of the evidence rather than a single measured quantity, and no family dominates the rest.}
\label{tab:capability}
\small
\setlength{\tabcolsep}{5pt}
\renewcommand{\arraystretch}{1.25}
\begin{tabular}{@{}lcccccc@{}}
\toprule
Family & \makecell{Expressive\\power} & \makecell{Scal-\\ability} & \makecell{Hetero-\\phily} & \makecell{Induct-\\ive} & \makecell{Interpret-\\ability} & \makecell{Dynamic\\support} \\
\midrule
Spectral / GCN        & \hbhalf  & \hbhalf  & \hbempty & \hbempty & \hbhalf  & \hbempty \\
Spatial / GraphSAGE   & \hbhalf  & \hbfull  & \hbempty & \hbfull  & \hbhalf  & \hbempty \\
Attention / GAT       & \hbhalf  & \hbhalf  & \hbhalf  & \hbfull  & \hbfull  & \hbempty \\
Isomorphism / GIN     & \hbfull  & \hbempty & \hbempty & \hbfull  & \hbempty & \hbempty \\
Graph transformer     & \hbfull  & \hbempty & \hbfull  & \hbfull  & \hbhalf  & \hbempty \\
Relational / R-GCN    & \hbhalf  & \hbhalf  & \hbhalf  & \hbhalf  & \hbhalf  & \hbempty \\
Temporal / dynamic    & \hbhalf  & \hbhalf  & \hbempty & \hbhalf  & \hbempty & \hbfull  \\
\bottomrule
\end{tabular}
\end{table}

\subsection{Locating methods in the space}
\label{subsec:taxonomymap}

The taxonomy and the design space meet in \cref{tab:taxonomymap}, which places a set of representative methods against the three top-level axes. The table is deliberately small, since its purpose is to calibrate the vocabulary rather than to catalogue the literature, and each row shows how a familiar method decomposes into a propagation choice, a structural target, and a training paradigm. Read across, the rows reveal that methods cluster: the most cited node-classification models share a spatial, homogeneous, supervised profile, which is exactly the profile the standard benchmarks reward, and the methods that depart from it usually do so because an application forced a different structure or a different source of supervision. That clustering reflects the field's incentives as much as the structure of the problem, and it is worth holding in view whenever a method designed for one profile is reported to transfer poorly to another.

\begin{table}[t]
\centering
\caption[Representative methods located in the taxonomy.]{A small set of representative methods decomposed along the three taxonomy axes of \cref{fig:taxonomy}. The table calibrates the vocabulary rather than cataloguing the literature.}
\label{tab:taxonomymap}
\small
\begin{tabularx}{\textwidth}{@{}lXXX@{}}
\toprule
Method & Propagation & Structure & Training \\
\midrule
GCN \cite{kipf2017gcn}                 & spectral, first order & homogeneous   & supervised \\
GraphSAGE \cite{hamilton2017graphsage} & spatial, sampled      & homogeneous   & supervised \\
GAT \cite{velickovic2018gat}           & attention             & homogeneous   & supervised \\
GIN \cite{xu2019gin}                    & spatial, sum          & homogeneous   & supervised \\
Graphormer \cite{ying2021graphormer}   & global attention      & homogeneous   & supervised \\
R-GCN \cite{schlichtkrull2018rgcn}     & spatial, per-relation & heterogeneous & supervised \\
GraphCL \cite{you2020graphcl}          & spatial               & homogeneous   & self-supervised \\
\bottomrule
\end{tabularx}
\end{table}

\subsection{Capability profiles at a glance}
\label{subsec:radar}

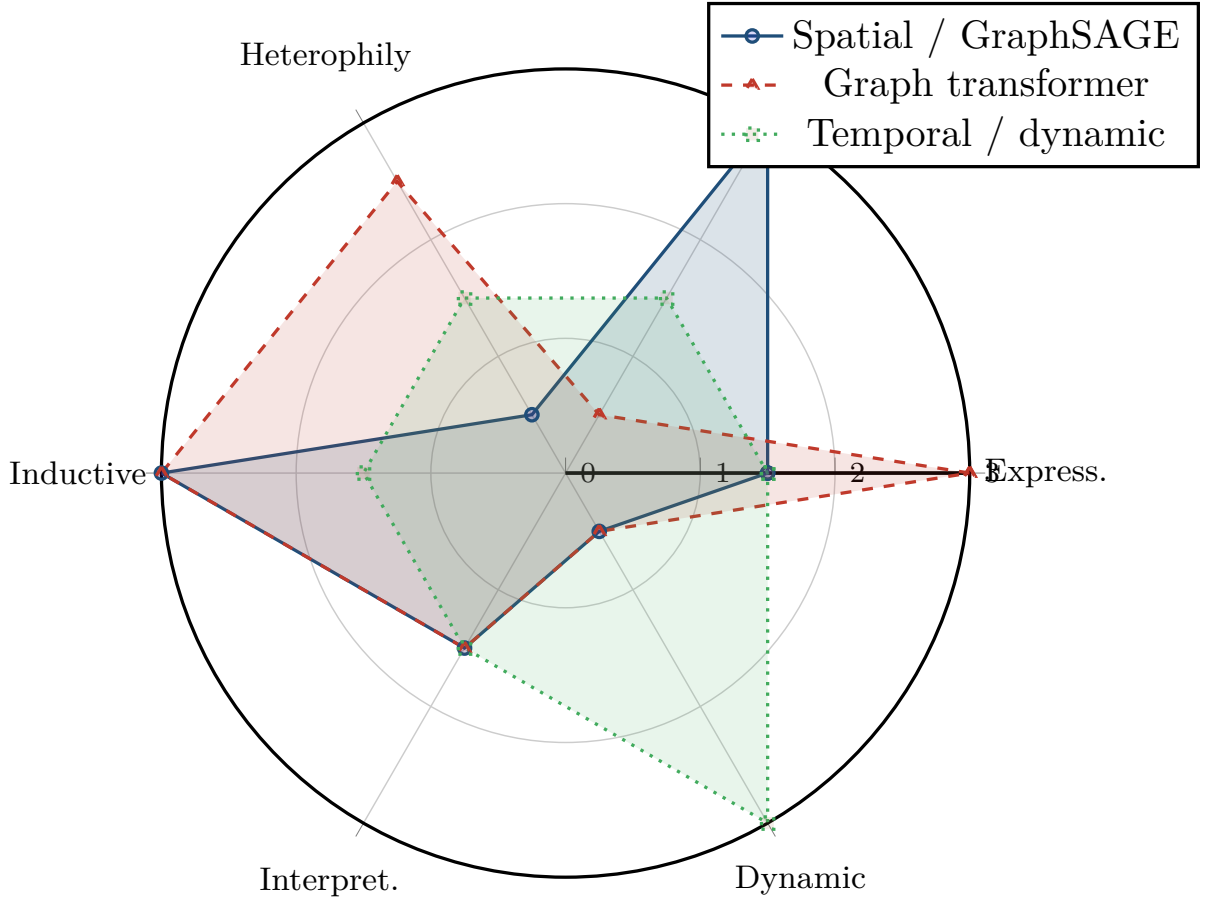
\begin{figure}[t]
\centering
\resizebox{\textwidth}{!}{%
\begin{tikzpicture}
\begin{polaraxis}[
  width=9.4cm,
  xtick={0,60,120,180,240,300},
  xticklabels={Express.,Scale,Heterophily,Inductive,Interpret.,Dynamic},
  xticklabel style={font=\footnotesize},
  ytick={0,1,2,3}, yticklabel style={font=\scriptsize, anchor=west},
  ymin=0, ymax=3, ymajorgrids=true, xmajorgrids=true,
  grid style={gnngray!35}, line width=0.9pt, mark size=1.7pt]
\addplot+[mark=*, color=gnnblue, fill=gnnblue, fill opacity=0.16]
  coordinates {(0,1.5)(60,3)(120,0.5)(180,3)(240,1.5)(300,0.5)(360,1.5)};
\addplot+[mark=triangle*, color=gnnred, dashed, fill=gnnred, fill opacity=0.13]
  coordinates {(0,3)(60,0.5)(120,2.5)(180,3)(240,1.5)(300,0.5)(360,3)};
\addplot+[mark=square*, color=gnngreen, dotted, fill=gnngreen, fill opacity=0.12]
  coordinates {(0,1.5)(60,1.5)(120,1.5)(180,1.5)(240,1.5)(300,3)(360,1.5)};
\legend{Spatial / GraphSAGE, Graph transformer, Temporal / dynamic}
\end{polaraxis}
\end{tikzpicture}}
\caption[Illustrative capability profiles of three architecture families.]{Illustrative capability profiles for three families across six axes, on a zero-to-three scale matching the symbolic ratings of \cref{tab:capability}. Values are a qualitative, literature-informed synthesis rather than measured benchmarks and should be read only for the shape of each profile, not for precise magnitudes.}
\label{fig:radarfamilies}
\end{figure}

\Cref{fig:radarfamilies} renders the same comparison as overlaid capability profiles, with three families traced across the axes of \cref{tab:capability}. The figure is illustrative: the radial positions express a qualitative synthesis of the literature and the authors' judgement rather than a measured benchmark, and the disclaimer in its caption should be read before any quantitative interpretation is attempted. Its purpose is to make the shape of a family's strengths visible at a glance, so that a model with a balanced but unspectacular profile can be told apart from one that is excellent on a single axis and weak elsewhere. The two shapes call for different decisions in an application, and the distinction between them is easy to lose in a grid of symbols.

\subsection{A task that cuts across the taxonomy}
\label{subsec:linkpred}

Some tasks sit orthogonal to the architecture taxonomy because almost any encoder can serve them, and link prediction is the clearest example, recurring in recommendation, knowledge graphs, and biological networks alike. \Cref{alg:linkpred} states the procedure in the general form these applications share. An encoder, drawn from any branch of \cref{fig:taxonomy}, produces a representation for every node; a scoring function then maps a pair of representations to a likelihood that an edge joins them, a common choice being the bilinear form
\begin{equation}\label{eq:bilinear}
s_{uv}=\hv_u^{\top}\,\Wgt_s\,\hv_v,
\end{equation}
with a learned matrix $\Wgt_s$ that need not be symmetric when the graph is directed. Training then proceeds against observed edges as positives and sampled non-edges as negatives, since the absence of an edge is never directly observed and must be approximated. The negative sampling step is where most of the practical difficulty lies, because the sample defines what absence means, and a careless sampler yields a model that scores well against easy negatives and fails against plausible ones. This single template, instantiated with different encoders and different samplers, underlies a large share of the application results this survey reviews.

\begin{algorithm}[t]
\caption{Generic graph link prediction}
\label{alg:linkpred}
\KwIn{graph $\graph=(\vset,\eset,\Feat)$, encoder $f_\theta$, score $s$, negative ratio $k$}
\KwOut{trained parameters $\theta$ and pairwise edge scores}
\For{each training step}{
  $\Hid \leftarrow f_\theta(\Adj,\Feat)$\tcp*{encode all nodes once}
  sample a batch of observed edges $B^{+}\subseteq\eset$\;
  \ForEach{$(u,v)\in B^{+}$}{
    draw $k$ non-edges $(u,v^{-})\notin\eset$\tcp*{negative sampling}
  }
  score positive and negative pairs by \cref{eq:bilinear}\;
  $\mathcal{L}\leftarrow -\!\!\sum_{(u,v)\in B^{+}}\!\!\log\sigma(s_{uv})\;-\;\sum_{\text{neg}}\log\!\big(1-\sigma(s_{uv^{-}})\big)$\;
  update $\theta$ by descending $\nabla_\theta\mathcal{L}$\;
}
\Return{$\theta$ and scores $s$}
\end{algorithm}

The taxonomy, the design axes, and the capability comparison together give the rest of this survey a fixed vocabulary. Describing a method by the region of the design space it occupies, and by where the capabilities it offers line up or fail to line up with the capabilities a problem needs, is more informative than naming it, and it is the description used throughout the chapters that follow.

\FloatBarrier
\section{Social networks and recommendation}
\label{sec:social}

Social systems were among the first settings in which graph neural networks moved from research benchmarks into production, and recommendation is the application that has drawn the most sustained industrial effort. Both rest on the same observation: the data is already a graph, and the relationships it records carry as much signal as the entities themselves. Some social ties are signed rather than merely present, recording trust or antagonism, and message passing has been extended to such signed graphs by propagating differently along positive and negative edges \cite{derr2018signedgcn}. A social network is a graph in the most literal sense, with people as nodes and their ties as edges, while a recommendation problem becomes a graph once users and the items they interact with are treated as two kinds of node joined by observed interactions. This section follows the template used for every domain that follows, describing how the graph is built, what is predicted on it, which architectures dominate, and where the approach succeeds and fails, before closing with a comparative reading of the families involved.

\subsection{Graph construction}
\label{subsec:social-graph}

Two graph shapes recur. The first is the social graph proper, an ordinarily homogeneous graph whose edges encode friendship, following, or communication, sometimes directed when the relation is asymmetric, as a follow is, and sometimes weighted by the frequency or recency of interaction. The second is the user-item bipartite graph at the centre of recommendation, in which edges record that a user clicked, rated, purchased, or watched an item, and in which no edge ever joins two users or two items directly. Real systems rarely use either shape in isolation. Side information turns the bipartite graph heterogeneous, adding nodes for item attributes, categories, or knowledge-graph entities, so that an item inherits signal from others that share a brand or a genre. Social recommendation fuses the two shapes outright, attaching a user-user social graph to the user side of the interaction graph so that a friend's preferences can inform a recommendation for someone with little history of their own. A further variant, the session graph, is built per user from a short sequence of recent actions and encodes order rather than long-run taste, which suits settings where intent shifts within a single visit. Multi-relational social graphs push further still, distinguishing edge types such as friending, messaging, and blocking, since collapsing them into a single undirected tie discards the sign and direction that often matter most. The choice among these constructions is the first and frequently the most consequential design decision in the domain, because it fixes what counts as a neighbour before any learning begins, and a model that aggregates over the wrong graph cannot be repaired by a better layer.

\subsection{Tasks}
\label{subsec:social-tasks}

The dominant task is recommendation itself, which graph methods almost always cast as link prediction on the user-item graph following the template of \cref{alg:linkpred}: score every candidate user-item pair and rank items for each user by that score. Rating prediction, a regression variant, instead predicts the value a user would assign and reads naturally as edge-weight prediction rather than edge existence. On the social graph, friend recommendation is again link prediction, while node-level tasks include attribute inference, such as estimating a user's interests from their position in the network, and the detection of bots or fraudulent accounts, a problem whose methods overlap with those of a later chapter on fraud. Influence and diffusion prediction asks how information or behaviour spreads along edges and underlies influence-maximization formulations that select a small seed set to reach the widest audience. Community detection seeks groups of densely connected nodes and serves both as an end in itself and as a preprocessing step for other tasks. Graph networks have been applied to community detection in both supervised and generative forms, learning to recover communities from structure and attributes rather than optimizing a fixed modularity objective \cite{chen2019lgnn,shchur2019nocd}, with generative models jointly inferring communities and node representations \cite{jia2019communitygan,sun2019vgraph}. A closely related cluster of work targets the integrity of the information ecosystem, detecting coordinated misinformation and inauthentic behaviour: rumor and fake-news detection cast the propagation tree or the article-source-user graph as the object to classify \cite{bian2020rumor,monti2019fakenews}, automated-account and malicious-entity detection exploit the relational footprint that bots and fraudulent accounts share \cite{feng2021botrgcn,liu2018malicious}, and co-attention over the propagation graph adds a measure of explainability to the verdict \cite{lu2020gcan}. The common thread is that almost every task reduces to predicting nodes or edges on a graph the system already maintains, which is precisely the setting graph networks were designed for, and it explains why the domain adopted them so quickly.

\subsection{Collaborative filtering as propagation}
\label{subsec:social-cf}

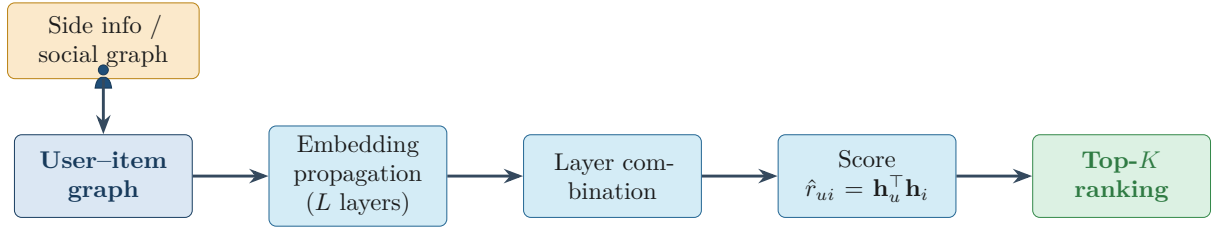
\begin{figure}[t]
\centering
\resizebox{\textwidth}{!}{%
\begin{tikzpicture}[
  proc/.style={rounded corners=3pt, draw=gnnteal!80!black, fill=tintteal, align=center,
               font=\small, text=gnnink, inner sep=4pt, minimum height=12mm, text width=23mm},
  io/.style={rounded corners=3pt, draw=gnnblue!80!black, fill=tintblue, align=center,
             font=\small\bfseries, text=gnnblue!75!black, inner sep=4pt, minimum height=12mm, text width=23mm},
  outbox/.style={rounded corners=3pt, draw=gnngreen!80!black, fill=tintgreen, align=center,
              font=\small\bfseries, text=gnngreen!72!black, inner sep=4pt, minimum height=12mm, text width=23mm},
  side/.style={rounded corners=3pt, draw=gnnamber!80!black, fill=tintamber, align=center,
               font=\small, text=gnnink, inner sep=4pt, minimum height=11mm, text width=25mm},
  ar/.style={-{Stealth[length=2.8mm]}, draw=gnnslate, line width=1.2pt}]
\node[io] (g) {User--item graph};
\node[proc, right=11mm of g] (prop) {Embedding propagation ($L$ layers)};
\node[proc, right=11mm of prop] (comb) {Layer combination};
\node[proc, right=11mm of comb] (score) {Score $\hat{r}_{ui}=\hv_u^{\top}\hv_i$};
\node[outbox, right=11mm of score] (rank) {Top-$K$ ranking};
\node[side, above=8mm of g] (sideinfo) {Side info / social graph};
\pic[scale=0.6] at ([yshift=8.5mm]g.north) {ic person={gnnblue}};
\draw[ar] (sideinfo) -- (g);
\draw[ar] (g) -- (prop);
\draw[ar] (prop) -- (comb);
\draw[ar] (comb) -- (score);
\draw[ar] (score) -- (rank);
\end{tikzpicture}}
\caption[Graph-based recommendation pipeline.]{The pipeline shared by graph collaborative-filtering models. A user--item interaction graph (blue), optionally enriched with side information or a social graph (amber), drives several rounds of embedding propagation (teal); the per-layer representations are combined, used to score user--item pairs, and turned into a ranked list (green).}
\label{fig:recpipeline}
\end{figure}

The most influential idea in the area is that collaborative filtering, long dominated by matrix factorization, can be recast as message passing on the user-item graph. Matrix factorization represents each user and item by a vector and scores a pair by their inner product,
\begin{equation}\label{eq:recscore}
\hat{r}_{ui}=\hv_u^{\top}\hv_i,
\end{equation}
but it treats every interaction independently and never lets a user's representation depend on the items reached two or more hops away, which is exactly the high-order structure a graph makes available. Graph convolutional matrix completion reframed the problem as link prediction on a bipartite graph and let representations propagate across observed interactions, recovering classical factorization as a one-layer special case \cite{berg2017gcmc}. Neural graph collaborative filtering made the high-order signal explicit, stacking propagation layers so that a user representation absorbs information from the items they touched, from the other users who touched those items, and onward through the graph, with an element-wise term that models the affinity between a user and an item as messages pass \cite{wang2019ngcf}. The decisive simplification came from the observation that much of the apparatus inherited from general graph networks is unnecessary here. Stripping the feature transformation and the nonlinearity from each layer and keeping only neighbourhood aggregation,
\begin{equation}\label{eq:lightgcn}
\hv_u^{(l+1)}=\!\!\sum_{i\in\nbr(u)}\frac{1}{\sqrt{|\nbr(u)|}\,\sqrt{|\nbr(i)|}}\,\hv_i^{(l)},
\end{equation}
then combining the representations from every layer into a final embedding,
\begin{equation}\label{eq:lightgcncomb}
\hv_u=\sum_{l=0}^{L}\alpha_l\,\hv_u^{(l)},
\end{equation}
matches or exceeds the accuracy of the heavier models while training faster and with fewer parameters \cite{he2020lightgcn}. The result is the recommendation-domain echo of the caution raised in the architectures chapter by the simplified graph convolution: on these graphs the measurable benefit comes from propagation, not from the depth and nonlinearity that more elaborate designs add, and the layer-combination step matters because it lets the model blend the sharp signal of shallow propagation with the smoothed signal of deeper propagation instead of committing to one depth. A later design pushed the simplification to its limit, approximating the effect of infinitely many propagation layers with a constraint that pulls connected embeddings together and dispensing with explicit message passing during training, trading a small change in accuracy for a further gain in speed \cite{mao2021ultragcn}. \Cref{fig:recpipeline} sets out the shared pipeline these methods instantiate, from graph construction through propagation and layer combination to scoring and ranking. Training typically optimizes a pairwise ranking objective that prefers an observed item over an unobserved one for each user,
\begin{equation}\label{eq:bpr}
\mathcal{L}_{\text{rank}}=-\!\!\sum_{(u,i,j)}\log\sigma\!\big(\hat{r}_{ui}-\hat{r}_{uj}\big),
\end{equation} The scoring, light propagation, layer-combination, and ranking objectives just given, \cref{eq:recscore,eq:lightgcn,eq:lightgcncomb,eq:bpr}, are shared by most graph recommenders.
with $i$ an item the user interacted with and $j$ a sampled item they did not, which places the negative-sampling concern raised in the previous chapter at the centre of recommendation quality: the sampled non-interactions define what the model treats as a negative signal, and a sampler that draws only easy negatives produces a model that ranks well in training and poorly in deployment.

\subsection{Social signal, sequence, and scale}
\label{subsec:social-methods}

Beyond pure collaborative filtering, three directions have shaped the domain. Social recommendation injects the user-user graph into the model, on the premise that a user resembles their friends, and architectures in this vein aggregate over both the interaction graph and the social graph and then reconcile the two views into a single user representation, which helps most precisely where collaborative filtering is weakest, for users whose own interaction history is too thin to support a reliable embedding \cite{fan2019graphrec}. Session-based recommendation builds a small graph from the current sequence of actions, with edges between consecutive items, and applies a graph network with a gated update and an attention readout to capture transitions within the session, so that the model recommends a next item from short-term intent rather than long-run history \cite{wu2019srgnn}. Scale is the third direction, and it is where graph recommendation proved itself in production: a web-scale system at Pinterest combined importance-based neighbourhood sampling, which fixes the number of neighbours a node attends to regardless of its degree, with localized convolutions and a producer-consumer inference scheme, training on a graph of billions of edges and showing that the approach survives contact with industrial data volumes \cite{ying2018pinsage}. Self-supervision arrived more recently, with contrastive objectives that build augmented views of the interaction graph by dropping nodes or edges and ask the model to agree across them, reducing sensitivity to the sparsity and popularity skew that afflict real interaction data \cite{wu2021sgl}. Recent work continues along these lines, with multi-component and structure-aware designs that target the cold-start and sparsity regimes directly \cite{bahi2026mycgnn,bahi2026richgnn,bahi2025benchrec}, feature-refinement schemes that reweight side information before it enters propagation so that uninformative attributes do not dilute the signal \cite{bahi2026sfnn,bahi2025siamesediv,bahi2023homenc}, and attention mechanisms that fuse heterogeneous user signals into a single ranking model \cite{huang2026radar}.

\subsection{Knowledge graphs and influence}
\label{subsec:social-kg}

Two further uses of graph structure deserve mention because they connect this domain to others in the survey. Item-side knowledge graphs supply semantic relationships that the bare interaction graph lacks, linking items through shared attributes, brands, or entities, and methods that propagate over the joint user-item-entity graph let an item inherit signal from related items a user has never seen. One line attaches an attention mechanism to the combined graph so that the most informative relations dominate aggregation \cite{wang2019kgat}, while another aggregates over an item's knowledge-graph neighbourhood to enrich its representation before it reaches the interaction graph at all \cite{wang2019kgcn}. These techniques anticipate the knowledge-graph chapter that follows, where the same machinery serves reasoning rather than ranking. On the social graph proper, influence prediction asks whether a user will adopt a behaviour given the recent actions of their neighbours, and framing the question as node classification on each user's local neighbourhood lets a graph network learn the structural conditions under which influence spreads rather than assuming a fixed diffusion rule \cite{qiu2018deepinf}. Both uses share the domain's defining move, treating an existing graph as the substrate for prediction, and both extend it by enriching the graph with relations that an interaction log alone omits.

\subsection{Datasets}
\label{subsec:social-data}

Evaluation in the area draws on a small set of public interaction logs, summarized by family in \cref{tab:recdatasets}, and their characteristics matter for interpreting any result. Ratings data such as the MovieLens collection is relatively dense and carries explicit graded feedback, which favours rating-prediction metrics and supports studies of how depth behaves on a well-connected graph. Review and check-in data such as the Amazon and Gowalla collections is sparse and implicit, recording only that an interaction occurred, which favours ranking metrics and stresses a model's ability to generalize from few signals per user. Review data also carries text, which makes these collections the natural testbed for methods that fuse language with the interaction graph. The Pinterest data sits at the large end of the scale and is used to test whether a method survives web-scale training rather than to discriminate fine accuracy differences. The table deliberately avoids reporting interaction counts, since these vary by release and preprocessing and are easy to cite incorrectly; what is stable across releases, and what actually governs which methods are appropriate, is the type of feedback and the relative scale.

\begin{table}[t]
\centering
\caption[Representative recommendation and social datasets.]{Families of public datasets used in graph-based recommendation, described by feedback type and relative scale. Interaction counts are omitted deliberately, since they vary with release and preprocessing.}
\label{tab:recdatasets}
\small
\begin{tabularx}{\textwidth}{@{}lllX@{}}
\toprule
Dataset family & Feedback & Scale & Typical use \\
\midrule
MovieLens (ratings) & explicit, graded & small--medium & rating prediction, dense-graph depth studies \\
Amazon (reviews)    & implicit, with text & medium--large & sparse ranking, side-information and text fusion \\
Yelp (reviews)      & implicit, with text & medium & social- and review-aware recommendation \\
Gowalla (check-ins) & implicit & medium & implicit-feedback ranking, location signals \\
Pinterest (pins)    & implicit & large & web-scale training and inference \\
\bottomrule
\end{tabularx}
\end{table}

\subsection{Strengths, weaknesses, and open problems}
\label{subsec:social-tradeoffs}

The central strength of graph methods here is access to high-order connectivity. A user-item-user-item path encodes a recommendation signal that matrix factorization cannot see, and exploiting it improves accuracy on sparse data while softening the cold-start problem, since a new user with few interactions still inherits signal from the items they did touch and from the users those items connect to. Folding heterogeneous side information and social ties into the model through additional nodes and edges is natural in a way it is not for factorization models, and it is one reason the graph formulation displaced the older approach so thoroughly in research. \Cref{tab:socialrec} compares the main families along the capabilities the domain values, and the trade-off pattern from the taxonomy chapter reappears: no family is strong everywhere, and the methods that scale are not the methods that handle sequence or social signal best. The weaknesses are equally clear. Over-smoothing limits useful depth, which is why the strongest collaborative-filtering models are deliberately shallow and why much of the architectural machinery from the methods chapter is removed rather than added; a recommendation model that stacks many layers tends to wash out the very preference distinctions it is meant to rank. Popularity bias is pervasive, because propagation reinforces the visibility of items that are already well connected, and several of the contrastive and reweighting schemes are in part attempts to counter the feedback loop in which popular items are recommended, interacted with, and thereby made more popular still. Scalability, though demonstrated at the billion-edge scale, remains an engineering burden that favours the simplest propagation rules and penalizes designs that need the full graph in memory.

Evaluation is a quieter but corrosive problem. A substantial part of the reported progress in the area has been measured with sampled ranking metrics, which score a held-out item against a small random sample of negatives rather than the full catalogue, and the reliability of those metrics has been questioned because the sampling can reorder methods relative to a full evaluation. The gap between offline ranking accuracy and online engagement compounds the difficulty, since a model that wins on a static split need not win when it shapes the very interactions that will train its successor. These measurement issues do not negate the domain's progress, but they do mean that improvements reported in fractions of a percentage point deserve more scepticism here than in domains with stable, fully ranked test sets.

The open problems follow from these limits. Integrating language models with the interaction graph, so that item text and user reviews inform recommendation directly, is an active direction taken up in the next chapter and one for which the text-rich review datasets are well suited. Fairness-aware and bias-corrected recommendation is moving from an afterthought to a design constraint, driven both by the popularity-bias mechanism above and by external pressure for systems that do not narrow what users are shown. Dynamic formulations that treat the interaction graph as a stream rather than a snapshot remain comparatively underdeveloped despite their obvious fit to systems whose data never stops arriving, and they connect to the temporal architectures that the methods chapter treated only in outline. Two further directions sit alongside these. Multi-behaviour recommendation treats the distinct signals a system records, a click, an add-to-cart, a purchase, a rating, as different edge types rather than collapsing them into a single notion of interaction, on the reasoning that a purchase says more about preference than a click and that the rarer, stronger signals should not be drowned by the abundant weak ones. Diversity and calibration form the other, since a model tuned only for ranking accuracy tends to converge on a narrow, popular slate, and treating the variety and fairness of a recommendation list as objectives in their own right, rather than as constraints bolted on afterwards, is increasingly seen as part of the problem rather than a refinement of it.

A fair summary is that recommendation is the domain where graph networks have delivered the clearest practical value and also the domain that most sharply exposes their limits. The value is real and deployed at scale, on graphs with billions of edges serving live traffic. The limits, over-smoothing that caps depth, popularity bias that propagation amplifies, and an evaluation culture that flatters offline numbers, are equally real, and the most durable contributions have come from methods that respect those limits, by simplifying propagation and correcting bias, rather than from those that add capacity in the hope of overriding them.

\begin{table}[t]
\centering
\caption[Capabilities of graph-based recommendation families.]{Graph-based recommendation families compared on the capabilities the domain values, using the symbolic scale defined in the text. Each row names a representative method.}
\label{tab:socialrec}
\small
\setlength{\tabcolsep}{5pt}
\renewcommand{\arraystretch}{1.25}
\begin{tabular}{@{}lccccc@{}}
\toprule
Family (representative) & \makecell{High-order\\signal} & \makecell{Social /\\side info} & \makecell{Scal-\\ability} & \makecell{Sequen-\\tial} & \makecell{Self-\\supervised} \\
\midrule
GC-MC \cite{berg2017gcmc}       & \hbhalf & \hbhalf  & \hbempty & \hbempty & \hbempty \\
NGCF \cite{wang2019ngcf}        & \hbfull & \hbempty & \hbempty & \hbempty & \hbempty \\
LightGCN \cite{he2020lightgcn}  & \hbfull & \hbempty & \hbhalf  & \hbempty & \hbempty \\
LR-GCCF \cite{chen2020lrgccf}   & \hbfull & \hbempty & \hbhalf  & \hbempty & \hbempty \\
DGCF \cite{wang2020dgcf}        & \hbfull & \hbempty & \hbhalf  & \hbempty & \hbempty \\
PinSage \cite{ying2018pinsage}  & \hbhalf & \hbhalf  & \hbfull  & \hbempty & \hbempty \\
GraphRec \cite{fan2019graphrec} & \hbhalf & \hbfull  & \hbempty & \hbempty & \hbempty \\
DiffNet \cite{wu2019diffnet}    & \hbhalf & \hbfull  & \hbempty & \hbempty & \hbempty \\
MMGCN \cite{wei2019mmgcn}       & \hbhalf & \hbfull  & \hbempty & \hbempty & \hbempty \\
SR-GNN \cite{wu2019srgnn}       & \hbhalf & \hbempty & \hbhalf  & \hbfull  & \hbempty \\
SGL \cite{wu2021sgl}            & \hbfull & \hbempty & \hbhalf  & \hbempty & \hbfull  \\
SimGCL \cite{yu2022simgcl}      & \hbfull & \hbempty & \hbhalf  & \hbempty & \hbfull  \\
LightGCL \cite{cai2023lightgcl} & \hbfull & \hbempty & \hbhalf  & \hbempty & \hbfull  \\
HCCF \cite{xia2022hccf}         & \hbfull & \hbempty & \hbhalf  & \hbempty & \hbfull  \\
\bottomrule
\end{tabular}
\end{table}

\FloatBarrier
\section{Knowledge graphs, language, and large models}
\label{sec:kg}

A knowledge graph is the canonical relational graph, and it is where the relational architectures of the methods chapter were first needed and first proved their worth. Entities are nodes, facts are typed edges, and the data is heterogeneous and multi-relational by construction. This domain has also become the meeting point of two research traditions that developed separately, graph representation learning and language modelling, and the convergence is now among the most active areas in the field. The section treats three threads in turn: reasoning over knowledge graphs with graph networks, graph networks applied to language tasks, and the integration of graphs with large language models that culminates in retrieval-augmented generation over structured knowledge. As in the other domain chapters, the account moves from how the graph is built through the dominant methods to a reckoning with strengths and limits.

\subsection{Knowledge graphs as relational data}
\label{subsec:kg-data}

\begin{figure}[t]
\centering
\resizebox{0.74\textwidth}{!}{%
\begin{tikzpicture}[
  ent/.style 2 args={rounded corners=3pt, draw=#1!80!black, fill=#2, font=\small\bfseries,
            text=#1!72!black, inner sep=4pt, align=center, minimum height=9mm, minimum width=16mm},
  rel/.style={font=\footnotesize, fill=white, text=gnnslate, inner sep=1.5pt, rounded corners=1pt},
  ar/.style={-{Stealth[length=2.6mm]}, draw=gnnslate, line width=1pt}]
\node[ent={gnnblue}{tintblue}]     (mc)    at (0.6,0.2)   {Marie\\Curie};
\node[ent={gnnteal}{tintteal}]     (pc)    at (7.2,0.2)   {Pierre\\Curie};
\node[ent={gnngreen}{tintgreen}]   (pol)   at (-2.4,-3.2) {Poland};
\node[ent={gnnpurple}{tintpurple}] (phys)  at (3.1,-3.5)  {Physics};
\node[ent={gnnamber}{tintamber}]   (nobel) at (7.9,-3.2)  {Nobel\\Prize};
\draw[ar] (mc) -- node[rel,above]{spouse} (pc);
\draw[ar] (mc) -- node[rel,sloped,pos=0.52]{born in} (pol);
\draw[ar] (mc) -- node[rel,sloped,pos=0.40]{field} (phys);
\draw[ar] (mc) -- node[rel,sloped,pos=0.74]{won} (nobel);
\draw[ar] (pc) -- node[rel,sloped,pos=0.30]{field} (phys);
\draw[ar] (pc) -- node[rel,sloped,pos=0.52]{won} (nobel);
\end{tikzpicture}}
\caption[A small knowledge-graph fragment.]{A fragment of a knowledge graph: entities are nodes (coloured by type) and typed relations are directed, labelled edges, so the data is a heterogeneous multi-relational graph. Completion asks a model to predict edges that hold but are absent, such as a missing field or award.}
\label{fig:kgschema}
\end{figure}
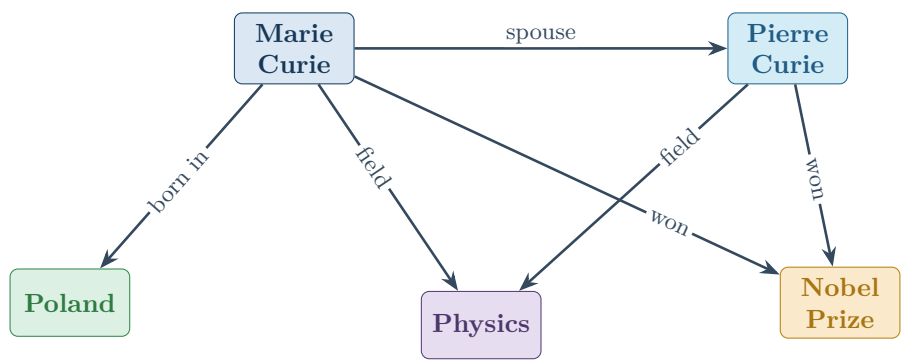

A knowledge graph records facts as triples, each naming a head entity, a relation, and a tail entity, so that a single edge such as the one joining a scientist to an award carries both endpoints and the type of the connection. \Cref{fig:kgschema} shows a small fragment. Collected across many relations, the graph is described by the family of relation-specific adjacency matrices introduced earlier in \cref{eq:reladj}, one matrix per relation type, and this representation is what distinguishes a knowledge graph from the homogeneous graphs of the social domain. Two properties shape every method built on these graphs. The first is heterogeneity: a model must treat relations as first-class objects rather than collapsing them, because the meaning of a path depends on the sequence of relation types along it, not merely on connectivity. The second is incompleteness. Real knowledge graphs are built from sources that never cover every fact, so the central task is completion, the prediction of edges that hold in the world but are missing from the graph, framed as link prediction on a multi-relational graph. A scoring function assigns each candidate triple a plausibility,
\begin{equation}\label{eq:triplescore}
s_{(h,r,t)}=f\big(\hv_h,\hv_r,\hv_t\big)\in\real,
\end{equation}
and the families of methods differ in how they build the entity and relation representations that this function reads, and in the form of $f$ itself.

\subsection{Construction and tasks}
\label{subsec:kg-tasks}

Knowledge graphs reach a model from two kinds of source. Curated graphs are assembled by hand or from structured databases, and the large general-purpose graphs that anchor much of the research, built from collaborative encyclopaedias and structured community projects, fall in this category, as do the domain-specific graphs that organize biomedical, geographic, or commercial knowledge. Extracted graphs are built automatically from text by identifying entities and the relations between them, which scales to corpora no curator could cover but inherits every error of the extraction. Most production graphs blend the two, curating a high-value core and extending it automatically. On whichever graph results, several prediction tasks recur and share machinery even when their framing differs. Completion, the prediction of missing triples, is the most studied and the one against which methods are usually compared, and it splits into tail, head, and relation prediction according to which element of a triple is withheld. Triple classification asks the simpler yes-or-no question of whether a given triple holds. Entity classification assigns a type or label to a node from its position and relations. Entity alignment, deciding that two nodes in different graphs denote the same real-world entity, is what lets separately built graphs be merged. Question answering translates a natural-language question into a walk or query over the graph, and its multi-hop variant, in which the answer lies several relations from the entities named in the question, is the setting where structural reasoning matters most and where the link to language models is tightest. These tasks recur in altered vocabulary across the application chapters, since completion is link prediction, entity classification is node classification, and alignment is graph matching, so the methods developed here transfer more widely than their framing suggests.

\subsection{Reasoning and completion with graph networks}
\label{subsec:kg-reasoning}

For most of the field's history, knowledge-graph completion was dominated by embedding methods that map entities and relations to vectors and score a triple by a fixed algebraic form, translational when a relation is modelled as a vector offset, bilinear when it is modelled as a matrix interaction, and neural when a small network scores the combination. The translational view, in which a relation translates a head embedding toward its tail, is simple and easy to interpret but struggles with relations that are one-to-many or symmetric, since a single offset cannot point to many tails at once. The bilinear view, in which a relation is a matrix or a vector of multiplicative interactions, captures a richer set of relational patterns at the cost of more parameters and a stronger tendency to overfit. The neural view scores a triple with a small convolutional or feedforward network and is the most expressive of the three, though its scores are the hardest to interpret and it inherits the data hunger of any learned scorer. These methods are strong baselines and remain widely used, but they share a structural weakness: each triple is scored in isolation, and the representation of an entity does not depend on the wider neighbourhood it sits in. Graph networks address exactly this gap by letting an entity's representation be assembled from its relational neighbourhood. The relational graph convolutional network introduced relation-specific message passing, applying a separate transformation for each relation type and aggregating across them, with a basis decomposition that shares parameters across relations to keep the model trainable when the relation set is large \cite{schlichtkrull2018rgcn}. Its propagation rule, given earlier as \cref{eq:rgcn}, is the template the later methods refine. Composition-based graph convolution made the relation a participant in the message rather than only a selector of weights, embedding entities and relations jointly and forming each message by composing a neighbour's representation with the relation's through a vector operation,
\begin{equation}\label{eq:compgcn}
\hv_i^{(l+1)}=g\!\Big(\!\!\sum_{(j,r)\in\nbr(i)}\!\!\Wgt^{(l)}_{\lambda(r)}\,\phi\big(\hv_j^{(l)},\hv_r^{(l)}\big)\Big),
\end{equation}
where $\phi$ is a composition such as subtraction, multiplication, or circular correlation and $\lambda(r)$ selects a direction-dependent weight, a formulation that recovers several earlier models as special cases \cite{vashishth2020compgcn}. Attention entered the same way it did elsewhere, with a model that weights relational neighbours by learned importance before aggregating, so that the most informative facts dominate an entity's representation \cite{nathani2019kbgat}, while a structure-aware convolution paired weighted aggregation with a convolutional decoder tuned for the scoring step \cite{shang2019sacn}. Training optimizes the plausibility of observed triples against corrupted ones, commonly through a softmax over candidate tails,
\begin{equation}\label{eq:kgrank}
\mathcal{L}_{\text{KG}}=-\!\!\sum_{(h,r,t)}\log\frac{\exp\!\big(s_{(h,r,t)}\big)}{\sum_{t'}\exp\!\big(s_{(h,r,t')}\big)},
\end{equation} The composition update and the ranking objective, \cref{eq:compgcn,eq:kgrank}, underlie the knowledge-graph methods compared above.
which again makes the construction of negatives, here the corrupted tails $t'$, a determinant of quality.

The limitation that the embedding methods and the early graph methods share is that both learn a fixed vector for every entity seen in training and have nothing to say about an entity that appears only at test time. Inductive reasoning removes this assumption by scoring a candidate triple from the structure of the subgraph around it rather than from entity identities, which lets a model transfer to entities, and even to entirely new graphs, that it never saw during training \cite{teru2020grail}. This line reaches its current conclusion in a model that learns transferable representations of relations themselves and reasons on an arbitrary knowledge graph without retraining, a development that belongs as much to the foundation-model chapter as to this one \cite{galkin2024ultra}.

Two further ideas round out the reasoning picture. Path-based methods reason explicitly over chains of relations rather than over a single aggregated neighbourhood, learning which sequences of relation types support a conclusion, and they connect graph reasoning to the older tradition of logical rules, where a rule states that one relation follows from a path of others. A graph network can be read as learning soft, weighted versions of such rules from data rather than receiving them by hand, which trades the transparency of an explicit rule for the coverage of a learned one. Entity alignment addresses a different need, the fusion of graphs built independently, by embedding two graphs into a shared space so that nodes denoting the same entity land close together, which lets a model carry facts from one graph to another and is a prerequisite for assembling large graphs from many sources. Both ideas reinforce the section's theme, that the value of a graph network on relational data comes from its use of structure beyond the immediate edge, whether that structure is a multi-hop path, a logical pattern, or a correspondence between two separate graphs.

\Cref{tab:kgmethods} sets these methods side by side, and \cref{alg:kgreason} states the shared reasoning procedure, relational propagation followed by triple scoring, that underlies the graph-based approaches.

\begin{table}[t]
\centering
\caption[Graph neural network methods for knowledge graphs.]{Representative graph neural network methods for knowledge-graph reasoning and the related tasks of link prediction and entity alignment, showing how each handles relations and whether it generalizes to entities unseen during training.}
\label{tab:kgmethods}
\small
\begin{tabularx}{\textwidth}{@{}lXcX@{}}
\toprule
Method & Relation handling & Inductive & Task / decoder \\
\midrule
R-GCN \cite{schlichtkrull2018rgcn}  & per-relation weights with basis sharing & no  & factorization decoder \\
CompGCN \cite{vashishth2020compgcn} & entity--relation composition operators   & no  & translational or convolutional \\
KBGAT \cite{nathani2019kbgat}       & attention over relational neighbours     & no  & convolutional decoder \\
SACN \cite{shang2019sacn}           & weighted structure-aware convolution     & no  & convolutional decoder \\
GraIL \cite{teru2020grail}          & subgraph reasoning around a triple        & yes & subgraph scoring \\
SEAL \cite{zhang2018seal}           & enclosing-subgraph features               & yes & subgraph classification \\
NBFNet \cite{zhu2021nbfnet}         & learned path (Bellman--Ford) formulation  & yes & path-based scoring \\
RDGCN \cite{wu2019rdgcn}            & relation-aware dual graph                 & no  & entity alignment \\
AliNet \cite{sun2020alinet}         & gated multi-hop aggregation               & no  & entity alignment \\
ULTRA \cite{galkin2024ultra}        & transferable relation representations      & yes & relation-conditioned scoring \\
\bottomrule
\end{tabularx}
\end{table}

\begin{algorithm}[t]
\caption{Knowledge-graph completion by relational message passing}
\label{alg:kgreason}
\KwIn{relations $\{\Adj_r\}_{r\in\mathcal{R}}$, entity features $\Feat$, score $f$, candidate $(h,r,t)$}
\KwOut{plausibility of the candidate triple}
\For{layer $l=0$ to $L-1$}{
  \ForEach{entity $i$}{
    aggregate relation-specific messages from neighbours by \cref{eq:rgcn}\;
  }
}
read entity embeddings $\hv_h,\hv_t$ and relation embedding $\hv_r$\;
$s \leftarrow f(\hv_h,\hv_r,\hv_t)$ by \cref{eq:triplescore}\tcp*{score the triple}
\Return{plausibility $s$}
\end{algorithm}

\subsection{Graph networks for language}
\label{subsec:kg-nlp}

Language is not obviously a graph, but much of its structure is relational, and a productive line of work makes that structure explicit so a graph network can use it. Several graph constructions recur. A dependency or constituency parse turns a sentence into a tree whose edges carry syntactic roles; a co-occurrence graph links words that appear together across a corpus; and a document-word graph joins documents to the words they contain, turning a collection into a single heterogeneous graph. Graph networks have been applied across these constructions: over dependency trees for relation extraction \cite{zhang2018cgcn} and semantic role labeling \cite{marcheggiani2017srl}, as syntax-aware encoders for machine translation \cite{bastings2017nmt}, over aspect-specific dependency graphs for sentiment classification \cite{zhang2019asgcn,huang2019syntaxgat}, and over document-word and co-occurrence graphs for text classification, including tensor and hypergraph variants \cite{liu2020tensorgcn,ding2020hypergat} and hybrids with pretrained language models such as the combination of a graph network with BERT \cite{lin2021bertgcn}. Gated graph sequence models extended the early recurrent approach to tasks with sequential output \cite{li2016ggnn}, and jointly learning entity and relation representations supports alignment across knowledge graphs \cite{wu2019entityalign}, while label-smoothness regularization sharpens knowledge-aware recommendation \cite{wang2019kgnnls}. The last construction underlies a model that performs text classification by propagating over a graph of documents and words and reading off document labels, treating classification as node classification in the transductive setting \cite{yao2019textgcn}. Syntactic structure proved most valuable where the relationship between distant words matters. Aspect-based sentiment analysis, which asks how an opinion attaches to a particular target in a sentence, benefits from routing information along the dependency tree so that an opinion word reaches the aspect it modifies even when the two are far apart in linear order, and a relational graph attention model over the parse does exactly this \cite{wang2020rgat}. The same template, a graph built from linguistic structure and a graph network propagating over it, has been applied to relation extraction, semantic role labelling, summarization, and machine translation, in each case to inject structure that a purely sequential model would have to rediscover.

The range of these applications is worth spelling out, because it shows how many language problems carry a latent graph. Relation extraction, which identifies how two entities in a text are related, benefits from a graph that connects candidate entities through the syntactic paths between them, so the model attends to the words that actually mediate the relation rather than to every word in between. Document-level understanding builds a graph across sentences, linking mentions of the same entity so that information about a referent accumulates instead of resetting at each sentence boundary. Abstract meaning representation encodes a sentence's semantics directly as a graph, and parsing text into such a graph and generating text out of it are themselves graph problems. Dialogue and discourse add edges between utterances to track how a conversation's topics connect. Across these tasks the graph supplies a structural inductive bias, and the recurring empirical finding is that the bias helps most when training data is limited, because a model with the right structure needs fewer examples to generalize, and helps least when data is plentiful enough for a flexible sequence model to learn the structure on its own.

The trajectory of this line is instructive and a little sobering. Graph methods for language were ascendant precisely when sequence models struggled to capture long-range structure, and they offered a principled way to supply it. The rise of pretrained transformer language models changed the calculus, because a model trained on enough text learns much of the relevant structure implicitly, and the marginal value of an explicit graph fell in many tasks. The lesson is not that structure stopped mattering but that the bar for an explicit structural prior rose: a graph helps language tasks when it encodes information the language model does not already have, such as an external knowledge graph or a document collection's global organization, rather than syntax the model has effectively internalized. This reframing is what connects the NLP thread to the third and most active thread of the section.

\subsection{Large language models and graphs}
\label{subsec:kg-llm}

\begin{figure}[t]
\centering
\resizebox{0.94\textwidth}{!}{%
\begin{tikzpicture}[
  llm/.style={rounded corners=3pt, draw=gnnpurple!80!black, fill=tintpurple, font=\small\bfseries, align=center,
            text=gnnpurple!72!black, inner sep=4pt, minimum height=9.5mm, text width=18mm},
  gnn/.style={rounded corners=3pt, draw=gnnteal!80!black, fill=tintteal, font=\small\bfseries, align=center,
            text=gnnteal!75!black, inner sep=4pt, minimum height=9.5mm, text width=18mm},
  res/.style={rounded corners=3pt, draw=gnngreen!80!black, fill=tintgreen, font=\small\bfseries, align=center,
            text=gnngreen!72!black, inner sep=4pt, minimum height=9.5mm, text width=18mm},
  tr/.style={rounded corners=3pt, draw=gnnamber!80!black, fill=tintamber, font=\small, align=center,
            text=gnnink, inner sep=4pt, minimum height=9.5mm, text width=20mm},
  lab/.style={font=\footnotesize\itshape, text=gnnslate},
  tag/.style={font=\footnotesize\bfseries, text=gnnslate},
  ar/.style={-{Stealth[length=2.4mm]}, draw=gnnslate, line width=1pt}]
\node[llm] (l1) {LLM};
\node[gnn, right=34mm of l1] (g1) {GNN};
\node[res, right=11mm of g1] (p1) {Prediction};
\draw[ar] (l1) -- node[lab,above]{text features, labels} (g1);
\draw[ar] (g1) -- (p1);
\node[tag, left=2mm of l1] {(a)};

\node[gnn, below=11mm of l1] (g2) {GNN};
\node[llm, right=34mm of g2] (l2) {LLM};
\node[res, right=11mm of l2] (p2) {Answer};
\draw[ar] (g2) -- node[lab,above]{graph tokens} (l2);
\draw[ar] (l2) -- (p2);
\node[tag, left=2mm of g2] {(b)};

\node[tr, below=11mm of g2] (s3) {Graph $\to$ text};
\node[llm, right=34mm of s3] (l3) {LLM};
\node[res, right=11mm of l3] (p3) {Answer};
\draw[ar] (s3) -- node[lab,above]{serialized structure} (l3);
\draw[ar] (l3) -- (p3);
\node[tag, left=2mm of s3] {(c)};
\end{tikzpicture}}
\caption[Modes of combining language models and graph networks.]{Three ways language models and graph networks are combined: (a) a language model supplies textual features or labels that a graph network consumes; (b) a graph network encodes structure into tokens a language model reasons over; (c) the graph is serialized to text and read directly by a language model. The modes trade structural fidelity against the language model's general competence.}
\label{fig:llmgnn}
\end{figure}
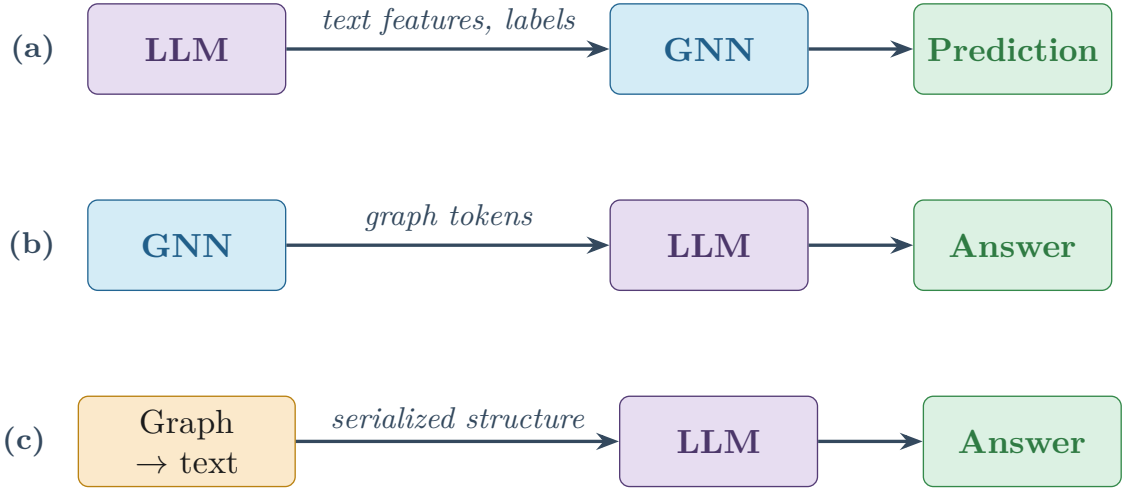

\begin{figure}[t]
\centering
\begin{tikzpicture}
\begin{axis}[width=8.8cm, height=4.6cm, ybar, bar width=10pt,
  symbolic x coords={2021,2022,2023,2024,2025},
  xtick=data, ymin=0, enlarge x limits=0.16,
  ylabel={Relative activity}, ylabel style={font=\footnotesize},
  yticklabel style={font=\scriptsize}, xticklabel style={font=\footnotesize},
  ymajorgrids=true, grid style={gnngray!25},
  every axis plot/.append style={fill=gnnteal, draw=gnnteal!80!black}]
\addplot coordinates {(2021,1)(2022,2)(2023,5)(2024,11)(2025,16)};
\end{axis}
\end{tikzpicture}
\caption[Illustrative growth of work combining graphs and language models.]{Illustrative, literature-informed trend in research that combines graph neural networks with large language models, normalised to the earliest year shown. The values are schematic and convey only the rapid recent growth, not a precise count from any single bibliographic source.}
\label{fig:llmgraphtrend}
\end{figure}
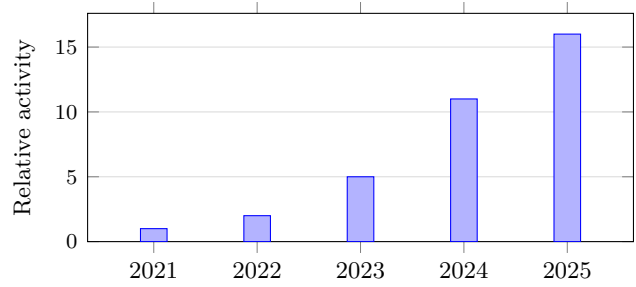

The integration of graphs with large language models runs in two directions, and \cref{fig:llmgnn} sketches the main modes. In one direction the language model serves the graph. It can act as an enhancer, generating textual features or labels for nodes that a graph network then consumes, which is attractive when nodes carry rich text such as paper abstracts or product descriptions. It can also act as a predictor, taking a graph that has been serialized or tokenized and producing an answer directly; methods in this vein design ways to present graph structure to a language model so that its general reasoning can be applied to graph tasks \cite{tang2024graphgpt}, and a complementary approach has a graph network encode structure into tokens that the language model reads, combining structural fidelity with linguistic competence \cite{chen2024llaga}. In the other direction the graph serves the language model, supplying the grounded, structured knowledge that a parametric model lacks. Constraining a language model's multi-hop reasoning to follow paths in a knowledge graph makes the reasoning both more accurate and inspectable, since the supporting path can be read off and checked \cite{luo2024rog}. \Cref{tab:llmgraph} organizes these roles, and the surge of activity along all of them, sketched in \cref{fig:llmgraphtrend}, has been rapid enough that any catalogue dates quickly; the durable content is the set of roles, not the particular systems filling them.

Underneath the taxonomy of roles lies a hard technical problem that none of the modes fully resolves, the mismatch between a graph and the token sequence a language model expects. A graph has no canonical order, yet serializing it into text imposes one, and a model's answer can shift with the order in which nodes and edges are listed, which is the permutation sensitivity that graph networks were designed to avoid reappearing in a new guise. Encoding structure into tokens with a graph network sidesteps the ordering problem but raises an alignment problem instead, since the structural tokens and the language model's word tokens occupy different spaces and must be reconciled, usually by training a projection on paired examples. The empirical picture is genuinely mixed and deserves to be reported as such. Language models show a real ability to perform graph tasks posed in text, including reading off neighbours and tracing short paths, but their accuracy falls as the graph grows and as the required reasoning lengthens, and controlled studies find that they often rely on surface patterns rather than on a faithful internal model of the structure. The reasonable reading is that current language models hold a shallow competence with explicit graphs that is useful for small instances and for supplying priors, but that they do not yet replace a graph network on tasks where structure must be tracked exactly. This is why the most effective systems combine the two rather than choosing between them, using the language model to interpret text and generate fluent output and a graph component to propagate information faithfully over structure.

The ambition behind much of this work is a single model that transfers across graphs and tasks, the graph foundation model that the survey returns to in its own chapter \cite{mao2024gfm}.

\begin{table}[t]
\centering
\caption[Roles in combining language models with graphs.]{A taxonomy of how language models and graph networks are combined, with representative methods named where they fit cleanly.}
\label{tab:llmgraph}
\small
\begin{tabularx}{\textwidth}{@{}lXl@{}}
\toprule
Role & Description & Example \\
\midrule
LLM as enhancer    & language model generates node text features or labels for a graph network & GLEM \cite{zhao2023glem} \\
LLM as predictor   & graph is serialized or tokenized and the language model predicts directly & GraphGPT, InstructGLM \cite{tang2024graphgpt,ye2024instructglm} \\
LLM as reasoner    & graph posed in natural language and the model prompted to reason over it & GPT4Graph, GraphLLM \cite{guo2023gpt4graph,chai2023graphllm} \\
GNN as encoder     & graph network encodes structure into tokens the language model reasons over & LLaGA \cite{chen2024llaga} \\
Graph as grounding & knowledge-graph paths constrain and justify language-model reasoning & RoG, GNN-RAG \cite{luo2024rog,mavromatis2024gnnrag} \\
Structured access  & the model iteratively reads structured data through tools & StructGPT \cite{jiang2023structgpt} \\
Unified model      & one model targets transfer across graphs and tasks & GFM \cite{mao2024gfm} \\
\bottomrule
\end{tabularx}
\end{table}

\subsection{Retrieval-augmented generation over graphs}
\label{subsec:kg-graphrag}

The most consequential application of the graph-language combination addresses a known failure of large language models, their tendency to state unsupported claims when asked about specific or recent facts. Retrieval-augmented generation grounds a model by fetching relevant material and conditioning generation on it, and the question of what to retrieve and how to organize it is where graph structure earns its place. Flat retrieval pulls text passages ranked by similarity to the query, which serves single-fact lookups but handles neither multi-hop questions, whose answer requires chaining several facts, nor corpus-wide questions, whose answer requires synthesizing across an entire collection. Graph retrieval-augmented generation builds a knowledge graph from the corpus, organizes it into communities with precomputed summaries, and retrieves over that structure, which lets it answer the global queries that flat retrieval cannot, by reading community summaries rather than scattered passages \cite{edge2024graphrag}. A complementary design retrieves a query-relevant subgraph and passes it to the language model, scoring candidate subgraphs by their relevance to the query,
\begin{equation}\label{eq:subgraphscore}
\mathrm{rel}(q,S)=\mathbf{q}^{\top}\mathbf{s},
\end{equation}
with $\mathbf{q}$ and $\mathbf{s}$ embeddings of the query and the candidate subgraph, so that the model reasons over a compact structured context rather than a flat list of passages \cite{he2024gretriever}.

The construction phase is where most of the engineering lives. Building the graph from a corpus means extracting entities and relations with a language model, resolving mentions that refer to the same entity so the graph does not fragment into near-duplicates, and often extracting short claims attached to entities so that retrieval can return evidence rather than only structure. The retrieval phase then offers two modes that suit different questions. A local mode gathers the entities and relations near those named in the query and answers narrow, specific questions; a global mode reasons over community summaries that cover the whole corpus and answers broad questions about themes and patterns that no single passage contains. A benefit easy to overlook is provenance, since an answer assembled from identified entities, relations, and source-linked claims can show its work in a way a flat model cannot, which matters wherever an answer must be audited or defended. These advantages are real, and so is the cost, because extracting and maintaining a high-quality graph over a large and changing corpus is expensive, and a graph that drifts out of date grounds the model in stale facts as confidently as a current one grounds it in correct facts.

\begin{figure}[t]
\centering
\resizebox{\textwidth}{!}{%
\begin{tikzpicture}[
  proc/.style={rounded corners=3pt, draw=gnnteal!80!black, fill=tintteal, font=\small, align=center,
            text=gnnink, inner sep=4pt, minimum height=12mm, text width=22mm},
  io/.style={rounded corners=3pt, draw=gnnblue!80!black, fill=tintblue, font=\small\bfseries, align=center,
             text=gnnblue!75!black, inner sep=4pt, minimum height=12mm, text width=15mm},
  qry/.style={rounded corners=3pt, draw=gnnamber!80!black, fill=tintamber, font=\small\bfseries, align=center,
             text=gnnrust, inner sep=4pt, minimum height=11mm, text width=15mm},
  ans/.style={rounded corners=3pt, draw=gnngreen!80!black, fill=tintgreen, font=\small\bfseries, align=center,
             text=gnngreen!72!black, inner sep=4pt, minimum height=12mm, text width=15mm},
  ar/.style={-{Stealth[length=2.8mm]}, draw=gnnslate, line width=1.1pt}]
\node[io] (corpus) {Corpus};
\node[proc, right=9mm of corpus] (kg) {Extract KG (entities, relations)};
\node[proc, right=9mm of kg] (comm) {Communities and summaries};
\node[proc, right=9mm of comm] (ret) {Retrieve subgraph / summaries};
\node[proc, right=9mm of ret] (gen) {LLM generation};
\node[ans, right=9mm of gen] (ansn) {Answer};
\node[qry, above=8mm of ret] (q) {Query};
\pic[scale=0.6] at ([yshift=8mm]corpus.north) {ic doc={gnnblue}};
\draw[ar] (corpus) -- (kg);
\draw[ar] (kg) -- (comm);
\draw[ar] (comm) -- (ret);
\draw[ar] (q) -- (ret);
\draw[ar] (ret) -- (gen);
\draw[ar] (gen) -- (ansn);
\end{tikzpicture}}
\caption[A graph retrieval-augmented generation pipeline.]{A graph retrieval-augmented generation pipeline. A knowledge graph is extracted from a corpus and organized into communities with summaries; a query retrieves a relevant subgraph or set of summaries that grounds the language model's answer. The structure supports multi-hop and corpus-wide questions that flat retrieval handles poorly.}
\label{fig:graphrag}
\end{figure}
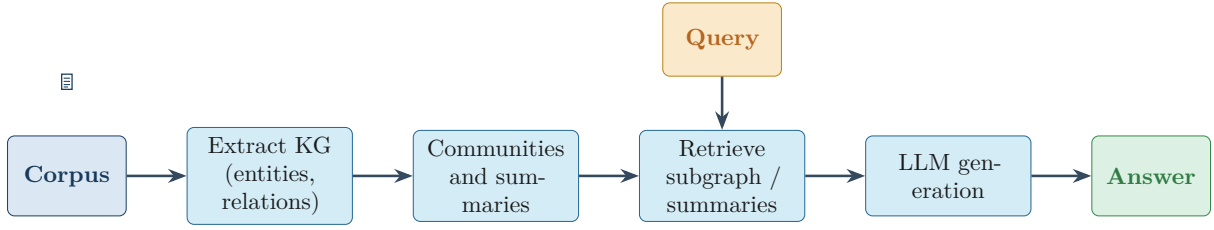

\Cref{fig:graphrag} traces the shared pipeline and \cref{alg:graphrag} states the procedure, with an offline phase that builds and organizes the graph and an online phase that retrieves and generates. \Cref{tab:graphragcompare} compares the strategies on the capabilities that distinguish them.

\begin{algorithm}[t]
\caption{Graph retrieval-augmented generation}
\label{alg:graphrag}
\KwIn{corpus $\mathcal{C}$, query $q$, language model $g$}
\KwOut{grounded answer $a$}
\textbf{Offline:}\;
extract entities and relations from $\mathcal{C}$ to build a knowledge graph $\graph$\;
partition $\graph$ into communities and summarize each\;
\textbf{Online:}\;
retrieve the subgraph or summaries $S$ most relevant to $q$ by \cref{eq:subgraphscore}\;
$a \leftarrow g(q,S)$\tcp*{generate grounded answer}
\Return{answer $a$}
\end{algorithm}

\begin{table}[t]
\centering
\caption[Retrieval strategies for grounding language models.]{Retrieval strategies compared on the capabilities that distinguish them, using the symbolic scale defined in the text. Graph retrieval gains multi-hop and corpus-wide reach at the cost of building and maintaining the graph.}
\label{tab:graphragcompare}
\small
\setlength{\tabcolsep}{5pt}
\renewcommand{\arraystretch}{1.25}
\begin{tabular}{@{}lccccc@{}}
\toprule
Approach & \makecell{Multi-hop\\reach} & \makecell{Global\\queries} & \makecell{Ground-\\ing} & \makecell{Low setup\\cost} & \makecell{Easy\\update} \\
\midrule
Flat text RAG                              & \hbempty & \hbempty & \hbhalf  & \hbfull  & \hbfull  \\
Subgraph retrieval \cite{he2024gretriever} & \hbfull  & \hbhalf  & \hbfull  & \hbhalf  & \hbhalf  \\
Adaptive graph use \cite{dong2026usegraph} & \hbhalf  & \hbhalf  & \hbfull  & \hbhalf  & \hbhalf  \\
Community GraphRAG \cite{edge2024graphrag} & \hbfull  & \hbfull  & \hbfull  & \hbempty & \hbempty \\
Hierarchical causal KG \cite{wang2026hugrag} & \hbfull & \hbfull  & \hbfull  & \hbempty & \hbhalf  \\
\bottomrule
\end{tabular}
\end{table}

\subsection{Strengths, weaknesses, and open problems}
\label{subsec:kg-tradeoffs}

The strength of graph methods across this domain is that they match the shape of the data. Knowledge is relational, and a relational graph network reasons over it without the awkward flattening that sequence and table models require; inductive variants extend that reasoning to entities and graphs never seen in training, which is the property that turns a completion model into something closer to a reusable reasoner. Grounding a language model in a structured graph improves both accuracy and accountability on questions whose answers must be traced, and the organization a graph imposes on a corpus is what makes corpus-wide and multi-hop questions answerable at all. The surveyed knowledge-graph literature, summarized in part by a dedicated review \cite{ji2021kgsurvey}, supports these claims with a decade of completion benchmarks.

Beyond the benchmarks, knowledge graphs earn their place in deployed systems. Web search engines use them to answer factual queries directly and to assemble the information panels that accompany results, virtual assistants consult them to resolve questions about people, places, and times, and recommendation systems draw on them for the side information discussed in the previous chapter. In the sciences the same structure organizes biomedical knowledge, linking genes, proteins, diseases, and drugs into graphs that the next chapter's methods mine for new associations, while commercial systems encode product catalogues and supply relationships as graphs that downstream models query. These deployments share a demand the research benchmarks understate, the need to keep the graph current as the world changes, and that demand is what makes the temporal case more than an academic refinement. A temporal knowledge graph attaches validity intervals to facts, recording not only that a relation held but when, and reasoning over it must respect that a fact true last year may be false today. Methods built for the static case do not transfer cleanly, because they carry no representation of time, and the temporal extensions that exist are less mature and less standardized than their static counterparts. The gap matters precisely because the highest-value applications, search and assistance over current events, are the ones where stale facts do the most damage, which is one reason the integration with language models, facing the same currency problem from the other side, has drawn so much attention.

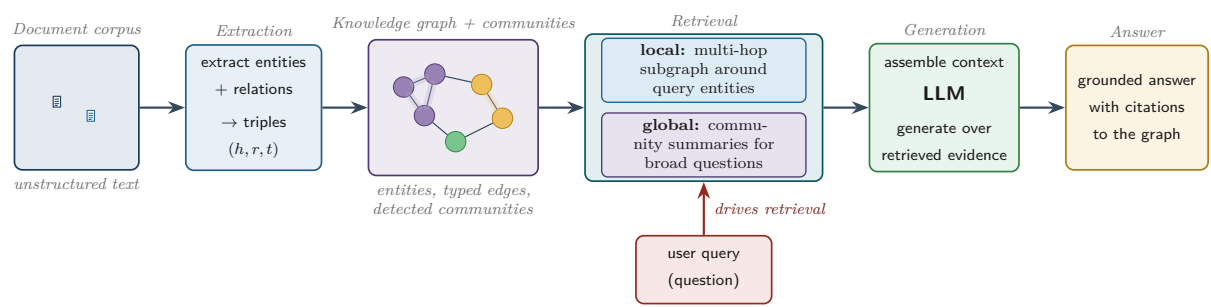
\begin{figure}[tb]
\centering
\resizebox{\textwidth}{!}{%
\begin{tikzpicture}[
  font=\sffamily, >=Stealth,
  stage/.style={rounded corners=4pt, draw=#1!75!black, fill=#1!12, line width=0.85pt,
                text=gnnink, align=center, inner sep=6pt, minimum height=22mm},
  knode/.style={circle, draw=#1!80!black, fill=#1!70, inner sep=0pt, minimum size=3.6mm},
  pill/.style={rounded corners=3pt, draw=#1!80!black, fill=#1!16, line width=0.7pt,
               align=center, font=\scriptsize, text=gnnink, inner sep=3pt},
  flow/.style={-{Stealth[length=3mm]}, line width=1.1pt, draw=gnnslate},
  clab/.style={font=\scriptsize\itshape, text=gnngray, align=center}]

% ---- 1: corpus ----
\node[stage=gnnblue, minimum width=22mm] (corpus) at (0,0) {};
\node[clab, anchor=south] at (corpus.north) {Document corpus};
\pic[scale=0.5] at ($(corpus.center)+(-0.35,0.1)$){ic doc={gnnblue}};
\pic[scale=0.5] at ($(corpus.center)+(0.25,-0.15)$){ic doc={gnnteal}};
\node[clab, anchor=north, text width=22mm] at (corpus.south) {unstructured text};

% ---- 2: extraction ----
\node[stage=gnnteal, minimum width=24mm, right=8mm of corpus] (extract)
  {\scriptsize extract entities\\\scriptsize + relations\\[3pt]\scriptsize $\to$ triples\\\scriptsize $(h,r,t)$};
\node[clab, anchor=south] at (extract.north) {Extraction};

% ---- 3: knowledge graph + communities ----
\node[stage=gnnpurple, minimum width=30mm, minimum height=24mm, right=8mm of extract] (kg) {};
\node[clab, anchor=south] at (kg.north) {Knowledge graph + communities};
\begin{scope}[shift={($(kg.center)+(-0.05,-0.05)$)}, scale=0.92]
  \node[knode=gnnpurple] (k1) at (-0.9,0.45) {};
  \node[knode=gnnpurple] (k2) at (-0.3,0.7)  {};
  \node[knode=gnnpurple] (k3) at (-0.5,-0.1) {};
  \node[knode=gnnamber]  (k4) at (0.6,0.5)   {};
  \node[knode=gnnamber]  (k5) at (1.0,-0.15) {};
  \node[knode=gnngreen]  (k6) at (0.1,-0.6)  {};
  \draw[gnnslate,line width=0.6pt] (k1)--(k2) (k1)--(k3) (k2)--(k3) (k4)--(k5) (k2)--(k4) (k3)--(k6) (k5)--(k6);
  \draw[gnnpurple!45,line width=4pt,line cap=round,opacity=0.30] (k1)--(k2)--(k3)--(k1);
  \draw[gnnamber!55,line width=4pt,line cap=round,opacity=0.30] (k4)--(k5);
\end{scope}
\node[clab, anchor=north, text width=30mm] at (kg.south) {entities, typed edges, detected communities};

% ---- 4: retrieval (two modes) ----
\node[stage=gnnocean, minimum width=42mm, minimum height=26mm, right=8mm of kg] (ret) {};
\node[clab, anchor=south] at (ret.north) {Retrieval};
\node[pill=gnnteal, text width=34mm]   (loc) at ($(ret.center)+(0,0.66)$)  {\textbf{local:} multi-hop subgraph around query entities};
\node[pill=gnnpurple, text width=34mm] (glo) at ($(ret.center)+(0,-0.66)$) {\textbf{global:} community summaries for broad questions};

% ---- 5: LLM ----
\node[stage=gnngreen, minimum width=22mm, minimum height=22mm, right=8mm of ret] (llm)
  {\scriptsize assemble context\\[3pt]\textbf{\small LLM}\\[3pt]\scriptsize generate over\\\scriptsize retrieved evidence};
\node[clab, anchor=south] at (llm.north) {Generation};

% ---- 6: answer ----
\node[stage=gnnamber, minimum width=22mm, right=8mm of llm] (ans)
  {\scriptsize grounded answer\\\scriptsize with citations\\\scriptsize to the graph};
\node[clab, anchor=south] at (ans.north) {Answer};

% ---- query entering retrieval from below ----
\node[stage=gnnred, minimum width=24mm, minimum height=12mm] (q) at ($(ret.south)+(0,-1.55)$)
  {\scriptsize user query\\\scriptsize (question)};
\draw[flow, draw=gnnred!75!black] (q.north) -- node[clab, right=1pt, text=gnnred!60!black]{drives retrieval} (ret.south);

% ---- main flow ----
\draw[flow] (corpus)--(extract);
\draw[flow] (extract)--(kg);
\draw[flow] (kg)--(ret);
\draw[flow] (ret)--(llm);
\draw[flow] (llm)--(ans);
\end{tikzpicture}}
\caption[End-to-end GraphRAG architecture.]{The end-to-end GraphRAG architecture for grounding a language model in a corpus. An offline stage turns unstructured text into a knowledge graph by extracting entity-relation triples, then detects communities that summarise the graph at several levels. At query time the system retrieves evidence in one of two modes: a local multi-hop subgraph around the entities named in the query, or, for broad questions, the community summaries that cover the corpus. The retrieved evidence is assembled into context for the language model, which generates an answer grounded in the graph and able to cite it. The construction and retrieval steps are where graph structure earns its place over flat text retrieval.}
\label{fig:e2egraphrag}
\end{figure}

The weaknesses are as structural as the strengths. A knowledge graph is only as good as its construction, and graphs extracted from text inherit every error and omission of the extraction, so a reasoning model can be defeated by a graph that is wrong rather than by a method that is weak. Scalability constrains reasoning, since exhaustive multi-hop search over a large graph is expensive and the subgraph-retrieval and community-summary schemes are in part responses to that cost. \Cref{fig:e2egraphrag} lays out the end-to-end GraphRAG architecture, from building the knowledge graph offline to retrieving local subgraphs or global community summaries at query time and generating a grounded answer. Evaluation is the least settled part of the newest work: retrieval-augmented generation over graphs is assessed with a patchwork of question-answering benchmarks and human judgements, and there is no agreed standard that isolates the contribution of the graph from the contribution of the underlying language model. That last point sharpens into the open question that hangs over the whole third thread, whether explicit graph structure adds durable value or whether a sufficiently capable language model will absorb the structure as it absorbed syntax. The honest answer is that the question is unresolved, and that the strongest evidence for structure comes precisely from the tasks, multi-hop and corpus-wide reasoning with traceable support, where a flat model has the least to stand on.

The open problems follow this contour. Temporal and dynamic knowledge graphs, in which facts have validity intervals and the graph changes over time, are underserved by methods built for a static snapshot. Constructing high-quality graphs from text at scale remains a bottleneck that limits everything downstream. A unified model of graph and language, rather than a pipeline that bolts one onto the other, is the destination much of the current work is aimed at, and it is the subject the survey takes up directly in the chapter on foundation models. A fair summary of the domain is that knowledge graphs gave graph networks their first natural home and their first hard problem, and that the same relational structure is now the most promising route to grounding language models, with the caveat that the field has not yet built the evaluations that would let it prove the point.

\FloatBarrier
\section{Drug discovery, molecules, and biology}
\label{sec:drug}

If recommendation is the domain where graph networks reached the largest audience, molecular science is the domain where they fit the data most exactly. A molecule is a graph without any modelling decision: atoms are nodes, bonds are edges, and the properties chemists care about are functions of that structure. The field also gave the methods chapter its organizing idea, since the message-passing framework was first crystallized as a unification of several molecular models, and it has produced some of the clearest real-world results in the whole survey, including the discovery of new drugs. This section treats three connected areas, the prediction of molecular properties, the use of graphs across drug discovery, and applications in structural biology, and it follows the familiar path from how the graph is built to where the approach succeeds and where it strains.

\subsection{Molecules as graphs}
\label{subsec:mol-graph}

\begin{figure}[t]
\centering
\resizebox{0.6\textwidth}{!}{%
\begin{tikzpicture}[
  atom/.style={circle, draw=#1!75!black, fill=#1, text=white, font=\normalsize\bfseries,
               inner sep=1.2pt, minimum size=11mm}, atom/.default=gnnslate,
  bond/.style={draw=gnnslate, line width=1.6pt},
  dbond/.style={draw=gnnslate, line width=1.6pt, double, double distance=2.2pt}]
\node[atom=gnnblue]  (n1) at (0,0)      {N};
\node[atom=gnnslate] (c1) at (1.7,0)    {C};
\node[atom=gnnslate] (c2) at (3.4,0)    {C};
\node[atom=gnnred]   (o1) at (4.7,1.15) {O};
\node[atom=gnnred]   (o2) at (4.7,-1.15){O};
\draw[bond]  (n1) -- (c1);
\draw[bond]  (c1) -- (c2);
\draw[dbond] (c2) -- (o1);
\draw[bond]  (c2) -- (o2);
\node[anchor=north, font=\footnotesize, text=gnngray, align=center] at (2.35,-1.7)
  {nodes $=$ atoms (typed by element)\qquad edges $=$ bonds (single / double)};
\end{tikzpicture}}
\caption[A molecule represented as a graph.]{A small molecule as a graph: atoms are nodes typed and coloured by element (nitrogen, carbon, oxygen) and bonds are edges typed by order, with the double line marking a double bond. Hydrogen atoms are omitted by convention. A property-prediction model reads features off this structure rather than off a hand-designed descriptor.}
\label{fig:molgraph}
\end{figure}
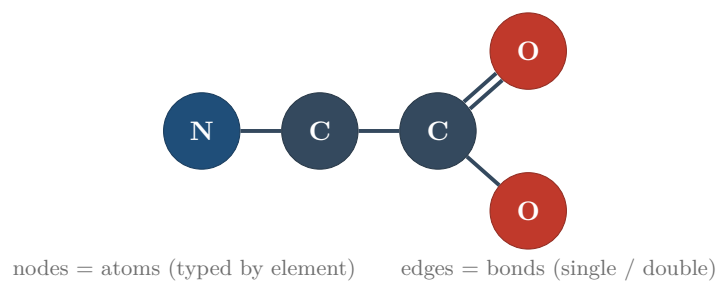

A molecular graph attaches a feature vector to every atom, recording its element, charge, and hybridization, and a feature vector to every bond, recording its order and whether it lies in an aromatic ring. \Cref{fig:molgraph} shows a small example. Two representations coexist and serve different purposes. The two-dimensional graph captures connectivity alone, which suffices for many property-prediction tasks and is cheap to compute, while the three-dimensional representation places atoms at spatial coordinates and is necessary when a property depends on geometry, as quantum-mechanical energies do. The distinction matters because a two-dimensional graph cannot distinguish stereoisomers, molecules with identical connectivity but different spatial arrangements that can behave very differently in the body, and a method that ignores geometry inherits that blindness. Molecular graphs are small, rarely exceeding a hundred atoms, which removes the scalability pressure that dominates the social and knowledge-graph domains and shifts the difficulty entirely onto accuracy: the graphs are easy to store and the properties are hard to predict. The atom and bond features deserve a word, because they carry chemistry the bare graph does not. An atom is typically described by its element, its degree, its formal charge, its hybridization state, whether it sits in an aromatic ring, and its chirality, and a bond by its order, its conjugation, and its ring membership. These features are supplied rather than learned, and a model's accuracy depends on them as much as on its architecture, which is one reason simple message-passing networks with good features often match more elaborate ones with poor features. An alternative input, the linear string notation chemists use to write molecules as text, can be fed to a sequence model, but the string imposes an arbitrary traversal order on a structure that has none, and the graph representation avoids that arbitrariness, the same argument that recurs whenever structure is flattened into a sequence.

\subsection{Tasks}
\label{subsec:mol-tasks}

The tasks in this domain span all three levels of the methods chapter and add concerns specific to chemistry. Property prediction is a graph-level task, mapping a molecule to a scalar or a class: aqueous solubility and lipophilicity as regression, toxicity and blood-brain-barrier permeability as classification, and quantum-mechanical energies as high-precision regression. Drug-target interaction asks whether and how strongly a candidate molecule binds a protein, a problem that joins two biological objects and is naturally posed as prediction over a pair. Related work embeds protein-interaction networks to predict molecular quantities downstream, joining network structure with per-node prediction \cite{dai2021piker2p}. Drug-drug interaction and the prediction of polypharmacy side effects are link-prediction tasks on a multi-relational graph of drugs and their targets. Molecular generation inverts property prediction, asking a model to produce novel molecules with desired characteristics rather than to score given ones, which raises questions of validity and synthesizability that scoring tasks never face. Structural biology adds protein-centred tasks, predicting a protein's function, the sites where it binds other molecules, and its interactions with other proteins. Reaction prediction and retrosynthesis, which plan how to make a target molecule, round out the set. What unites these is that each reduces to a prediction on a graph whose nodes and edges have direct physical meaning, so an error in the graph is an error about chemistry rather than about data formatting. Two groupings of these tasks drive most industrial interest. The first is the prediction of the properties that decide whether a molecule can become a drug, collected under the heading of absorption, distribution, metabolism, excretion, and toxicity, since a compound that binds its target but is toxic or cannot reach it is useless. The second is virtual screening, in which a model ranks a large library so that only the most promising compounds are tested, and lead optimization, in which a known active molecule is modified to improve it. On the quantum side the properties of interest are precise physical quantities such as the gap between a molecule's highest occupied and lowest unoccupied orbitals or its atomization energy, where the target accuracy is set by chemistry rather than by convention and is demanding enough that geometry cannot be ignored. Reaction prediction and retrosynthesis add a different flavour, since predicting the product of a reaction or the steps that synthesize a target is naturally posed as editing a graph, adding and removing bonds, rather than as scoring a fixed one. This reframing, from evaluating molecules to generating them, is what connects graph learning to drug design rather than only drug screening. Generating a valid molecule is harder than scoring one, because the output must satisfy the constraints of chemistry, correct valences and realizable structures, that a random graph would violate, and the methods that do it well build these constraints into the generation rather than hoping a model learns them. The payoff is the ability to search the vast space of possible molecules for ones with desired properties, rather than only ranking a fixed library, which is the difference between proposing new candidates and selecting among existing ones, and it is where the relational view of a molecule contributes most directly to discovery.

\subsection{Property prediction and message passing}
\label{subsec:mol-property}

Property prediction is the task that shaped the methods chapter. Several early molecular models, each proposed independently, were shown to be instances of a single message-passing scheme in which atoms exchange information along bonds and a graph-level readout produces the prediction, and that unification is the message-passing framework the survey has used throughout. A molecular property is read from the atom representations through a permutation-invariant pooling,
\begin{equation}\label{eq:molprop}
\hat{y}_{\graph}=\rho\!\Big(\sum_{v\in\vset}\hv_v^{(L)}\Big),
\end{equation}
with $\rho$ a small prediction head, and \cref{alg:molprop} states the full procedure, atom-feature initialization, message passing along bonds, readout, and prediction. For properties that depend on geometry, connectivity is not enough, and a second line of work makes the spatial arrangement of atoms part of the model. One approach forms messages from the distances and angles between bonded atoms, so that the bending of a molecule informs its predicted energy \cite{klicpera2020dimenet}, and a more general principle builds networks that are equivariant to rotation and translation, guaranteeing that a molecule and its rotated copy receive the same prediction by construction rather than by data augmentation \cite{satorras2021egnn}. Such a geometric message can be written compactly as a function of two atoms and the distance between them,
\begin{equation}\label{eq:geommsg}
\msg_{uv}=\phi\big(\hv_u,\hv_v,\lVert\xv_u-\xv_v\rVert\big),
\end{equation}
and the equivariant variants extend this to use relative positions while preserving the symmetry. The geometric models matter because two molecules can share a connectivity graph yet differ in shape, and only a model that sees coordinates can tell them apart; the directional approach conditions each message on the angle a bond makes with its neighbours, and the equivariant approach transforms coordinate-derived features so they rotate with the molecule rather than being memorized in a fixed frame. Expressiveness sets a quieter limit on the connectivity-only models. Because a message-passing network distinguishes structures exactly as well as the Weisfeiler-Leman test, it cannot separate certain pairs of molecules the test deems equivalent, and when those pairs have different properties no amount of training fixes the error, which has motivated higher-order and substructure-aware models that count rings and other motifs the basic network cannot. The available benchmarks, summarized in \cref{tab:moldatasets}, span solubility, toxicity, permeability, and quantum properties, drawn largely from a standard molecular collection and a large-scale graph benchmark.

\begin{algorithm}[t]
\caption{Molecular property prediction}
\label{alg:molprop}
\KwIn{molecular graph $\graph$ with atom features $\xv_v$ and bond features $\ev_{uv}$}
\KwOut{predicted property $\hat{y}$}
initialize $\hv_v^{(0)}\leftarrow \xv_v$ for every atom $v$\;
\For{layer $l=0$ to $L-1$}{
  \ForEach{atom $v$}{
    exchange messages with bonded neighbours by \cref{eq:designtemplate}\;
  }
}
$\hv_{\graph}\leftarrow \sum_{v\in\vset}\hv_v^{(L)}$\tcp*{permutation-invariant readout}
$\hat{y}\leftarrow \rho(\hv_{\graph})$ by \cref{eq:molprop}\;
\Return{property $\hat{y}$}
\end{algorithm}

\begin{table}[t]
\centering
\caption[Representative molecular datasets and benchmarks.]{Representative molecular datasets, by prediction level and the property each measures. The collections are drawn from a standard molecular benchmark suite and a large-scale graph benchmark.}
\label{tab:moldatasets}
\small
\begin{tabularx}{\textwidth}{@{}lllX@{}}
\toprule
Dataset & Level & Task & Property measured \\
\midrule
ESOL, FreeSolv \cite{wu2018moleculenet} & graph & regression     & aqueous solubility, hydration free energy \\
Tox21, ClinTox \cite{wu2018moleculenet} & graph & classification & toxicity and clinical toxicity \\
BBBP, BACE \cite{wu2018moleculenet}     & graph & classification & blood-brain-barrier permeability, enzyme inhibition \\
QM7, QM9 \cite{wu2018moleculenet}       & graph & regression     & quantum-mechanical properties \\
OGB molecular \cite{hu2020ogb}          & graph & both           & large-scale prediction with scaffold splits \\
\bottomrule
\end{tabularx}
\end{table}

A recurring caution applies with special force here. The scaffold split, which places molecules with different core structures into different folds, tests whether a model generalizes to chemistry it has not seen, and performance under a scaffold split is routinely and substantially worse than under a random split. Reported numbers that do not state the split are therefore hard to compare, and a model that looks strong on a random split can be close to useless on the out-of-distribution chemistry that matters in a real screening campaign. The molecular benchmarks are also small and noisy by the standards of other domains, since each label is a measurement from a physical assay rather than a click, and a careful treatment of the experimental uncertainty is often missing from the comparison. There is also a baseline question this domain has confronted more honestly than most. Before graph networks, molecules were represented by fixed circular fingerprints, hand-designed bit vectors recording which substructures a molecule contains, and these fingerprints fed to a gradient-boosted or random-forest model remain a strong and stubborn baseline. On several property tasks the learned graph representation provides only a modest gain over this classical pipeline, and on small datasets it can lose, because a flexible model has too little data to beat a good fixed featurization. The graph network's advantage is clearest on large datasets, on tasks where the relevant substructure is not in the fingerprint vocabulary, and where pretraining can be brought to bear, and stating that advantage precisely, rather than assuming it, is part of an honest account of the domain. The comparison with classical descriptors is worth making concrete. For decades, molecules were represented by fingerprints, fixed-length encodings of which substructures a molecule contains, fed to a conventional classifier, and these remain a strong baseline because they capture much of what determines a property without learning a representation at all. A graph network can in principle do better by learning features suited to the task rather than using a fixed vocabulary of substructures, and on many targets it does, but the margin is often modest and occasionally absent, which is why a careful study reports both the targets where the learned representation clearly helps and those where the fingerprint baseline is not meaningfully beaten. Stating this honestly is what separates a genuine advance from a demonstration that confirms what was already achievable.

\subsection{Drug discovery}
\label{subsec:mol-drug}

The clearest demonstration that graph networks can change practice came from antibiotic discovery, where a model trained to predict antibacterial activity screened a large chemical library and identified a compound, structurally unlike existing antibiotics, that was then confirmed in the laboratory to kill resistant bacteria \cite{stokes2020antibiotic}. The result is worth dwelling on because it inverts the usual relationship between benchmark and reality: the model's value was established not by a held-out metric but by a wet-lab confirmation of a molecule no chemist had flagged. The discovery also illustrates the screening pattern that graph models enable, training a cheap predictor on known actives and then applying it to a library far larger than could be tested directly, so that experimental effort is spent only where the model is most confident. The drug-target side adds the difficulty that a protein is not as easily graphed as a small molecule, and methods differ in whether they encode the protein as an amino-acid sequence read by a convolution or as a structural graph, a choice that trades the availability of sequence against the richer signal of structure. Beyond screening, several drug-discovery tasks have graph formulations. Drug-target binding affinity is predicted by encoding the drug as a graph and the protein as a sequence or graph and combining the two into an affinity,
\begin{equation}\label{eq:affinity}
\hat{a}=\psi\big(\hv_{\text{drug}},\hv_{\text{prot}}\big),
\end{equation} The geometric message and the affinity readout, \cref{eq:geommsg,eq:affinity}, recur across the molecular models discussed here.
with $\psi$ a learned interaction head \cite{nguyen2021graphdta}. Polypharmacy, the prediction of side effects that arise when drugs are combined, is modelled as link prediction on a multimodal graph that joins drugs to the proteins they act on, so that a predicted edge is a predicted interaction effect \cite{zitnik2018decagon}, and related work organizes drug knowledge into a graph and reasons over it to predict interactions \cite{lin2020kgnn}. Molecular generation addresses the inverse problem of designing new molecules, and graph-based generators build a molecule node by node or through a normalizing flow over graph structure, optimizing toward target properties while trying to keep the output a valid, synthesizable molecule \cite{shi2020graphaf,luo2021graphdf}. Generation is usually steered toward a goal, by conditioning on a desired property, by optimizing a property score with reinforcement learning, or by searching the model's latent space, and the central tension is between exploring chemical space widely enough to find something new and staying close enough to known chemistry that the result can actually be made. \Cref{tab:molmethods} places the representative methods side by side by how they represent a molecule and what they are built to do. A useful way to read the landscape of molecular methods is by the level of structure each commits to. The simplest treat a molecule as a graph of atoms and bonds and learn from that connectivity alone, which suffices for many properties. Others add three-dimensional geometry, the positions of atoms in space, because some properties depend on shape in ways the bond graph does not capture. Others still incorporate quantum-mechanical information or model the molecule's interaction with a target rather than the molecule in isolation. The progression buys accuracy on the properties that need it at the cost of data and computation, and the appropriate level is the one the target property actually requires, a judgment that recurs throughout the molecular and materials domains and that an honest method selection has to make explicitly.

\begin{table}[t]
\centering
\caption[Graph neural network methods for molecules.]{Graph neural network methods for molecular tasks, distinguished by how they represent a molecule, whether they use three-dimensional geometry, and their primary use. The list spans two-dimensional and geometric property predictors and the main families of molecular generators.}
\label{tab:molmethods}
\small
\begin{tabularx}{\textwidth}{@{}lXll@{}}
\toprule
Method / family & Representation & Geometry & Primary use \\
\midrule
Neural fingerprints \cite{duvenaud2015fingerprints} & 2D atom--bond graph     & no  & differentiable fingerprints \\
Molecular graph conv \cite{kearnes2016molconv}       & 2D atom--bond graph     & no  & property prediction \\
MPNN \cite{gilmer2017mpnn}                           & 2D atom--bond graph     & no  & property prediction \\
D-MPNN \cite{yang2019dmpnn}                          & 2D directed-bond graph  & no  & property prediction \\
AttentiveFP \cite{xiong2020attentivefp}              & 2D graph with attention & no  & property prediction \\
SchNet \cite{schutt2018schnet}                       & 3D coordinates          & yes & quantum properties \\
DimeNet \cite{klicpera2020dimenet}                   & 3D graph with angles    & yes & quantum properties \\
GemNet \cite{gasteiger2021gemnet}                    & 3D directional graph    & yes & quantum properties, forces \\
PaiNN \cite{schutt2021painn}                         & 3D equivariant graph    & yes & tensorial properties \\
SphereNet \cite{liu2022spherenet}                    & 3D spherical messages   & yes & quantum properties \\
E($n$)-equivariant \cite{satorras2021egnn}           & 3D coordinates          & yes & geometry-aware prediction \\
GROVER \cite{rong2020grover}                         & pretrained 2D graph     & no  & transfer to scarce-label tasks \\
JT-VAE \cite{jin2018jtvae}                           & junction-tree graph     & no  & molecule generation \\
GCPN \cite{you2018gcpn}                              & generative graph (RL)   & no  & goal-directed generation \\
MolGAN \cite{decao2018molgan}                        & generative graph (GAN)  & no  & small-molecule generation \\
GraphAF, MoFlow \cite{shi2020graphaf,zang2020moflow} & generative flow         & no  & de novo generation \\
\bottomrule
\end{tabularx}
\end{table}

\subsection{Structural biology and bioinformatics}
\label{subsec:mol-bio}

Proteins extend the molecular picture to a larger scale and a different graph construction. A protein is built from amino-acid residues, and the standard graph places a node at each residue and an edge between residues that lie close together in the folded structure, turning a contact map into a graph on which a network can operate. Protein function prediction reads a functional label from this structural graph, learning the structural motifs that signal what a protein does \cite{gligorijevic2021deepfri}, and interface prediction marks the residues at which one protein binds another, a node-classification task on the residue graph \cite{fout2017protein}. \Cref{tab:bioapps} collects these biological applications with their graph constructions. The field was reshaped by accurate structure prediction from sequence, which made reliable three-dimensional structures available at scale, and graph methods that operate on structure now benefit from predicted structures where experimental ones are missing, a dependency that ties this line of work to advances outside the graph literature. The breakthrough in question, accurate prediction of a protein's folded structure from its sequence, used attention over a graph-like representation of residues and effectively removed a bottleneck that had limited structure-based methods for decades, and the flood of predicted structures it produced is now an input to graph models rather than a competitor. Two further protein tasks have natural graph formulations. Inverse folding, the design of a sequence that will fold into a desired shape, is generation on a residue graph constrained by geometry, and molecular docking, the prediction of how a small molecule sits in a protein's binding pocket, can be cast as reasoning over a graph that joins the two. These tasks share the domain's defining feature, that the graph's nodes and edges are physical objects and physical contacts, so a structural error is a claim about biology. Other biological networks fit the same template, including gene regulatory networks, protein-protein interaction networks, and the graphs built from single-cell measurements, in each case treating a biological system as a graph and a biological question as node, edge, or graph prediction. These wider biological graphs differ from molecules in scale and in noise. A gene regulatory network or a protein interaction network is large, incompletely measured, and assembled from many experiments of varying reliability, which makes them closer in character to the knowledge graphs of the previous chapter than to the small, clean graphs of a single molecule, and the methods that work on them borrow as much from relational reasoning as from molecular modelling. Disease and patient-level graphs push further in this direction and shade into the healthcare applications taken up next, where the graph is built not from a molecule but from a population or a physiological system.

\begin{table}[t]
\centering
\caption[Biological applications of graph neural networks.]{Biological applications, with the graph each builds and the task it poses. A representative method is named where one fits cleanly. The applications span drug interaction and docking, protein function and structure, and single-cell analysis.}
\label{tab:bioapps}
\small
\begin{tabularx}{\textwidth}{@{}lXll@{}}
\toprule
Application & Graph construction & Task & Example \\
\midrule
Drug--target affinity   & drug graph with protein sequence & regression           & GraphDTA \cite{nguyen2021graphdta} \\
Drug--target interaction & drug and protein graphs         & classification       & \cite{torng2019gcndti} \\
Polypharmacy effects    & multimodal drug--protein graph   & link prediction      & Decagon \cite{zitnik2018decagon} \\
Drug--drug interaction  & drug knowledge graph             & link prediction      & KGNN \cite{lin2020kgnn} \\
Drug--drug substructure & paired substructure graphs       & interaction prediction & SSI-DDI \cite{nyamabo2021ssiddi} \\
Molecular docking       & ligand and protein graphs        & pose prediction      & EquiBind, DiffDock \cite{stark2022equibind,corso2023diffdock} \\
Protein function        & residue contact graph            & graph classification & DeepFRI \cite{gligorijevic2021deepfri} \\
Protein function (large) & residue graph                   & multi-label classification & \cite{you2021deepgraphgo,lai2022gatgo} \\
Protein interface       & paired residue graphs            & node classification  & \cite{fout2017protein,reau2023deeprankgnn} \\
Protein interaction     & interaction network              & link prediction      & \cite{lv2021gnnppi,jha2022ppignn} \\
Protein representation  & 3D structure graph               & pretraining          & GearNet \cite{zhang2023gearnet} \\
Single-cell analysis    & cell similarity graph            & clustering           & scGNN \cite{wang2021scgnn} \\
\bottomrule
\end{tabularx}
\end{table}

\subsection{Pretraining and self-supervision}
\label{subsec:mol-pretrain}

Labelled molecular data is scarce in a way that labelled images and text are not, because each label is the outcome of a physical experiment that costs time and money, and the scarcity is the binding constraint on supervised molecular models. Self-supervised pretraining is the natural response, learning general molecular representations from the very large collections of unlabelled molecules that chemistry has catalogued, then fine-tuning on a small labelled set for the task at hand. The unlabelled collections are large, numbering in the millions to billions of catalogued compounds, and the pretraining objectives mirror those used elsewhere, masking atoms or bonds and asking the model to recover them, predicting the local context a substructure sits in, or, for geometric models, denoising perturbed coordinates so the model acquires a sense of physically plausible structure. The approach is not free of hazard, since a pretraining task poorly matched to the downstream property can transfer negatively and leave the model worse than one trained from scratch, a failure that mirrors the negative-transfer risk seen in other domains. One approach pretrains a transformer-style graph network on molecules with self-supervised objectives that predict masked structure \cite{rong2020grover}, and a contrastive approach builds augmented views of a molecule and trains the model to recognize that they describe the same compound \cite{wang2022molclr}, applying to chemistry the contrastive principle the methods chapter set out in general terms. The payoff is largest exactly where it is needed, on the small, hard, out-of-distribution property tasks where supervised models have too few labels to generalize, and the approach connects directly to the foundation-model ambitions discussed later in the survey.

\subsection{Strengths, weaknesses, and open problems}
\label{subsec:mol-tradeoffs}

The strength of graph methods in this domain is the tightness of the fit between model and object. A molecule is a graph, a protein contact map is a graph, and a drug-target system is a graph, so the inductive bias of a graph network is not an approximation but a match, and the antibiotic result shows that the match can translate into discoveries that conventional screening missed. Geometric and equivariant networks add the spatial fidelity that quantum and structural tasks demand, and pretraining addresses the data scarcity that would otherwise cap the approach. These are real advantages, supported by a decade of benchmarks and a growing record of laboratory confirmation.

The weaknesses are specific and consequential. A two-dimensional graph discards stereochemistry, and even a three-dimensional model can be defeated by the expressiveness limits the methods chapter described, since a standard message-passing network cannot always tell apart molecules that differ in ways the Weisfeiler-Leman test cannot see, and some of those differences are chemically real. Generalization across scaffolds is poor, which is the gap between benchmark accuracy and screening utility, and the benchmarks themselves are small and carry experimental noise that the headline numbers usually ignore. Generation raises its own difficulties, because a generated graph can be a valid graph and an impossible or unsynthesizable molecule, and the metrics that score validity and novelty do not capture whether a chemist could actually make the compound. A subtler failure is the activity cliff, a pair of molecules nearly identical in structure but very different in activity, which violates the smoothness a graph network implicitly assumes and which it therefore tends to predict poorly. Interpretability cuts the other way and is a genuine strength when it works, since the attribution methods discussed later in the survey can sometimes point to the substructure responsible for a predicted property, giving a chemist a hypothesis rather than only a number. Above all, the distance between an in-silico prediction and a wet-lab result is wide, and the literature's habit of reporting the former as if it settled the latter is the domain's version of the calibration problem that runs through the survey.

The open problems follow from these limits. Geometric foundation models that pretrain on three-dimensional structure, generation that respects synthesizability rather than only validity, the integration of quantum-mechanical priors into learned models, and multimodal models that combine structure with assay text and experimental context are all active directions. The most important, and the least technical, is data quality: molecular machine learning is bottlenecked by the size, noise, and reproducibility of experimental datasets, and progress on benchmarks will continue to overstate progress in the laboratory until that bottleneck is addressed. A fair summary is that molecular science is where graph networks fit the data best and have already produced tangible results, and also where the gap between a good metric and a real discovery is most visible and most in need of honest accounting.

\FloatBarrier
\section{Healthcare, brain networks, and medicine}
\label{sec:health}

Medicine produces relational data of several distinct kinds, and graph networks have been applied to each. The brain is a network of regions, a patient population can be organized by similarity, medical knowledge is a hierarchy of concepts, and an electronic health record is a structured history of coded events. This breadth makes healthcare a natural target for graph methods, and it is also the domain where the cautions running through this survey carry the most weight, because the datasets are small, the stakes are high, the need for interpretation and trust is acute, and a model that works in one hospital may fail in the next. The section surveys the principal graph constructions in medicine, the tasks they serve, and the methods built on them, and it is deliberately attentive to the distance between a strong benchmark number and a clinically useful tool.

\subsection{Graphs in medicine}
\label{subsec:health-graphs}

Several graph constructions recur, each answering a different clinical question. The brain network, or connectome, places a node at each brain region and an edge between regions whose activity correlates or which a white-matter tract connects, turning a scan into a graph on which a disorder can be classified. The population graph inverts the usual figure-ground relationship by making each patient a node and each edge a measure of similarity between patients, so that a diagnosis can spread across a cohort by semi-supervised learning. The medical ontology is a graph of diagnoses, procedures, and concepts arranged in hierarchies, used to give structure to the sparse, high-dimensional codes that fill a health record. The patient-specific graph built from a record links visits, diagnoses, and medications over time. Medical images supply a fifth construction, in which superpixels, anatomical regions, or, in histopathology, individual cells become nodes joined by spatial proximity. These constructions differ enough that methods rarely transfer between them, and the choice of which to build is the first and most consequential modelling decision in any medical application.

\subsection{Brain networks}
\label{subsec:health-brain}

\begin{figure}[t]
\centering
\resizebox{0.66\textwidth}{!}{%
\begin{tikzpicture}[
  reg/.style={circle, draw=gnnpink!80!black, fill=tintpink, line width=0.8pt, minimum size=7.5mm},
  con/.style={draw=gnnpurple}]
\draw[gnnpink!55, line width=1.1pt, fill=gnnpink!6] (0,0.25) ellipse (3.6 and 2.1);
\pic[scale=1.4] at (-3.0,1.95) {ic brain={gnnpink}};
\node[reg] (a) at (-2.2,1.0){};
\node[reg] (b) at (-1.4,1.65){};
\node[reg] (c) at (-2.4,0.0){};
\node[reg] (d) at (-1.6,-0.9){};
\node[reg] (e) at (-0.7,0.45){};
\node[reg] (f) at (0.7,0.45){};
\node[reg] (g) at (1.6,-0.9){};
\node[reg] (h) at (2.4,0.0){};
\node[reg] (i) at (1.4,1.65){};
\node[reg] (j) at (2.2,1.0){};
\draw[con,line width=1.7pt] (a)--(b); \draw[con,line width=0.6pt](a)--(c);
\draw[con,line width=1.1pt] (b)--(e); \draw[con,line width=0.6pt](c)--(d);
\draw[con,line width=1.4pt] (c)--(e); \draw[con,line width=0.7pt](d)--(e);
\draw[con,line width=2.0pt] (e)--(f);
\draw[con,line width=1.1pt] (f)--(h); \draw[con,line width=0.6pt](f)--(i);
\draw[con,line width=1.4pt] (h)--(g); \draw[con,line width=0.7pt](g)--(j);
\draw[con,line width=1.5pt] (i)--(j); \draw[con,line width=0.6pt](h)--(j);
\end{tikzpicture}}
\caption[A brain network as a graph.]{A brain network, or connectome, as a graph: nodes are brain regions (pink) and weighted edges are their functional or structural connections (purple), with thicker edges marking stronger connectivity. Disease classification reads a label from the whole graph, and interpretable models try to name the regions and connections that drive the prediction.}
\label{fig:braingraph}
\end{figure}
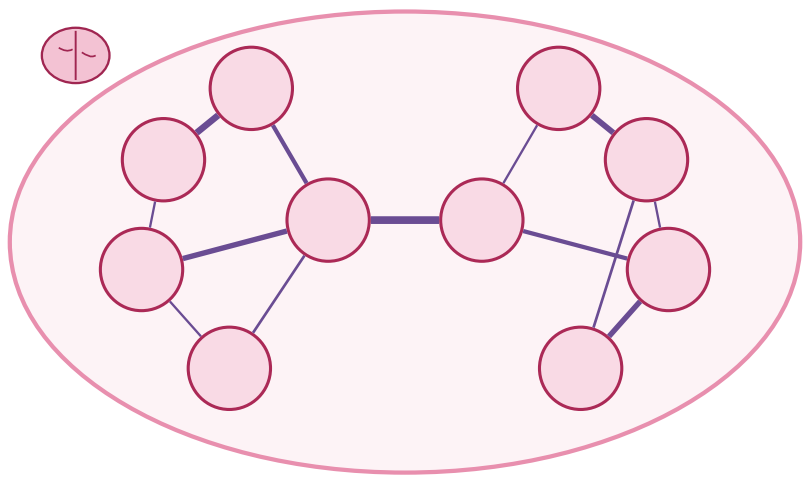

The connectome is the most graph-like object in medicine, and \cref{fig:braingraph} shows its form: regions as nodes, their connections as weighted edges. Two kinds of connectivity define two kinds of graph, a functional graph in which an edge records that two regions activate together and a structural graph in which an edge records an anatomical tract between them. The dominant task is classification of a neurological or psychiatric condition, such as a developmental disorder or a degenerative disease, from the pattern of connectivity, with secondary tasks predicting age or cognitive scores and identifying the connections that mark a condition. A representative method augments graph pooling with region awareness so that the model not only classifies a brain but reports which regions drove the decision, an interpretability that is close to mandatory in a clinical setting \cite{li2021braingnn}. Classification reads a label from the whole graph through a pooling and prediction head,
\begin{equation}\label{eq:braincls}
\hat{y}_{\graph}=\softmax\!\big(\rho(\mathrm{POOL}(\{\hv_v^{(L)}\}))\big),
\end{equation}
and the pooling step is where region-level interpretability is built in. The difficulties here are characteristic of the whole domain. Sample sizes are small, often a few hundred subjects, which invites overfitting and makes reported differences fragile; individual variability is large, so a pattern learned on one cohort may not hold in another; and the graph itself is not given but constructed, since turning continuous correlations into edges requires a threshold whose choice measurably changes the result and is rarely justified. A few specifics sharpen the picture. Functional connectivity is usually estimated from functional magnetic resonance imaging by correlating the activity time series of regions defined by a parcellation atlas, while structural connectivity is traced from diffusion imaging, and the two need not agree. The connectome is also not static, since functional coupling changes over seconds and minutes, and a dynamic formulation that treats the brain as a sequence of graphs captures states a single averaged graph hides. Neuroimaging machine learning has had a public reckoning with reproducibility, as effects reported on small samples have repeatedly failed to replicate on larger ones, and graph models inherit this exposure, because the combination of small cohorts, many analytic choices, and flexible models is exactly the recipe for findings that look strong and do not hold. The methodological hazards here deserve to be named rather than glossed, because they recur across medical graph applications. When the number of subjects is in the hundreds and the model has many parameters, the risk of fitting noise that happens to separate the training groups is severe, and a result not validated on a genuinely independent cohort should be treated as provisional. The construction of the brain graph compounds the concern, since the choice of how to define regions and how to threshold connections offers many degrees of freedom that, if explored against the outcome, inflate apparent performance. The discipline the domain requires, independent validation, fixed preprocessing chosen in advance, and honest reporting of how construction choices were made, is exactly the discipline the small-sample setting makes both hardest and most necessary.

\subsection{Population graphs and disease prediction}
\label{subsec:health-population}

A distinctive and influential idea in medical graph learning is to predict disease over a graph whose nodes are subjects rather than anatomical parts. Each subject carries imaging-derived features as a node attribute, and edges encode similarity drawn from non-imaging information such as demographics or acquisition site, so that the model combines two complementary signals, the imaging features on the nodes and the phenotypic similarity in the structure \cite{parisot2018disease}. An edge weight in such a graph multiplies a phenotypic affinity by an imaging similarity,
\begin{equation}\label{eq:popedge}
\Adj_{ij}=k_{\mathrm{ph}}(c_i,c_j)\cdot \mathrm{sim}\big(\xv_i,\xv_j\big),
\end{equation} The graph-classification readout and the population-graph edge construction, \cref{eq:braincls,eq:popedge}, recur across the medical applications above.
where $c$ denotes the non-imaging phenotype and $\xv$ the imaging features, and disease prediction becomes semi-supervised node classification on the resulting cohort graph. The construction fits a fixed study population well, but it carries two liabilities that recur in the domain. The similarity that defines the edges is a modelling choice with no obviously correct form, and a different choice yields a different graph and a different result; and because the graph is built around a fixed cohort, extending the model to a new patient who was not part of the original graph is awkward, which is the transductive-to-inductive gap the foundations chapter described, now with clinical consequences. Fairness adds a third concern, since edges drawn from demographic similarity can entrench demographic structure in the predictions. Later work has tried to remove the dependence on a hand-chosen similarity by learning the graph structure jointly with the classifier, so the edges are optimized for the task rather than fixed in advance, and by moving to inductive formulations that embed a new patient without rebuilding the cohort graph. These refinements ease the two liabilities but do not remove the deeper one, that a population graph encodes assumptions about which patients resemble which, and those assumptions deserve the same scrutiny as any other part of a clinical model. The population-graph idea also runs into the transductive obstacle the synthesis chapter identified, since a model that classifies patients as nodes in a fixed similarity graph does not straightforwardly accommodate a new patient who was absent when the graph was built. In a research setting where the cohort is fixed this is tolerable, but in a clinical setting where patients arrive continuously it is a real limitation, and addressing it requires either rebuilding the graph and retraining or adopting an inductive formulation that can place a new patient without redefining the whole structure. This tension between the transductive convenience of a fixed similarity graph and the inductive demands of deployment is one the domain has not fully resolved.

\subsection{Health records and clinical prediction}
\label{subsec:health-ehr}

An electronic health record is sparse, irregular, and high-dimensional, a long history of coded diagnoses, procedures, and medications recorded at uneven intervals, and graph structure helps in two ways. The first uses a medical ontology to relate codes that a flat model would treat as unrelated, so that a rare diagnosis borrows representation from its more common ancestors in the hierarchy and is not crippled by having few training examples; an attention mechanism over the ontology graph learns how much to rely on each ancestor \cite{choi2017gram}. The second builds a graph among the entities a record contains and reasons over it for a clinical decision, as in medication recommendation, where a model combines a patient's history with a graph of drug interactions to suggest a safe and effective combination rather than a single drug in isolation \cite{shang2019gamenet}. The tasks in this area include predicting a future diagnosis, estimating the risk of mortality or readmission, recommending treatment, and assigning phenotypes, and they share a set of obstacles: data is missing in ways that are themselves informative, events are irregularly timed, records are governed by privacy constraints that limit sharing, and coding practice differs enough between institutions that a model trained at one hospital can degrade sharply at another. The temporal character of records is central and underexploited. A record is a sequence of events at irregular intervals, and models that respect the timing, by combining graph structure over codes with a temporal component over visits, capture progression that a static snapshot misses. The unstructured part of a record, the free-text clinical notes, carries information the codes omit, and connecting the language methods of the knowledge-graph chapter to the structured record is an active direction that mirrors, in a clinical setting, the graph-and-language convergence discussed earlier.

\subsection{Medical imaging}
\label{subsec:health-imaging}

Medical imaging is dominated by convolutional models that treat an image as a grid, but several problems have a relational structure a grid misses, and graphs supply it. Segmenting an image into superpixels or anatomical regions and connecting them yields a graph on which a network can reason about the relationships between parts rather than about pixels in isolation. Surface meshes of organs are graphs by construction. The clearest case is histopathology, where a tissue sample is turned into a cell graph whose nodes are individual cells and whose edges join cells that lie close together, so that a model can read the tissue architecture, the spatial organization of cells, that pathologists use to grade cancer and that a pixel-level model represents only indirectly. The appeal in each case is the same: a graph encodes which parts relate to which, and in medicine that relational structure often carries the diagnostic signal. Specific methods in this area are less consolidated than in the connectome and record settings, and the line is best understood as an emerging application of the graph constructions the survey has already described rather than as a settled body of architectures. Two newer directions extend the imaging picture. Spatial transcriptomics measures gene expression at many locations within a tissue while preserving their spatial arrangement, which is naturally a graph of locations joined by proximity and annotated with expression, and graph models are a fit for relating molecular state to tissue structure. Cortical surface analysis represents the folded brain surface as a mesh and studies its geometry, a setting where the graph is the anatomy itself. Both reinforce the domain's pattern, that the most promising medical uses of graphs arise where the relational or spatial structure is the signal of interest rather than a convenience.

\subsection{Epidemics and public health}
\label{subsec:health-epi}

A different medical use of graphs operates at the level of populations rather than individuals. Disease spreads through contact, and the contact and mobility networks along which it travels are graphs, so forecasting an epidemic becomes a prediction problem on a graph whose nodes are regions or individuals and whose edges carry movement or contact. Spatio-temporal graph models, which the transportation chapter develops in detail, transfer directly to this setting, combining propagation over the spatial graph with a temporal component that tracks how case counts evolve. The approach drew particular attention during recent global outbreaks, when mobility data made the relevant graphs observable at scale. The same difficulties that beset other medical applications apply, since contact networks are measured incompletely and shift as behaviour changes, and a forecast grounded in a stale network misleads with the same confidence as one built on a current one. The public-health setting also raises the privacy concerns of the record setting in sharper form, because the data that makes the graph observable is the data that tracks where people go.

\subsection{Datasets, strengths, and weaknesses}
\label{subsec:health-tradeoffs}

\Cref{tab:healthcare} collects the main medical applications with their graph constructions and representative methods. Evaluation across the domain draws on a handful of public resources, brain-imaging cohorts assembled for the study of specific disorders and large de-identified records of intensive-care stays, supplemented by institutional datasets that cannot be shared. These resources are small and heterogeneous compared with the benchmarks of other domains, and that scarcity shapes everything about how medical graph models should be read.

The strengths are real. Medical data is relational in ways that match graph methods closely, a connectome is a graph, patient similarity is a graph, and a medical ontology is a graph, and the semi-supervised formulations let a model extract signal from the small labelled cohorts that are all clinical studies usually provide. Interpretability-oriented designs of the kind used for brain networks address a requirement that is optional in most domains and essential here, since a prediction a clinician cannot interrogate is a prediction a clinician cannot act on.

The weaknesses are where this domain demands more honesty than most. Small samples make overfitting the rule rather than the exception, and a model that reports a strong score on a few hundred subjects may be fitting the cohort rather than the condition. The graph constructions are under-justified, since the thresholding of a connectome and the choice of a patient-similarity metric are modelling decisions that change results and are rarely subjected to the scrutiny they deserve. Generalization across sites, scanners, and populations is poor, and a model validated at one institution can fail at another for reasons that have nothing to do with the disease. Above all, the distance between a benchmark metric and clinical utility is wider in medicine than anywhere else in this survey, because a model that improves an area-under-the-curve by a small margin may still be unusable in care, and the field's tendency to report the former as if it implied the latter is the calibration problem at its most consequential.

The open problems follow directly. Learning across institutions without moving private data, through federated and privacy-preserving methods, is needed before medical graph models can be trained at the scale that would make them reliable. Quantifying uncertainty is a precondition for clinical trust, since a confident wrong prediction is more dangerous than an honest abstention. Fairness across demographic groups, principled rather than incidental graph construction, and the prospect of foundation models for medical graphs are all active directions. A fair summary is that healthcare is a domain where the relational fit of graph methods is genuine and the potential is large, and also the domain where the gap between method and deployment is widest, the samples smallest, and the need for methodological discipline greatest.

\begin{table}[t]
\centering
\caption[Healthcare applications of graph neural networks.]{Medical applications, with the graph each builds and the task it poses. A representative method is named where the line of work has consolidated around one. The applications span brain networks, EEG, electronic health records, and medication recommendation.}
\label{tab:healthcare}
\small
\begin{tabularx}{\textwidth}{@{}lXll@{}}
\toprule
Application & Graph (nodes / edges) & Task & Example \\
\midrule
Brain disorder analysis   & regions / connectivity            & graph classification & BrainGNN \cite{li2021braingnn} \\
Brain (hierarchical)      & regions / connectivity            & graph classification & Hi-GCN \cite{jiang2020higcn} \\
Disease prediction        & patients / phenotypic similarity  & node classification  & population graph \cite{parisot2018disease} \\
EEG emotion recognition   & electrodes / functional links     & graph classification & RGNN, DGCNN \cite{zhong2020rgnn,song2020dgcnneeg} \\
EEG (learned structure)   & electrodes / learned links        & classification       & \cite{li2019stgcneeg} \\
Diagnosis from records    & medical codes / ontology          & sequence prediction  & GRAM \cite{choi2017gram} \\
EHR structure learning    & codes / learned structure         & prediction           & GCT \cite{choi2020gct} \\
Medication recommendation & drugs and visits / interactions   & recommendation       & GAMENet \cite{shang2019gamenet} \\
Medication (safety)       & drug molecular graphs             & recommendation       & SafeDrug \cite{yang2021safedrug} \\
Medication (pretraining)  & codes / EHR graph                 & recommendation       & G-BERT \cite{shang2019gbert} \\
\bottomrule
\end{tabularx}
\end{table}

\FloatBarrier
\section{Computer vision, scene graphs, and point clouds}
\label{sec:vision}

Computer vision is dominated by convolutional networks and, more recently, by transformers, and graph networks are not a replacement for either. They enter vision where the data or the desired output is irregular or explicitly relational in a way a grid of pixels or a sequence of patches does not capture, and in those niches they are a natural fit. Three settings account for most of the work. A scene graph turns an image into objects and their relationships, a structured output that is a graph by definition. A point cloud from a depth sensor is an unordered set of points in space, which a graph on spatial neighbours can model without forcing it onto a grid. And relational and few-shot problems ask a model to reason about how regions, objects, or examples relate, which is again a graph. This section treats these settings and is candid about a tension specific to vision, that the attention mechanism at the heart of the transformer is itself a kind of learned graph, and has absorbed much of the relational modelling that once motivated explicit graph networks.

\subsection{Where graphs enter vision}
\label{subsec:vision-where}

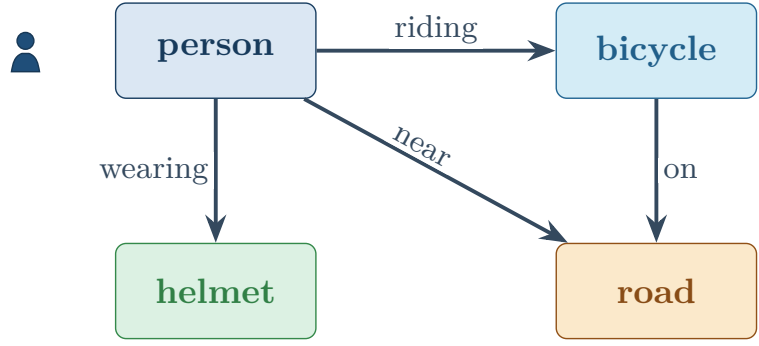
\begin{figure}[t]
\centering
\resizebox{0.62\textwidth}{!}{%
\begin{tikzpicture}[
  obj/.style 2 args={rounded corners=3pt, draw=#1!80!black, fill=#2, font=\small\bfseries,
            text=#1!75!black, inner sep=4pt, align=center, minimum height=9.5mm, minimum width=20mm},
  rel/.style={font=\footnotesize, fill=white, text=gnnslate, inner sep=1.5pt},
  ar/.style={-{Stealth[length=2.8mm]}, draw=gnnslate, line width=1.1pt}]
\node[obj={gnnblue}{tintblue}]    (person) at (0,0)     {person};
\node[obj={gnnteal}{tintteal}]    (bike)   at (4.4,0)   {bicycle};
\node[obj={gnngreen}{tintgreen}]  (helmet) at (0,-2.4)  {helmet};
\node[obj={gnnrust}{tintamber}]   (road)   at (4.4,-2.4){road};
\pic[scale=0.7] at (-1.9,0) {ic person={gnnblue}};
\draw[ar] (person) -- node[rel,above]{riding}  (bike);
\draw[ar] (person) -- node[rel,left]{wearing}  (helmet);
\draw[ar] (bike)   -- node[rel,right]{on}       (road);
\draw[ar] (person) -- node[rel,pos=0.42,above,sloped]{near} (road);
\end{tikzpicture}}
\caption[A scene graph extracted from an image.]{A scene graph: detected objects are nodes (coloured by category) and their pairwise relationships are labelled, directed edges, giving a structured summary of an image's content. Generating one is object detection followed by relationship prediction, and the result is, in effect, a knowledge graph of a single scene.}
\label{fig:scenegraph}
\end{figure}

The default representations in vision are the pixel grid, on which convolution slides, and the patch sequence, over which a transformer attends, and both are regular structures. Graphs become the better choice in four situations. The first is when the desired output is itself a graph, as a scene graph is. The second is when the input is geometric and irregular, as a point cloud is, so that no grid fits it without distortion. The third is relational reasoning, in which the relationships between detected objects carry the answer, as in visual question answering about how things in a scene relate. The fourth is few-shot learning, where the relationships among a handful of labelled examples and the unlabelled queries form a small graph on which labels can propagate. \Cref{fig:scenegraph} illustrates the first of these, the scene graph, which doubles as a useful mental model for the whole section: a graph laid over visual content.

\subsection{Scene graphs}
\label{subsec:vision-scene}

A scene graph represents an image as objects, their attributes, and the pairwise relationships between them, each relationship a subject-predicate-object triple of the same form a knowledge graph uses. The connection is exact: a scene graph is a knowledge graph extracted from a single image, and the methods of the knowledge-graph chapter apply once the graph exists. Generating the graph is the hard part, and it decomposes into detecting objects and predicting the relationship between each pair, where the relationship probability is read from the two object representations and their joint context,
\begin{equation}\label{eq:scenerel}
p\big(r_{ij}\big)=\softmax\!\big(\psi(\hv_i,\hv_j,\hv_{ij})\big),
\end{equation}
with $\hv_{ij}$ a representation of the pair's union region. Graph networks improve this by letting the object and relationship proposals inform one another through message passing, so that a detected bicycle raises the probability of a riding relation to a nearby person and vice versa, rather than predicting each in isolation \cite{yang2018graphrcnn}. A complementary observation is that scene graphs are highly regular, since a few relationships dominate and objects strongly predict their likely relations, and a model that exploits these statistical motifs is hard to beat \cite{zellers2018neuralmotifs}. Scene graphs feed image captioning, visual question answering, image retrieval, and image generation, in each case providing a structured intermediate that is easier to reason over than raw pixels. The intermediate is most valuable when the downstream task is itself relational. Image generation runs the pipeline in reverse, synthesizing an image from a scene graph so a user can specify content by its structure, and visual question answering benefits because a question about how two objects relate maps onto an edge to be read rather than a pattern to be found in pixels. The data that drives this line annotates images densely with objects, attributes, and relationships, and its long-tailed relationship distribution is both what makes the statistical regularities exploitable and what makes rare relationships hard, a tension that surfaces in every scene-graph result. The long-tailed distribution of relationships is the defining difficulty of scene-graph generation. A handful of relations, such as on or near, account for most of the labelled instances, while the relations that carry the most information, the specific and unusual ones, are rare, so a model trained to maximize overall accuracy learns to predict the common relations and neglect the informative ones. This is the same imbalance problem the fraud domain faced, in a different guise, and the responses are similar, reweighting the rare cases or evaluating with metrics that do not let the common relations dominate the score. Until this is handled, a scene-graph model can post a high average accuracy while failing at exactly the relationships that would make the graph useful, which is why the metric a scene-graph result reports matters as much as the number it attains.

\subsection{Point clouds and three-dimensional data}
\label{subsec:vision-3d}

Depth sensors and laser scanners produce point clouds, unordered sets of points in three-dimensional space, and the absence of a grid is exactly what makes them awkward for convolution and natural for graphs. An early and influential alternative processed each point independently and pooled the results, which respects the permutation invariance of a point set but ignores local geometry. Graph methods recover that geometry by building a graph on each point's spatial neighbours and passing messages along it. One widely used approach constructs the neighbourhood graph dynamically in feature space at every layer and aggregates edge features that encode the difference between a point and its neighbours,
\begin{equation}\label{eq:edgeconv}
\hv_v^{(l+1)}=\max_{u\in\nbr(v)}\,h_{\Theta}\!\big(\hv_v^{(l)},\,\hv_u^{(l)}-\hv_v^{(l)}\big),
\end{equation} The relationship score and the edge-convolution update, \cref{eq:scenerel,eq:edgeconv}, recur across the vision tasks discussed here.
so that the graph adapts as the representation changes and captures local shape \cite{wang2019dgcnn}. The same idea scales to detection, where a graph built on raw laser points supports three-dimensional object detection for autonomous driving \cite{shi2020pointgnn}, a task that connects directly to the transportation chapter. Depth has also been the setting for adapting the residual and dense connections that let convolutional networks grow deep, transferring those tricks to graph networks so they can be stacked far beyond the usual handful of layers without collapsing \cite{li2019deepgcns}, a direct response to the over-smoothing the methods chapter described. Point-cloud tasks span classification, part and scene segmentation, detection, and registration, and in each the graph supplies the geometric locality that a permutation-invariant set model alone omits. The design space here mirrors the broader one. A point cloud can be voxelized onto a grid and handed to a three-dimensional convolution, processed as a raw set with shared per-point networks, or treated as a graph on spatial neighbours, and the three trade memory, resolution, and geometric fidelity differently, with the graph approach preserving locality without the memory cost of a dense voxel grid. Hierarchical variants build the neighbourhood graph at several scales so a model captures both fine surface detail and coarse shape, and continuous-kernel methods generalize the convolution itself to operate directly on point coordinates. For autonomous driving and robotics the dominant tasks are segmentation, which labels every point, and detection, which localizes objects in three dimensions, both supported directly by the geometric locality of a graph. The point-cloud setting is a clean illustration of when a graph is the right representation and when it is merely available. A point cloud has no inherent connectivity, only positions in space, so the graph is constructed by joining nearby points, and the construction works because proximity in space genuinely corresponds to relevance for the local geometry the task cares about. This is constructed-graph reasoning of the kind the survey returns to, and its success here, where the constructed edges track a real geometric relationship, contrasts with the constructed graphs of other domains, where the edges encode a similarity whose relevance is assumed rather than evident, a contrast that helps explain why graph methods are on firmer ground for point clouds than for some of the similarity graphs built elsewhere.

\subsection{Relational reasoning and few-shot learning}
\label{subsec:vision-relational}

Two further uses of graphs in vision concern reasoning rather than perception. Relational reasoning models the relationships between detected objects or image regions, which is what questions about a scene often turn on, and a graph over regions lets a model aggregate the context a single region lacks. Few-shot learning has a particularly clean graph formulation. Given a few labelled examples and a query to classify, one can build a small graph whose nodes are all the examples and whose edges encode similarity, then propagate label information from the labelled support to the unlabelled query, which casts few-shot classification as transductive node classification on an episode graph \cite{garcia2018fewshot}. A refinement labels the edges rather than the nodes, learning to predict whether two examples belong to the same class and assembling the classification from those pairwise decisions \cite{liu2019egnn}. The appeal in both cases is that the relationships among examples carry information that treating each example independently discards, which is the same argument that recurs throughout the survey, here applied to the structure of a learning episode. Visual question answering is the relational task that has drawn the most attention, because a question such as whether one object lies to the left of another is answered not by recognizing objects in isolation but by reasoning over their arrangement, and a graph over detected objects gives a model the relational substrate to do so. The few-shot formulation also clarifies a connection to metric learning, since predicting whether two examples share a class is learning a similarity, and the graph turns a collection of such pairwise similarities into a coherent labelling rather than a set of independent comparisons. Both uses are transductive, exploiting the test-time structure of the episode or the scene, which is where graph methods hold a particular advantage. The few-shot setting illustrates the point with unusual clarity. When only a handful of labelled examples are available, the question is how a new instance relates to them, and casting the examples and the query as nodes in a graph whose edges encode similarity turns classification into a problem of propagating labels across that graph. This is a natural fit for graph methods, since the relational structure among the few examples is exactly the information a few-shot learner must exploit, and the graph makes that structure explicit rather than leaving it implicit in a distance computation. The advantage is real but bounded, since when examples are plentiful the relational framing adds less, which is consistent with the survey's broader finding that graph methods help most where relational structure carries signal that abundant data would otherwise supply.

\subsection{Video and skeleton-based action recognition}
\label{subsec:vision-action}

Human action recognition from skeletons is the clearest spatio-temporal graph problem in vision. A pose is a set of body joints with a natural skeleton connecting them, so a node is a joint, an edge is a bone, and stacking poses over time adds temporal edges between the same joint in consecutive frames, producing a spatio-temporal graph. A graph convolution over the skeleton combined with a temporal convolution along the time edges recognizes actions from the evolving pose \cite{yan2018stgcn}, an approach that ties this section to the temporal architectures of the methods chapter and to the traffic forecasting of the next. Video understanding more broadly can be cast as reasoning over a graph of region proposals across space and time, linking the same object through a clip so that its trajectory and interactions inform the prediction, though this line is less consolidated than the skeleton case. Even within the skeleton setting the modelling choices matter, since a fixed skeleton graph encodes anatomical connectivity but not the longer-range coordination between distant joints that some actions require, and methods that learn additional edges or attend across the whole skeleton recover that coordination at the cost of the clean structural prior. The temporal side admits the same range of choices the methods chapter described, from simple convolutions along the time edges to recurrent and attention-based readers of the joint trajectories.

\subsection{Strengths, weaknesses, and open problems}
\label{subsec:vision-tradeoffs}

\Cref{tab:cvapps} collects these applications with their graph constructions. The strength of graph methods in vision is specific and genuine. Where the data is irregular geometry, as in a point cloud, a graph on spatial neighbours is the right structure and outperforms forcing the data onto a grid; where the output is relational, as in a scene graph, a graph is the output, and message passing improves it by letting objects and relations constrain each other. In these niches graphs are not a stylistic choice but a fit to the problem.

The weaknesses are equally specific and worth stating plainly. In most of vision the graph is a component rather than the system, and the heavy lifting is done by a convolutional or transformer backbone that extracts features, with the graph network adding a relational refinement on top; the contribution is real but bounded, and it is easy to overstate by reporting the full system's accuracy as if the graph produced it. Scene-graph generation is bottlenecked by object detection, since a relationship cannot be predicted between objects that were not detected, and it is biased by the long-tailed statistics of the training data, so that common relations are predicted well and rare ones poorly. The deepest tension is that the transformer's attention is itself a learned, fully connected graph, and for many vision tasks it has absorbed the relational modelling that once required an explicit graph network, which narrows the remaining advantage of explicit graphs to settings where the structure is irregular geometry or an explicitly structured output rather than implicit relations a transformer can learn on its own.

\begin{table}[t]
\centering
\caption[Computer-vision applications of graph neural networks.]{Vision applications, with the graph each builds and the role the graph plays. A representative method is named for each. The list spans scene understanding, visual reasoning, geometry, and skeleton-based action recognition.}
\label{tab:cvapps}
\small
\begin{tabularx}{\textwidth}{@{}lXll@{}}
\toprule
Task & Graph construction & Role of the graph & Example \\
\midrule
Scene graph generation       & objects / candidate relations        & relational refinement     & Graph R-CNN \cite{yang2018graphrcnn} \\
Scene graph (message passing) & objects / candidate relations       & iterative refinement      & \cite{xu2017sgg} \\
Visual question answering     & objects / spatial-semantic relations & relational reasoning      & \cite{teney2017graphvqa,wang2023vqagnn} \\
Point-cloud analysis         & points / spatial neighbours          & geometric aggregation     & DGCNN \cite{wang2019dgcnn} \\
3D object detection          & laser points / proximity             & detection over geometry   & Point-GNN \cite{shi2020pointgnn} \\
3D shape analysis            & mesh vertices / faces                & surface convolution       & FeaStNet \cite{verma2018feastnet} \\
Few-shot classification      & support and query / similarity       & label propagation         & \cite{garcia2018fewshot} \\
Zero-shot classification     & classes / knowledge graph            & semantic propagation      & \cite{wang2018zsl,kampffmeyer2019dgp} \\
Action recognition           & joints / bones and time              & spatio-temporal modelling & ST-GCN \cite{yan2018stgcn} \\
Action (adaptive topology)   & joints / learned bones               & adaptive aggregation      & 2s-AGCN \cite{shi2019asgcn} \\
Action (refined topology)    & joints / channel-wise topology       & topology refinement       & CTR-GCN \cite{chen2021ctrgcn} \\
Action (unified, efficient)  & joints / multi-scale, shift          & efficient modelling       & \cite{liu2020msg3d,cheng2020shiftgcn} \\
\bottomrule
\end{tabularx}
\end{table}

\FloatBarrier
\section{Transportation and traffic forecasting}
\label{sec:traffic}

Transportation is the domain where space and time meet most cleanly on a graph. A road network is a graph in the most literal sense, and the traffic moving over it is a signal that lives on the nodes and changes from one moment to the next, so a transportation problem is almost always a problem of forecasting a graph signal forward in time. This is the setting in which spatio-temporal graph networks were developed, and it is one of the clearest cases of practical value in the survey, since the forecasts feed navigation, arrival-time estimation, and traffic management that millions of people rely on. The defining difficulty is that the prediction depends jointly on two kinds of structure, the spatial dependence encoded by the network, along which congestion propagates, and the temporal dependence of the signal, with its daily and weekly rhythms, and a method must model both at once.

\subsection{Roads, sensors, and the spatio-temporal graph}
\label{subsec:traffic-graph}

\begin{figure}[t]
\centering
\resizebox{0.82\textwidth}{!}{%
\begin{tikzpicture}[
  sensor/.style={circle, draw=gnnocean!80!black, fill=tintocean, line width=0.9pt,
                 minimum size=9mm, font=\scriptsize\bfseries, text=gnnocean},
  road/.style={draw=gnnslate, line width=1.7pt}]
\node[sensor] (a) at (0,2.2)   {$v_1$};
\node[sensor] (b) at (2.2,2.5) {$v_2$};
\node[sensor] (c) at (4.4,2.2) {$v_3$};
\node[sensor] (d) at (0.5,0.4) {$v_4$};
\node[sensor] (e) at (2.4,0.7) {$v_5$};
\node[sensor] (f) at (4.5,0.3) {$v_6$};
\draw[road] (a)--(b) (b)--(c) (a)--(d) (b)--(e) (c)--(f) (d)--(e) (e)--(f);
\pic[scale=0.7] at ($(a)!0.5!(b)$) {ic car={gnnred}};
\pic[scale=0.7] at ($(e)!0.55!(f)$) {ic car={gnnamber}};
% time-series inset
\begin{scope}[shift={(6.3,0.7)}]
  \draw[->,gnngray,line width=0.7pt] (0,0)--(2.0,0) node[right,font=\scriptsize]{$t$};
  \draw[->,gnngray,line width=0.7pt] (0,-0.05)--(0,1.35);
  \draw[gnnblue,line width=1.3pt] (0,1.0) sin (0.45,0.5) cos (0.9,0.2) sin (1.4,0.7) cos (1.85,1.05);
  \node[font=\scriptsize,text=gnnblue] at (0.95,-0.42) {speed at $v_5$};
\end{scope}
\draw[gnnocean,dashed,-{Stealth[length=2.4mm]},line width=0.9pt] (e) to[bend left=12] (6.2,0.7);
\end{tikzpicture}}
\caption[A road network as a spatio-temporal graph.]{A road network as a graph: sensors or road segments are nodes (teal) and road connectivity gives the edges, while each node carries a time series such as speed or flow (inset). Forecasting predicts the next values at every node from recent history, using both the spatial structure of the network and the temporal pattern of the signal.}
\label{fig:roadnet}
\end{figure}
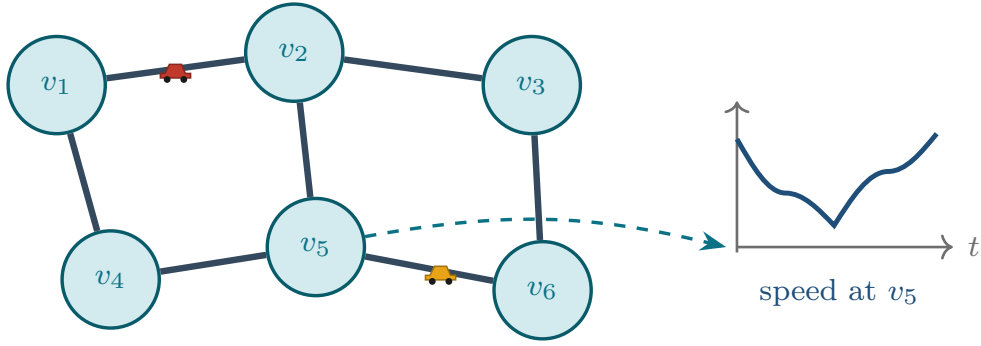

The standard construction places a node at each sensor, road segment, or intersection and an edge wherever the network connects two of them, weighting edges by road distance or adjacency. \Cref{fig:roadnet} shows the form, with a time series of speed, flow, or occupancy attached to every node. The data is therefore a sequence of graph signals, and the forecasting task is to predict the next several steps at every node from a window of recent history,
\begin{equation}\label{eq:stforecast}
\widehat{\Feat}^{(t+1:t+H)}=f\big(\Feat^{(t-T+1:t)};\graph\big),
\end{equation}
where $\Feat^{(t-T+1:t)}$ stacks the last $T$ observations on the graph and $H$ is the forecast horizon. The reason a graph helps is concrete: congestion does not appear at random but spreads along roads from where it starts, so a jam at one sensor raises the near-future readings at its downstream neighbours, and a model that knows the network can anticipate this propagation that a per-sensor time-series model cannot. The graph is built in more than one way, and the choice matters. Nodes can be sensors, fixed-length road segments, or whole intersections, and edges can be set to the binary road adjacency or weighted by a decaying function of road distance with a threshold that drops the weakest connections, a construction that injects a prior about how far influence reaches. Direction is a further choice, since traffic on a one-way road propagates downstream but not up, and a directed graph captures this asymmetry that an undirected one discards. None of these decisions is forced by the data, and each shapes what the model can learn before training begins, which is one reason the field eventually turned to learning the graph rather than fixing it.

\subsection{Tasks}
\label{subsec:traffic-tasks}

Forecasting traffic speed, flow, or travel time is the dominant task and the one most benchmarks measure, but the domain is wider. Demand prediction estimates how many trips will originate in each region of a city, the problem ride-hailing and bike-sharing services must solve to position vehicles ahead of need. Trajectory prediction forecasts the future paths of pedestrians or vehicles in an interactive scene, where each agent's motion depends on the others and the set of agents is itself a small graph. This view has produced a distinct line of trajectory and motion-forecasting models in which agents are nodes and edges encode social or physical influence, from pedestrian prediction with social spatio-temporal convolutions \cite{mohamed2020socialstgcnn} to probabilistic multi-agent models \cite{ivanovic2019trajectron}, and, in autonomous driving, vectorized encodings of agents and map elements \cite{gao2020vectornet} and lane-graph representations built for motion forecasting \cite{liang2020lanegcn}. Origin-destination estimation has likewise been cast as a graph problem, predicting the matrix of flows between regions from a multi-perspective view of the network \cite{shi2020odflow}, and heterogeneous dynamic graphs have been used for finer-grained urban prediction such as crime forecasting \cite{zhao2024hdmgnn}. Arrival-time estimation predicts how long a specific route will take, combining forecasts along its segments. Beyond prediction, graphs support traffic-signal control, incident and anomaly detection, and the estimation of origin-destination flows that summarize how a population moves. What unites these is a graph that the transportation system already defines and a quantity that evolves on it, which is the spatio-temporal pattern graph networks were built to model. Several of these tasks reward a closer look. Trajectory prediction is distinctive because the graph is not a road network but the transient set of agents in a scene, with edges encoding which agents can influence each other, and the prediction must respect that people and vehicles react to one another rather than moving independently. Origin-destination estimation predicts a matrix of flows between every pair of regions, a structured output far larger than a per-node forecast and one that should respect conservation constraints. Traffic-signal control is a sequential decision problem rather than a prediction, usually posed as reinforcement learning over a graph of intersections, where a controller must account for how its choices at one junction ripple to the next. The multi-step nature of forecasting cuts across all of these, since predicting one step ahead is far easier than predicting an hour ahead, and error accumulates over the horizon in ways that separate the methods.

\subsection{Spatio-temporal architectures}
\label{subsec:traffic-arch}

\begin{figure}[tb]
\centering
\resizebox{\textwidth}{!}{%
\begin{tikzpicture}[
  font=\sffamily, >=Stealth,
  stage/.style={rounded corners=4pt, draw=#1!75!black, fill=#1!12, line width=0.8pt,
                text=gnnink, align=center, inner sep=6pt},
  block/.style={rounded corners=4pt, draw=gnnocean!80!black, fill=tintocean, line width=0.9pt,
                align=center, inner sep=5pt, minimum width=30mm, minimum height=30mm},
  sub/.style={rounded corners=3pt, draw=#1!80!black, fill=#1!16, line width=0.7pt,
              align=center, font=\scriptsize, text=gnnink, minimum width=26mm, minimum height=10mm},
  snode/.style={circle, draw=#1!80!black, fill=#1!70, inner sep=0pt, minimum size=3.6mm},
  flow/.style={-{Stealth[length=3mm]}, line width=1.1pt, draw=gnnslate},
  clab/.style={font=\scriptsize\itshape, text=gnngray, align=center}]

% ---- Stage 1: sensor network + signals ----
\node[stage=gnnblue, minimum width=30mm, minimum height=30mm] (sensors) at (0,0) {};
\node[clab, anchor=south] at (sensors.north) {Sensor network + signals};
\begin{scope}[shift={(-1.0,0.45)}, scale=0.9]
  \node[snode=gnnblue]  (s1) at (0,0)      {};
  \node[snode=gnnteal]  (s2) at (1.0,0.35) {};
  \node[snode=gnngreen] (s3) at (1.4,-0.6) {};
  \node[snode=gnnamber] (s4) at (0.2,-0.85){};
  \draw[gnnslate,line width=0.7pt] (s1)--(s2) (s1)--(s4) (s2)--(s3) (s3)--(s4);
  \pic[scale=0.34] at (s2.center){ic sensor={gnnteal}};
\end{scope}
% small time series
\begin{scope}[shift={(-1.15,-1.35)}, scale=0.42]
  \draw[gnngray,line width=0.4pt] (0,0)--(3.4,0);
  \draw[gnnblue,line width=0.8pt] (0,0.4) sin (0.85,0.8) cos (1.7,0.4) sin (2.55,0.0) cos (3.4,0.4);
  \draw[gnnamber,line width=0.8pt] (0,0.1) cos (0.85,0.45) sin (1.7,0.15) cos (2.55,0.5) sin (3.4,0.2);
\end{scope}
\node[clab, anchor=north, text width=30mm] at (sensors.south) {$T$ past steps per node, $\Feat_t$};

% ---- Stage 2: graph construction ----
\node[stage=gnnteal, minimum width=22mm, minimum height=24mm, right=9mm of sensors] (build)
  {\scriptsize build graph\\[2pt]$\Adj$ from\\\scriptsize proximity\\\scriptsize or similarity};
\node[clab, anchor=south] at (build.north) {Graph construction};

% ---- Stage 3: stacked ST blocks ----
\node[block, right=9mm of build] (b1) {};
\node[clab, anchor=south] at (b1.north) {ST block 1};
\node[sub=gnnteal]  (sp1) at ($(b1.center)+(0,0.62)$) {Spatial graph conv\\(mix neighbours)};
\node[sub=gnnpurple](tp1) at ($(b1.center)+(0,-0.62)$) {Temporal module\\(GRU / 1D conv)};
\draw[-{Stealth[length=2mm]},line width=0.8pt,draw=gnnocean!75!black] (sp1)--(tp1);

\node[block, right=7mm of b1] (b2) {};
\node[clab, anchor=south] at (b2.north) {ST block 2};
\node[sub=gnnteal]  (sp2) at ($(b2.center)+(0,0.62)$) {Spatial graph conv};
\node[sub=gnnpurple](tp2) at ($(b2.center)+(0,-0.62)$) {Temporal module};
\draw[-{Stealth[length=2mm]},line width=0.8pt,draw=gnnocean!75!black] (sp2)--(tp2);
\node[clab, anchor=north, text width=68mm] at ($(b1.south)!0.5!(b2.south)$)
  {each block interleaves a spatial pass over $\graph$ with a temporal pass over time};

% ---- Stage 4: forecast ----
\node[stage=gnngreen, minimum width=26mm, minimum height=28mm, right=9mm of b2] (fc) {};
\node[clab, anchor=south] at (fc.north) {Forecast horizon $H$};
\begin{scope}[shift={($(fc.center)+(-0.95,0.15)$)}, scale=0.5]
  \draw[gnngray,line width=0.4pt] (0,-0.8)--(0,0.9) (0,0)--(3.2,0);
  \draw[gnnblue,line width=0.9pt] (0,0.3) cos (0.8,0.7) sin (1.6,0.35) cos (2.4,0.75) sin (3.2,0.4);
  \draw[gnngreen,line width=1.0pt,densely dashed] (1.6,0.35) cos (2.4,0.62) sin (3.2,0.3);
  \node[font=\tiny, text=gnngray] at (2.4,-0.5) {$t{+}1..t{+}H$};
\end{scope}
\node[clab, anchor=north, text width=26mm] at (fc.south) {$\hat{\Feat}_{t+1:t+H}$};

\draw[flow] (sensors)--(build);
\draw[flow] (build)--(b1);
\draw[flow] (b1)--(b2);
\draw[flow] (b2)--(fc);
\end{tikzpicture}}
\caption[End-to-end spatio-temporal graph neural network architecture.]{The end-to-end architecture that recurs in traffic and other spatio-temporal forecasting. Signals observed at the nodes of a sensor network over $T$ past steps are arranged on a graph built from proximity or similarity, then passed through a stack of spatio-temporal blocks. Each block interleaves a spatial pass, a graph convolution that mixes neighbouring sensors, with a temporal pass, a recurrent or one-dimensional convolutional module that models the time axis. The final block feeds a head that predicts the next $H$ steps at every node. The spatial and temporal operators vary across models, but the interleaved structure is shared.}
\label{fig:e2estgnn}
\end{figure}
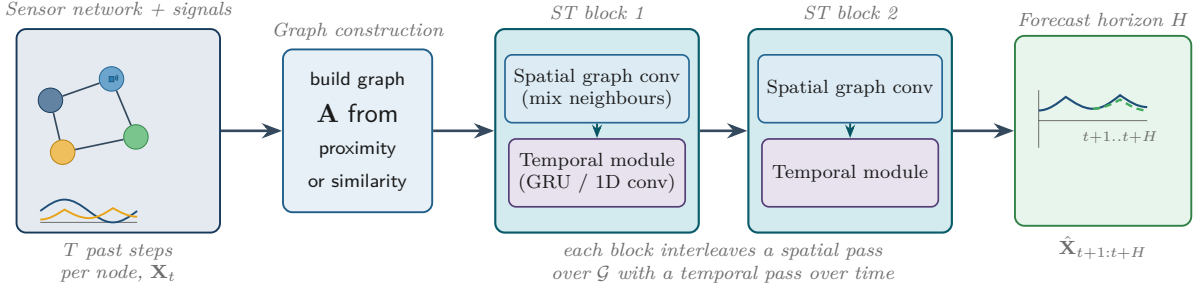

Every architecture in this area answers two questions, how to model spatial dependence and how to model temporal dependence, and then how to interleave the two. \Cref{fig:e2estgnn} shows the end-to-end architecture that results, a stack of blocks that alternate a spatial graph convolution with a temporal module before a forecasting head. The spatial side is some form of graph convolution or graph attention; the temporal side is a recurrent network, a temporal convolution, or attention over time; and a spatio-temporal layer composes them, alternating propagation over the graph with propagation over time,
\begin{equation}\label{eq:stlayer}
\Hid^{(l+1)}=\mathrm{TConv}\!\big(\mathrm{GConv}(\Hid^{(l)};\Adj)\big),
\end{equation} The forecasting objective and the spatio-temporal layer, \cref{eq:stforecast,eq:stlayer}, define the template shared by the traffic models above.
where the order and exact form vary by method. The temporal choice carries real trade-offs. A recurrent network processes the sequence step by step and can accumulate error over a long horizon, a temporal convolution sees a fixed window in parallel and trains faster but must widen its receptive field through dilation to reach far into the past, and attention over time can in principle connect any two moments at a computational cost that grows with the window. The spatial choice carries the analogous trade-off between a fixed propagation rule and a learned or attention-based one. A further refinement makes the graph itself change over time, on the reasoning that the relevant dependencies differ between free flow and congestion, so a model that adapts its graph to the current regime can capture what a static graph cannot. Some methods also decompose the signal into a smooth trend and a periodic component before modelling, exploiting the strong daily and weekly cycles traffic exhibits. A landmark model treated propagation as diffusion, a random walk over the road graph, and paired it with a recurrent network over time, casting forecasting as sequence-to-sequence prediction on a graph \cite{li2018dcrnn}. The diffusion view is worth making explicit, because it models traffic as spreading both with and against the direction of flow, capturing that a jam influences the road behind it as well as the road ahead, and it grounds the spatial operation in the random-walk picture the foundations chapter introduced rather than in a spectral filter. A contemporaneous design replaced the recurrence with gated temporal convolutions, keeping the whole model convolutional and therefore faster to train while stacking spatial and temporal convolutions in blocks \cite{yu2018stgcn}. An influential idea followed from a simple observation, that the road map is an incomplete description of spatial dependence, since sensors can be statistically coupled without being adjacent and adjacent without being coupled. Learning an adaptive adjacency matrix from data, rather than fixing it to the road graph, lets the model discover the dependencies the map omits,
\begin{equation}\label{eq:adaptadj}
\Atil_{\text{adp}}=\softmax\!\big(\relu(\mathbf{E}_1\mathbf{E}_2^{\top})\big),
\end{equation}
with $\mathbf{E}_1$ and $\mathbf{E}_2$ learned node embeddings, and pairing this learned graph with dilated temporal convolutions produced a model that improved markedly on its predecessors \cite{wu2019graphwavenet}. Attention entered on both axes, weighting which other locations and which past times matter for a given prediction \cite{guo2019astgcn}, and an encoder-decoder built entirely from spatial and temporal attention pushed the idea further \cite{zheng2020gman}. The learned-graph principle also connects to the broader treatment of multivariate time series, where the same machinery learns a graph among arbitrary correlated series rather than among roads \cite{wu2020mtgnn}. \Cref{tab:trafficmethods} sets these methods side by side, and \cref{alg:traffic} states the shared forecasting procedure.

\begin{table}[t]
\centering
\caption[Spatio-temporal methods for traffic forecasting.]{Representative spatio-temporal methods, by how each models space and time and whether it learns the graph rather than fixing it to the road network.}
\label{tab:trafficmethods}
\small
\begin{tabularx}{\textwidth}{@{}lXll@{}}
\toprule
Method & Spatial module & Temporal module & Learned graph \\
\midrule
DCRNN \cite{li2018dcrnn}                & diffusion convolution    & recurrent (GRU)            & no  \\
STGCN \cite{yu2018stgcn}                & graph convolution        & gated temporal convolution & no \\
Graph WaveNet \cite{wu2019graphwavenet} & graph convolution        & dilated convolution        & yes \\
ASTGCN \cite{guo2019astgcn}             & spatial attention        & temporal attention         & no  \\
GMAN \cite{zheng2020gman}               & spatial attention        & temporal attention         & no  \\
STSGCN \cite{song2020stsgcn}            & synchronous localized    & synchronous localized      & no  \\
AGCRN \cite{bai2020agcrn}               & node-adaptive convolution & recurrent (GRU)           & yes \\
MTGNN \cite{wu2020mtgnn}                & graph convolution        & dilated convolution        & yes \\
GTS \cite{shang2021gts}                 & graph convolution        & recurrent                  & yes \\
STGODE \cite{fang2021stgode}            & graph ODE                & continuous dynamics        & no  \\
ST-MGCN \cite{geng2019stmgcn}           & multi-graph convolution  & recurrent                  & no  \\
\bottomrule
\end{tabularx}
\end{table}

\begin{algorithm}[t]
\caption{Spatio-temporal traffic forecasting}
\label{alg:traffic}
\KwIn{road graph $\graph$, history $\Feat^{(t-T+1:t)}$ of node signals, horizon $H$}
\KwOut{forecast $\widehat{\Feat}^{(t+1:t+H)}$}
\textbf{optional:} learn an adaptive adjacency $\Atil_{\text{adp}}$ by \cref{eq:adaptadj}\;
$\Hid \leftarrow$ encode the history $\Feat^{(t-T+1:t)}$\;
\For{layer $l=0$ to $L-1$}{
  $\Hid \leftarrow$ spatial graph convolution over $\graph$ and $\Atil_{\text{adp}}$\;
  $\Hid \leftarrow$ temporal convolution or recurrence along time\;
}
$\widehat{\Feat}^{(t+1:t+H)} \leftarrow$ decode $\Hid$\;
\Return{forecast $\widehat{\Feat}^{(t+1:t+H)}$}
\end{algorithm}

\subsection{Demand, ride-hailing, and urban mobility}
\label{subsec:traffic-demand}

A second family of problems forecasts demand rather than flow, and it changes the graph. Instead of road sensors, the nodes are regions of a city, and several graphs can connect them at once, one by spatial adjacency, another by similarity of function, another by transport connectivity, so that two distant districts with similar activity can inform each other's demand. A multi-graph, multi-modal approach of this kind predicts ride-hailing demand by combining these complementary views of how regions relate \cite{ke2021multimodal}. The broader study of urban mobility takes the same step from roads to regions and from flow to movement, modelling how a population travels across a city and how those patterns shift, work that treats mobility itself as a graph signal to be learned and forecast \cite{bahi2025urbanmobility}. The distinction between a road graph and a region graph is more than cosmetic, because the two encode different notions of proximity, and a method tuned for one does not transfer unchanged to the other. The demand setting raises problems the forecasting setting does not. A newly added region has no history, a cold-start the region graph can partly cover by borrowing from similar districts, and the demand signal is sparser and burstier than highway flow, which stresses models tuned on smooth speed series. Bike-sharing adds a rebalancing problem, since predicting demand is only useful if vehicles can be moved to meet it. Human-mobility studies draw on movement traces and check-in data that make the relevant graphs observable at the scale of a whole city, and they inherit a privacy concern highway sensors avoid, because the data that reveals how a population moves also reveals where individuals go, a tension that recurs wherever fine-grained mobility is modelled. It is worth stepping back to note how general this spatio-temporal template is. The same combination of a spatial graph and a temporal model recurs in forecasting disease spread over a contact network, electricity demand over a grid, and weather over a sensor field, all of which the survey touches elsewhere, and the architectures developed for traffic transfer to those settings with little change. Traffic is the domain where the template was sharpened, but the template itself is the durable contribution, and recognizing a new problem as spatio-temporal forecasting on a graph is often most of the work of solving it.

\subsection{Datasets and the learned-graph question}
\label{subsec:traffic-data}

\Cref{tab:trafficdatasets} lists the datasets that recur in the area, a small set of sensor networks from a few metropolitan regions together with the region-level demand data used for mobility studies. Their narrowness is itself a methodological concern, since a field that compares almost everything on the same two or three highway sensor networks risks tuning to those networks rather than to traffic in general. The datasets also motivate the question that runs through the domain. The most effective methods learn the graph rather than taking it from the road map, which is an admission that the obvious graph, the road network, is not the graph the prediction needs, and it raises a sharper question about how much the spatial structure contributes at all. Careful studies have found that strong purely temporal baselines, and even a well-constructed historical average that simply returns the typical value for the time of day and day of week, are surprisingly hard to beat, and that the margin elaborate spatio-temporal architectures add over them is often smaller than the headline comparisons suggest. The spatial prior is real and useful, but its contribution is sometimes overstated, and an honest account of the domain separates the gain from modelling the network from the gain from modelling time well. The evaluation protocols compound the difficulty. Reported results in the area have varied with preprocessing, with the exact split, and with the horizon at which error is measured, and comparisons that hold these fixed sometimes shrink the gaps between methods considerably. The lesson is not that spatio-temporal modelling fails, since the deployed systems clearly work, but that the benchmark culture here, like that in recommendation, has at times rewarded architectural elaboration over honest comparison against simple, strong baselines.

\begin{table}[t]
\centering
\caption[Datasets for traffic and mobility forecasting.]{Datasets that recur in traffic and mobility forecasting, described by signal and relative scale. Exact sensor counts are omitted, since they vary by release.}
\label{tab:trafficdatasets}
\small
\begin{tabularx}{\textwidth}{@{}lllX@{}}
\toprule
Dataset & Signal & Scale & Note \\
\midrule
METR-LA \cite{li2018dcrnn}  & highway speed             & hundreds of sensors & Los Angeles loop detectors \\
PEMS-BAY \cite{li2018dcrnn} & highway speed             & hundreds of sensors & Bay Area loop detectors \\
PEMS04, PEMS08             & flow, speed, occupancy    & hundreds of sensors & California freeway sensors \\
Ride-hailing zones \cite{ke2021multimodal} & demand per region & city regions & multi-modal demand data \\
\bottomrule
\end{tabularx}
\end{table}

\subsection{Strengths, weaknesses, and open problems}
\label{subsec:traffic-tradeoffs}

The strengths here are among the most concrete in the survey. A road network is a genuine graph, congestion genuinely propagates along it, and spatio-temporal graph networks capture that propagation in a way that improves forecasts people use, in navigation and arrival-time estimation deployed at scale. The learned-graph idea is a real contribution that extends beyond traffic, and the encoder-decoder attention models give accurate multi-step forecasts that earlier methods could not.

The weaknesses temper this without erasing it. The predefined graph is incomplete, which is why the best methods learn their own, and the baseline question is unavoidable, since a well-tuned temporal model captures much of the predictable structure on its own and the spatial gain, though real, is often modest and sometimes overstated. The benchmarks are few and regionally narrow, so reported progress may reflect adaptation to a handful of sensor networks rather than a general advance. And the models are trained on normal conditions, which means they forecast the recurring patterns well and the non-recurring events that matter most, an incident, a storm, a closure, a large gathering, poorly, precisely because those events are rare in the training data and break the regularities the model learned. Scale is a further practical limit. A city-wide network has tens of thousands of nodes, and a model that must hold the whole graph and a window of history in memory while producing forecasts fast enough to be useful faces a computational burden that favours the simplest spatial and temporal modules and penalizes the all-to-all attention that performs best on small benchmarks. The tension between accuracy on a few hundred sensors and tractability on a whole city is one the deployed systems resolve toward tractability. A related limit is that traffic patterns drift, as road works, new developments, and changing travel habits alter the regularities a model learned, so a forecaster trained once and left in place degrades, and keeping it current requires online updating that most benchmark evaluations, which train and test on a fixed period, never exercise.

The open problems follow from these gaps. Forecasting under non-recurring events, transfer across cities so a model trained in one place is useful in another, the integration of exogenous signals such as weather and scheduled events, uncertainty estimates that a traffic manager can act on, and the prospect of foundation models for spatio-temporal data are the active directions. A fair summary is that transportation is one of the domains where graph methods fit the data most naturally and deliver the clearest practical value, and also one where an honest accounting must separate the genuine contribution of the spatial graph from the strong performance a good temporal model achieves on its own.

\FloatBarrier
\section{Power systems and renewable energy}
\label{sec:power}

The electrical grid is a graph that carries physics. Substations, generators, and loads sit at its nodes, transmission and distribution lines form its edges, and the quantities that flow over it obey laws as exact as any in engineering. This combination makes power systems a distinctive domain for graph learning, because the graph is not a convenient abstraction laid over messy data but a faithful model of a physical network whose behaviour is governed by known equations. The domain has also acquired new urgency, as the integration of variable renewable generation, the spread of distributed resources, and the frequent reconfiguration of the network stress methods built for a stable, centrally generated grid. Graph networks have been applied both to the operation of the grid and to the forecasting of the renewable generation now feeding it, and this section treats each, with attention to a question sharper here than elsewhere: what a learned model can offer in a domain where the governing physics is already known and the cost of error is a blackout.

\subsection{The grid as a graph}
\label{subsec:power-graph}

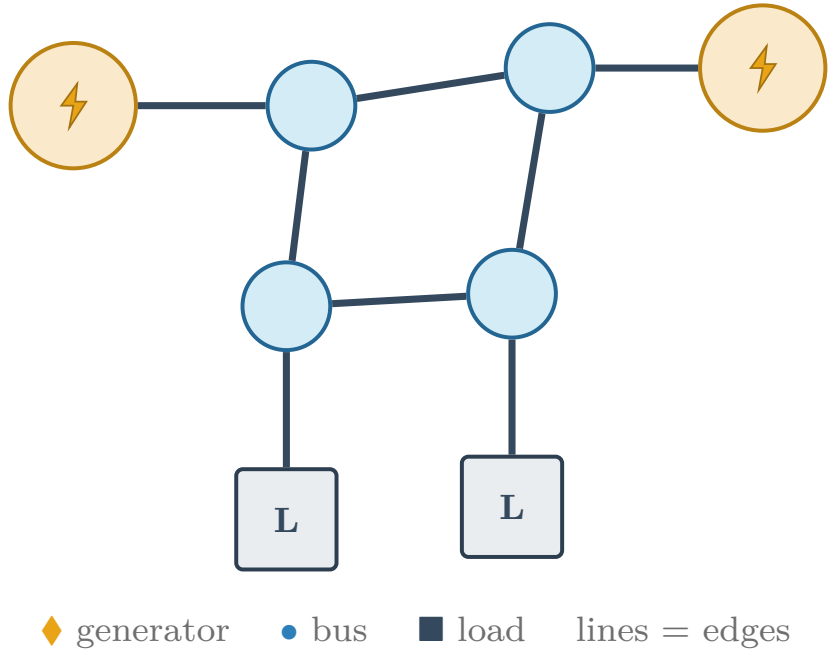
\begin{figure}[t]
\centering
\resizebox{0.68\textwidth}{!}{%
\begin{tikzpicture}[
  gen/.style={circle, draw=gnnamber!80!black, fill=tintamber, line width=0.9pt, minimum size=10mm},
  bus/.style={circle, draw=gnnteal!80!black, fill=tintteal, line width=0.9pt, minimum size=7mm},
  load/.style={rectangle, rounded corners=1.5pt, draw=gnnslate!85!black, fill=tintgray,
               line width=0.9pt, minimum size=8mm, font=\scriptsize\bfseries, text=gnnslate},
  pline/.style={draw=gnnslate, line width=1.6pt}]
\node[gen]  (g1) at (0,2.2)    {};
\node[bus]  (b1) at (1.9,2.2)  {};
\node[bus]  (b2) at (3.8,2.5)  {};
\node[bus]  (b3) at (3.5,0.7)  {};
\node[bus]  (b4) at (1.7,0.6)  {};
\node[gen]  (g2) at (5.5,2.5)  {};
\node[load] (l1) at (1.7,-1.1) {L};
\node[load] (l2) at (3.5,-1.0) {L};
\draw[pline] (g1)--(b1) (b1)--(b2) (b2)--(g2) (b1)--(b4) (b2)--(b3) (b3)--(b4) (b4)--(l1) (b3)--(l2);
\pic[scale=0.8] at (g1) {ic bolt={gnnamber}};
\pic[scale=0.8] at (g2) {ic bolt={gnnamber}};
\node[anchor=west,font=\scriptsize,text=gnngray] at (-0.4,-2.0)
  {\textcolor{gnnamber}{$\blacklozenge$}~generator \quad
   \textcolor{gnnteal}{$\bullet$}~bus \quad
   \textcolor{gnnslate}{$\blacksquare$}~load \quad lines $=$ edges};
\end{tikzpicture}}
\caption[A power grid as a graph.]{A power grid as a graph: generators (amber, marked with a bolt), buses (teal), and loads (grey) are nodes and transmission lines are edges, with node quantities such as voltage and power injection and edge quantities such as line impedance. Unlike a data-only graph, the grid obeys known physical laws, which both constrain and inform the models built on it.}
\label{fig:smartgrid}
\end{figure}

\Cref{fig:smartgrid} shows the basic construction, with generators, buses, and loads as nodes and lines as edges, each node carrying quantities such as voltage magnitude and power injection and each edge carrying an impedance. What sets this graph apart is that the relationship between these quantities is fixed by physical law rather than learned from data. The power injected at a bus is determined by the voltages across the network through the power-flow equations, which for the active power at bus $i$ take the form
\begin{equation}\label{eq:powerflow}
P_i=\sum_{j}|V_i|\,|V_j|\big(G_{ij}\cos\theta_{ij}+B_{ij}\sin\theta_{ij}\big),
\end{equation}
with $G_{ij}$ and $B_{ij}$ the conductance and susceptance of the line between buses and $\theta_{ij}$ the difference in their voltage angles. These equations are nonlinear and coupled across the whole network, and solving them is the core computation of grid operation. The grid is also dynamic in a structural sense, since switching, maintenance, and faults change which lines are in service, so the graph itself varies and a model that assumes a fixed topology is brittle in exactly the situations operators care about most. Two distinctions refine the picture. The transmission grid that carries bulk power over long distances and the distribution grid that delivers it locally differ in scale, in how meshed they are, and in how well observed they are, and a method suited to one does not automatically suit the other. The full alternating-current power flow of \cref{eq:powerflow} is sometimes replaced by a linearized direct-current approximation that is faster but less accurate, a recurring trade-off a learned surrogate inherits. Observability matters too, since the state is inferred from sensor measurements of varying quality, from the slow telemetry of supervisory control systems to the fast, precise readings of phasor measurement units, and state estimation, the recovery of the true network state from noisy and incomplete measurements, is where many grid applications begin.

\subsection{Tasks}
\label{subsec:power-tasks}

The tasks divide into operating the existing grid and forecasting what flows into it. Operating tasks include solving the power flow to find the network state from a set of injections, estimating that state from noisy measurements, optimizing generator dispatch to meet demand at least cost subject to the physical constraints, detecting and locating faults, assessing the grid's stability and its reliability under the loss of components, and controlling voltage and power in real time. Forecasting tasks predict the demand the grid must serve and, increasingly, the output of the wind and solar generation feeding it, both of which vary in time and across space. Graph methods have been applied directly to this forecasting problem, predicting photovoltaic output across many sites by exploiting their spatial correlation \cite{simeunovic2022pv,bahi2026solarrec} and jointly forecasting behind-the-meter load and generation where the two are entangled \cite{khodayar2021pv}. The two halves connect, since a forecast of renewable output is an input to the dispatch decision, and the appeal of graph methods spans both: an operating task is computation on the grid graph, and a forecasting task is a spatio-temporal prediction on a graph of distributed resources. A few tasks deserve naming individually because they recur. State estimation and the detection of bad or tampered measurements underpin everything downstream, since an operator acts on the estimated state and a corrupted estimate leads to a wrong action, a vulnerability that connects to the grid's cyber-physical security and to the false-data-injection attacks the fraud chapter touches. Unit commitment and economic dispatch decide which generators run and at what output, a combinatorial optimization the renewable forecast feeds. Load forecasting predicts the demand the grid must meet, a spatio-temporal problem in its own right. And the rise of microgrids and distributed resources adds local, semi-autonomous subgraphs that must coordinate with the wider network, multiplying the topologies a method must handle.

\subsection{Operating the grid}
\label{subsec:power-operation}

The clearest use of graph networks in grid operation is as a fast surrogate for an expensive physical computation. Solving the power-flow equations requires an iterative numerical method, and doing it repeatedly, for many scenarios or within an optimization loop, is a computational bottleneck. A graph network can be trained to map injections directly to the resulting voltages,
\begin{equation}\label{eq:surrogate}
\widehat{\mathbf{V}}=f_\theta(\mathbf{P},\mathbf{Q};\graph),
\end{equation} The learned power-flow surrogate, \cref{eq:surrogate}, illustrates the role graph models play in grid analysis.
approximating the solver in a single forward pass, and a model built to mirror the structure of the iterative solution learns to perform the calculation on the grid graph \cite{donon2020graphsolver}. The analogy to the numerical solver is close, since the standard iterative method passes information between connected buses until the voltages settle, which is itself a kind of message passing on the grid graph, so a graph network that unrolls this process inherits a structure suited to the problem. The physics-informed training that makes such surrogates trustworthy adds a penalty for violating the power-flow relationships, pulling the model toward physically consistent outputs rather than mere curve-fitting and narrowing the gap between a fast approximation and a feasible solution. The payoff is sharpest in contingency analysis, where an operator evaluates the grid's response to the loss of each component in turn, a sweep of many near-identical power-flow solves that a surrogate can accelerate dramatically, turning an overnight study into one that runs in near real time. Optimal power flow, which wraps the power-flow constraints in an optimization over dispatch, is even more expensive, and graph networks have been used to approximate its solutions so that a near-optimal dispatch can be produced quickly enough for repeated use \cite{owerko2020opf}. Because the grid carries known physics, a productive design constrains the learned model to respect that physics rather than discarding it, training the network to satisfy the power-flow relationships and using the graph structure to encode the network's connectivity. Operation under a changing network is a further concern that graph methods are suited to, since a model that reasons over topology can assess the grid's risk as lines enter and leave service rather than assuming a fixed configuration \cite{zhang2024gridrisk,bahi2025smartgrid,bahi2025greendc}, and the real-time control of grid quantities such as voltage is a further operating application graph methods address. \Cref{alg:gridfault} states a generic procedure for the related task of localizing a fault from the measured state of the network.

\begin{algorithm}[t]
\caption{Grid fault localization on the network graph}
\label{alg:gridfault}
\KwIn{grid graph $\graph$, bus measurements $\Feat$ (voltages, currents), trained model $f_\theta$}
\KwOut{suspected fault location}
$\Hid \leftarrow f_\theta(\Adj,\Feat)$\tcp*{encode the measured grid state}
\ForEach{bus or line element $e$}{
  score the anomaly of $e$ from its learned embedding\;
}
\Return{the element with the highest anomaly score}
\end{algorithm}

\subsection{Renewable generation forecasting}
\label{subsec:power-renewable}

\begin{figure}[t]
\centering
\resizebox{0.78\textwidth}{!}{%
\begin{tikzpicture}[
  wind/.style={circle, draw=gnnteal!80!black, fill=tintteal, minimum size=7.5mm, font=\scriptsize\bfseries, text=gnnteal},
  solar/.style={circle, draw=gnnamber!80!black, fill=tintamber, minimum size=7.5mm, font=\scriptsize\bfseries, text=gnnrust},
  ln/.style={draw=gnnslate, line width=1pt},
  io/.style={rounded corners=3pt, draw=gnnblue!80!black, fill=tintblue, font=\small, text=gnnblue!75!black, inner sep=4pt},
  fcast/.style={rounded corners=3pt, draw=gnngreen!80!black, fill=tintgreen, font=\small, text=gnngreen!72!black, inner sep=4pt}]
\node[wind]  (p1) at (0,1.4)   {W};
\node[solar] (p2) at (1.4,2.1) {S};
\node[wind]  (p3) at (1.6,0.6) {W};
\node[solar] (p4) at (3.0,1.3) {S};
\draw[ln] (p1)--(p2) (p1)--(p3) (p2)--(p3) (p3)--(p4) (p2)--(p4);
\node[io] (w) at (1.4,3.4) {weather};
\draw[-{Stealth[length=2.4mm]},gnnblue!70,line width=1pt] (w) -- (p2);
\draw[-{Stealth[length=2.4mm]},gnnblue!70,line width=1pt] (w) to[bend right=12] (p1);
\node[fcast] (out) at (5.7,1.35) {forecast $\widehat{\mathbf{y}}^{(t+1:t+H)}$};
\draw[-{Stealth[length=2.8mm]},gnnslate,line width=1.1pt] (p4) -- (out);
\end{tikzpicture}}
\caption[Renewable forecasting as a spatio-temporal graph problem.]{Forecasting renewable generation: wind (W, teal) and solar (S, amber) plants are nodes joined by spatial proximity, each carrying a generation time series, with a shared weather signal driving them. As in traffic forecasting, the model predicts future output at every site from recent history and weather, exploiting the spatial correlation between nearby plants.}
\label{fig:renewable}
\end{figure}
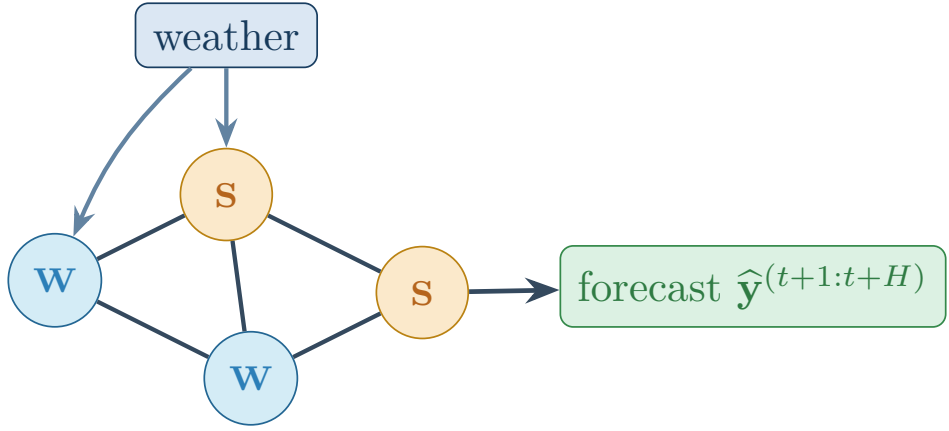

The variability of wind and solar generation is the central challenge of a decarbonizing grid, and forecasting that generation is a spatio-temporal problem of exactly the kind the transportation chapter described. Wind farms and solar installations are distributed across a region, their output is driven by weather that moves across that region, and nearby sites are correlated, so a graph of plants joined by spatial proximity, each carrying a generation time series and a shared weather signal, is the natural representation, sketched in \cref{fig:renewable}. A forecast predicts future output at every site from recent history and exogenous weather,
\begin{equation}\label{eq:reforecast}
\widehat{\mathbf{y}}^{(t+1:t+H)}=g\big(\mathbf{y}^{(t-T+1:t)},\mathbf{w};\graph\big),
\end{equation}
and graph models that combine spatial propagation with temporal modelling improve on per-site forecasts by exploiting the correlation between plants, as demonstrated for photovoltaic output prediction \cite{karimi2021pv}. Renewable forecasting also exposes a problem that the methods chapter treated only abstractly, distribution shift. A forecaster is trained on data from particular sites under particular conditions, and it must operate as new sites come online, as seasons turn, and as the climate itself changes, often without access to the original training data for privacy or practical reasons. Adapting a model continually to these shifts without revisiting its source data is a live concern, and a source-free adaptation approach for renewable forecasting addresses it directly \cite{bahi2026freegnn,bahi2025seqenergyrec,bahi2026llmenergy}, connecting this domain to the broader treatment of robustness and distribution shift later in the survey. Wind and solar pose different forecasting problems despite sharing a framework. Solar output follows a strong daily cycle modulated by cloud cover, so the hard part is short-term variability as clouds pass, while wind is driven by weather systems whose ramps, the sudden large changes in output, are both the most consequential events for grid balancing and the hardest to predict. Two features set renewable forecasting apart from the traffic case it resembles. The first is that uncertainty is not optional but central, since an operator balancing supply and demand needs the range of possible output, not only its expected value, which makes probabilistic forecasting that returns a distribution the appropriate target. The second is the role of numerical weather prediction, whose physical forecasts of wind and irradiance are a powerful exogenous input that a graph model can combine with the spatial correlations among plants. The relevant horizons span minutes for real-time control, hours for dispatch, and days for scheduling, and a method tuned for one horizon rarely serves another. \Cref{alg:reforecast} states the shared forecasting procedure.

\begin{algorithm}[t]
\caption{Renewable generation forecasting}
\label{alg:reforecast}
\KwIn{plant graph $\graph$, history $\mathbf{y}^{(t-T+1:t)}$, weather covariates $\mathbf{w}$, horizon $H$}
\KwOut{forecast $\widehat{\mathbf{y}}^{(t+1:t+H)}$}
$\Hid \leftarrow$ encode history and weather over the plant graph\;
\For{layer $l=0$ to $L-1$}{
  $\Hid \leftarrow$ spatial graph convolution over nearby plants\;
  $\Hid \leftarrow$ temporal convolution along time\;
}
$\widehat{\mathbf{y}}^{(t+1:t+H)} \leftarrow$ decode $\Hid$ by \cref{eq:reforecast}\;
\Return{forecast $\widehat{\mathbf{y}}^{(t+1:t+H)}$}
\end{algorithm}

\subsection{The wider energy system}
\label{subsec:power-wider}

The grid is only the backbone of a larger energy system that is itself becoming more graph-like. Distributed energy resources, rooftop solar, home batteries, and controllable loads turn passive consumers into active participants whose coordination is a problem on a graph of grid-edge devices. Electric-vehicle charging adds a mobile, time-varying demand whose graph couples the power network to the transportation network of the previous chapter, since where and when vehicles charge depends on how they move. Demand response, in which loads are shifted in time to match supply, is a control problem over this graph of flexible resources. These developments share a direction, the decentralization of a system built around a few large generators into one with many small, interacting participants, and that direction is exactly the one that makes graph structure more relevant rather than less, because the coordination problem grows with the number of nodes that must act together.

\subsection{Datasets, strengths, and open problems}
\label{subsec:power-tradeoffs}

\Cref{tab:powerapps} collects the main applications. Evaluation in grid operation relies heavily on standardized synthetic test systems of varying size that stand in for real networks, supplemented by the limited real grid data that utilities can release, while renewable forecasting draws on generation and weather records. The reliance on synthetic test systems is itself a limitation, since a method that performs well on a textbook network need not transfer to the scale and irregularity of a real grid, and the broader power-systems literature, surveyed in a dedicated review \cite{liao2021powersurvey}, returns to this gap repeatedly. The synthetic systems in question are small reference networks of a few dozen to a few hundred buses that have served power engineering for decades, and their convenience is also their weakness, since a real grid has orders of magnitude more components, irregular structure, and operating conditions no textbook case reproduces. The field lacks the large, standardized, openly shared benchmarks that drove progress in vision and language, a scarcity that follows directly from the sensitivity of grid data, and one consequence is that reported gains are harder to compare across studies and easier to overstate than in domains with common test sets.

The strengths of graph methods here are real but specific. The grid is a genuine graph with genuine physics, and a graph network that respects its structure is a faithful model rather than an analogy. The clearest value is acceleration: a learned surrogate that approximates a power-flow or optimal-power-flow solution in a single pass can be orders of magnitude faster than the iterative computation it replaces, which matters when the calculation must be run thousands of times for contingency analysis or within a control loop. Topology awareness lets a model reason about a reconfiguring grid, and the spatio-temporal forecasting of renewable output is a direct and useful transfer of the transportation toolkit. The stakes give this work a significance beyond the technical, since integrating variable renewable generation at the scale decarbonization requires depends on forecasting that generation accurately and operating a more complex, more dynamic grid reliably, and both are problems where the grid's graph structure is part of the solution. A method that lets operators run more contingency studies, integrate more renewable capacity, or balance supply and demand more finely contributes to a transition whose difficulty is as much operational as it is political.

The weaknesses sharpen into a question the domain cannot avoid. Because the physics is known and exact, a learned model does not compete with a principled solver on correctness, only on speed, and a surrogate that produces a physically impossible state is worse than useless in a system where a wrong answer can trip a network. The bar for replacing a guaranteed method with an approximate one is therefore high, and it is met only where the approximation is fast enough to enable something the exact method cannot do in time and reliable enough to be trusted or checked. Data is scarce and sensitive, since real grid topologies and measurements are critical infrastructure that utilities guard, which pushes the field toward synthetic systems and limits how well results generalize. Transfer across grids is hard, because a model trained on one network's topology need not work on another, and the high stakes demand reliability guarantees that learned models do not naturally provide. This is the domain where the survey's recurring tension between a flexible learned model and a principled alternative is sharpest, because the alternative is not a weaker heuristic but exact physics, and the learned model's job is not to be more accurate but to be fast enough and trustworthy enough to use in the loop. The verification question, how to certify that a learned surrogate will not produce a dangerous output, is therefore central here in a way it is not where errors are merely costly rather than catastrophic, and it is one reason the most credible work pairs learning with constraints or with a physics-based check rather than replacing the physics outright.

The open problems follow. Physics-informed and provably constrained graph networks that cannot output an infeasible state, transfer across grid topologies, calibrated uncertainty that an operator can act on, the integration of forecasting with operation into a single decision pipeline, and the prospect of foundation models for energy systems are the active directions. A fair summary is that power systems are a domain where the graph is unusually faithful and the physics unusually well understood, so the value of graph learning lies less in prediction accuracy, which physics already provides, than in acceleration, topology awareness, and the forecasting of the variable generation that the physics cannot supply on its own, all under a reliability bar that the safety-critical setting makes uncommonly strict.

\begin{table}[t]
\centering
\caption[Power-systems applications of graph neural networks.]{Power and energy applications, with the graph each builds and the task it poses. A representative method is named for each.}
\label{tab:powerapps}
\small
\begin{tabularx}{\textwidth}{@{}lXll@{}}
\toprule
Application & Graph (nodes / edges) & Task & Example \\
\midrule
Power flow                 & buses / lines                & fast state solving        & Graph Neural Solver \cite{donon2020graphsolver} \\
Optimal power flow         & buses / lines                & optimization surrogate    & \cite{owerko2020opf} \\
Operational risk           & buses / evolving topology    & reliability assessment    & \cite{zhang2024gridrisk} \\
Grid control               & buses and devices / lines    & voltage and dispatch control & --- \\
PV and wind forecasting    & plants / spatial proximity   & spatio-temporal forecasting & \cite{karimi2021pv} \\
Distribution-shift adaptation & plants / proximity        & source-free adaptation    & FreeGNN \cite{bahi2026freegnn} \\
\bottomrule
\end{tabularx}
\end{table}

\FloatBarrier
\section{Internet of Things, wireless, and 6G networks}
\label{sec:iot}

A communication network is a graph in the engineering sense as much as the mathematical one. Transmitters, receivers, base stations, and devices are nodes, and the channels that carry signal between them, along with the interference one transmission imposes on another, are edges. This makes resource allocation, routing, and network management problems on a graph, and it explains a surge of interest in graph networks for wireless systems and the Internet of Things, sharpened by the vision of sixth-generation networks that are denser, more dynamic, and more autonomous than anything deployed today. The domain is distinctive in that the properties practitioners most want from a learned solution, that it ignore the arbitrary labelling of devices, that it scale to networks of different sizes, and that it run in a decentralized way from local information, are exactly the properties a graph network provides by construction, which makes the match between method and problem unusually clean.

\subsection{Networks and devices as graphs}
\label{subsec:iot-graph}

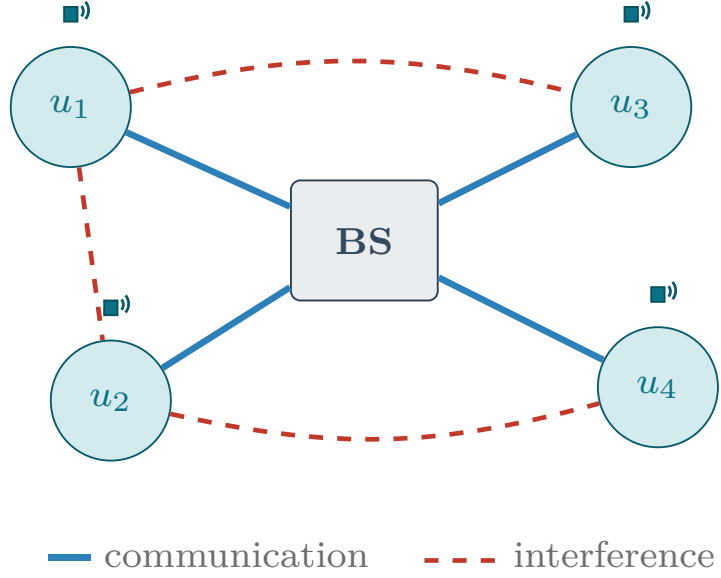
\begin{figure}[t]
\centering
\resizebox{0.6\textwidth}{!}{%
\begin{tikzpicture}[
  dev/.style={circle, draw=gnnocean!80!black, fill=tintocean, minimum size=9mm, font=\scriptsize\bfseries, text=gnnocean},
  bs/.style={rectangle, rounded corners=2pt, draw=gnnslate!85!black, fill=tintgray, inner sep=4pt,
             font=\scriptsize\bfseries, text=gnnslate, minimum height=9mm, minimum width=11mm},
  comm/.style={draw=gnnteal, line width=1.4pt},
  intf/.style={draw=gnnred, dashed, line width=1pt}]
\node[bs]  (b)  at (2.2,1.5)  {BS};
\node[dev] (d1) at (0,2.5)    {$u_1$};
\node[dev] (d2) at (0.3,0.3)  {$u_2$};
\node[dev] (d3) at (4.2,2.5)  {$u_3$};
\node[dev] (d4) at (4.4,0.4)  {$u_4$};
\foreach \n in {d1,d2,d3,d4}{\pic[scale=0.5] at ([yshift=7mm]\n) {ic sensor={gnnocean}};}
\draw[comm] (b)--(d1) (b)--(d2) (b)--(d3) (b)--(d4);
\draw[intf] (d1) to[bend left=14]  (d3);
\draw[intf] (d2) to[bend right=14] (d4);
\draw[intf] (d1) -- (d2);
\node[anchor=west,font=\scriptsize,text=gnngray] at (-0.3,-0.85)
  {\textcolor{gnnteal}{\rule[0.4ex]{9pt}{1.3pt}}~communication \quad
   \tikz[baseline]{\draw[intf](0,0.4ex)--(0.55,0.4ex);}~interference};
\end{tikzpicture}}
\caption[A wireless or IoT network as a graph.]{A wireless or IoT network as a graph: devices ($u_i$, marked with a sensor) and a base station (BS) are nodes, solid teal edges are communication links, and dashed red edges mark interference between transmissions. Resource allocation, deciding each device's power or spectrum, is a problem on this graph, and the right policy ignores how the devices are labelled and generalizes across networks of different sizes.}
\label{fig:iiotgraph}
\end{figure}

\Cref{fig:iiotgraph} shows the basic construction for a wireless network, with devices and a base station as nodes, communication links as edges, and interference relationships as a second kind of edge. The interference graph is the key structure for resource allocation, since whether two transmissions conflict depends on whether an edge joins them, and the allocation that resolves the conflict is therefore a computation on the graph. For the Internet of Things the construction is similar, with devices or sensors as nodes and communication or physical proximity as edges, often at a scale of many thousands of resource-constrained nodes. Three properties make graph networks a natural fit beyond the bare presence of a graph. The allocation a network needs should not depend on how the devices happen to be numbered, which is permutation equivariance; a policy trained on a small network should apply to a larger one, which is the size generalization that follows from sharing the same local rule across all nodes; and execution should be possible from local information at each node, which message passing supports directly. These are not incidental conveniences but the central design goals of practical wireless systems, and they are the inductive biases a graph network supplies. The wireless graphs themselves come in several forms. A cellular network connects users to base stations in a roughly star-shaped pattern with interference between cells; a device-to-device or ad-hoc network connects peers directly in a mesh with no central coordinator; and a vehicular network adds mobility, so the graph changes as vehicles move. Each carries node features describing channel quality and edge features describing the strength of a link or the interference it causes, and the channel state, the collection of these quantities, is what a resource-allocation policy must read. The diversity of these constructions is one reason a method's ability to generalize across topologies, rather than fitting a single fixed network, matters so much in this domain.

\subsection{Wireless resource allocation and network optimization}
\label{subsec:iot-wireless}

The central problem is to allocate limited resources, transmit power, spectrum, or beamforming directions, so as to optimize a network objective subject to interference. A canonical instance maximizes the total communication rate by choosing a power for each transmitter,
\begin{equation}\label{eq:resalloc}
\max_{\mathbf{p}}\ \sum_{i}\log\!\big(1+\mathrm{SINR}_i(\mathbf{p})\big)\quad\text{subject to}\quad 0\le p_i\le p_{\max},
\end{equation}
where $\mathrm{SINR}_i$ is the signal-to-interference-and-noise ratio at receiver $i$ and depends on the powers of all the interfering transmitters. This optimization is non-convex and hard, and the classical approach solves it with iterative algorithms that are accurate but slow and must be rerun whenever the channel changes. A graph network instead learns a policy that maps the channel state to an allocation in a single pass,
\begin{equation}\label{eq:gnnpolicy}
\mathbf{p}=\pi_\theta(\mathbf{H};\graph),
\end{equation} The resource-allocation objective and the graph policy it induces, \cref{eq:resalloc,eq:gnnpolicy}, frame the wireless problems discussed here.
and because the policy is a function on the interference graph it inherits the permutation equivariance the problem demands. Treating the wireless channel as a random graph and learning a policy that generalizes across its realizations gives a model that adapts to changing channels without resolving the optimization each time \cite{eisen2020regnn}. The size-generalization property has been established with theoretical analysis, showing that a policy trained on small networks transfers to larger ones because it applies the same local rule everywhere \cite{shen2021rrm}, and the specific problem of transmit-power control has a scalable graph formulation of the same kind \cite{shen2019powercontrol}. Decentralization follows naturally, since a policy expressed as message passing can be executed by each node from its local neighbourhood rather than from a global view, which matches how a real network must operate \cite{wang2022decentralized}. The maturing of this line from theoretical results to practical systems is the subject of a dedicated survey \cite{shen2023theory2practice}. Several threads reward a closer look. The classical iterative methods these policies approximate, of which the weighted minimum-mean-square-error algorithm for power control is the best known, provide both the training targets and the baseline to beat, and a productive design unrolls such an algorithm into a graph network so the learned model inherits the structure of the optimization it replaces rather than starting from nothing. Beamforming, the shaping of a transmitted signal across an antenna array, and link scheduling, the combinatorial choice of which transmissions to activate, extend the same framework to richer action spaces. The theoretical results that distinguish this domain establish not only that a policy generalizes across network sizes but that it is stable, in the sense that a small change in the channel produces a small change in the allocation, a property that matters because a policy operating on a constantly varying channel must not swing wildly between decisions. \Cref{tab:iotwireless} collects the main applications.

\begin{table}[t]
\centering
\caption[Wireless and IoT applications of graph neural networks.]{Wireless and IoT applications, with the graph each builds and the task it poses. A representative method is named for each.}
\label{tab:iotwireless}
\small
\begin{tabularx}{\textwidth}{@{}lXll@{}}
\toprule
Application & Graph (nodes / edges) & Task & Example \\
\midrule
Resource allocation       & transceivers / interference   & power and spectrum control & Random-edge GNN \cite{eisen2020regnn} \\
Radio resource management & users / channel relations     & scalable allocation        & \cite{shen2021rrm} \\
Power control             & transmitters / interference   & transmit-power policy      & \cite{shen2019powercontrol} \\
Decentralized allocation  & nodes / local neighbourhood   & distributed control        & \cite{wang2022decentralized} \\
IoT anomaly detection     & devices / communication       & distributed detection      & \cite{protogerou2021iot} \\
\bottomrule
\end{tabularx}
\end{table}

\subsection{IoT systems and sensor networks}
\label{subsec:iot-systems}

The Internet of Things shifts the emphasis from optimizing a network to monitoring and securing one made of many small devices. The graph here is the system itself, and a common task is detecting anomalies or intrusions from the pattern of device behaviour and communication, posed as a problem on the device graph so that a compromised or malfunctioning node is identified by how it relates to its neighbours rather than in isolation, an approach that suits the distributed nature of these systems \cite{protogerou2021iot}. Sensor networks add a spatial graph of measurement nodes used for data fusion, event detection, and localization, where the geometry of the deployment is the graph. Network management at a higher level is also a graph problem. Predicting the performance of a network, the delay and loss a configuration will produce, from its topology and traffic lets an operator evaluate changes before making them, and a graph model that learns the relationship between configuration and performance can serve as a fast what-if tool. Routing, the choice of paths through the network, and the placement of functions and resources are further problems on the same graph, and they share the domain's defining advantage, that the network the model reasons over is a real, observable graph rather than an inferred one. Two constraints distinguish the IoT setting from the wireless-optimization one. The devices are resource-limited in compute and energy, so a model that runs on them must be small, and the network is dynamic and adversarial, since devices join and leave and the system is a large attack surface, which makes security a first-order concern that connects this material to the fraud and intrusion-detection methods of the next chapter. Two further considerations shape IoT applications. The energy and compute limits of devices push toward small, efficient models and toward compressing or distilling a trained network so it can run at the edge, a constraint absent from the data-center setting most graph research assumes. And privacy, together with the sheer volume of distributed data, motivates federated approaches in which devices collaborate to train a shared model without sending their raw data to a central server, a setting whose communication pattern is itself a graph and whose constraints align with the decentralized execution graph networks already support.

\subsection{Toward sixth-generation networks}
\label{subsec:iot-6g}

The interest in graph methods for communication is driven partly by what the next generation of networks is expected to demand. Sixth-generation systems are envisioned as ultra-dense and highly dynamic, serving enormous numbers of devices, partitioning themselves into virtual slices for different services, and integrating sensing with communication, all of which are problems on large, changing graphs. The properties that make graph networks attractive for current wireless problems, scalability across network sizes, decentralized execution, and adaptability to changing topology, are precisely the properties such networks will require, which is why graph-based learning features prominently in research toward them. The gap between this expectation and a deployed capability is wide, and it is the same gap that tempers the rest of the domain: most results exist in simulation, the move to real radios and real traffic is largely unproven, and the real-time and hardware constraints of a deployed network are not captured by the simulations on which methods are developed. A few of the envisioned capabilities are explicitly graph problems. Network slicing partitions a shared physical network into virtual networks for different services, an allocation over a graph of resources and demands. Integrated sensing and communication uses the same infrastructure to perceive the environment and to communicate, coupling a sensing graph to a communication one. Reconfigurable surfaces that steer signals add controllable edges to the wireless graph. And the notion of a digital twin of the network, a continuously updated model used to plan and optimize, is a graph-learning problem at the scale of an entire operator's infrastructure. These are aspirations more than deployed systems, but they indicate why graph methods are expected to matter as networks grow more complex.

\subsection{Strengths, weaknesses, and open problems}
\label{subsec:iot-tradeoffs}

The strengths here are among the most principled in the survey. A communication network is a genuine graph, and the inductive biases a graph network brings, equivariance to device relabelling, generalization across network sizes, and decentralized execution from local information, are not approximations to what the problem needs but exactly what it needs, which is why some of this work comes with theoretical guarantees rather than only empirical results. The theoretical grounding is worth dwelling on, because it is rarer in applied graph learning than the empirical successes of other chapters. Results in this area connect the equivariance of the policy to its ability to transfer across networks, and the stability of message passing to the robustness of the allocation under channel variation, giving a principled account of why the approach works rather than only a demonstration that it does. This is the kind of analysis the methods chapter called for in general and that few application domains have achieved, and it is possible here precisely because the problem's structure, an optimization over a graph with a clear objective and clear symmetries, is clean enough to reason about. The practical payoff is the same acceleration seen in power systems, since a learned policy that produces an allocation in one pass can replace an iterative optimization that is too slow to rerun as channels change, and the decentralized formulation matches how networks must actually operate.

The weaknesses are equally clear and largely about validation rather than concept. Most results live in simulation, and the leap from a simulated channel to a deployed network, with its hardware limits, real interference, and unmodelled effects, is the field's central unproven step. As in power systems, the learned policy competes against strong classical optimization, so its advantage is speed and scalability rather than optimality, and a policy that is fast but meaningfully worse than the optimum may not be worth deploying. The resource limits of IoT devices constrain how large a model can be, distribution shift in channel conditions and network composition threatens policies trained on a fixed setting, and the domain is younger and less benchmarked than the established application areas, so comparisons are harder to trust.

The open problems follow from the validation gap. Transfer from simulation to deployment, robustness to shifting channels and topologies, on-device efficiency for resource-constrained nodes, integration with the architectures of next-generation networks, and the extension of the existing theoretical guarantees to more realistic settings are the active directions. A fair summary is that wireless and IoT networks are among the domains where the fit between a graph network's inductive biases and the problem's requirements is tightest and best understood theoretically, and also a domain where that promise has so far been demonstrated mostly in simulation, leaving the decisive test, performance in deployed networks, still to be passed.

\FloatBarrier
\section{Cybersecurity, finance, and fraud detection}
\label{sec:fraud}

Fraud and cyberattacks are relational by their nature. A fraudster rarely acts alone or in isolation, operating instead through shared devices, payment instruments, and addresses, coordinating in rings, and transacting with victims, so the trace of fraud is a pattern of connections rather than a property of any single account. An intrusion spreads from host to host, and network traffic is a graph of flows between machines. The same relational framing extends to software security, where graph embeddings of binary functions support cross-platform similarity detection for vulnerability search \cite{xu2017gemini}. This makes the domain a natural fit for graph methods, which can find the coordination and propagation that a model examining records one at a time cannot see. It also makes the domain a stress test, because the assumptions that serve graph networks well elsewhere are often false here: fraud is rare, so the classes are extremely imbalanced; adversaries adapt to evade detection, so the target moves; and fraudsters disguise themselves by connecting to legitimate entities, so connected nodes are frequently dissimilar rather than alike. These difficulties have driven some of the more interesting method development in applied graph learning, and this section treats both the financial and the security sides with attention to where the standard toolkit must be rebuilt.

\subsection{Why fraud and intrusion are graph problems}
\label{subsec:fraud-graph}

\begin{figure}[t]
\centering
\resizebox{0.72\textwidth}{!}{%
\begin{tikzpicture}[
  acct/.style 2 args={circle, draw=#1!80!black, fill=#2, line width=0.9pt, minimum size=9mm},
  fedge/.style={draw=gnnred, line width=1.4pt},
  camo/.style={draw=gnngray, line width=0.8pt, dash pattern=on 2.5pt off 1.8pt},
  ledge/.style={draw=gnnteal!75!black, line width=0.9pt}]
% fraud cluster (dense ring)
\node[acct={gnnred}{tintred}] (f1) at (2.3,1.8) {};
\node[acct={gnnred}{tintred}] (f2) at (3.4,2.1) {};
\node[acct={gnnred}{tintred}] (f3) at (3.6,1.0) {};
\node[acct={gnnred}{tintred}] (f4) at (2.6,0.7) {};
\begin{scope}[on background layer]
  \node[panel=gnnred, fit=(f1)(f2)(f3)(f4), inner sep=3.2mm] (fr) {};
\end{scope}
\foreach \n in {f1,f2,f3,f4}{\pic[scale=0.85] at (\n) {ic person={gnnred}};}
\draw[fedge] (f1)--(f2) (f2)--(f3) (f3)--(f4) (f4)--(f1) (f1)--(f3) (f2)--(f4);
\node[chip=gnnred] at (fr.north) {fraud ring (dense)};
% legitimate accounts
\node[acct={gnnteal}{tintteal}] (l1) at (0.2,2.6)  {};
\node[acct={gnnteal}{tintteal}] (l2) at (-0.1,1.1) {};
\node[acct={gnnteal}{tintteal}] (l3) at (0.6,-0.4) {};
\node[acct={gnnteal}{tintteal}] (l4) at (5.4,2.5)  {};
\node[acct={gnnteal}{tintteal}] (l5) at (5.8,1.0)  {};
\node[acct={gnnteal}{tintteal}] (l6) at (5.1,-0.4) {};
\foreach \n in {l1,l2,l3,l4,l5,l6}{\pic[scale=0.85] at (\n) {ic person={gnnteal}};}
% camouflage links (fraud disguising as legitimate)
\draw[camo] (f1)--(l1) (f1)--(l2) (f4)--(l3) (f2)--(l4) (f3)--(l5) (f3)--(l6);
% ordinary legitimate links
\draw[ledge] (l1)--(l2) (l2)--(l3) (l4)--(l5) (l5)--(l6);
% legend
\node[anchor=west,font=\scriptsize,text=gnngray] at (-0.5,-1.35)
  {\textcolor{gnnred}{$\bullet$} fraud \quad \textcolor{gnnteal}{$\bullet$} legitimate \quad
   \tikz[baseline]{\draw[camo](0,0.4ex)--(0.5,0.4ex);} camouflage link};
\end{tikzpicture}}
\caption[A fraud ring in a transaction graph.]{A transaction or entity graph: most accounts are legitimate (teal) and a small set of fraudulent accounts (red) form a densely connected ring while also forming camouflage links (dashed) to legitimate accounts to disguise themselves. The dense internal structure is signal a graph method can find, and the camouflage links are what defeat a naive one, which is why fraud-specific designs filter or reweight neighbours rather than aggregating them uniformly.}
\label{fig:cybergraph}
\end{figure}
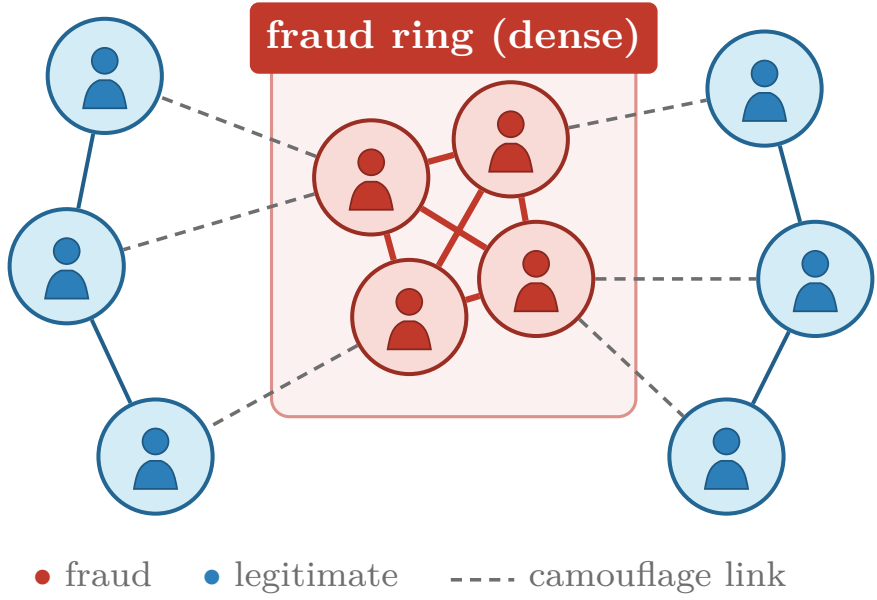

The constructions are varied but share a logic. A transaction graph makes accounts or transactions nodes and payments edges; a review graph joins users to the items they review; a device-sharing graph links accounts that used the same hardware; and a network-flow graph makes hosts nodes and connections edges. In each, the relational structure exposes coordination that per-entity features hide, since a fraud ring is dense where legitimate activity is sparse, and a compromised host behaves anomalously relative to its neighbours. \Cref{fig:cybergraph} shows the canonical picture, a small set of fraudulent nodes densely interconnected while also linking to legitimate nodes to blend in. The detection task is node or edge classification, separating fraudulent from legitimate or malicious from benign, and a node's fraud score is read from its learned representation,
\begin{equation}\label{eq:fraudscore}
s_v=\sigma\!\big(\mathbf{w}^\top\hv_v^{(L)}\big),
\end{equation}
where the representation is supposed to encode both the node's own attributes and the structure of its neighbourhood. The promise is that the structure carries signal the attributes lack, and the difficulty is that adversaries know this and shape the structure to mislead. The graphs in practice are usually heterogeneous and multi-relational, since two accounts can be related in several ways at once, through a shared device, a shared address, a common payment instrument, or a direct transaction, and each relation carries different evidential weight. A shared device is a stronger signal of collusion than a single transaction, and a model that collapses these relations into one loses that distinction, so fraud graphs are often modelled with multiple relation types the network treats separately. Time adds a further dimension, because fraud unfolds as a sequence of events and a sudden burst of activity can itself be the signal, which makes the transaction graph a dynamic object whose recent structure matters more than its distant history.

\subsection{The distinctive challenges}
\label{subsec:fraud-challenges}

Four difficulties define this domain and explain why a standard graph network underperforms on it. The first is class imbalance, since fraud is a tiny fraction of activity, often well under one percent, and a model trained without correction is overwhelmed by the legitimate majority and learns to predict that everything is legitimate. Addressing this requires balancing the training signal, for instance by sampling that gives the rare fraudulent nodes and their neighbourhoods adequate representation \cite{liu2021pcgnn}, and a class-weighted loss that upweights the rare positives,
\begin{equation}\label{eq:imbloss}
\mathcal{L}=-\sum_{v}\big(\alpha\,y_v\log s_v+(1-y_v)\log(1-s_v)\big),
\end{equation}
with $\alpha\gg1$, is a standard ingredient. The second is camouflage. Fraudsters actively disguise themselves, both by adopting features that resemble legitimate accounts and by connecting to legitimate nodes so that naive neighbourhood aggregation dilutes their signal, and a detector must resist this by choosing which neighbours to trust rather than averaging over all of them \cite{dou2020caregnn}\footnote{Reference implementation: \url{https://github.com/YingtongDou/CARE-GNN}}. The third is inconsistency, the failure of the homophily assumption that underlies most graph networks. Ordinary message passing presumes that connected nodes are similar and smooths their representations together, but a fraudster connected to many legitimate nodes violates this, so aggregation that assumes homophily actively erases the signal, and consistency-aware aggregation that accounts for the mismatch is needed \cite{liu2020graphconsis}. The fourth is label scarcity, since confirmed fraud labels are expensive and few, which pushes the field toward semi-supervised methods that propagate a little label information across a largely unlabelled graph \cite{xiang2023gtan}. A camouflage-resistant layer captures the common response to the middle two challenges, aggregating only over neighbours whose relevance to the node exceeds a threshold,
\begin{equation}\label{eq:fraudagg}
\hv_v^{(l+1)}=\UPD\!\Big(\hv_v^{(l)},\ \AGG\big\{\hv_u^{(l)}:u\in\nbr(v),\ \rho(v,u)>\tau\big\}\Big),
\end{equation}
so that disguising links are filtered rather than allowed to dominate. These four challenges connect directly to the heterophily and robustness discussions later in the survey, and they are the reason fraud is a domain where off-the-shelf methods fail and task-specific designs are essential. Two further properties deepen the difficulty. Graph networks are themselves vulnerable to structural manipulation, since an adversary who can add or remove a few edges, by creating accounts or transactions, can shift a model's prediction, and in fraud the adversary is precisely the party with both the incentive and the means to do so. And the setting demands explanation, because a flagged transaction triggers a costly action, a frozen account or a declined payment, and an analyst and often a regulator must understand why, which makes the opacity of a learned model a practical obstacle rather than only an aesthetic one. Both properties recur in the robustness and explainability chapter, and both are felt acutely here because the stakes are immediate and the adversary is real.

\subsection{Financial fraud detection}
\label{subsec:fraud-financial}

The financial applications span several settings. Credit-card fraud is detected from a transaction graph, where a semi-supervised model that combines transaction attributes with graph structure identifies fraudulent activity from limited labels \cite{xiang2023gtan}. Money laundering traces illicit value through a transaction network, a problem that has drawn particular attention in cryptocurrency, where the public ledger makes the transaction graph observable and the task is to separate illicit flows from legitimate ones. The laundering problem has a structure of its own, since the movement of illicit value tends to follow recognizable patterns, layering through chains of intermediaries and aggregating through hubs, and these typologies are subgraph patterns a graph model can be trained to recognize. A public benchmark of cryptocurrency transactions labelled as licit or illicit has made this a concrete test case and exposed the same imbalance and label-scarcity problems that define the domain. The regulatory context shapes the work as much as the data, because anti-money-laundering and fraud controls operate under compliance requirements that demand auditable decisions, which raises the premium on explanation and constrains how freely an opaque model can be deployed. Insurance and e-commerce add further variants, organized claims fraud and coordinated abuse of promotions, that share the relational, coordinated character and the same defining challenges. Review and opinion fraud, the manufacture of fake reviews to manipulate ratings, is detected by finding the coordinated fraudster accounts behind it in a user-review graph \cite{wang2019fdgars}. Across these settings the methods that work are the ones built for the domain's challenges, the camouflage-resistant, imbalance-aware, and consistency-aware designs already described, and \cref{tab:fraudmethods} organizes them by the difficulty each targets. \Cref{alg:fraud} states the shared procedure, neighbour filtering followed by scoring under a balanced loss.

\begin{table}[t]
\centering
\caption[Graph methods for fraud, anomaly, and intrusion detection.]{Representative methods, by target and by the difficulty each is designed to address. The recurring theme is that standard aggregation must be modified for this domain.}
\label{tab:fraudmethods}
\small
\begin{tabularx}{\textwidth}{@{}llXX@{}}
\toprule
Method & Target & Challenge addressed & Key idea \\
\midrule
CARE-GNN \cite{dou2020caregnn}      & fraud          & camouflage                & reinforced neighbour selection \\
PC-GNN \cite{liu2021pcgnn}          & fraud          & class imbalance           & label-balanced sampling \\
GraphConsis \cite{liu2020graphconsis} & fraud        & inconsistency, heterophily & consistency-aware aggregation \\
GTAN \cite{xiang2023gtan}           & credit-card fraud & label scarcity         & semi-supervised attribute graph \\
BWGNN \cite{tang2022bwgnn}          & anomaly        & smoothing hides anomalies & band-pass spectral filters \\
E-GraphSAGE \cite{lo2022egraphsage} & intrusion      & flow classification       & edge-feature message passing \\
GDroid \cite{gao2021gdroid}         & malware        & program structure         & call-graph representation \\
\bottomrule
\end{tabularx}
\end{table}

\begin{algorithm}[t]
\caption{Fraud detection with neighbour filtering}
\label{alg:fraud}
\KwIn{entity graph $\graph$, node features $\Feat$, labelled fraud and legitimate nodes}
\KwOut{fraud score for each node}
initialize $\hv_v^{(0)}\leftarrow\xv_v$\;
\For{layer $l=0$ to $L-1$}{
  \ForEach{node $v$}{
    select neighbours $u$ whose relevance $\rho(v,u)$ exceeds a threshold\;
    aggregate over the selected neighbours by \cref{eq:fraudagg}\;
  }
}
score each node $s_v$ by \cref{eq:fraudscore}\;
train with the class-balanced loss of \cref{eq:imbloss}\;
\Return{fraud scores $\{s_v\}$}
\end{algorithm}

\subsection{Anomaly detection on graphs}
\label{subsec:fraud-anomaly}

Fraud is a special case of the broader problem of anomaly detection on attributed graphs, the identification of nodes or edges that deviate from the normal pattern in structure or features, and that broader problem has yielded one of the more illuminating methodological results in applied graph learning. Anomalies, by definition, differ from their surroundings, which means that in the spectral terms of the foundations chapter they carry high-frequency energy, the rapid variation between a node and its neighbours that a smooth signal lacks. A standard graph network is a low-pass filter that smooths neighbouring representations together, and so it attenuates exactly the high-frequency signal that marks an anomaly, which means the default inductive bias of the whole field is not merely unhelpful but actively wrong for this task. The remedy is to design filters that retain high-frequency components, treating anomaly detection as a problem requiring band-pass or high-pass response rather than the low-pass smoothing that suits homophilous classification \cite{tang2022bwgnn}. This is a clarifying point that reaches beyond fraud, since it shows concretely that the smoothing the methods chapter described as a limitation in the form of over-smoothing is, in some tasks, the wrong objective from the first layer, and it connects the practical problem of catching fraud to the theoretical treatment of expressiveness and heterophily elsewhere in the survey. The broader anomaly-detection problem also comes in several shapes the fraud framing can obscure. An anomaly may be a single node whose attributes or connections are unusual, an edge that should not exist between two communities, or a whole subgraph whose collective pattern is suspicious even when each member looks ordinary, and the last of these, the anomalous subgraph, is exactly the fraud ring and is the hardest to detect because the signal is distributed across many nodes. The setting also varies in how much supervision is available, from fully unsupervised detection that must define normality from the data alone, often by learning to reconstruct the graph and flagging what reconstructs poorly, to the semi-supervised case where a few labels guide the search, and the right method depends heavily on which regime holds.

\subsection{Cybersecurity: intrusion and malware}
\label{subsec:fraud-cyber}

The security applications mirror the financial ones with different graphs. Network intrusion detection models traffic as a graph of flows between hosts, and because the malicious signal often lives in the flows rather than the hosts, an effective approach classifies edges using message passing that incorporates edge features describing each flow \cite{lo2022egraphsage}, with related systems building intrusion detectors on the network graph \cite{sun2024gnnids} and emphasizing robustness to the evasion attempts that detectors must withstand \cite{pujolperich2022robust}. Malware detection turns a program into a graph, since the structure of a program's calls captures behaviour that surface features miss, and classifying Android applications from a graph of their interactions detects malicious software more reliably than flat features \cite{gao2021gdroid}, with work in the Internet-of-Things setting adding explicit defences against adversarial manipulation of the input \cite{yumlembam2023iot}. \Cref{alg:cyber} states the intrusion procedure as edge classification on the flow graph. The adversarial dimension is unavoidable throughout, since an attacker who understands the detector will craft traffic or code to evade it, which makes robustness to adversarial manipulation not an afterthought but a defining requirement, and ties this material to the robustness discussion the survey takes up next. A few specifics fill out the security picture. Network intrusion detection draws its edge features from flow records summarizing each connection's duration, volume, and protocol, and a recurring target is lateral movement, the spread of an intruder from an initial foothold across internal hosts, which appears as an unusual path through the host graph. Provenance graphs, which record how processes, files, and network connections relate on a system, support the detection of sophisticated multi-stage intrusions whose individual steps are innocuous but whose combination is not. On the malware side, a program can be represented by its control-flow graph, its function-call graph, or its dependencies, each capturing a different aspect of behaviour, and the adversarial pressure is concrete, since malware authors deliberately obfuscate and restructure code to evade graph-based detectors, which is why defences against such manipulation are integral rather than optional.

\begin{algorithm}[t]
\caption{Intrusion detection on a network-flow graph}
\label{alg:cyber}
\KwIn{flow graph $\graph$ with edge features (flow statistics), trained model $f_\theta$}
\KwOut{benign or malicious label for each flow}
embed nodes and edges by message passing that incorporates edge features\;
\ForEach{flow (edge) $e$}{
  classify $e$ as benign or malicious from its edge embedding\;
}
\Return{per-flow labels}
\end{algorithm}

\subsection{Strengths, weaknesses, and open problems}
\label{subsec:fraud-tradeoffs}

The strength of graph methods here is that fraud and intrusion are genuinely relational, and the graph exposes the coordination and propagation that define them and that a per-record model cannot see. The domain has also been unusually productive for the field, forcing innovations, in imbalance handling, camouflage resistance, heterophily-aware aggregation, and spectral anomaly detection, that have improved the understanding of graph networks generally, and it has clear and immediate practical value in financial institutions and security operations.

The weaknesses are sharp and specific. The most important is that the standard assumptions of graph learning are frequently wrong in this domain, since homophily fails when fraudsters disguise themselves and smoothing erases the high-frequency signal anomalies carry, so an off-the-shelf network underperforms and the value comes entirely from task-specific design. The second is the adversarial arms race, because detectors and evaders co-evolve, and a method that scores well on a static benchmark may be defeated quickly once adversaries adapt to it, which makes benchmark performance an especially weak predictor of durability. Concept drift compounds this, as fraud and attack patterns change over time and a model trained once degrades. Evaluation is genuinely hard, since at a base rate well below one percent the usual accuracy is meaningless and the choice of metric, the trade-off between catching fraud and flagging legitimate activity, encodes a costly business decision that benchmarks rarely make explicit. Data is private and sensitive, limiting public benchmarks, and the need to explain why a case was flagged, both for analysts and for affected customers, is a requirement that opaque models do not meet. Deployment imposes constraints the benchmarks ignore. Detection must often run in real time, scoring a transaction in the moment it is attempted, which limits how much of a large graph a model can consult per decision. The output feeds human analysts who can review only so many alerts, so a detector that is accurate but floods the queue with false positives is unusable, and alert fatigue is a real failure mode rather than a hypothetical one. These operational realities mean the measure a deployed system is judged by, useful alerts per unit of analyst effort under a latency budget, bears little resemblance to the accuracy a benchmark reports.

The open problems follow directly. Adversarially robust detection that degrades gracefully as attackers adapt, methods that handle concept drift through online or continual updating, heterophily-aware and spectral designs suited to the domain's broken homophily, explanations for flagged cases, evaluation protocols that reflect realistic imbalance and adversarial pressure, and privacy-preserving detection across institutions are the active directions. A fair summary is that fraud and cybersecurity are domains where graph methods address a genuinely relational problem and have driven real methodological progress, and also where the field's default assumptions are most often inverted and where the adversarial, drifting setting makes the gap between a benchmark result and a durable deployed system wider than almost anywhere else in the survey.

\FloatBarrier
\section{Industrial systems, prognostics, and digital twins}
\label{sec:industrial}

Industrial systems generate relational data of several kinds. A machine is a set of interacting components watched by many sensors, a production line is a network of stages, a physical asset can be modelled as a mesh of interacting elements, and a supply chain is a network of firms and flows. Graph methods have been applied across these, to diagnosing faults and predicting failures from sensor data, to simulating the physics of an asset so that a digital twin can mirror it, and to reasoning about the supply networks that connect industrial activity. This domain is less consolidated than the others the survey has treated, with fewer shared benchmarks and more proprietary data, but two of its threads have real substance, fault diagnosis and prognostics on instrumented machinery and graph-based physical simulation, and the account here concentrates on those while treating the supply-chain thread as the emerging area it is.

\subsection{Sensors, machines, and networks as graphs}
\label{subsec:ind-graph}

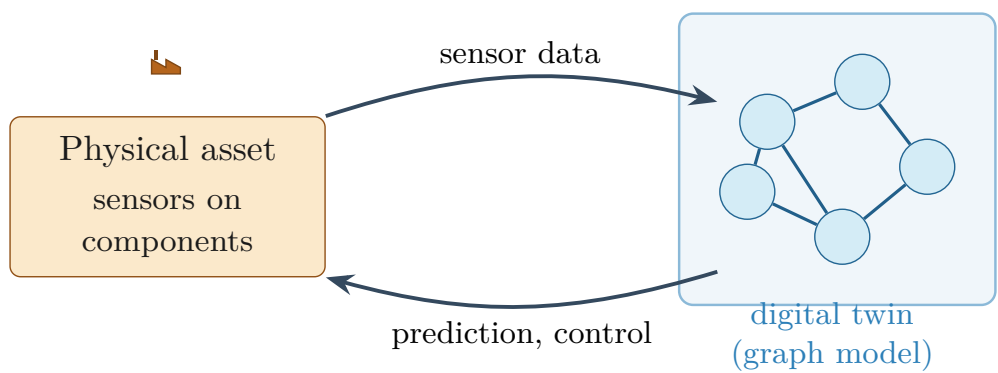
\begin{figure}[t]
\centering
\resizebox{0.82\textwidth}{!}{%
\begin{tikzpicture}[
  box/.style={rounded corners=3pt, draw=gnnrust!80!black, fill=tintamber, align=center, font=\small,
              text=gnnink, inner sep=5pt, minimum height=16mm, text width=28mm},
  gn/.style={circle, draw=gnnteal!80!black, fill=tintteal, inner sep=0pt, minimum size=5.5mm},
  ge/.style={draw=gnnteal!75!black, line width=0.9pt},
  ar/.style={-{Stealth[length=2.8mm]}, draw=gnnslate, line width=1.1pt}]
\node[box] (phys) at (0,0) {Physical asset\\[2pt]\footnotesize sensors on components};
\pic[scale=0.8] at ([yshift=5.5mm]phys.north) {ic factory={gnnrust}};
\begin{scope}[shift={(6.0,0.15)}]
  \node[gn] (a) at (0,0.6)   {};
  \node[gn] (b) at (0.95,1.0){};
  \node[gn] (c) at (1.6,0.15){};
  \node[gn] (d) at (0.75,-0.55){};
  \node[gn] (e) at (-0.2,-0.1){};
  \draw[ge] (a)--(b) (b)--(c) (c)--(d) (d)--(e) (e)--(a) (a)--(d);
  \begin{scope}[on background layer]
    \node[panel=gnnteal, fit=(a)(b)(c)(d)(e), inner sep=4mm] (dt){};
  \end{scope}
  \node[font=\footnotesize, align=center, text=gnnteal] at (0.65,-1.55) {digital twin\\(graph model)};
\end{scope}
\draw[ar] (phys.north east) to[bend left=16] node[above,font=\footnotesize]{sensor data} (5.5,0.95);
\draw[ar] (5.5,-0.75) to[bend left=16] node[below,font=\footnotesize]{prediction, control} (phys.south east);
\end{tikzpicture}}
\caption[A digital twin as a graph model of a physical asset.]{A digital twin pairs a physical asset with a graph model kept synchronized with it: sensor data flows from the asset to the model, and the model returns predictions and control decisions. The graph represents the asset's interacting components, and a fast model lets the twin run in step with the real system.}
\label{fig:digitaltwin}
\end{figure}

The constructions span scales. At the level of a single machine, the multiple sensors monitoring it, measuring vibration, temperature, current, and the like, are physically coupled because the components they watch interact, so a graph over sensors captures dependencies that treating each signal independently would miss. At the level of a physical asset, the object itself can be discretized into particles or a mesh whose elements interact locally, which is a graph suited to simulation. Learned mesh-based simulators of this kind, which predict the next state of a system by message passing over its mesh, have matched the behaviour of traditional solvers across fluids and structural mechanics at a fraction of the cost \cite{pfaff2021meshgraphnets}. At the level of an enterprise, a supply chain is a graph of suppliers, manufacturers, and distributors joined by the flow of goods. \Cref{fig:digitaltwin} illustrates the idea that unifies much of this work, the digital twin, a graph model kept synchronized with a physical asset and used to monitor it, predict its behaviour, and test interventions before applying them. The common thread is that industrial systems are made of interacting parts, and the interactions are the graph. How the sensor graph is built is itself a choice. Edges can be set from the physical layout of the machine, joining sensors on mechanically connected components, or learned from the correlations among signals, in which case the construction recovers the learned-graph idea from traffic forecasting in an industrial setting. Either way the data is a collection of synchronized time series with a relational structure over them, which makes industrial condition monitoring a spatio-temporal problem of the kind earlier chapters formalized, with sensors in place of road segments and a fault in place of a traffic event.

\subsection{Fault diagnosis and prognostics}
\label{subsec:ind-prognostics}

Condition monitoring detects and diagnoses faults in machinery from sensor data, and it is the most developed industrial application of graph networks. Because the sensors on a machine are coupled, a fault changes several signals in related ways, and a graph over the sensors captures the joint pattern that a per-sensor model fragments. Fault diagnosis classifies the type or location of a fault, and a benchmark and guideline study has organized the emerging methods and the datasets on which they are compared \cite{li2022gnnfaultbench}, with a review surveying the area more broadly \cite{chen2021faultreview}. Anomaly detection extends diagnosis to the case where the fault type is not known in advance, and graph-based detection with severity estimation has been applied to specific machinery such as three-phase induction motors \cite{bentrad2025gnnase}. Prognostics goes further still, predicting not whether a machine has failed but how long it has left, the remaining useful life, which a graph model estimates by reading a degradation trend from the evolving sensor graph,
\begin{equation}\label{eq:rul}
\widehat{r}=\psi\big(\mathrm{READOUT}(\{\hv_v^{(L)}\})\big),
\end{equation}
as demonstrated for bearing life prediction with an adaptive graph convolution \cite{wei2023rulgcn}. \Cref{alg:faultdiag} states the shared procedure. The difficulties echo those of the fraud chapter, since failures are rare, which makes the data imbalanced, and high-quality labelled failure data is expensive and proprietary, which makes it scarce, so the methods that work must handle both conditions. A few specifics ground the picture. Rotating machinery fails in characteristic ways, through bearing wear, gear damage, shaft imbalance, and misalignment, each of which leaves a signature in the vibration spectrum, and a diagnostic model learns to associate these signatures with fault types. A practical obstacle the sensor graph helps with but does not remove is the dependence on operating conditions, since a machine's signals differ with its speed and load, and a model trained at one operating point can fail at another, a distribution-shift problem that mirrors the cross-site difficulty seen in healthcare and the cross-grid difficulty seen in power systems. Prognostics adds the challenge of modelling degradation, since remaining useful life is not measured directly but inferred from how the sensor signals trend toward a failure threshold, and the inference must extrapolate beyond the data it has seen, which is intrinsically uncertain and argues for predictions that report a range rather than a single number.

\begin{algorithm}[t]
\caption{Fault diagnosis from a sensor graph}
\label{alg:faultdiag}
\KwIn{sensor graph $\graph$, multivariate sensor signals $\Feat$, trained model $f_\theta$}
\KwOut{fault type, or remaining useful life}
build node features from each sensor's recent signal window\;
$\Hid \leftarrow f_\theta(\Adj,\Feat)$\tcp*{message passing over coupled sensors}
$\hv_{\graph}\leftarrow\mathrm{READOUT}(\{\hv_v\})$\;
\Return{classified fault type, or remaining useful life by \cref{eq:rul}}
\end{algorithm}

\subsection{Physical simulation and digital twins}
\label{subsec:ind-simulation}

The most striking industrial use of graph networks is as learned simulators of physical systems, which is the computational core of a digital twin. A physical system can be represented as a graph whose nodes are particles or mesh points and whose edges join elements that interact, and a graph network can learn the system's dynamics by predicting how each element evolves from the state of its neighbours. An early formulation learned the dynamics of objects and their relations directly from data \cite{battaglia2016interaction}, and a later one scaled the idea to complex physics, simulating fluids, deformable materials, and granular media by predicting each node's motion from local interactions,
\begin{equation}\label{eq:simstep}
\xv_v^{(t+1)}=\xv_v^{(t)}+\Delta t\cdot\phi\Big(\AGG\big\{\msg(\xv_v^{(t)},\xv_u^{(t)}):u\in\nbr(v)\big\}\Big),
\end{equation} The learned simulation step, \cref{eq:simstep}, is the object that the mesh-based simulators above approximate.
and rolling the update forward in time to produce a trajectory \cite{sanchez2020simulate}. The appeal is the same that motivated learned surrogates in power systems, since a learned simulator can be faster than a traditional numerical solver and is differentiable, which makes it usable inside optimization and control, and a fast simulator is what lets a digital twin run alongside the physical asset rather than lagging it. The same caution applies as well, because the physics of these systems is often known, so a learned simulator competes with a principled solver on accuracy rather than displacing it, and it faces a difficulty specific to simulation, that small per-step errors accumulate over a long rollout and can drive the simulation away from physical plausibility, so stability over long horizons is as important as accuracy at a single step. Two extensions broaden the reach of learned simulators. Operating on a mesh rather than on free particles lets a model simulate structured systems such as deforming solids and flows around objects, and because the mesh is a graph the same message-passing machinery applies, with the added ability to adapt resolution where the physics is most active. Generalization across geometries is the property that makes such simulators useful for engineering, since a model that transfers from the shapes it trained on to new ones can explore designs without resimulating each from scratch with a slow solver. This connects industrial simulation to the broader programme of scientific machine learning, where graph networks approximate the solutions of partial differential equations, and it is one of the clearer cases in which a learned model offers something a solver does not, a fast, differentiable surrogate that can be placed inside a design-optimization loop. The long-rollout stability problem remains the binding constraint, and techniques that add noise during training, enforce conservation laws, or correct the simulation against physics are the active responses to it.

\subsection{Supply chains and manufacturing}
\label{subsec:ind-supplychain}

\begin{figure}[t]
\centering
\resizebox{0.72\textwidth}{!}{%
\begin{tikzpicture}[
  sup/.style={circle, draw=gnnblue!80!black, fill=tintblue, minimum size=7.5mm, font=\scriptsize\bfseries, text=gnnblue},
  man/.style={circle, draw=gnnteal!80!black, fill=tintteal, minimum size=7.5mm, font=\scriptsize\bfseries, text=gnnteal},
  dis/.style={circle, draw=gnnamber!80!black, fill=tintamber, minimum size=7.5mm, font=\scriptsize\bfseries, text=gnnrust},
  ret/.style={circle, draw=gnngreen!80!black, fill=tintgreen, minimum size=7.5mm, font=\scriptsize\bfseries, text=gnngreen},
  fl/.style={-{Stealth[length=2.2mm]}, draw=gnnslate, line width=1pt}]
\node[sup] (s1) at (0,2.2) {S}; \node[sup] (s2) at (0,1.1) {S}; \node[sup] (s3) at (0,0) {S};
\node[man] (m1) at (2.1,1.65){M}; \node[man] (m2) at (2.1,0.55){M};
\node[dis] (d1) at (4.1,1.1) {D};
\node[ret] (r1) at (6.0,1.75){R}; \node[ret] (r2) at (6.0,0.55){R};
\draw[fl] (s1)--(m1) (s2)--(m1) (s2)--(m2) (s3)--(m2) (m1)--(d1) (m2)--(d1) (d1)--(r1) (d1)--(r2);
\foreach \x/\t in {0/supplier,2.1/manufacturer,4.1/distributor,6.0/retailer}{
  \node[font=\scriptsize,text=gnngray] at (\x,-0.85) {\t};}
\end{tikzpicture}}
\caption[A supply network as a graph.]{A supply network as a directed graph: suppliers (S), manufacturers (M), distributors (D), and retailers (R) are nodes coloured by tier, and the flow of goods gives the edges. A disruption at one node propagates downstream along these flows, which makes risk assessment a propagation problem on the graph.}
\label{fig:supplychain}
\end{figure}
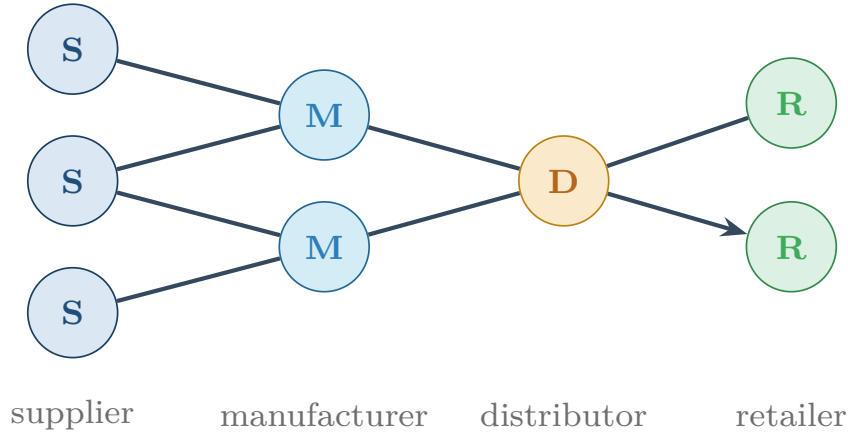

A supply chain is a graph of firms, facilities, parts, and the flows between them, and several of its problems are graph problems, sketched in \cref{fig:supplychain}. Demand must be forecast across a network of products and locations, a spatio-temporal prediction of the kind earlier chapters described. Risk and disruption propagate through the network, since a failure at one supplier cascades to everyone who depends on it, which is a propagation problem on the supply graph and one that recent global disruptions made vivid. Optimization of where to source, hold, and route inventory is an allocation over the same graph. The application of graph networks here is earlier and less consolidated than the prognostics and simulation threads, with fewer established methods and benchmarks, and it is best understood as a promising direction in which the relational structure of supply networks is a natural fit for graph reasoning rather than as a mature body of results. The problems have well-known structure graph methods are positioned to address. The bullwhip effect, in which small fluctuations in retail demand amplify into large swings upstream, is a propagation phenomenon on the supply graph, and multi-echelon inventory optimization, deciding how much stock to hold at each tier, is an allocation over it. Recent global disruptions sharpened interest in modelling how shocks cascade, since a shortage of one component can halt production far downstream in ways that are obvious only when the network is viewed as a whole. The binding obstacle is data, because a complete supply graph spans many firms that regard their supplier relationships as confidential and do not share them, so no single party observes the whole network, which limits both the modelling and its evaluation. \Cref{tab:industrial} collects the domain's applications, established and emerging alike.

\begin{table}[t]
\centering
\caption[Industrial applications of graph neural networks.]{Industrial applications, with the graph each builds and the task it poses. A representative method is named where the line has consolidated around one.}
\label{tab:industrial}
\small
\begin{tabularx}{\textwidth}{@{}lXll@{}}
\toprule
Application & Graph (nodes / edges) & Task & Example \\
\midrule
Fault diagnosis        & sensors / correlation or layout & classification          & benchmark \cite{li2022gnnfaultbench} \\
Machine anomaly        & sensors / coupling              & detection and severity  & GNN-ASE \cite{bentrad2025gnnase} \\
Remaining useful life  & sensors / correlation           & regression              & \cite{wei2023rulgcn} \\
Physical simulation    & particles or mesh / proximity   & dynamics prediction     & \cite{sanchez2020simulate} \\
Supply networks        & firms and parts / flows         & risk and forecasting    & --- \\
\bottomrule
\end{tabularx}
\end{table}

\subsection{Strengths, weaknesses, and open problems}
\label{subsec:ind-tradeoffs}

The strengths of graph methods here are genuine. Industrial systems are relational in fact rather than by analogy, since the sensors on a machine are physically coupled, the elements of a simulated asset interact locally, and a supply chain is literally a network, so the inductive bias of a graph network matches the structure. Fault diagnosis and prognostics benefit from modelling the correlations among sensors, learned physical simulators offer the speed and differentiability that make real-time digital twins possible, and the economic value of predicting failures before they happen and of reducing downtime is clear and direct.

The weaknesses are largely those of an immature field. The domain is less benchmarked than the established application areas, with fewer shared datasets and evaluation protocols, which makes results harder to compare and progress harder to measure. Failure data is rare and proprietary, since manufacturers neither experience many failures by design nor share the data when they do, so models are trained on little data that is hard to obtain. The physical-simulation thread, which is the most exciting, faces the speed-versus-accuracy tension common to physics-governed domains, compounded by the long-horizon stability problem that distinguishes simulation from one-step prediction. The supply-chain thread is early. And industrial settings are conservative, demanding reliability and interpretability before a learned model is trusted near expensive or safety-critical equipment, which widens the gap between a research result and a deployed system. The conservatism is not irrational, since a false alarm that halts a production line is costly and a missed fault that destroys equipment is worse, so the asymmetric and large costs of error in an industrial setting raise the bar for trusting an automated decision. This is why much industrial practice still pairs a learned model with established condition-monitoring methods and human expertise rather than replacing them, and why the most credible path to deployment is augmentation, a model that flags candidates for an engineer to confirm, rather than full automation.

The open problems follow from this immaturity. Standardized benchmarks that would let the field measure itself, transfer across machines and systems so a model trained on one asset informs another, stable long-horizon simulation, the integration of known physics into learned models, principled supply-chain resilience modelling, and the careful validation that conservative industrial deployment requires are the active directions. A fair summary is that industrial systems are a domain where the relational fit of graph methods is real and the economic stakes are high, where two threads, prognostics and physical simulation, have genuine substance, and where the main obstacles are the scarcity of shared data and benchmarks and the distance, familiar from the physics-governed domains, between a fast learned model and a trusted one.

\FloatBarrier
\section{Materials science and climate}
\label{sec:science}

Two scientific domains show graph methods at their most consequential. In materials science, a crystal is a graph of atoms, and graph networks both predict material properties and serve as fast, accurate models of interatomic forces that accelerate the simulation of matter. In climate and weather, the atmosphere can be discretized into a graph spanning the globe, and graph networks now forecast the weather at planetary scale, recently matching and in places surpassing the physics-based numerical models that have dominated the field for half a century. Both are cases where graph learning has produced results of real scientific and practical weight rather than incremental benchmark gains, and both share the tension, familiar from the power and industrial chapters, between a learned model and the known physics it competes with or accelerates.

\subsection{Crystals and the periodic graph}
\label{subsec:sci-crystal}

\begin{figure}[t]
\centering
\resizebox{0.5\textwidth}{!}{%
\begin{tikzpicture}[
  atom/.style={circle, draw=gnnpurple!80!black, fill=tintpurple, inner sep=0pt, minimum size=6.5mm},
  b/.style={draw=gnnpurple!75!black, line width=1.1pt},
  pb/.style={draw=gnnpurple!55, dashed, line width=0.9pt}]
\draw[gnnslate!55, line width=1.1pt, fill=gnnpurple!5] (0,0) rectangle (3,3);
\node[font=\scriptsize, text=gnnslate] at (1.5,3.35) {unit cell};
\node[atom] (a) at (0.8,2.2){};
\node[atom] (b) at (2.2,2.4){};
\node[atom] (c) at (1.5,1.0){};
\node[atom] (d) at (0.5,0.6){};
\draw[b] (a)--(b) (a)--(c) (b)--(c) (c)--(d);
\draw[pb] (b) -- (3.7,2.4);
\draw[pb] (a) -- (-0.7,2.2);
\draw[pb] (d) -- (0.5,-0.6);
\node[font=\scriptsize, text=gnnpurple!70] at (4.05,2.4)  {$\cdots$};
\node[font=\scriptsize, text=gnnpurple!70] at (-1.05,2.2) {$\cdots$};
\node[font=\scriptsize, text=gnnpurple!70] at (0.5,-0.9){$\vdots$};
\end{tikzpicture}}
\caption[A crystal as a periodic graph.]{A crystal as a periodic graph: the atoms of a repeating unit cell are nodes (purple), solid edges join neighbours within the cell, and dashed edges connect to atoms in adjacent periodic images, so the finite graph represents an infinite structure. Unlike a molecular graph, the construction must respect the crystal's periodicity.}
\label{fig:crystalgraph}
\end{figure}
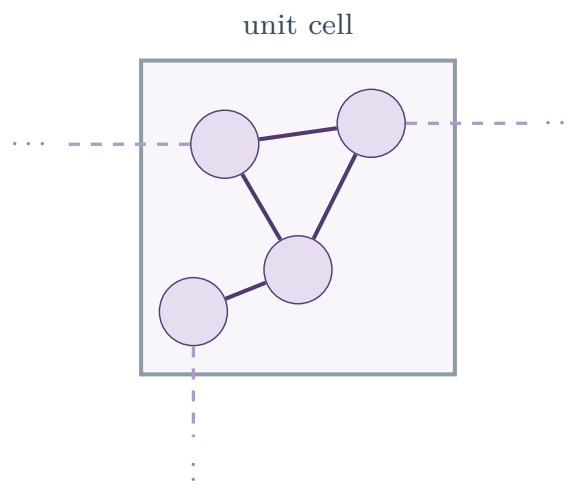

A crystal differs from a molecule in a way that matters for the graph. It is a periodic structure, in principle infinite, built by repeating a unit cell through space, and the graph must capture that periodicity rather than treating the cell as an isolated molecule. The construction, shown in \cref{fig:crystalgraph}, places the atoms of the unit cell as nodes, joins neighbours within the cell, and adds edges to atoms in adjacent periodic images, so a finite graph represents the infinite solid. As with molecules, nodes carry element identity and edges carry interatomic distances, and geometry is essential because a material's properties depend on the precise three-dimensional arrangement of its atoms. The task is to predict those properties, such as formation energy, band gap, or elastic response, from the structure, and the prediction is read from the atom representations through the same permutation-invariant readout the molecular chapter used in \cref{eq:molprop}, so the property-prediction procedure of \cref{alg:molprop} carries over with the crystal graph in place of the molecular one. Two construction choices recur. Neighbours are usually defined by a cutoff radius, joining atoms within a fixed distance, which keeps the graph finite but makes the cutoff a parameter that affects what the model can see, and some methods use several distance shells to capture interactions at different ranges. The contrast with a molecular graph is instructive, since a molecule has a definite boundary and a crystal does not, so the periodic edges that wrap across the unit cell are not an optional refinement but the feature that distinguishes the construction. Crystal symmetry adds further structure, because the space group of a crystal constrains its properties, and a model that respects the relevant symmetries needs less data to generalize, the same argument that makes equivariance valuable for molecules and, as the next subsection shows, indispensable for potentials.

\subsection{Materials property prediction}
\label{subsec:sci-property}

The foundational model in this area introduced the crystal graph and a convolution over it that predicts material properties accurately and with enough interpretability to suggest which structural features drive a property \cite{xie2018cgcnn}\footnote{Reference implementation: \url{https://github.com/txie-93/cgcnn}}. A unifying framework followed that treats molecules and crystals on the same footing, applying one architecture across both and incorporating global state alongside atom and bond features \cite{chen2019megnet}. The geometric line of the molecular chapter applies here as well, with continuous-filter convolutions that respect the spatial arrangement of atoms serving materials as they serve molecules \cite{schutt2018schnet}, and accuracy improves further when bond angles are brought in through a line-graph construction that makes the angles between neighbouring bonds first-class objects \cite{choudhary2021alignn}. These models are trained on the large materials databases assembled from quantum-mechanical calculations, and they inherit the molecular chapter's caution about generalization, since a model fitted to known materials need not extrapolate to the novel chemistries that materials discovery is actually about. The databases that train these models are large repositories of computed properties assembled by high-throughput quantum-mechanical calculation, and the workflow they enable is screening, in which a cheap graph model ranks a vast space of candidate materials so that only the most promising are computed in full or synthesized, a pattern that mirrors virtual screening in drug discovery. The inverse problem, generating new materials with desired properties rather than scoring given ones, is an active and harder direction that imports the generative methods of the molecular chapter into the periodic setting. The same baseline question the molecular chapter raised applies here too, since well-chosen physical descriptors fed to a conventional model remain a competitive baseline on some property tasks, and an honest comparison states where the learned graph representation genuinely helps rather than assuming it does. \Cref{tab:materials} collects the methods.

\subsection{Interatomic potentials and molecular dynamics}
\label{subsec:sci-potentials}

The most consequential materials application is the use of graph networks as interatomic potentials, the functions that give the energy of an atomic configuration and the forces on its atoms. Simulating the motion of atoms, in molecular dynamics, requires evaluating these forces at every step, and the field has long faced a stark trade-off: quantum-mechanical calculations are accurate but so expensive that only small systems and short times are feasible, while hand-crafted classical potentials are fast but too inaccurate for many purposes. A graph network trained on quantum data learns a potential that is both fast and close to quantum accuracy, reading the energy from the atom representations and obtaining the forces as its gradient,
\begin{equation}\label{eq:potential}
E=\sum_{v\in\vset}\epsilon\big(\hv_v^{(L)}\big),\qquad \mathbf{F}_v=-\nabla_{\xv_v}E,
\end{equation}
where the gradient is taken with respect to atomic positions. Equivariance is essential here in a way it is merely helpful elsewhere, because a force is a vector that must rotate with the system, and an equivariant network guarantees this by construction. A data-efficient equivariant potential demonstrated that high accuracy is achievable from modest training data \cite{batzner2022nequip}, and a universal potential trained across the periodic table extended the approach to arbitrary compositions, making large-scale, near-quantum simulation of materials practical \cite{chen2022m3gnet}. The significance is that this accelerates molecular dynamics by orders of magnitude while retaining accuracy, opening simulations of materials, reactions, and dynamics that were previously infeasible, and it ties this domain to the learned-simulator thread of the industrial chapter, with the same promise and the same caution about behaviour outside the training distribution. A few details convey why this thread has had such impact. A molecular-dynamics simulation evaluates forces at every one of millions of time steps, so the speed of the potential determines what is simulable, and a learned potential thousands of times faster than a quantum calculation while retaining its accuracy changes which systems and which timescales are within reach. The training data for these potentials is itself generated by quantum calculation, often through an active-learning loop in which the simulation proposes configurations, the most uncertain are computed and added to the training set, and the potential is refined, which is how a model trained on modest data can cover the configurations a long simulation visits. The message passing has a physical reading, since exchanging information between nearby atoms mirrors the local interactions that determine forces, and the universal potentials that span the periodic table point toward a future in which a single model serves as a general-purpose engine for simulating matter, much as a foundation model serves as a general engine for language.

\subsection{Weather and climate}
\label{subsec:sci-climate}

Weather forecasting has been dominated for decades by numerical weather prediction, which solves the physical equations of the atmosphere on a grid and is accurate but enormously expensive, requiring some of the largest computers in the world. The graph approach reframes the problem as learning. The globe is represented as a mesh of points joined into a graph, and a graph network learns to map the current atmospheric state to the next, rolling the prediction forward to produce a forecast,
\begin{equation}\label{eq:weatherstep}
\widehat{\mathbf{S}}^{(t+1)}=f\big(\mathbf{S}^{(t)};\graph_{\text{globe}}\big),
\end{equation} The interatomic-potential prediction and the learned weather step, \cref{eq:potential,eq:weatherstep}, anchor the materials and climate models discussed here.
which is the spatio-temporal forecasting of the transportation chapter at planetary scale. An early demonstration established that a graph network could forecast global weather at all \cite{keisler2022weather}, and a subsequent system produced skillful medium-range global forecasts that matched and in many respects exceeded the leading numerical model while running orders of magnitude faster \cite{lam2023graphcast}. The speed is not a mere convenience, since it makes large forecast ensembles affordable and brings high-quality forecasting within reach of far more institutions, and the result stands as a landmark for machine learning in the physical sciences. Climate, as distinct from weather, operates on far longer timescales and asks a model to project conditions outside the range of recent experience, which is a harder and less settled problem that the weather results do not directly resolve. The architecture behind the weather results is worth noting, since it follows the encode-process-decode pattern, mapping the gridded atmospheric state onto a mesh, propagating information across the mesh through many rounds of message passing, and decoding back to the grid, with a multi-scale mesh that lets information travel both locally and across the globe in few steps. The models are trained on decades of reanalysis, a physically consistent reconstruction of past atmospheric states, and the autoregressive rollout that produces a multi-day forecast is also the source of their main weakness, since errors compound step by step and forecasts blur or drift at longer ranges. Probabilistic forecasting, which the speed of these models makes affordable through large ensembles, is an active response, as is nowcasting, the very short-range prediction of precipitation, where graph and related learned models have also been competitive. The rapidity with which operational forecasting centres have adopted and extended these methods is itself a measure of the result's significance.

\subsection{Strengths, weaknesses, and open problems}
\label{subsec:sci-tradeoffs}

The strengths here are as strong as anywhere in the survey. Crystals and molecules are genuine graphs, learned interatomic potentials are a real breakthrough that delivers both speed and accuracy and enables science that was out of reach, and graph-based weather forecasting has matched physics-based models that took decades to build while running far faster, which is a result of the first rank. Equivariant designs that respect physical symmetry are not stylistic choices but correctness requirements that the field has learned to meet, and the practical impact, in materials discovery and in forecasting, is direct and large. There is also a democratizing dimension worth naming. Both breakthroughs replace a computation that required exceptional resources, supercomputer-scale numerical weather prediction and quantum-chemical simulation, with a learned model that runs on far more modest hardware, which puts capabilities that were the preserve of a few large institutions within reach of many more researchers. This broadening of access is part of why these results have propagated so quickly through their fields, and it is a kind of impact that benchmark accuracy alone does not capture.

The weaknesses share a single root, the difficulty of extrapolation, which is sharper here than elsewhere because extrapolation is exactly what science demands. A property model fitted to known materials may mispredict the novel chemistries that discovery targets. An interatomic potential can become unstable or unphysical when a simulation wanders outside the configurations it was trained on, the same long-rollout stability problem the industrial chapter described, now with the added danger that an unphysical trajectory can look plausible. A weather model is trained on decades of reanalysis data and inherits its biases, carries no guarantee of respecting physical conservation laws, and offers no assurance that skill on past weather transfers to a changing climate, which is the unproven leap from weather to climate. In each case the learned model is reliable in the regime it has seen and uncertain beyond it, and beyond it is where the scientific value lies.

The open problems follow directly. Out-of-distribution generalization to novel materials, potentials that remain stable and physically consistent far from their training data, the incorporation of physical constraints and conservation laws into weather and climate models, the genuine extension from weather to climate, calibrated uncertainty that a scientist can act on, and foundation models spanning atomistic and Earth systems are the active directions. A fair summary is that materials and climate are among the domains where graph methods have produced the field's most striking results, in learned potentials and in weather forecasting, and also where the defining challenge is the one science cares about most, reliable extrapolation beyond the training distribution, which remains unsolved and which separates an impressive interpolating model from a trustworthy scientific instrument.

\begin{table}[t]
\centering
\caption[Materials and climate applications of graph neural networks.]{Applications in the physical sciences, with the graph each builds and the task it poses. A representative method is named for each.}
\label{tab:materials}
\small
\begin{tabularx}{\textwidth}{@{}lXll@{}}
\toprule
Application & Graph (nodes / edges) & Task & Example \\
\midrule
Crystal property prediction  & atoms / periodic neighbours & regression              & CGCNN \cite{xie2018cgcnn} \\
Molecules and crystals       & atoms / neighbours          & regression              & MEGNet \cite{chen2019megnet} \\
Geometry-aware prediction    & atoms / distances           & regression              & SchNet \cite{schutt2018schnet} \\
Interatomic potentials       & atoms / distances           & energy and forces       & NequIP \cite{batzner2022nequip} \\
Universal potential          & atoms / distances           & periodic-table coverage & M3GNet \cite{chen2022m3gnet} \\
Global weather               & grid points / mesh          & spatio-temporal forecasting & GraphCast \cite{lam2023graphcast} \\
\bottomrule
\end{tabularx}
\end{table}

\FloatBarrier
\section{Cross-domain synthesis}
\label{sec:cross}

The twelve application areas the survey has treated differ in subject as widely as molecules differ from money, yet the chapters describing them returned again and again to the same handful of decisions and the same handful of tensions. This chapter steps back to make those commonalities explicit. The domains share a small vocabulary of graph constructions, a structural divide between fixed and transferable graphs, a recurring spatio-temporal template, a consistent pattern in where learned models genuinely win as opposed to merely accelerate, and a spread of maturity and data availability that shapes how their results should be read. Drawing these together yields lessons more general than any single domain teaches, and it sets up the question of whether the domains are alike enough that one model might serve many, which the chapter on foundation models takes up directly.

\begin{figure}[t]
\centering
\resizebox{\textwidth}{!}{%
\begin{tikzpicture}[
  root/.style={rounded corners=4pt, fill=gnnink, text=white, font=\small\bfseries, align=center, minimum width=2.2cm, minimum height=1.1cm, inner sep=4pt},
  area/.style 2 args={rounded corners=3pt, fill=#1!15, draw=#1!75!black, text=#1!72!black, font=\footnotesize\bfseries, align=center, text width=2.6cm, minimum height=1.0cm, inner sep=3pt},
  leaf/.style 2 args={rounded corners=2pt, fill=white, draw=#1!55, text=#1!78!black, font=\scriptsize, anchor=west, text width=3.5cm, minimum height=0.42cm, inner sep=2.5pt},
  lk/.style={draw=#1!55, line width=0.7pt}]
\node[root] (root) at (0,-0.3) {GNN\\applications\\across domains};
\node[area={gnnblue}{}]   (a1) at (4.7, 5.0)  {Recommendation \& social};
\node[area={gnnteal}{}]   (a2) at (4.7, 3.0)  {Knowledge \& language};
\node[area={gnngreen}{}]  (a3) at (4.7, 1.0)  {Molecules \& health};
\node[area={gnnpurple}{}] (a4) at (4.7,-1.0)  {Vision \& geometry};
\node[area={gnnamber}{}]  (a5) at (4.7,-3.2)  {Spatiotemporal \& infrastructure};
\node[area={gnnrust}{}]   (a6) at (4.7,-5.4)  {Security, industry \& materials};
\node[leaf={gnnblue}{}]   (l11) at (9.4, 5.62) {LightGCN~\cite{he2020lightgcn}};
\node[leaf={gnnblue}{}]   (l12) at (9.4, 5.00) {PinSAGE~\cite{ying2018pinsage}};
\node[leaf={gnnblue}{}]   (l13) at (9.4, 4.38) {GraphRec~\cite{fan2019graphrec}};
\node[leaf={gnnteal}{}]   (l21) at (9.4, 3.62) {R-GCN~\cite{schlichtkrull2018rgcn}};
\node[leaf={gnnteal}{}]   (l22) at (9.4, 3.00) {CompGCN~\cite{vashishth2020compgcn}};
\node[leaf={gnnteal}{}]   (l23) at (9.4, 2.38) {GraphRAG~\cite{edge2024graphrag}};
\node[leaf={gnngreen}{}]  (l31) at (9.4, 1.62) {MPNN~\cite{gilmer2017mpnn}};
\node[leaf={gnngreen}{}]  (l32) at (9.4, 1.00) {SchNet~\cite{schutt2018schnet}};
\node[leaf={gnngreen}{}]  (l33) at (9.4, 0.38) {BrainGNN~\cite{li2021braingnn}};
\node[leaf={gnnpurple}{}] (l41) at (9.4,-0.38) {DGCNN~\cite{wang2019dgcnn}};
\node[leaf={gnnpurple}{}] (l42) at (9.4,-1.00) {Point-GNN~\cite{shi2020pointgnn}};
\node[leaf={gnnpurple}{}] (l43) at (9.4,-1.62) {Graph R-CNN~\cite{yang2018graphrcnn}};
\node[leaf={gnnamber}{}]  (l51) at (9.4,-2.58) {DCRNN~\cite{li2018dcrnn}};
\node[leaf={gnnamber}{}]  (l52) at (9.4,-3.20) {Graph Neural Solver~\cite{donon2020graphsolver}};
\node[leaf={gnnamber}{}]  (l53) at (9.4,-3.82) {REGNN~\cite{eisen2020regnn}};
\node[leaf={gnnrust}{}]   (l61) at (9.4,-4.78) {CARE-GNN~\cite{dou2020caregnn}};
\node[leaf={gnnrust}{}]   (l62) at (9.4,-5.40) {GTAN~\cite{xiang2023gtan}};
\node[leaf={gnnrust}{}]   (l63) at (9.4,-6.02) {CGCNN~\cite{xie2018cgcnn}};
\foreach \a in {a1,a2,a3,a4,a5,a6} \draw[gnnslate!55, line width=0.8pt] (root.east) to[out=0,in=180] (\a.west);
\foreach \a/\l in {a1/l11,a1/l12,a1/l13} \draw[lk={gnnblue}] (\a.east) to[out=0,in=180] (\l.west);
\foreach \a/\l in {a2/l21,a2/l22,a2/l23} \draw[lk={gnnteal}] (\a.east) to[out=0,in=180] (\l.west);
\foreach \a/\l in {a3/l31,a3/l32,a3/l33} \draw[lk={gnngreen}] (\a.east) to[out=0,in=180] (\l.west);
\foreach \a/\l in {a4/l41,a4/l42,a4/l43} \draw[lk={gnnpurple}] (\a.east) to[out=0,in=180] (\l.west);
\foreach \a/\l in {a5/l51,a5/l52,a5/l53} \draw[lk={gnnamber}] (\a.east) to[out=0,in=180] (\l.west);
\foreach \a/\l in {a6/l61,a6/l62,a6/l63} \draw[lk={gnnrust}] (\a.east) to[out=0,in=180] (\l.west);
\end{tikzpicture}}
\caption[Taxonomy of representative GNN methods across the application domains.]{A taxonomy of the survey's application domains, grouped into six areas, each with representative methods discussed in the corresponding sections. The figure structures the references the survey builds on and mirrors the per-domain architecture choices recorded in \Cref{tab:archbydomain}.}
\label{fig:appmindmap}
\end{figure}
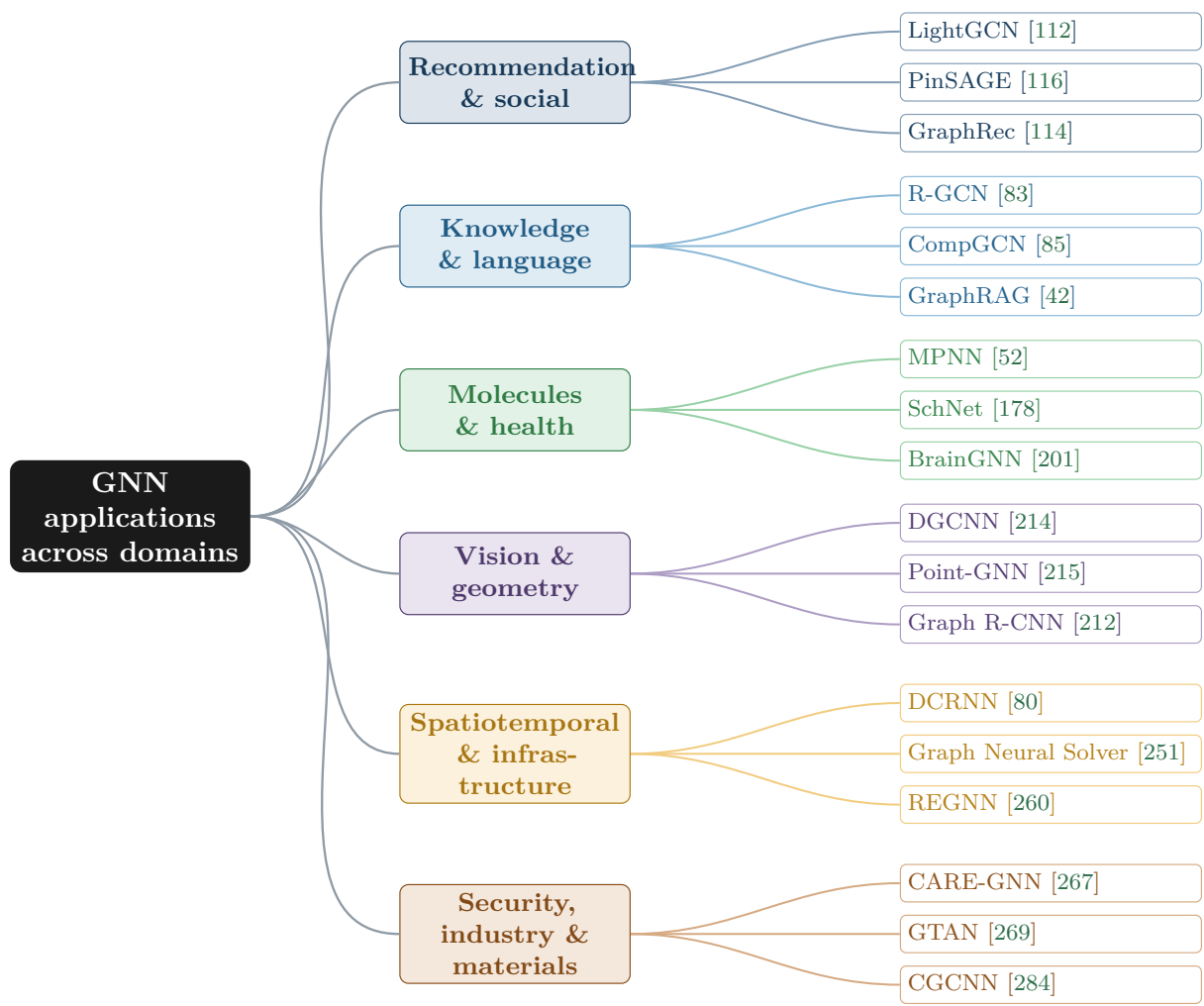

Two views summarise the comparison that follows. \Cref{fig:appmindmap} organises the methods cited across the twelve domains into six areas, and \Cref{tab:archbydomain} records, for each domain, the architecture family it leans on and the failure mode that most often bounds it.

\begin{table}[t]
\centering
\caption[GNN architecture and failure mode by application domain.]{Which architecture family each domain leans on, the graph it is applied to, and where it falls short. One or two representative methods are named per domain; the failure column states the limitation that most often bounds accuracy in that setting.}
\label{tab:archbydomain}
\footnotesize
\begin{tabularx}{\textwidth}{@{}p{1.95cm} p{2.55cm} p{2.35cm} p{2.45cm} X@{}}
\toprule
\textbf{Domain} & \textbf{Representative method(s)} & \textbf{GNN backbone} & \textbf{Graph (nodes / edges)} & \textbf{Where it falls short} \\
\midrule
Recommendation \& social & LightGCN~\cite{he2020lightgcn}, PinSAGE~\cite{ying2018pinsage} & simplified, degree-normalised GCN; sampled aggregation & users, items / interactions (bipartite) & popularity bias; reported gains often shrink against well-tuned matrix-factorisation baselines. \\
\addlinespace[2pt]
Knowledge graphs \& LMs & R-GCN~\cite{schlichtkrull2018rgcn}, CompGCN~\cite{vashishth2020compgcn} & relational message passing & entities / typed relations (multi-relational) & per-relation parameters scale with the relation count; weak on long-tail and unseen entities. \\
\addlinespace[2pt]
Molecules \& drug discovery & MPNN~\cite{gilmer2017mpnn}, D-MPNN~\cite{yang2019dmpnn} & message passing; geometric variants~\cite{schutt2018schnet} & atoms / bonds & 2D graphs miss stereochemistry; quantum-accurate targets need 3D geometry and equivariance. \\
\addlinespace[2pt]
Healthcare \& brain networks & BrainGNN~\cite{li2021braingnn} & region-graph convolution and attention & brain regions / functional or structural links & small, heterogeneous cohorts; scanner and site effects break generalisation. \\
\addlinespace[2pt]
Computer vision & DGCNN~\cite{wang2019dgcnn}, Point-GNN~\cite{shi2020pointgnn} & dynamic edge convolution on $k$-NN graphs & points or detected objects / proximity & the graph refines a CNN or transformer backbone; scene-graph quality is capped by detection. \\
\addlinespace[2pt]
Transportation \& traffic & DCRNN~\cite{li2018dcrnn}, STGCN~\cite{yu2018stgcn} & diffusion and spatio-temporal convolution with recurrence & sensors / road adjacency & gains are modest over strong temporal baselines; accuracy degrades under incidents and regime shift. \\
\addlinespace[2pt]
Power \& energy & Graph Neural Solver~\cite{donon2020graphsolver} & physics-aligned message passing & buses / transmission lines & earns speed, not accuracy, where the physics is known; brittle to unseen topologies. \\
\addlinespace[2pt]
Wireless \& 6G & REGNN~\cite{eisen2020regnn} & random-edge graph convolution & transceivers / interference and fading & fast-varying channels and topology; tight on-device compute, energy, and latency budgets. \\
\addlinespace[2pt]
Cybersecurity \& fraud & CARE-GNN~\cite{dou2020caregnn}, PC-GNN~\cite{liu2021pcgnn} & neighbour-selecting, camouflage-resistant GNN & accounts, devices / transactions & engineered heterophily (camouflage) defeats smoothing; severe class imbalance. \\
\addlinespace[2pt]
Industrial prognostics & GTAN~\cite{xiang2023gtan}, E-GraphSAGE~\cite{lo2022egraphsage} & temporal and heterogeneous GNN & sensors, assets / physical or flow links & label scarcity; distribution shift across machines and operating regimes. \\
\addlinespace[2pt]
Materials science & CGCNN~\cite{xie2018cgcnn}, MEGNet~\cite{chen2019megnet} & crystal-graph convolution & atoms in a unit cell / periodic neighbours & needs geometry, equivariance, and periodic boundaries; poor extrapolation beyond trained chemistries. \\
\addlinespace[2pt]
Climate \& simulation & Learned simulators~\cite{sanchez2020simulate} & encode--process--decode GNN & mesh or particle nodes / proximity & long-horizon rollouts drift; resolution and compute bound the achievable fidelity. \\
\addlinespace[2pt]
\bottomrule
\end{tabularx}
\end{table}

\subsection{How the graph is built}
\label{subsec:cross-construction}

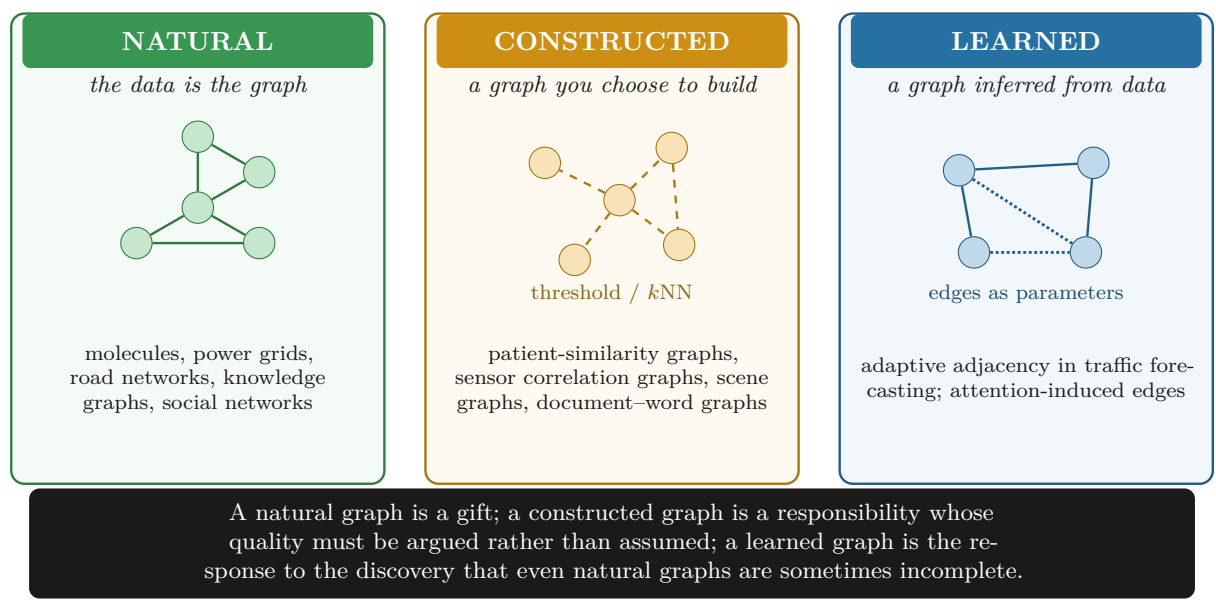
\begin{figure}[tbp]\centering
\resizebox{\textwidth}{!}{%
\begin{tikzpicture}[
  panel/.style={rounded corners=5pt, draw=#1!72!black, fill=#1!6, line width=0.9pt, minimum width=5.0cm, minimum height=6.3cm},
  hdr/.style={rounded corners=3pt, text=white, font=\bfseries\small, minimum width=4.7cm, minimum height=0.72cm, inner sep=2pt},
  sub/.style={font=\itshape\footnotesize, text=gnnink},
  ex/.style={font=\scriptsize, align=center, text width=4.5cm, text=gnnink},
  gn/.style={circle, draw=#1!70!black, fill=#1!30, minimum size=4.2mm, inner sep=0pt},
  thesis/.style={rounded corners=4pt, fill=gnnink, text=white, font=\footnotesize, align=center, text width=15.2cm, inner sep=6pt}]

% panels
\node[panel={gnngreen}]  (pN) at (2.45,0)  {};
\node[panel={gnnamber}]  (pC) at (8.0,0)   {};
\node[panel={gnnteal}]   (pL) at (13.55,0) {};
% headers
\node[hdr, fill=gnngreen!88!black] at (2.45,2.78)  {NATURAL};
\node[hdr, fill=gnnamber!88!black] at (8.0,2.78)   {CONSTRUCTED};
\node[hdr, fill=gnnteal!88!black]  at (13.55,2.78) {LEARNED};
% subtitles
\node[sub] at (2.45,2.18)  {the data is the graph};
\node[sub] at (8.0,2.18)   {a graph you choose to build};
\node[sub] at (13.55,2.18) {a graph inferred from data};

% --- natural mini-graph: fixed ring+center, solid edges ---
\begin{scope}[shift={(2.45,0.55)}]
  \node[gn={gnngreen}] (n0) at (0,0) {};
  \node[gn={gnngreen}] (n1) at (90:0.95) {};
  \node[gn={gnngreen}] (n2) at (210:0.95) {};
  \node[gn={gnngreen}] (n3) at (330:0.95) {};
  \node[gn={gnngreen}] (n4) at (30:0.95) {};
  \draw[gnngreen!70!black,line width=0.9pt] (n0)--(n1) (n0)--(n2) (n0)--(n3) (n0)--(n4) (n1)--(n4) (n2)--(n3);
\end{scope}
% --- constructed mini-graph: scattered nodes, dashed threshold edges ---
\begin{scope}[shift={(8.0,0.55)}]
  \node[gn={gnnamber}] (c1) at (-0.9,0.6) {};
  \node[gn={gnnamber}] (c2) at (0.8,0.8) {};
  \node[gn={gnnamber}] (c3) at (-0.5,-0.7) {};
  \node[gn={gnnamber}] (c4) at (0.9,-0.5) {};
  \node[gn={gnnamber}] (c5) at (0.1,0.1) {};
  \draw[gnnamber!75!black,dashed,line width=0.9pt] (c5)--(c1) (c5)--(c2) (c5)--(c3) (c5)--(c4) (c2)--(c4);
  \node[font=\scriptsize,text=gnnamber!60!black] at (0,-1.15) {threshold / $k$NN};
\end{scope}
% --- learned mini-graph: nodes + learned (dashed wavy) edges from a matrix glyph ---
\begin{scope}[shift={(13.55,0.55)}]
  \node[gn={gnnteal}] (d1) at (-0.9,0.5) {};
  \node[gn={gnnteal}] (d2) at (0.9,0.6) {};
  \node[gn={gnnteal}] (d3) at (-0.7,-0.6) {};
  \node[gn={gnnteal}] (d4) at (0.8,-0.6) {};
  \draw[gnnteal!75!black,line width=0.9pt] (d1)--(d2) (d1)--(d3) (d2)--(d4);
  \draw[gnnteal!75!black,densely dotted,line width=1.1pt] (d3)--(d4) (d1)--(d4);
  \node[font=\scriptsize,text=gnnteal!60!black] at (0,-1.15) {edges as parameters};
\end{scope}

% example lists
\node[ex] at (2.45,-1.75)  {molecules, power grids, road networks, knowledge graphs, social networks};
\node[ex] at (8.0,-1.75)   {patient-similarity graphs, sensor correlation graphs, scene graphs, document--word graphs};
\node[ex] at (13.55,-1.75) {adaptive adjacency in traffic forecasting; attention-induced edges};

% thesis strip
\node[thesis] at (8.0,-3.95) {A natural graph is a gift; a constructed graph is a responsibility whose quality must be argued rather than assumed; a learned graph is the response to the discovery that even natural graphs are sometimes incomplete.};
\end{tikzpicture}}
\caption[How a problem yields its graph.]{How a problem yields its graph. The structure is either natural and given by the data, constructed as a modelling choice the practitioner controls, or learned from data; a natural graph matches the inductive bias of a graph network, a constructed graph must justify its edges, and a learned graph addresses the cases where even a natural graph is incomplete.}
\label{fig:graphkinds}
\end{figure}

The first and most consequential decision in every domain is the graph itself, and across the survey three situations recur, set side by side in \Cref{fig:graphkinds}. In the first, the graph is natural, given by the data with no modelling choice: a molecule's atoms and bonds, a crystal's periodic lattice, a power grid's buses and lines, a road network, a social network, a knowledge graph, a communication network, a supply chain. Here the graph is a faithful representation rather than an analogy, and the inductive bias of a graph network is a genuine match to the structure. In the second, the graph is constructed, a modelling choice the practitioner controls: a population graph joining similar patients, a sensor graph built from correlations, a scene graph extracted from an image, an episode graph for few-shot learning, a document-word graph for text. Here the construction is a liability as much as a tool, because an arbitrary or poorly justified choice, the threshold that turns a brain's correlations into edges or the metric that defines patient similarity, silently determines the result. In the third, the graph is learned from data, as in the adaptive adjacency that traffic forecasting discovered it needed because the road map was an incomplete description of spatial dependence. \Cref{tab:crossdomain} classifies the domains along these lines. The lesson is compact: a natural graph is a gift, a constructed graph is a responsibility whose quality must be argued rather than assumed, and a learned graph is the field's response to the discovery that even natural graphs are sometimes incomplete. Two cross-cutting properties complicate the construction further. Many domains have heterogeneous graphs, with several types of node or edge, a knowledge graph's typed relations, a fraud graph's distinct shared-device and shared-address links, a healthcare graph's mix of patients and codes, and a model that collapses these types loses the information the types carry. Many domains also have dynamic graphs that change over time, a transaction graph that grows, a power grid that reconfigures, a social network that evolves, a wireless network whose links shift with mobility, and a model built for a static snapshot misrepresents them. The general point is that the distinction between natural, constructed, and learned graphs is the first cut, and that heterogeneity and dynamics are second cuts most real domains require, so the graph a problem actually presents is usually richer than the homogeneous static graph textbook methods assume.

\begin{table}[t]
\centering
\caption[The application domains classified by graph construction and task.]{The application domains compared by the origin of their graph, their primary task, and the characteristic that most shapes method design in each.}
\label{tab:crossdomain}
\small
\begin{tabularx}{\textwidth}{@{}lllX@{}}
\toprule
Domain & Graph origin & Primary task & Defining characteristic \\
\midrule
Social, recommendation   & natural             & link prediction          & homophily, scale \\
Knowledge graphs, NLP    & natural             & link prediction          & heterogeneous relations \\
Drugs, molecules         & natural             & graph regression         & small graphs, geometry \\
Healthcare, brain        & constructed         & node classification      & small samples, high stakes \\
Vision, point clouds     & constructed         & varied                   & graph as a component \\
Traffic                  & natural and learned & forecasting              & spatio-temporal \\
Power, energy            & natural             & surrogate, forecasting   & known physics \\
Wireless, IoT            & natural             & allocation               & equivariance, scale \\
Fraud, cybersecurity     & natural             & node, edge classification & heterophily, adversarial \\
Industrial               & constructed         & diagnosis, simulation    & scarce data \\
Materials                & natural             & graph regression         & periodicity, geometry \\
Climate, weather         & constructed mesh    & forecasting              & spatio-temporal, scale \\
\bottomrule
\end{tabularx}
\end{table}

\subsection{The transductive and inductive divide}
\label{subsec:cross-transductive}

A second structural pattern cuts across the domains and determines what generalization even means. In the transductive setting there is a single fixed graph, and the task is to predict labels for held-out nodes within it, as in classifying papers in a citation network, predicting disease across a fixed patient cohort, or completing a single knowledge graph. In the inductive setting a model must generalize to graphs or nodes it never saw in training, as when each molecule is a new graph, each image a new scene graph, each wireless deployment a new network, or each newly arrived account a node absent from the training graph. The divide governs which methods apply and how performance should be read, and it explains some of the difficulties the domain chapters reported, since the population-graph approach to disease prediction and classical knowledge-graph completion both struggle precisely because they are transductive and a new patient or entity does not fit, while molecular and wireless models generalize naturally because they were inductive from the start. The ambition of a graph foundation model is, in these terms, inductive generalization carried to its limit, a single model that transfers not only across graphs but across domains. The practical weight of this divide is easy to underestimate. A transductive model must be retrained, or at least rerun, whenever the graph gains a node, which is workable for a citation network updated occasionally but a serious limitation for a payment network that adds accounts continuously or a recommender that onboards users in real time. Inductive methods avoid this by learning a rule that applies to any node, which is why the field has moved steadily toward inductive formulations even where a transductive setup is natural. The divide also interacts with data, since an inductive model can be trained on many small graphs and deployed on new ones, whereas a transductive model is bound to the single graph it was fitted to, which limits how much data it can draw on.

\subsection{The spatio-temporal template}
\label{subsec:cross-spatiotemporal}

The most striking recurrence in the survey is a single architectural template appearing under many names. Traffic forecasting, renewable-generation forecasting, epidemic prediction, weather forecasting, and industrial condition monitoring are, structurally, the same problem: a signal living on the nodes of a graph and evolving over time, to be forecast by combining propagation over the graph with modelling of the temporal dynamics. The machinery transfers among these settings with little more than a change of vocabulary, road sensors becoming wind farms becoming atmospheric grid points, and the refinements developed in one, the learned graph of traffic forecasting and the physics-informed training of the energy and climate models, are refinements of the shared template rather than separate inventions. The practical lesson is worth stating plainly, that recognizing a new problem as spatio-temporal forecasting on a graph is most of the work of solving it, because the architectures are already in hand. The spatio-temporal template is the clearest case of a more general phenomenon, the transfer of methodological innovations across domains. Several of the field's important ideas were born in one application and then spread. The learned graph emerged from traffic forecasting and now serves any setting where the natural graph is incomplete. Equivariance was developed for molecules and is now indispensable for materials potentials and central to wireless resource allocation. Heterophily-aware aggregation was forced by fraud detection and clarifies the limits of smoothing everywhere. Physics-informed training arose in power systems and recurs in materials and climate. The lesson for a practitioner entering a new domain is that the relevant innovation has often already been worked out elsewhere under a different name, and that surveying the methods of adjacent domains is frequently more productive than inventing from scratch.

\subsection{Where learned models win, and where they only accelerate}
\label{subsec:cross-value}

\begin{figure}[t]
\centering
\begin{tikzpicture}
\begin{axis}[
  xbar, width=8.8cm, height=6.6cm,
  xmin=0, xmax=10,
  xlabel={Illustrative relational gain over a strong non-relational baseline},
  xlabel style={font=\footnotesize},
  symbolic y coords={Traffic,Recommendation,{Mol. property},Vision,Materials,{Knowledge graphs},Weather,Fraud,{Mol. discovery}},
  ytick=data,
  yticklabel style={font=\scriptsize},
  xticklabel style={font=\scriptsize},
  bar width=8pt,
  enlarge y limits=0.07,
  every axis plot/.append style={fill=gnnteal, draw=gnnteal!80!black},
  xmajorgrids=true, grid style={gnngray!25}]
\addplot coordinates {(3,Traffic)(3,Recommendation)(4,{Mol. property})(5,Vision)(5,Materials)(7,{Knowledge graphs})(8,Weather)(9,Fraud)(9,{Mol. discovery})};
\end{axis}
\end{tikzpicture}
\caption[Illustrative relational gain across domains.]{Illustrative, qualitative ranking of how much the graph structure adds over a strong non-relational baseline, by domain. The bars are schematic and express only the broad pattern argued in the text: graph methods add most where relational structure carries signal a non-relational model cannot access, such as coordination in fraud or geometry in discovery, and least where a strong non-relational baseline already captures most of the signal, as in traffic and recommendation. No measured quantity is plotted.}
\label{fig:domainshare}
\end{figure}
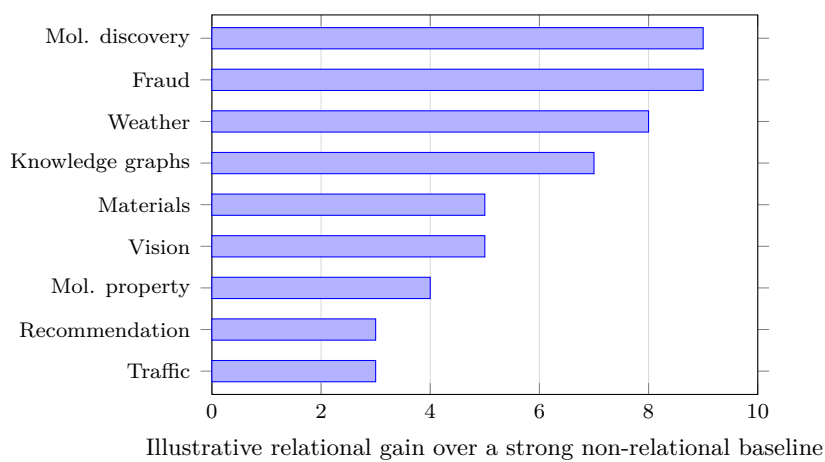

The most important synthesis concerns value, because the contribution of a graph network varies sharply across the domains and is easy to overstate. A prediction in the message-passing framework reads from both a node's own features and its aggregated neighbourhood,
\begin{equation}\label{eq:relgain}
\hat{y}_v=g\big(\xv_v,\ \textstyle\AGG_{u\in\nbr(v)}\hv_u\big),
\end{equation} The relational-gain decomposition, \cref{eq:relgain}, makes precise the comparison drawn across domains in this section.
and the graph earns its place exactly when the second argument carries signal the first lacks. Three regimes follow, sketched qualitatively in \cref{fig:domainshare}. In the first, the graph reveals something a non-relational model genuinely cannot see, and the gain is large: the coordination of a fraud ring is invisible in any single account, the discovery of a new antibiotic turned on structure no per-molecule descriptor captured, weather forecasting matched physics-based models by propagating information across the globe, and relational structure is the entire content of a scene graph. In the second regime the underlying physics is known and exact, and the learned model competes not on accuracy but on speed: power-flow surrogates, interatomic potentials, and learned physical simulators are valuable because they are orders of magnitude faster than the principled computations they approximate and because they are differentiable, not because they are more correct, and the reliability bar they must clear is correspondingly high. A consideration cutting across the regimes is scale, since a method that wins on a small benchmark may be infeasible on a real graph of millions of nodes, so the practical value of a graph network depends not only on its accuracy but on whether it runs at the size the application demands, a point the next chapter develops. In the third regime a strong non-relational baseline already captures most of the signal, and the relational gain, though real, is modest: a tuned temporal model rivals elaborate spatio-temporal networks on traffic benchmarks, matrix factorization remains competitive in recommendation, and well-chosen fingerprints rival graph networks on some molecular properties. The pattern is consistent and worth internalizing, that graph methods add most where relational structure carries inaccessible signal and least where a strong non-relational baseline suffices, and that honest evaluation in the third regime is what separates a real contribution from a restatement of what simpler models already achieve.

\subsection{Assumptions, maturity, and data}
\label{subsec:cross-maturity}

\begin{figure}[t]
\centering
\resizebox{0.82\textwidth}{!}{%
\begin{tikzpicture}[
  pt/.style={draw=#1!80!black, fill=#1, circle, inner sep=0pt, minimum size=8pt}, pt/.default=gnnteal,
  lbl/.style={font=\scriptsize, text=gnnslate}]
\draw[-{Stealth[length=2.6mm]},gnnslate,line width=1.1pt] (0,0)--(11.6,0) node[right,font=\footnotesize]{maturity};
\draw[-{Stealth[length=2.6mm]},gnnslate,line width=1.1pt] (0,0)--(0,7.8) node[above,font=\footnotesize]{impact};
\foreach \x in {2,4,6,8,10} \draw[gnngray!15] (\x,0.05)--(\x,7.4);
\foreach \y in {2,4,6} \draw[gnngray!15] (0.05,\y)--(11.0,\y);
% mature + high impact (green)
\node[pt=gnngreen] (tra) at (8.6,6.6){};  \node[lbl,above=2.5pt] at (tra){traffic};
\node[pt=gnngreen] (rec) at (9.4,5.3){};  \node[lbl,right=3pt] at (rec){recommendation};
\node[pt=gnngreen] (kg)  at (9.1,4.1){};  \node[lbl,right=3pt] at (kg){knowledge graphs};
\node[pt=gnngreen] (mol) at (7.6,5.7){};  \node[lbl,above=2.5pt] at (mol){molecules};
% high impact, mid maturity (teal / amber)
\node[pt=gnnamber] (wea) at (4.9,6.7){};  \node[lbl,above=2.5pt] at (wea){weather};
\node[pt=gnnteal]  (fra) at (6.4,6.1){};  \node[lbl,above=2.5pt] at (fra){fraud};
\node[pt=gnnteal]  (mat) at (6.3,4.3){};  \node[lbl,below=2.5pt] at (mat){materials};
\node[pt=gnnteal]  (vis) at (6.5,3.3){};  \node[lbl,right=3pt] at (vis){vision};
% lower maturity / impact (amber / red)
\node[pt=gnnamber] (pow) at (4.5,4.5){};  \node[lbl,left=3pt] at (pow){power};
\node[pt=gnnamber] (wir) at (3.0,3.7){};  \node[lbl,left=3pt] at (wir){wireless, 6G};
\node[pt=gnnred]   (hea) at (3.7,2.3){};  \node[lbl,right=3pt] at (hea){healthcare};
\node[pt=gnnred]   (ind) at (2.4,1.9){};  \node[lbl,below=2.5pt] at (ind){industrial};
\end{tikzpicture}}
\caption[Illustrative placement of domains by maturity and impact.]{Illustrative placement of the application domains by the maturity of their methods (horizontal) against their realized real-world impact (vertical), coloured from early-stage (red) through emerging (amber, teal) to mature-and-deployed (green). Positions are the survey's qualitative judgement, not measured coordinates, and convey only the broad landscape: some areas are both mature and deployed, weather forecasting is newer yet already high in impact, and several areas remain early on both axes. No quantity is measured.}
\label{fig:domainbubble}
\end{figure}
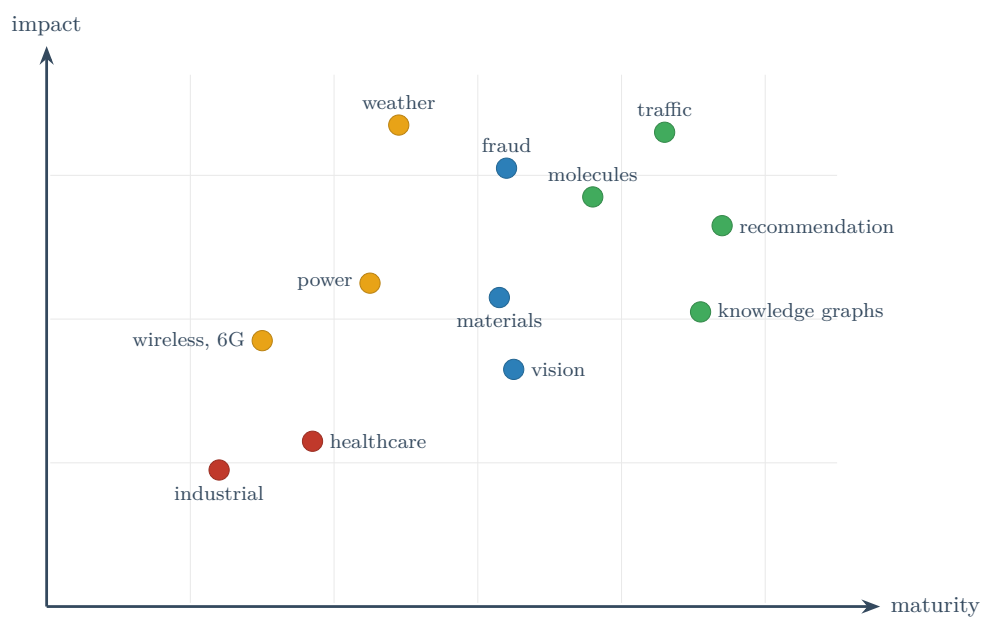

Three further axes separate the domains. The first is whether the homophily assumption holds, that connected nodes are similar, which underlies the smoothing that ordinary graph networks perform. It holds in social and citation networks, where the default methods work well, and it fails in fraud and anomaly detection, where adversaries connect to legitimate nodes to disguise themselves and where the smoothing that helps elsewhere actively erases the signal, so off-the-shelf methods underperform and task-specific designs are mandatory. The second axis is data. Some domains enjoy rich public benchmarks, molecular property suites, citation networks, traffic sensor archives, knowledge-graph completion sets, and they progress quickly, though they risk overfitting to the same few datasets; others, healthcare, power grids, industrial systems, and to a degree fraud, work with scarce and proprietary data, which slows measurable progress and widens the uncertainty around deployment. The benchmark-rich domains carry a subtler risk worth naming, that a field measuring itself on the same few datasets for years can mistake adaptation to those datasets for genuine progress, optimizing architectures against a fixed target until the gains reflect the benchmark's quirks rather than the underlying problem. The molecular and citation-network areas have both confronted versions of this concern, and the corrective, harder splits, larger and more varied datasets, and evaluation that mimics deployment, is itself an active line of work. The third axis is maturity, ranging from settled areas with deployed systems, recommendation, molecular property prediction, traffic forecasting, knowledge-graph completion, to emerging ones, supply-chain modelling, climate as distinct from weather, and sixth-generation networks. \Cref{tab:domainmaturity} records a qualitative assessment of the domains along these axes, and \cref{fig:domainbubble} places them by maturity against real-world impact. Running through all three axes is the calibration theme the survey has pressed in every chapter, that the gap between a benchmark number and a deployed system is widest exactly where the stakes are highest, in healthcare, in safety-critical power systems, and in the adversarial setting of fraud, and that naming this gap honestly is a recurring obligation rather than a one-time caveat.

\begin{table}[t]
\centering
\caption[A qualitative assessment of the domains by data, maturity, and deployment.]{A qualitative assessment of the domains, using the symbolic scale defined in the text. The ratings reflect the survey's reading of each area rather than a measured quantity, and the final column rates how well current benchmarks reflect deployment conditions.}
\label{tab:domainmaturity}
\small
\setlength{\tabcolsep}{5pt}
\renewcommand{\arraystretch}{1.2}
\begin{tabular}{@{}lcccc@{}}
\toprule
Domain & \makecell{Public\\data} & \makecell{Method\\maturity} & \makecell{Real\\deployment} & \makecell{Benchmark\\realism} \\
\midrule
Social, recommendation & \hbfull  & \hbfull  & \hbfull  & \hbhalf  \\
Knowledge graphs, NLP  & \hbfull  & \hbfull  & \hbfull  & \hbhalf  \\
Drugs, molecules       & \hbfull  & \hbfull  & \hbhalf  & \hbhalf  \\
Healthcare, brain      & \hbhalf  & \hbhalf  & \hbempty & \hbempty \\
Vision, point clouds   & \hbfull  & \hbhalf  & \hbhalf  & \hbhalf  \\
Traffic                & \hbfull  & \hbfull  & \hbfull  & \hbhalf  \\
Power, energy          & \hbempty & \hbhalf  & \hbhalf  & \hbempty \\
Wireless, IoT          & \hbhalf  & \hbhalf  & \hbempty & \hbempty \\
Fraud, cybersecurity   & \hbhalf  & \hbfull  & \hbfull  & \hbempty \\
Industrial             & \hbempty & \hbempty & \hbhalf  & \hbempty \\
Materials              & \hbfull  & \hbfull  & \hbhalf  & \hbhalf  \\
Climate, weather       & \hbhalf  & \hbhalf  & \hbhalf  & \hbhalf  \\
\bottomrule
\end{tabular}
\end{table}

\Cref{tab:datasets} consolidates the benchmark datasets that recur across these domains, listing the approximate scale, the task, and the metric usually reported for each.

\begin{table}[t]
\centering
\caption[Representative benchmark datasets across domains.]{Representative benchmark datasets used across the application domains, with approximate scale, the task they pose, and the metric usually reported. Scales are order-of-magnitude rather than exact, since releases vary. The table consolidates into one place the datasets named throughout the domain chapters.}
\label{tab:datasets}
\footnotesize
\begin{tabularx}{\textwidth}{@{}lllXl@{}}
\toprule
Dataset & Domain & Scale & Task & Metric \\
\midrule
Cora, Citeseer, Pubmed              & citation            & $10^3$--$10^4$ nodes      & node classification    & accuracy \\
ogbn-arxiv, -products \cite{hu2020ogb} & citation, co-purchase & $10^5$--$10^6$ nodes   & node classification    & accuracy \\
Reddit \cite{hamilton2017graphsage} & social              & $\sim$233K nodes         & node classification    & micro-F1 \\
PPI \cite{hamilton2017graphsage}    & biology             & 24 graphs                & multi-label node cls.  & micro-F1 \\
MUTAG, PROTEINS, NCI1               & molecules, bio      & $10^2$--$10^4$ graphs    & graph classification   & accuracy \\
QM9 \cite{gilmer2017mpnn}           & molecules           & $\sim$134K molecules     & property regression    & MAE \\
ZINC                                & molecules           & $\sim$250K molecules     & property regression    & MAE \\
MoleculeNet \cite{wu2018moleculenet} & molecules          & varies                   & property prediction    & RMSE / AUC \\
FB15k-237, WN18RR                   & knowledge graphs    & $10^4$ entities          & link prediction        & MRR, Hits@$k$ \\
ogbl-citation2 \cite{hu2020ogb}     & citation            & $\sim$3M nodes           & link prediction        & MRR \\
Visual Genome                       & vision (scenes)     & $\sim$108K images        & scene graph generation & Recall@$k$ \\
NTU RGB+D                           & vision (action)     & $\sim$57K clips          & action recognition     & accuracy \\
METR-LA, PEMS-BAY \cite{li2018dcrnn} & traffic            & 207 / 325 sensors        & forecasting            & MAE, RMSE \\
PEMS03/04/07/08 \cite{song2020stsgcn} & traffic           & $10^2$--$10^3$ sensors   & forecasting            & MAE \\
MovieLens, Amazon                   & recommendation      & $10^6$--$10^8$ interact. & rating, ranking        & NDCG, Recall \\
Elliptic, Yelp, Amazon              & fraud               & $10^4$--$10^6$ nodes     & node classification    & F1, AUC \\
Materials Project \cite{xie2018cgcnn} & materials         & $\sim$$10^5$ materials   & property regression    & MAE \\
OQMD, JARVIS \cite{chen2019megnet}  & materials           & $10^5$--$10^6$ entries   & property regression    & MAE \\
ABIDE, ADNI                         & healthcare (brain)  & $10^2$--$10^3$ subjects  & disease classification & accuracy \\
ERA5, WeatherBench \cite{lam2023graphcast} & climate      & global grid              & forecasting            & RMSE, ACC \\
\bottomrule
\end{tabularx}
\end{table}

\subsection{Lessons for practitioners}
\label{subsec:cross-lessons}

The domain chapters support a short list of practical guidance. Begin with the graph construction, since it is the most consequential choice and the one most often made by default; a constructed graph in particular deserves an argument for its edges rather than an arbitrary threshold. Establish strong non-relational baselines before claiming a relational gain, because in a large share of domains a well-tuned model that ignores the graph captures most of the signal, and the relational contribution must be measured against it rather than presumed. Determine whether the domain is homophilous, in which case standard methods apply, or heterophilous, in which case they must be redesigned. Recognize when the governing physics is known, because then a graph network's value is speed rather than accuracy and the bar for trusting it is high. Look for spatio-temporal structure, since finding it lets the shared template be reused. Match the method to the transductive or inductive setting the problem actually presents. And state the gap between benchmark and deployment plainly, especially where the stakes are high, since the survey's repeated finding is that this gap is the rule rather than the exception. One further lesson sits beneath the others. The choice of whether to use a graph network at all is itself a decision to be made deliberately rather than assumed, since the survey's domains show that a graph model is sometimes the clear right answer, sometimes a fast approximation to a principled method, and sometimes an elaborate way to match what a simpler model already does, and asking which of these a given problem is, before reaching for the most sophisticated architecture, is the single most useful habit the cross-domain view recommends.

\subsection{Toward generalization across domains}
\label{subsec:cross-generalization}

The synthesis points to a single conclusion that frames the remainder of the survey. The domains are more alike than their surface differences suggest, sharing a small set of graph constructions, a common spatio-temporal template, a recurring transductive-inductive divide, and a consistent pattern of where relational structure helps. This commonality is the premise behind graph foundation models, the proposition that if the domains share structure then a single model might serve many of them, transferring across graphs and tasks the way a language model transfers across text. The obstacles are equally real and equally rooted in the synthesis, since the domains differ in graph type, in whether they are transductive or inductive, and in domain-specific assumptions such as homophily that no single model trivially spans. There is also a deeper obstacle the synthesis exposes. A language model succeeds in part because text is a single modality with a shared vocabulary, whereas graphs are a family of objects with no shared feature space, a molecule's atom types and a social network's user attributes having nothing in common, so a graph foundation model must solve a feature-alignment problem a language model never faces. 

\FloatBarrier
\section{Challenges and open problems}
\label{sec:challenges}

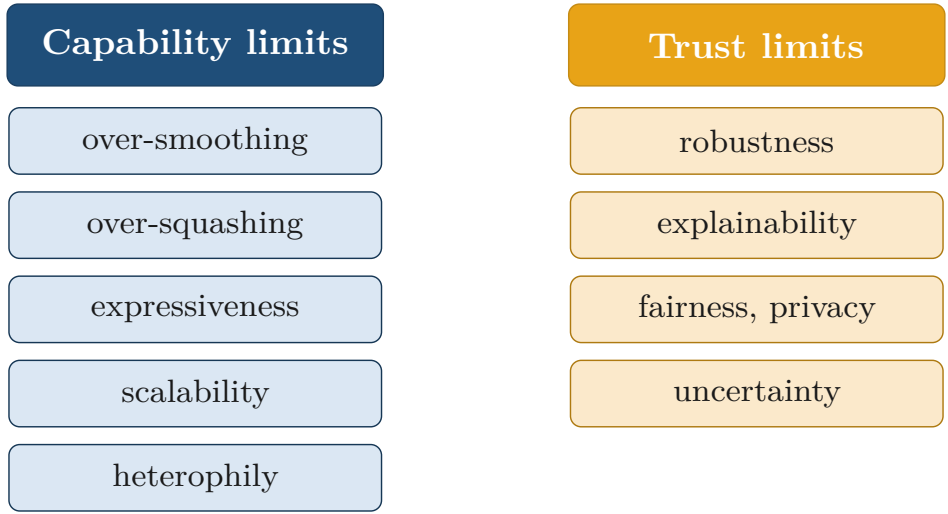
\begin{figure}[t]
\centering
\resizebox{0.78\textwidth}{!}{%
\begin{tikzpicture}[
  hdrC/.style={rounded corners=3pt, draw=gnnblue!80!black, fill=gnnblue, text=white, font=\small\bfseries, inner sep=4pt, text width=34mm, align=center, minimum height=8mm},
  itemC/.style={rounded corners=3pt, draw=gnnblue!70!black, fill=tintblue, font=\footnotesize, inner sep=3.5pt, text width=34mm, align=center, minimum height=6.5mm, text=gnnink},
  hdrT/.style={rounded corners=3pt, draw=gnnamber!85!black, fill=gnnamber, text=white, font=\small\bfseries, inner sep=4pt, text width=34mm, align=center, minimum height=8mm},
  itemT/.style={rounded corners=3pt, draw=gnnamber!75!black, fill=tintamber, font=\footnotesize, inner sep=3.5pt, text width=34mm, align=center, minimum height=6.5mm, text=gnnink}]
\node[hdrC] (cap) at (0,0) {Capability limits};
\node[itemC, below=2mm of cap]  (i1) {over-smoothing};
\node[itemC, below=1.6mm of i1] (i2) {over-squashing};
\node[itemC, below=1.6mm of i2] (i3) {expressiveness};
\node[itemC, below=1.6mm of i3] (i4) {scalability};
\node[itemC, below=1.6mm of i4] (i5) {heterophily};
\node[hdrT, right=18mm of cap] (tr) {Trust limits};
\node[itemT, below=2mm of tr]   (j1) {robustness};
\node[itemT, below=1.6mm of j1] (j2) {explainability};
\node[itemT, below=1.6mm of j2] (j3) {fairness, privacy};
\node[itemT, below=1.6mm of j3] (j4) {uncertainty};
\end{tikzpicture}}
\caption[A map of the challenges facing graph neural networks.]{The challenges grouped into two families: limits on what graph networks can compute (blue) and limits on whether they can be trusted (amber). The families are connected, since the smoothing behind over-smoothing also underlies the difficulty with heterophily, and the trust limits interact throughout.}
\label{fig:challenge}
\end{figure}

The application chapters showed graph networks succeeding across a wide range of domains, but they also returned, again and again, to the same set of limitations. This chapter gathers those limitations into a systematic treatment of the technical challenges that constrain the field. They divide naturally into two families, sketched in \cref{fig:challenge}: limits on what graph networks can compute, including the difficulties of depth, the bounds on expressiveness, the obstacles to scale, and the failure of the homophily assumption, and limits on whether they can be trusted, including their vulnerability to attack, their opacity, their potential unfairness, and their poor calibration. The two families are connected rather than separate, since the smoothing that causes the depth problem also underlies the difficulty with heterophily, and the trust limits interact with one another throughout, so progress on the field's larger ambitions, deployment in high-stakes settings and the foundation models of the next chapter, depends on the whole connected set. It is worth placing these difficulties in perspective. The field's first phase established that graph networks worked, often spectacularly, across the domains the survey has covered, and the present phase is increasingly about understanding why they fail when they do and what they cannot do at all. The challenges below are the product of that maturing, since each was identified by pushing the methods until they broke, and none is a reason to doubt the field's value so much as a map of where its current methods stop. The capability limits bound what is computable with today's architectures, and the trust limits bound what is deployable with today's guarantees, and a reader deciding whether to apply graph learning to a new problem is well served by knowing both.

\subsection{The limits of depth}
\label{subsec:ch-depth}

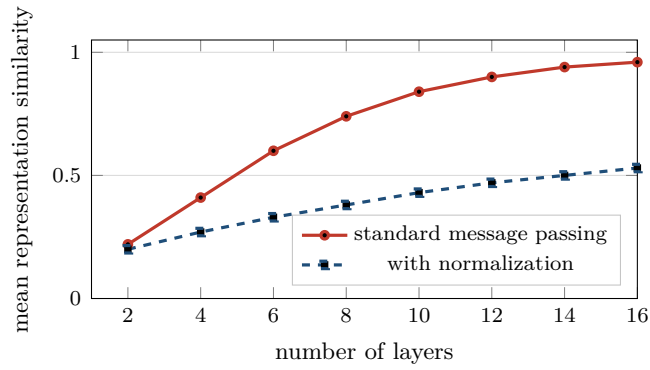
\begin{figure}[t]
\centering
\begin{tikzpicture}
\begin{axis}[width=8.8cm, height=5cm,
  xlabel={number of layers}, ylabel={mean representation similarity},
  xlabel style={font=\footnotesize}, ylabel style={font=\footnotesize},
  xticklabel style={font=\scriptsize}, yticklabel style={font=\scriptsize},
  xmin=1, xmax=16, ymin=0, ymax=1.05, ymajorgrids=true, grid style={gnngray!25},
  legend style={font=\scriptsize, at={(0.97,0.05)}, anchor=south east, draw=gnngray!40},
  every axis plot/.append style={line width=1.2pt}]
\addplot[mark=*, mark size=1.4pt, draw=gnnred] coordinates {(2,0.22)(4,0.41)(6,0.60)(8,0.74)(10,0.84)(12,0.90)(14,0.94)(16,0.96)};
\addlegendentry{standard message passing}
\addplot[mark=square*, mark size=1.4pt, draw=gnnblue, dashed] coordinates {(2,0.20)(4,0.27)(6,0.33)(8,0.38)(10,0.43)(12,0.47)(14,0.50)(16,0.53)};
\addlegendentry{with normalization}
\end{axis}
\end{tikzpicture}
\caption[Illustrative over-smoothing with depth.]{Illustrative growth in the average similarity between node representations as layers are added, the signature of over-smoothing: standard message passing (red) drives representations together quickly, while normalization (blue) slows the collapse. The curves are schematic and convey only the qualitative effect, not measured values.}
\label{fig:depthsmoothing}
\end{figure}

A graph network of $L$ layers lets information travel $L$ hops, so a task whose answer depends on distant nodes appears to demand a deep network. In practice depth is sharply limited by two phenomena. The first is over-smoothing. Each message-passing layer averages a node with its neighbours, which is a smoothing operation, and repeated smoothing drives the representations of all nodes toward a common value, erasing the distinctions a classifier needs. The effect can be measured by the Dirichlet energy of the representations, which sums the differences across edges,
\begin{equation}\label{eq:smoothenergy}
\mathcal{E}\big(\Hid^{(l)}\big)=\sum_{(u,v)\in\eset}\big\lVert\hv_u^{(l)}-\hv_v^{(l)}\big\rVert^2,
\end{equation}
and which decays toward zero as layers accumulate, so that node representations become indistinguishable; the expressive power of a message-passing network for node classification has been shown to decay exponentially with depth for this reason \cite{oono2020expressive}. The second phenomenon is over-squashing. The number of nodes within $L$ hops grows roughly exponentially, yet their information is compressed into a fixed-size vector at the receiving node, so signal from distant nodes is squeezed through bottlenecks and lost. The sensitivity of a node's representation to a distant node,
\begin{equation}\label{eq:squash}
\Big\lVert\frac{\partial\,\hv_v^{(L)}}{\partial\,\xv_u}\Big\rVert\ \longrightarrow\ 0\quad\text{as }d(u,v)\text{ grows},
\end{equation}
falls off with the distance between them, which means long-range dependencies are learned poorly however many layers are added. Together these produce a dilemma, since tasks that need long-range interaction need depth, and depth brings over-smoothing and over-squashing, and \cref{fig:depthsmoothing} illustrates how quickly representations collapse together as layers increase. The remedies treat the symptoms rather than dissolving the tension: residual and dense connections that carry earlier representations forward, normalization schemes that rescale representations to keep them apart, of which one tackles over-smoothing directly \cite{zhao2020pairnorm}, and the adaptation of deep-network engineering that let point-cloud models stack many layers. None removes the underlying difficulty, that the smoothing which makes message passing work is also what limits how far it can reach. Two further observations sharpen the picture. The spectral reading of over-smoothing is that repeated aggregation acts as a low-pass filter whose effect is to project representations onto the smoothest signal the graph supports, so that with enough layers every node collapses toward a function of its connected component alone, a rank collapse no amount of width repairs. Over-squashing has been linked to the geometry of the graph, specifically to edges of negative curvature that act as bottlenecks, which has motivated rewiring methods that add or modify edges to widen the channels through which distant information must pass, trading fidelity to the original graph for reach. The empirical consequence of both is a shallow sweet spot, since most successful models use only two to four layers, and the construction of benchmarks that genuinely require long-range interaction has become a way to measure progress against precisely the limitation shallow models cannot escape.

\subsection{Expressiveness}
\label{subsec:ch-expressive}

\begin{figure}[tbp]\centering
\resizebox{0.95\textwidth}{!}{%
\begin{tikzpicture}[
  box/.style={rounded corners=5pt, draw=#1!72!black, fill=#1!7, line width=0.9pt, minimum width=4.7cm, minimum height=4.6cm},
  nd/.style={circle, draw=#1!78!black, fill=#1!35, minimum size=4.6mm, inner sep=0pt},
  ttl/.style={font=\bfseries\small},
  verdict/.style={rounded corners=4pt, draw=gnnamber!78!black, fill=gnnamber!13, line width=1pt, align=center, font=\footnotesize, text width=3.0cm, inner sep=5pt},
  thesis/.style={rounded corners=4pt, fill=gnnink, text=white, font=\footnotesize, align=center, text width=14.4cm, inner sep=6pt}]
\node[box={gnnblue}] (BL) at (2.5,0) {};
\node[ttl, text=gnnblue!78!black] at (2.5,1.95) {Two triangles};
\begin{scope}[shift={(2.5,0.05)}]
  \node[nd={gnnblue}] (t1a) at (-0.9,1.0) {};
  \node[nd={gnnblue}] (t1b) at (-1.5,0.0) {};
  \node[nd={gnnblue}] (t1c) at (-0.3,0.0) {};
  \draw[gnnblue!70!black,line width=0.9pt] (t1a)--(t1b)--(t1c)--(t1a);
  \node[nd={gnnblue}] (t2a) at (0.95,-0.15) {};
  \node[nd={gnnblue}] (t2b) at (0.35,-1.15) {};
  \node[nd={gnnblue}] (t2c) at (1.55,-1.15) {};
  \draw[gnnblue!70!black,line width=0.9pt] (t2a)--(t2b)--(t2c)--(t2a);
\end{scope}
\node[font=\scriptsize, text=gnnink] at (2.5,-2.0) {two disjoint 3-cycles};
\node[box={gnnteal}] (BR) at (12.0,0) {};
\node[ttl, text=gnnteal!72!black] at (12.0,1.95) {One six-cycle};
\begin{scope}[shift={(12.0,0.05)}]
  \foreach \a/\i in {90/1,150/2,210/3,270/4,330/5,30/6} {\node[nd={gnnteal}] (h\i) at (\a:1.25) {};}
  \draw[gnnteal!70!black,line width=0.9pt] (h1)--(h2)--(h3)--(h4)--(h5)--(h6)--(h1);
\end{scope}
\node[font=\scriptsize, text=gnnink] at (12.0,-2.0) {a single 6-cycle};
\node[verdict] (V) at (7.25,0.15) {$f(G_1)=f(G_2)$\\[2pt]\scriptsize identical 1-WL colours};
\draw[-{Stealth[length=2.2mm]}, gnngray, line width=0.8pt] (BL.east) -- (V.west);
\draw[-{Stealth[length=2.2mm]}, gnngray, line width=0.8pt] (BR.west) -- (V.east);
\node[font=\scriptsize\itshape, text=gnnred!82!black] at (7.25,-1.45) {but $G_1\not\cong G_2$};
\node[thesis] at (7.25,-3.45) {Both graphs are 2-regular, so the one-dimensional Weisfeiler--Leman test, and any message-passing network whose power it bounds, gives every node the same colour and the two graphs the same embedding, although they are not isomorphic. Separating them needs power beyond 1-WL: higher-order message passing, or positional and structural features.};
\end{tikzpicture}}
\caption[Two graphs a message-passing network cannot tell apart.]{Two graphs a message-passing network cannot tell apart. Two disjoint triangles (left) and a single six-cycle (right) are both 2-regular, so the one-dimensional Weisfeiler--Leman test assigns every node the same colour in both graphs; a standard message-passing network, whose distinguishing power that test bounds \cite{xu2019gin}, therefore maps the two graphs to the same embedding even though they are not isomorphic. This is the expressiveness ceiling of \Cref{eq:wlbound}, and escaping it needs power beyond 1-WL, such as higher-order message passing or positional and structural features.}
\label{fig:expressivity}
\end{figure}
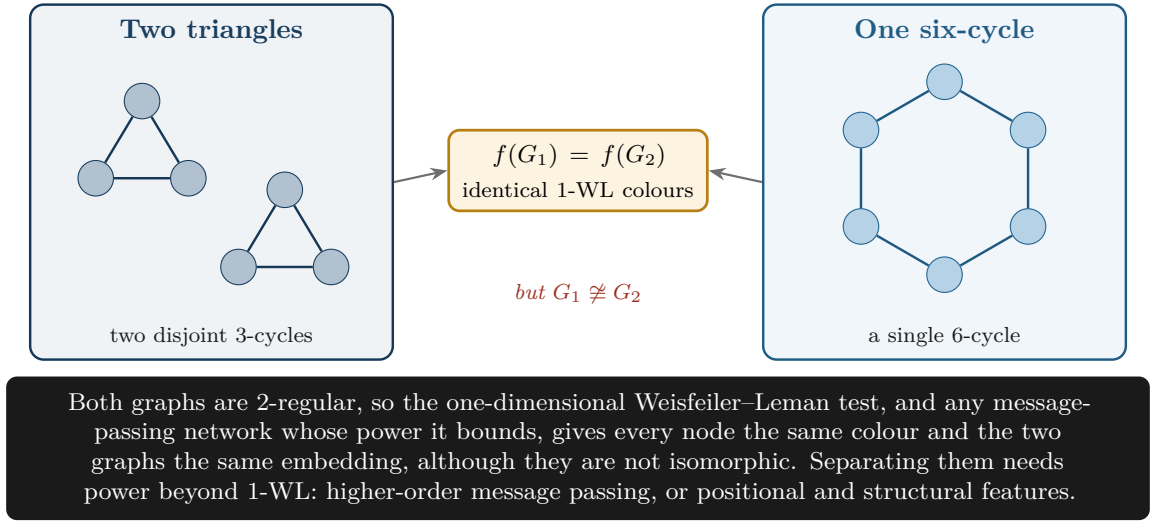

A separate limit concerns what message passing can compute even in principle. The foundations chapter noted that a standard message-passing network is at most as powerful as the first-order Weisfeiler-Leman test at distinguishing graphs, and the bound is exact: if a network assigns two graphs different representations, the test also separates them,
\begin{equation}\label{eq:wlbound}
f(G_1)\neq f(G_2)\ \Longrightarrow\ \mathrm{WL}(G_1)\neq \mathrm{WL}(G_2),
\end{equation}
so any two graphs the test deems equivalent are indistinguishable to the network. The consequences are concrete. Such a network cannot count many substructures, including triangles and longer cycles, cannot tell certain regular graphs apart, the standard case of which \Cref{fig:expressivity} shows, and therefore cannot represent properties that depend on these features, a limit the molecular chapter met in the form of molecules that differ chemically but look identical to the network. More powerful models exist, built to match higher-order variants of the test by operating on tuples of nodes rather than single nodes \cite{morris2019wlneural}, but their power comes at a cost in computation and memory that grows with the order, and the practical trade-off between expressiveness and efficiency is the subject of a dedicated survey \cite{sato2020expressivesurvey}. The lesson is that adding layers does not add expressive power in this sense, since the bound is on the architecture rather than on the depth, and that escaping it requires changing what the network operates on, not how deep it is. The ways of escaping the bound form a small zoo. Higher-order networks operating on tuples climb a hierarchy of ever more powerful tests at ever greater cost. Cheaper routes inject information that breaks the symmetry message passing cannot, attaching positional or structural encodings to nodes, adding random features that let otherwise identical nodes be told apart, or running the network on subgraphs around each node so that local structure becomes visible. A practical question tempers the theory, namely whether the tasks that matter actually demand power beyond the first-order test, since for many node-classification problems the limiting factor is not expressiveness but data or over-smoothing, while for others, including molecular properties that depend on counting rings, the expressiveness bound is the binding constraint. The useful stance is therefore to treat expressiveness as one possible bottleneck among several and to diagnose whether a given task is actually limited by it before paying for a more powerful and more expensive model.

\subsection{Scalability}
\label{subsec:ch-scalability}

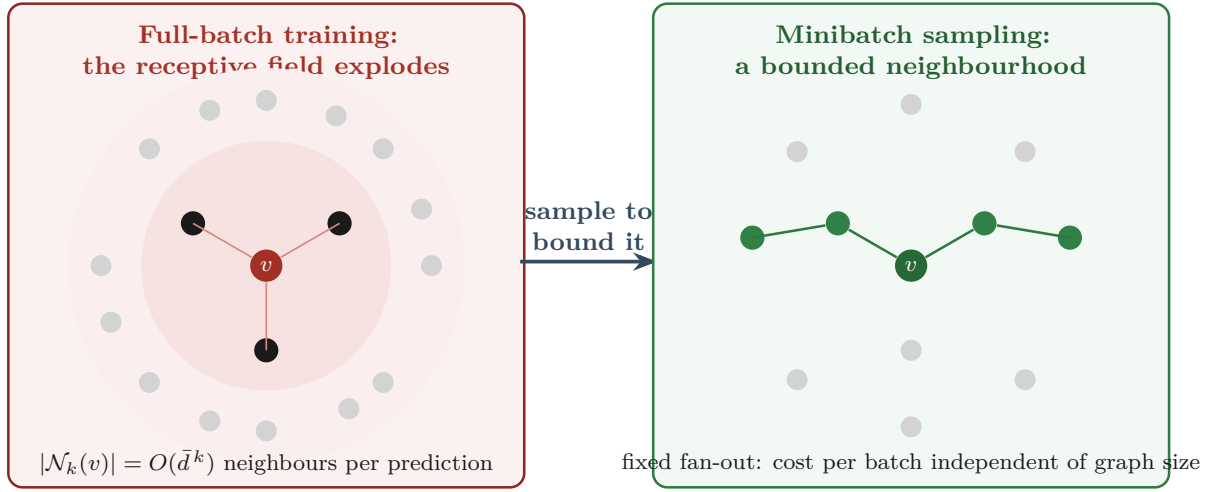
\begin{figure}[tbp]\centering
\resizebox{\textwidth}{!}{%
\begin{tikzpicture}[
  blk/.style={rounded corners=5pt, draw=#1!72!black, fill=#1!7, line width=0.9pt},
  nd/.style={circle, fill=gnnink, minimum size=3.0mm, inner sep=0pt},
  fade/.style={circle, fill=gnngray!30, minimum size=2.6mm, inner sep=0pt},
  hit/.style={circle, fill=gnngreen!75!black, minimum size=3.0mm, inner sep=0pt},
  tag/.style={rounded corners=3pt, draw=gnnslate!60, fill=white, font=\scriptsize, inner sep=3pt, align=center}]

% ---- left: full-batch explosion ----
\node[blk={gnnred}, minimum width=6.4cm, minimum height=6.0cm] (L) at (0,0) {};
\node[font=\bfseries\footnotesize, text=gnnred!85!black, align=center, text width=5.9cm] at (0,2.4) {Full-batch training:\\the receptive field explodes};
\begin{scope}[shift={(0,-0.25)}]
  \fill[gnnred!8] (0,0) circle (2.45);
  \fill[gnnred!14] (0,0) circle (1.55);
  \node[nd, fill=gnnred!85!black, minimum size=4mm] (v) at (0,0) {};
  \node[font=\scriptsize\bfseries, text=white] at (0,0) {$v$};
  \foreach \a in {30,150,270} {\node[nd] at (\a:1.05) {};}
  \foreach \a in {0,45,90,135,180,225,270,315,20,65,110,200,250,300} {\node[fade] at (\a:2.05) {};}
  \draw[gnnred!60,line width=0.6pt] (v)--(30:1.05) (v)--(150:1.05) (v)--(270:1.05);
\end{scope}
\node[font=\scriptsize, text=gnnink] at (0,-2.7) {$|\mathcal{N}_k(v)| = O(\bar d^{\,k})$ neighbours per prediction};

% ---- arrow ----
\node[font=\footnotesize\bfseries, text=gnnslate, align=center] at (4.0,0.25) {sample to\\ bound it};
\draw[-{Stealth[length=3mm]}, line width=1.3pt, gnnslate] (3.15,-0.2) -- (4.85,-0.2);

% ---- right: bounded sampling ----
\node[blk={gnngreen}, minimum width=6.4cm, minimum height=6.0cm] (R) at (8.0,0) {};
\node[font=\bfseries\footnotesize, text=gnngreen!55!black, align=center, text width=5.9cm] at (8.0,2.4) {Minibatch sampling:\\a bounded neighbourhood};
\begin{scope}[shift={(8.0,-0.25)}]
  \node[nd, fill=gnngreen!60!black, minimum size=4mm] (v2) at (0,0) {};
  \node[font=\scriptsize\bfseries, text=white] at (0,0) {$v$};
  \node[hit] (h1) at (30:1.05) {}; \node[hit] (h2) at (150:1.05) {};
  \node[fade] at (270:1.05) {};
  \node[hit] at (10:2.0) {}; \node[hit] at (170:2.0) {};
  \foreach \a in {45,90,135,225,270,315} {\node[fade] at (\a:2.0) {};}
  \draw[gnngreen!70!black,line width=0.9pt] (v2)--(h1) (v2)--(h2);
  \draw[gnngreen!70!black,line width=0.9pt] (h1)--(10:2.0) (h2)--(170:2.0);
\end{scope}
\node[font=\scriptsize, text=gnnink] at (8.0,-2.7) {fixed fan-out: cost per batch independent of graph size};

% ---- method tags below ----

\end{tikzpicture}}
\caption[The neighbourhood explosion and the sampling idea that bounds it.]{The neighbourhood explosion and the sampling idea that bounds it. A node's $k$-hop receptive field grows as $O(\bar{d}^{\,k})$ in the average degree $\bar{d}$ (left), so full-batch training over many layers becomes infeasible on large graphs; sampling a fixed neighbourhood (right) makes the cost of a minibatch independent of graph size.}
\label{fig:scaling}
\end{figure}

Many of the graphs that matter in practice have hundreds of millions or billions of nodes, and training a graph network on them is constrained by a problem specific to the setting. Computing a node's representation requires its neighbours, computing theirs requires their neighbours, and so the receptive field expands hop by hop, so that the number of nodes a single prediction depends on grows roughly as
\begin{equation}\label{eq:neighexp}
\big|\mathcal{N}_k(v)\big|=O\big(\bar{d}^{\,k}\big),
\end{equation}
with $\bar{d}$ the average degree, the neighbourhood explosion that makes full-batch training over many layers infeasible in memory; \Cref{fig:scaling} contrasts this explosion with the bounded neighbourhood that sampling produces. The responses are forms of sampling that bound the computation per node. Surveys of the area trace these methods from the algorithmic level through to the hardware accelerators that graph workloads increasingly target \cite{abadal2021computingsurvey}. One samples a fixed number of neighbours at each layer, replacing the full neighbourhood by a manageable subset \cite{hamilton2017graphsage}; another samples nodes at the level of whole layers to control the expansion \cite{chen2018fastgcn}; and a third samples subgraphs, training on densely connected clusters or on randomly sampled subgraphs so that each batch is a small graph \cite{chiang2019clustergcn,zeng2020graphsaint}. A sampled neighbourhood aggregation is rescaled to remain unbiased,
\begin{equation}\label{eq:sampling}
\tilde{\hv}_v=\frac{|\nbr(v)|}{|\mathcal{S}_v|}\sum_{u\in\mathcal{S}_v}\msg\big(\hv_u\big),
\end{equation}
with $\mathcal{S}_v$ the sampled subset, and \cref{alg:sampling} states the resulting training loop. \Cref{tab:scalability} compares the strategies. Sampling trades exactness for tractability, introducing variance that can slow or destabilize training, and it sits alongside complementary techniques, the distillation of a large model into a small one, the quantization of weights, and the precomputation of propagation, that together make graph learning feasible at scale without removing the underlying tension between the size of real graphs and the cost of message passing. A useful distinction separates training from inference, since the cost structures differ and so do the remedies. For training, the sampling methods above bound the work per update. For inference, and for some training regimes, a complementary idea precomputes the expensive propagation once and then learns a cheap model on the result, decoupling the graph operation from the learning and turning repeated message passing into a one-time cost. Coarsening the graph into a smaller summary that preserves its essential structure, partitioning it across machines for distributed training, and caching intermediate representations are further tools, each trading some combination of memory, time, and accuracy. The recurring theme is that scale forces a choice about where to spend a fixed budget, and that the right choice depends on whether the bottleneck is memory, training time, or inference latency, which differ across applications and are worth identifying before reaching for a particular method.

\begin{algorithm}[t]
\caption{Neighbourhood sampling for scalable training}
\label{alg:sampling}
\KwIn{graph $\graph$, features $\Feat$, sample sizes $\{s_l\}$, target nodes $\mathcal{B}$}
\KwOut{updated model parameters}
\ForEach{minibatch of target nodes $\mathcal{B}$}{
  \For{layer $l=L$ down to $1$}{
    sample $s_l$ neighbours for each required node\;
  }
  compute representations on the sampled subgraph by \cref{eq:sampling}\;
  update parameters from the loss on $\mathcal{B}$\;
}
\Return{trained parameters}
\end{algorithm}

\begin{table}[t]
\centering
\caption[Scalability strategies for graph neural networks.]{Strategies for training graph networks on large graphs, by what each samples and the trade-off it makes.}
\label{tab:scalability}
\small
\begin{tabularx}{\textwidth}{@{}llX@{}}
\toprule
Strategy & Samples & Trade-off \\
\midrule
Neighbour sampling \cite{hamilton2017graphsage} & fixed neighbours per layer & variance from subsampling \\
Layer sampling \cite{chen2018fastgcn}            & nodes per layer            & controls expansion, adds bias \\
Cluster subgraphs \cite{chiang2019clustergcn}    & densely connected clusters & misses between-cluster edges \\
Random subgraphs \cite{zeng2020graphsaint}       & sampled subgraphs          & needs normalization for unbiasedness \\
\bottomrule
\end{tabularx}
\end{table}

\subsection{Heterophily}
\label{subsec:ch-heterophily}

\begin{figure}[t]
\centering
\begin{tikzpicture}
\begin{axis}[width=8.8cm, height=5cm,
  xlabel={homophily ratio $h$}, ylabel={accuracy},
  xlabel style={font=\footnotesize}, ylabel style={font=\footnotesize},
  xticklabel style={font=\scriptsize}, yticklabel style={font=\scriptsize},
  xmin=0, xmax=1, ymin=0.4, ymax=1.0, ymajorgrids=true, grid style={gnngray!25},
  legend style={font=\scriptsize, at={(0.97,0.05)}, anchor=south east, draw=gnngray!40},
  every axis plot/.append style={line width=1.2pt}]
\addplot[mark=*, mark size=1.4pt, draw=gnnred] coordinates {(0.1,0.52)(0.3,0.60)(0.5,0.71)(0.7,0.83)(0.9,0.93)};
\addlegendentry{standard message passing}
\addplot[mark=square*, mark size=1.4pt, draw=gnnblue, dashed] coordinates {(0.1,0.80)(0.3,0.82)(0.5,0.85)(0.7,0.88)(0.9,0.93)};
\addlegendentry{heterophily-aware}
\end{axis}
\end{tikzpicture}
\caption[Illustrative effect of heterophily on accuracy.]{Illustrative accuracy as the homophily ratio varies: a standard network (red) degrades as homophily falls, since its smoothing averages dissimilar neighbours, while a heterophily-aware design (blue) holds up across the range. The curves are schematic and convey only the qualitative pattern.}
\label{fig:heterophily}
\end{figure}
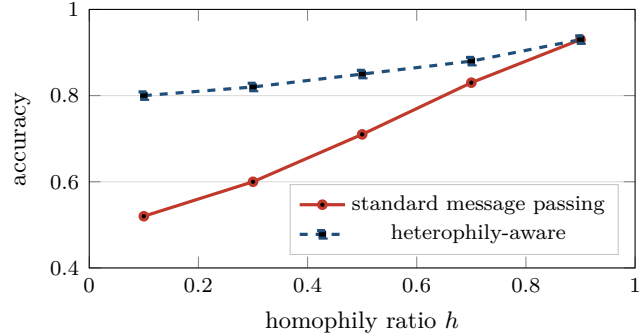

Most graph networks rest on an assumption the foundations chapter named and the fraud chapter saw fail: homophily, that connected nodes tend to share labels. The degree to which a graph satisfies it is captured by the homophily ratio,
\begin{equation}\label{eq:homophily}
h=\frac{1}{|\eset|}\sum_{(u,v)\in\eset}\mathbb{1}\big[y_u=y_v\big],
\end{equation}
the fraction of edges joining same-label nodes. When $h$ is high, as in citation and social networks, the smoothing that message passing performs is exactly right, since averaging a node with its neighbours sharpens a signal they share. When $h$ is low, as in fraud graphs where camouflage connects fraudsters to legitimate accounts, in certain web graphs, and in molecules where adjacent atoms play opposite roles, smoothing averages together nodes that should be kept apart and actively degrades performance. \Cref{fig:heterophily} sketches the resulting gap between standard networks, whose accuracy falls as homophily decreases, and heterophily-aware designs that hold up across the range. Those designs change what is aggregated and how: one separates a node's own representation from its neighbours' and aggregates over a geometry that can reach beyond immediate neighbours \cite{pei2020geomgcn}, and another learns generalized propagation weights that can become negative, turning the implicit low-pass filter into one that can emphasize differences rather than similarities \cite{chien2021gprgnn}. The connection to the rest of the chapter is direct, since heterophily is the over-smoothing problem seen from another angle, both rooted in the smoothing inductive bias, and it is the formal statement of why the fraud and anomaly domains had to rebuild the standard toolkit rather than apply it. The topic carries a live debate worth flagging. Careful study has questioned whether low homophily by itself is what hurts, suggesting that some heterophilous graphs are handled well by standard methods and that the harm depends on finer properties of how labels and structure relate, which has prompted better measures of homophily and more careful benchmarks. The methods that help span a spectrum, from filters that pass high frequencies, to aggregation that keeps a node's own representation separate from its neighbours', to schemes that learn a different combination of neighbourhood information per node, and several connect to the rewiring ideas raised for over-squashing, since both modify the effective graph over which information flows. The unsettledness of the question is itself instructive, a reminder that an intuition as basic as homophily resists a clean general theory.

\FloatBarrier
\subsection{Robustness and adversarial attacks}
\label{subsec:ch-robustness}

\begin{figure}[t]
\centering
\begin{tikzpicture}
\begin{axis}[width=8.8cm, height=5cm,
  xlabel={fraction of edges perturbed}, ylabel={accuracy},
  xlabel style={font=\footnotesize}, ylabel style={font=\footnotesize},
  xticklabel style={font=\scriptsize}, yticklabel style={font=\scriptsize},
  xmin=0, xmax=0.2, ymin=0.35, ymax=0.9, ymajorgrids=true, grid style={gnngray!25},
  legend style={font=\scriptsize, at={(0.97,0.95)}, anchor=north east, draw=gnngray!40},
  every axis plot/.append style={line width=1.2pt}]
\addplot[mark=*, mark size=1.4pt, draw=gnnred] coordinates {(0,0.85)(0.05,0.72)(0.10,0.58)(0.15,0.47)(0.20,0.40)};
\addlegendentry{standard message passing}
\addplot[mark=square*, mark size=1.4pt, draw=gnngreen, dashed] coordinates {(0,0.84)(0.05,0.80)(0.10,0.74)(0.15,0.68)(0.20,0.63)};
\addlegendentry{robust design}
\end{axis}
\end{tikzpicture}
\caption[Illustrative accuracy under adversarial perturbation.]{Illustrative accuracy as a growing fraction of edges is adversarially perturbed: a standard network (red) degrades sharply, while a robust design (green) degrades more gracefully. The curves are schematic and convey only the qualitative effect.}
\label{fig:robustdrop}
\end{figure}
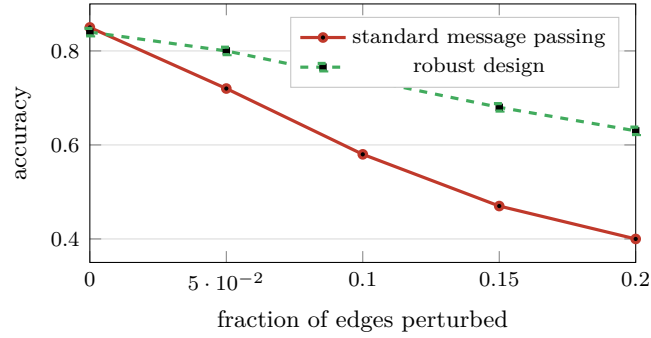

Graph networks are vulnerable to adversarial manipulation, and the vulnerability has a form peculiar to graphs. Surveys of this area catalogue the attacks and the defenses developed against them across the graph setting \cite{sun2022advsurvey}. Beyond the feature perturbations that affect any model, an attacker can alter the structure itself, adding or removing a small number of edges, and because message passing propagates information along edges a single well-chosen change can shift a prediction far from where it was made. Formally an attack searches for the perturbed graph within a budget that most increases the loss,
\begin{equation}\label{eq:advattack}
\max_{\graph'\in\mathcal{B}(\graph,\Delta)}\ \mathcal{L}\big(f_\theta(\graph')\big),
\end{equation}
where $\mathcal{B}(\graph,\Delta)$ is the set of graphs reachable from $\graph$ by perturbations of size at most $\Delta$. Such searches are strikingly effective. A targeted attack that modifies the edges and features around a chosen node can flip its prediction with very few changes \cite{zugner2018nettack}, and a poisoning attack that perturbs the training graph through meta-gradients can degrade a model's accuracy across the whole graph rather than at a single node \cite{zugner2019metattack}, with \cref{fig:robustdrop} sketching how quickly accuracy falls as the perturbation grows. The defenses, summarized in \cref{tab:robustness}, work by cleaning or distrusting the structure: one learns a corrected graph jointly with the model, exploiting the observation that adversarial edges tend to disturb properties real graphs exhibit \cite{jin2020prognn}, and another reweights message passing to discount edges between dissimilar nodes, which are where attacks concentrate \cite{zhang2020gnnguard}, with the broader landscape of attacks and defenses surveyed in dedicated reviews \cite{jin2021advreview}. The connection to the fraud and security domains is immediate, since there the adversary is not hypothetical but an actively adapting opponent, and the discrete, propagating nature of structural attacks is why robustness in those settings is a defining requirement rather than a refinement, and why certified guarantees, which bound the worst case rather than patching observed attacks, are a sought-after but still limited goal. The threat is best understood through its dimensions. An attack may strike at test time, perturbing the input to a fixed model, or at training time, poisoning the data the model learns from; it may target a single node or aim to degrade the model everywhere; and it may assume full knowledge of the model or operate from the outside with only its predictions. Each combination calls for different defenses and admits different guarantees. Certified approaches, including those that add random noise and certify a region around the input within which the prediction cannot change, give worst-case assurances but currently cover only restricted attack models and modest budgets. A question that should accompany any of this is how realistic the attacks are in a given setting, since the freedom to rewrite arbitrary edges that many attacks assume may not be available to a real adversary, so the practical importance of robustness varies from a central concern in the adversarial domains to a more theoretical one elsewhere.

\begin{table}[t]
\centering
\caption[Adversarial attacks and defenses on graph neural networks.]{Representative attacks and defenses, by type and central idea.}
\label{tab:robustness}
\small
\begin{tabularx}{\textwidth}{@{}llX@{}}
\toprule
Method & Type & Idea \\
\midrule
Nettack \cite{zugner2018nettack}       & targeted attack  & perturb edges and features near a node \\
Metattack \cite{zugner2019metattack}   & poisoning attack & meta-gradients degrade training \\
Pro-GNN \cite{jin2020prognn}           & defense          & learn a clean graph structure \\
GNNGuard \cite{zhang2020gnnguard}      & defense          & down-weight suspicious edges \\
\bottomrule
\end{tabularx}
\end{table}

\subsection{Explainability}
\label{subsec:ch-explainability}

A graph network is opaque in the way of any deep model, and the application chapters made clear that opacity is an obstacle to deployment wherever a prediction triggers a consequential action, in healthcare, in fraud, and in scientific use. Explaining a graph network poses a problem the image and text settings do not, because the relevant explanation is structural: not merely which features mattered but which nodes and edges of the input did, a subgraph rather than a saliency map. The dominant formulation seeks the small subgraph most informative about the prediction, maximizing the mutual information between the output and the retained structure under a size constraint,
\begin{equation}\label{eq:explainmi}
\max_{\mathcal{S}\subseteq\graph,\ |\mathcal{S}|\le k}\ \mathrm{MI}\big(Y;\,\mathcal{S}\big),
\end{equation}
which the original method optimized through a soft mask over edges and features \cite{ying2019gnnexplainer} and which \cref{alg:gnnexplain} states in outline. Later methods refine this in several directions, collected in \cref{tab:explainmethods}: one trains a parameterized explainer that produces explanations in a single pass and so transfers to new instances \cite{luo2020pgexplainer}, one searches for connected subgraphs guided by a game-theoretic measure of each part's contribution \cite{yuan2021subgraphx}, and one seeks counterfactual explanations, the smallest change to the graph that would alter the prediction \cite{lucic2022cfgnn}, alongside gradient-based attribution that traces the output back to inputs \cite{sanchez2020attribution}. A taxonomic survey organizes the area \cite{yuan2023explainsurvey}. The honest caveat is that explanations can be unstable, varying with small changes to the input or the method, and that judging whether an explanation is faithful to the model's actual reasoning rather than merely plausible to a human is itself an unsolved evaluation problem, so explainability remains a partial answer to the demand for trust rather than a settled one. Several distinctions organize the area. An explanation may be local, accounting for a single prediction, or global, describing the model's behaviour as a whole, and the two serve different purposes, the first for justifying a particular decision and the second for auditing a model before deployment. The deepest difficulty is the absence of ground truth, since outside synthetic settings there is rarely a known correct explanation against which to check a produced one, which is why faithfulness, whether an explanation reflects the model's actual computation, is so much harder to establish than plausibility, whether it looks reasonable to a person. In the scientific domains the relationship inverts in a productive way, since there an explanation is valuable precisely as a hypothesis, a claim about which substructure drives a property that a domain expert can test, and this reframing, from explanation as justification to explanation as discovery, is among the more promising uses of the machinery.

\begin{algorithm}[t]
\caption{Explaining a prediction by edge masking}
\label{alg:gnnexplain}
\KwIn{trained model $f_\theta$, node $v$ with prediction $\hat{y}_v$, its computation subgraph}
\KwOut{important subgraph explaining $\hat{y}_v$}
initialize a soft mask $M$ over the edges of the computation subgraph\;
\Repeat{converged}{
  form the masked graph by applying $M$\;
  update $M$ to keep $f_\theta$'s output near $\hat{y}_v$ while penalizing the mask size, following \cref{eq:explainmi}\;
}
\Return{the edges with the largest mask values}
\end{algorithm}

\begin{table}[t]
\centering
\caption[Explanation methods for graph neural networks.]{Representative explanation methods, by the form of explanation they produce and the idea behind them, spanning instance-level subgraph attributions, counterfactual and causal explanations, and model-level and self-explaining approaches.}
\label{tab:explainmethods}
\small
\begin{tabularx}{\textwidth}{@{}llX@{}}
\toprule
Method & Explanation & Idea \\
\midrule
GNNExplainer \cite{ying2019gnnexplainer}    & subgraph and features & maximize mutual information \\
PGExplainer \cite{luo2020pgexplainer}       & subgraph              & amortized, inductive masks \\
SubgraphX \cite{yuan2021subgraphx}          & connected subgraph    & contribution-guided search \\
PGM-Explainer \cite{vu2020pgmexplainer}     & probabilistic model   & local interpretable Bayesian model \\
GraphMask \cite{schlichtkrull2021graphmask} & edge relevance        & differentiable edge masking \\
CF-GNNExplainer \cite{lucic2022cfgnn}       & counterfactual        & minimal change to flip output \\
RC-Explainer \cite{wang2023rcexplainer}     & causal subgraph       & reinforced causal screening \\
XGNN \cite{yuan2020xgnn}                     & model-level           & generate prototypical patterns \\
SE-GNN \cite{dai2021selfexplain}            & self-explaining       & built-in nearest-neighbour rationale \\
\bottomrule
\end{tabularx}
\end{table}

\subsection{Fairness, privacy, and uncertainty}
\label{subsec:ch-fairness}

The final cluster of challenges concerns whether a graph network can be trusted to behave acceptably, and each member of the cluster has a graph-specific twist. Fairness is complicated by propagation, because message passing spreads information about a sensitive attribute through the graph, and when connections correlate with that attribute, as they often do, a model can reconstruct and act on it even when the attribute is withheld, so the structure itself is a channel for bias. A standard measure of the resulting disparity is demographic parity, the gap in positive-prediction rates across groups,
\begin{equation}\label{eq:fairparity}
\Delta_{\mathrm{DP}}=\big|\,P(\hat{y}{=}1\mid s{=}0)-P(\hat{y}{=}1\mid s{=}1)\,\big|,
\end{equation}
and methods to reduce it, gathered in \cref{tab:fairness}, address the propagation directly: one debiases representations adversarially while needing only limited sensitive labels \cite{dai2021fairgnn}, one combines fairness with stability through a contrastive objective \cite{agarwal2021nifty}, and one suppresses the feature channels through which the sensitive attribute leaks \cite{wang2022fairvgnn}. Privacy carries an analogous relational twist, since a graph encodes who is connected to whom and a model trained on it can leak that structure, and because a node's data implicates its neighbours, protecting one individual is entangled with others in a way that isolated records avoid, which complicates the federated and differentially private training that the privacy-sensitive domains require. Uncertainty is the least developed of the three, as graph networks are frequently overconfident and poorly calibrated, with their stated confidence diverging from their actual accuracy as measured by the calibration error,
\begin{equation}\label{eq:calib}
\mathrm{ECE}=\sum_{m}\frac{|B_m|}{n}\,\big|\mathrm{acc}(B_m)-\mathrm{conf}(B_m)\big|,
\end{equation} The formal statements of the challenges gathered here, \cref{eq:smoothenergy,eq:squash,eq:wlbound,eq:neighexp,eq:homophily,eq:advattack,eq:fairparity,eq:calib}, are referenced where each constraint is discussed in the domain chapters.
which matters acutely in the high-stakes domains, healthcare and power among them, where a model that cannot say when it is unsure cannot be safely deployed, and where conformal methods that attach statistical guarantees to predictions on graphs are an active but early response. A comprehensive survey draws privacy, robustness, fairness, and explainability together as the constituents of trustworthy graph learning \cite{dai2024trustworthy}. Each strand has further structure worth naming. Fairness divides into group notions, which equalize outcomes across demographic groups, and individual notions, which ask that similar individuals be treated similarly, and the two can conflict, as can either with accuracy, so a fairness intervention is a choice among trade-offs rather than a free improvement, and some of the most graph-specific interventions act on the structure itself, rewiring edges to reduce the propagation of bias. Privacy faces graph-specific attacks, including membership inference that asks whether a node was in the training data and link inference that reconstructs hidden edges, both exploiting the way a model encodes structure, and the entanglement of neighbours complicates the differential-privacy accounting that would bound such leakage. Uncertainty has two sources usually worth separating, the irreducible noise in the data and the model's ignorance away from its training distribution, the latter being exactly what high-stakes deployment needs to detect, and Bayesian and ensemble approaches that estimate it on graphs remain an open and active area.

\begin{table}[t]
\centering
\caption[Fairness methods for graph neural networks.]{Representative fairness methods, by the aspect of the problem each targets and the idea behind it.}
\label{tab:fairness}
\small
\begin{tabularx}{\textwidth}{@{}llX@{}}
\toprule
Method & Target & Idea \\
\midrule
FairGNN \cite{dai2021fairgnn}    & limited sensitive labels & adversarial debiasing \\
NIFTY \cite{agarwal2021nifty}    & fairness and stability   & counterfactual contrastive learning \\
FairVGNN \cite{wang2022fairvgnn} & feature leakage          & suppress sensitive-correlated channels \\
EDITS \cite{dong2022edits}       & data bias                & debias graph structure and node features \\
\bottomrule
\end{tabularx}
\end{table}

\subsection{Distribution shift and generalization}
\label{subsec:ch-shift}

A challenge that surfaced in nearly every application chapter, though under different names, is distribution shift: a model trained on one distribution of graphs degrades when the distribution changes. The forms it takes recur across the survey. Healthcare models trained at one hospital falter at another, power-system models trained on one grid transfer poorly to a different topology, traffic models trained on ordinary conditions fail during incidents, fraud models decay as patterns drift, and molecular models split by scaffold generalize worse than random splits suggest. Underlying these are a few distinct shifts, in the size of the graph, in its structure, in node features, and in the balance of labels, each of which can break a model that assumed the training distribution would persist. The responses, domain adaptation that aligns source and target, source-free and continual methods that adapt without revisiting the original data \cite{bahi2026freegnn}, and invariant learning that seeks features stable across environments, are the same in spirit across domains, which is the point: the many domain-specific generalization failures the survey reported are one challenge wearing many costumes, and treating it as such is more productive than solving it separately in each field. The unification has begun to take institutional form, with benchmarks that deliberately construct training and test splits differing in structure, size, or features so that out-of-distribution performance can be measured rather than assumed, and with theory that studies when a model trained on small graphs can be expected to generalize to larger ones. Test-time adaptation, which adjusts a model using the unlabelled target data it encounters at deployment, is one practical response, and the connection to the next chapter is direct, since a central argument for large-scale pretraining is that a model exposed to enough varied graphs might acquire representations that transfer across the shifts that defeat narrowly trained ones, making distribution shift not only a challenge in its own right but part of the motivation for graph foundation models.

\subsection{The connected nature of the challenges}
\label{subsec:ch-connected}

The challenges of this chapter are not independent, and \cref{tab:challengemap} recaps them alongside their causes and the domains they most affect. Over-smoothing and heterophily are the same smoothing bias seen from two angles. The analysis of over-smoothing as a collapse toward a low-frequency subspace was made early and sharply \cite{li2018deeperinsights}, and the heterophily response of designing filters that also pass high-frequency signal follows directly from it \cite{bo2021fagcn}. Expressiveness and the depth limits are both statements about what message passing can and cannot do as computation. Robustness, explainability, and fairness are facets of a single demand for trust, and they interact, since an explanation can expose a fairness violation, a fairness intervention can change a model's robustness, and a successful attack is a failure of the trust the other two try to establish. Scalability cuts across everything, because a remedy that is not feasible at the size of a real graph is not a remedy, and distribution shift cuts across everything else, because a model that cannot generalize beyond its training distribution fails regardless of how it scores within it. The interactions are not merely additive but can be adversarial, in the sense that progress on one axis sometimes costs another, since a more expressive model can be harder to scale, a more robust one can be less accurate, and a fairer one can be less performant, so the challenges define a space of trade-offs rather than a checklist of independently solvable problems. This interconnection is why the field's larger ambitions raise the stakes on the whole set at once. A graph foundation model, the subject of the next chapter, must be expressive enough to be useful, scalable enough to train, robust enough to deploy, fair enough to trust, and able to generalize across the very distribution shifts that defeat narrower models, and it must be all of these simultaneously rather than one at a time, which makes the challenges gathered here the precise obstacles that any general, trustworthy graph learning must overcome.

\begin{table}[t]
\centering
\caption[A summary of the challenges, their causes, and the domains they most affect.]{A recapitulation of the chapter's challenges, each with its underlying cause and the domains where it bites hardest.}
\label{tab:challengemap}
\small
\begin{tabularx}{\textwidth}{@{}lXl@{}}
\toprule
Challenge & Underlying cause & Most affected \\
\midrule
Over-smoothing      & repeated smoothing          & deep tasks \\
Over-squashing      & bottlenecked propagation    & long-range tasks \\
Expressiveness      & the 1-WL bound              & substructure tasks \\
Scalability         & neighbourhood explosion     & web-scale graphs \\
Heterophily         & the homophily assumption    & fraud, web graphs \\
Robustness          & discrete structure attacks  & fraud, security \\
Explainability      & opaque aggregation          & healthcare, science \\
Fairness, privacy   & propagation of attributes   & social, healthcare \\
Uncertainty         & poor calibration            & high-stakes domains \\
\bottomrule
\end{tabularx}
\end{table}

\FloatBarrier
\section{Graph foundation models}
\label{sec:gfm}

Foundation models, large models pretrained once on broad data and then adapted to many downstream tasks, reshaped natural language processing and computer vision, and the obvious question is whether graphs can have the same. The cross-domain synthesis gave the premise and the obstacle together. The premise is that the application domains share a small set of constructions, a common spatio-temporal template, and a consistent pattern of where relational structure helps, which suggests that one model might serve many. The obstacle is the feature-alignment problem, that graphs are a family of objects with no shared vocabulary, so a model pretrained on one kind of graph does not obviously transfer to another. This chapter assesses how far the ambition has been carried, beginning with the self-supervised pretraining that is its prerequisite, then the foundation models that genuinely exist within narrow families, then the convergence with large language models, and closing with an honest appraisal of what the term can and cannot yet mean for graphs.

\subsection{What a foundation model would mean for graphs}
\label{subsec:gfm-what}

\begin{figure}[t]
\centering
\resizebox{0.92\textwidth}{!}{%
\begin{tikzpicture}[
  box/.style={rounded corners=3pt, draw=gnnblue!80!black, fill=tintblue, align=center, font=\footnotesize,
              text=gnnink, inner sep=4pt, minimum height=11mm, text width=21mm},
  pre/.style={rounded corners=3pt, draw=gnnteal!80!black, fill=tintteal, align=center, font=\footnotesize,
              text=gnnink, inner sep=4pt, minimum height=11mm, text width=21mm},
  fm/.style={rounded corners=3pt, draw=gnnpurple!80!black, fill=tintpurple, align=center, font=\footnotesize\bfseries,
             text=gnnpurple!72!black, inner sep=4pt, minimum height=11mm, text width=21mm},
  task/.style={rounded corners=3pt, draw=gnngreen!80!black, fill=tintgreen, align=center, font=\scriptsize,
               text=gnngreen!72!black, inner sep=3.5pt, minimum height=7mm, text width=19mm},
  ar/.style={-{Stealth[length=2.6mm]}, draw=gnnslate, line width=1pt}]
\node[box] (data) {broad graph data};
\node[pre, right=9mm of data] (pre) {self-supervised pretraining};
\node[fm, right=9mm of pre] (model) {foundation model};
\node[task, right=13mm of model] (t2) {classification};
\node[task, above=2mm of t2] (t1) {link prediction};
\node[task, below=2mm of t2] (t3) {regression};
\draw[ar] (data)--(pre); \draw[ar] (pre)--(model);
\draw[ar] (model.east)--(t1.west); \draw[ar] (model.east)--(t2.west); \draw[ar] (model.east)--(t3.west);
\node[font=\scriptsize, align=center, text=gnnslate] at ($(model)!0.5!(t2)+(0,-1.4)$) {adapt, prompt,\\ zero-shot};
\end{tikzpicture}}
\caption[The foundation-model pipeline for graphs.]{The foundation-model pipeline: a model is pretrained on broad data with a self-supervised objective and then adapted to many downstream tasks. For graphs the difficulty is the pretraining data, since graphs across domains share no common vocabulary, which is the obstacle the rest of the chapter examines.}
\label{fig:fmecosystem}
\end{figure}
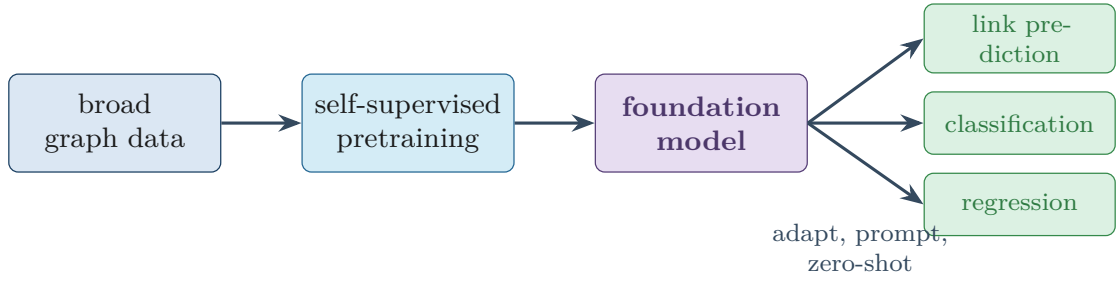

The defining pattern of a foundation model, illustrated in \cref{fig:fmecosystem}, is to pretrain on broad data and then adapt the result to many tasks by fine-tuning, prompting, or zero-shot use, which amortizes the cost of learning general structure and enables few-shot transfer to tasks with little labelled data. Stated abstractly, pretraining minimizes a self-supervised objective over a broad collection and adaptation specializes the result to a task,
\begin{equation}\label{eq:pretrain}
\theta^\star=\arg\min_{\theta}\ \mathcal{L}_{\mathrm{ssl}}\big(\theta;\,\mathcal{D}\big),\qquad
\theta_t=\mathrm{Adapt}\big(\theta^\star,\mathcal{D}_t\big).
\end{equation}
The graph case is harder than the text or image case for a structural reason. Text is a single modality with a shared vocabulary of tokens, and images share a pixel space, but a molecule's atom types and a social network's user attributes have nothing in common, and even the meaning of a node or an edge differs across domains, so there is no graph analogue of the token that holds steady from one domain to the next. The transductive-inductive divide compounds the difficulty, since a foundation model is inductive generalization carried to its limit, a single model expected to apply not only to new nodes but to new graphs and new domains. These two facts, the absence of a shared substrate and the demand for extreme inductive transfer, are why the graph version of the foundation-model idea is genuinely harder than the versions that preceded it. The appeal is nonetheless strong enough to drive the effort. A foundation model amortizes the expense of learning general structure across all the tasks that reuse it, brings few-shot and zero-shot capability to settings where labelled data is scarce, and broadens access by letting a practitioner adapt a pretrained model rather than train one from nothing. There is a spectrum rather than a binary here, since a model can be a foundation model in a weak sense, transferring across tasks on a fixed graph, or in a strong sense, transferring across graphs and domains, and much of the disagreement about whether graph foundation models exist is really disagreement about where on this spectrum the bar should sit. The same generality that is the goal also carries a risk worth noting, that a single widely reused model concentrates its biases and failures in everything built on it, so the homogenization a foundation model brings is a liability as much as a convenience.

\subsection{Self-supervised pretraining}
\label{subsec:gfm-pretrain}

The prerequisite for any foundation model is the ability to learn useful representations from unlabelled data, and self-supervised pretraining on graphs has become a substantial area in its own right, with methods falling into two broad families. The contrastive family trains a model to agree across different views of the same object while distinguishing it from others, optimizing a loss that pulls a representation toward a positive counterpart and away from negatives,
\begin{equation}\label{eq:contrastive}
\mathcal{L}_{\mathrm{con}}=-\log\frac{\exp\!\big(\mathrm{sim}(\mathbf{z}_i,\mathbf{z}_i^{+})/\tau\big)}{\sum_{j}\exp\!\big(\mathrm{sim}(\mathbf{z}_i,\mathbf{z}_j)/\tau\big)},
\end{equation}
where the views are produced by augmenting the graph. One foundational method maximizes the agreement between local node representations and a global summary of the graph \cite{velickovic2019dgi}, another contrasts augmented views of the graph directly \cite{you2020graphcl}, and a third designs the contrast to yield representations transferable across different graphs \cite{qiu2020gcc}. The generative family instead trains a model to reconstruct deliberately hidden parts of the graph, masking node attributes or edges and predicting them, with the reconstruction loss summed over the masked set,
\begin{equation}\label{eq:maskrecon}
\mathcal{L}_{\mathrm{rec}}=\sum_{v\in\mathcal{M}}\ell\big(\hat{\xv}_v,\xv_v\big),
\end{equation}
as in a generative pretraining that predicts masked attributes and structure \cite{hu2020gptgnn} and a masked autoencoder that reconstructs hidden features \cite{hou2022graphmae}. Beyond these two families a predictive line uses auxiliary tasks defined on the graph itself, such as predicting node degrees, distances, or the presence of particular motifs, as pretext objectives whose solution requires the structural understanding downstream tasks reward. The three families are complementary sources of supervisory signal rather than rivals, and the practical question for a given application is which pretext most resembles the eventual task, since the closer the match the more reliably the pretrained representation transfers. A careful study of pretraining strategies showed that combining node-level and graph-level objectives matters, and that naive pretraining can even hurt, which is a caution against assuming transfer comes for free \cite{hu2020pretrain}. Several difficulties specific to graphs complicate this picture. The augmentations contrastive learning relies on are well defined for images, where cropping or rotating clearly preserves content, but ambiguous for graphs, where dropping a node or an edge may change the very property a downstream task cares about, so designing augmentations that preserve meaning is an unsolved part of the recipe rather than a detail. Contrastive methods also depend on negative examples, and what counts as a negative on a graph, where two subgraphs may be more similar than they appear, is delicate. The molecular setting has been a particular proving ground, where pretraining on large unlabelled molecule collections, sometimes incorporating three-dimensional structure or chemically meaningful motifs, transfers usefully to property prediction precisely because the shared atom vocabulary lets a pretrained model apply to new molecules, and pretraining that injects three-dimensional information into a two-dimensional encoder can transfer geometric knowledge to settings where only the graph is available \cite{stark2022infomax3d}. The finding that careless pretraining can degrade rather than improve downstream performance, sometimes called negative transfer, is the most important practical lesson, since it shows that the benefits of pretraining are real but contingent on matching the pretext task to the downstream one. \Cref{alg:gfmpretrain} states the pretrain-then-adapt procedure these methods share, and \cref{tab:gfm} collects them alongside the foundation-model efforts discussed next.

\begin{algorithm}[t]
\caption{Self-supervised pretraining and adaptation}
\label{alg:gfmpretrain}
\KwIn{broad collection of unlabelled graphs, downstream task with limited labels}
\KwOut{adapted model for the task}
\tcp{pretraining}
\Repeat{converged}{
  sample a graph and form augmented or masked versions\;
  update the model to satisfy a self-supervised objective, by \cref{eq:contrastive} or \cref{eq:maskrecon}\;
}
\tcp{adaptation}
specialize the pretrained model to the task by fine-tuning or prompting, following \cref{eq:pretrain}\;
\Return{adapted model}
\end{algorithm}

\subsection{Foundation models within domains}
\label{subsec:gfm-within}

\begin{figure}[t]
\centering
\begin{tikzpicture}
\def\R{2.3}
\foreach \r in {0.25,0.5,0.75,1.0} {
  \draw[gnngray!28] (90:{\r*\R}) -- (162:{\r*\R}) -- (234:{\r*\R}) -- (306:{\r*\R}) -- (18:{\r*\R}) -- cycle;
}
\foreach \a in {90,162,234,306,18} \draw[gnngray!45] (0,0)--(\a:\R);
\node[font=\scriptsize, above]      at (90:\R)  {cross-task};
\node[font=\scriptsize, left]       at (162:\R) {cross-domain};
\node[font=\scriptsize, below left] at (234:\R) {few-shot};
\node[font=\scriptsize, below right]at (306:\R) {zero-shot};
\node[font=\scriptsize, right]      at (18:\R)  {data scale};
\draw[draw=gnnred, dashed, line width=1.2pt, fill=gnnred, fill opacity=0.14]
  (90:{0.8*\R}) -- (162:{0.2*\R}) -- (234:{0.4*\R}) -- (306:{0.2*\R}) -- (18:{0.5*\R}) -- cycle;
\draw[draw=gnnteal, line width=1.2pt, fill=gnnteal, fill opacity=0.16]
  (90:{0.85*\R}) -- (162:{0.8*\R}) -- (234:{0.8*\R}) -- (306:{0.7*\R}) -- (18:{0.6*\R}) -- cycle;
\node[font=\scriptsize, text=gnnred] at (-2.6,-2.55) {$- -$ domain-specific};
\node[font=\scriptsize, text=gnnteal] at (1.7,-2.55) {--- foundation aspiration};
\end{tikzpicture}
\caption[Illustrative capabilities of domain-specific and foundation models.]{Illustrative comparison along several capability axes: the dashed red region is a domain-specific model, strong within its task but weak across domains, and the solid teal region is the general foundation-model aspiration, balanced across axes. The shape is schematic and conveys only the qualitative gap, not measured values.}
\label{fig:gfmradar}
\end{figure}
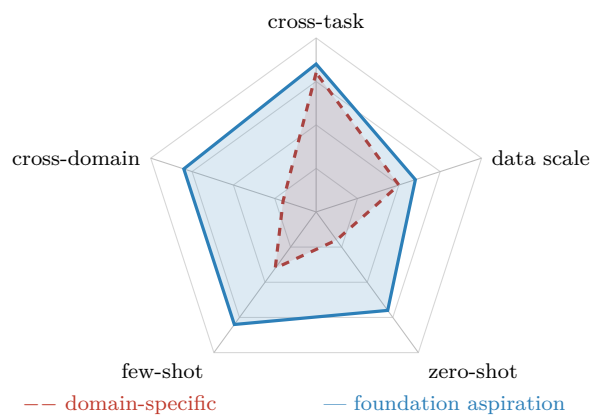

The genuine successes so far share a common shape: they hold either the domain or the task fixed, so that the feature-alignment problem does not arise in full. Within knowledge-graph reasoning, a model that learns relation-invariant representations transfers across knowledge graphs with different entities and relations, performing inference on graphs it never saw in training, which is a foundation model in the meaningful sense within that domain \cite{galkin2024ultra}. Within chemistry, the pretrained molecular models and the universal interatomic potentials of the materials chapter transfer across molecules and compositions precisely because the vocabulary of atoms is shared, so the obstacle is absent. A different strategy fixes the task rather than the domain, building a model that performs node classification on any graph regardless of its feature space by reducing the problem to a form that does not depend on a shared vocabulary \cite{zhao2024graphany}, or that performs in-context learning over graphs, adapting to a new task from a few examples supplied at inference without retraining \cite{huang2023prodigy}. The status of these results is genuinely contested. One position holds that graph foundation models are already here, pointing to exactly these within-domain and within-task successes \cite{mao2024gfm}, while a survey of the area takes the more guarded view that the general case, a single model across arbitrary graph types and tasks, remains open \cite{liu2023gfmsurvey}. Other efforts pursue generality through different routes: graph prompting recasts downstream tasks into a common pretext format so a single pretrained model can be adapted by prompts rather than fine-tuning \cite{sun2023allinone,liu2023graphprompt}, open graph models target transfer to entirely unseen graphs \cite{xia2024opengraph}, recent work asks whether such models generalize across architectures \cite{gutteridge2026gfmarch} and builds them on a shared geometric space across domains \cite{sun2026graphglue,yu2026rfm}, and progress is increasingly measured against dedicated benchmarks that test these abilities systematically \cite{yu2026gfmbench}. It is worth understanding why the within-domain successes work, since the reason illuminates the general obstacle. The knowledge-graph model transfers because it represents relations not by fixed embeddings but by their pattern of interaction with other relations, a meta-structure comparable across knowledge graphs even when the relations themselves differ, which is in effect a way of manufacturing a shared vocabulary where none was given. The molecular models transfer because the vocabulary, the periodic table, is shared to begin with. The task-fixed models transfer by reformulating node classification so that it depends on the relationships among labelled examples rather than on the raw feature space, removing the dependence on a common set of features. In every case the device is to find or construct something invariant across the graphs in scope, and the difficulty of the general problem is precisely that no such invariant is known to exist across all graphs at once. \Cref{fig:gfmradar} sketches the gap qualitatively, between what a domain-specific model achieves and what a general foundation model would.

\begin{table}[t]
\centering
\caption[Pretraining and foundation-model approaches for graphs.]{Self-supervised pretraining methods and foundation-model efforts, by category and central idea. The within-domain and any-graph entries succeed by fixing either the domain or the task.}
\label{tab:gfm}
\small
\begin{tabularx}{\textwidth}{@{}llX@{}}
\toprule
Approach & Category & Idea \\
\midrule
DGI \cite{velickovic2019dgi}      & contrastive pretraining   & maximize local--global agreement \\
GraphCL \cite{you2020graphcl}     & contrastive pretraining   & contrast augmented views \\
GCC \cite{qiu2020gcc}             & transferable pretraining  & contrastive coding across graphs \\
GPT-GNN \cite{hu2020gptgnn}       & generative pretraining    & predict masked attributes and edges \\
GraphMAE \cite{hou2022graphmae}   & generative pretraining    & masked feature reconstruction \\
ULTRA \cite{galkin2024ultra}      & within-domain foundation  & relation-invariant graph reasoning \\
GraphAny \cite{zhao2024graphany}  & any-graph foundation      & node classification on any graph \\
PRODIGY \cite{huang2023prodigy}   & in-context learning       & prompt with examples on graphs \\
GraphGPT \cite{tang2024graphgpt}  & language-model integration & graph instruction tuning \\
\bottomrule
\end{tabularx}
\end{table}

\subsection{Large language models and graphs}
\label{subsec:gfm-llm}

The most active recent direction joins graphs with large language models, in three distinct modes. In the first, the language model is the predictor, and a graph is rendered into text that the model reasons over, an approach whose effectiveness depends heavily on how the graph is encoded \cite{fatemi2024talklikegraph} and which the empirical literature finds workable for small graphs and structural reasoning but strained by large or intricate structure \cite{chen2024exploring}. In the second, the language model enhances a graph network on text-attributed graphs, where nodes carry text, by turning that text into rich features the graph network then propagates, as when language-model-generated explanations become node features that improve downstream prediction \cite{he2024tape}. In the third, the two are integrated more deeply, instruction-tuning a language model so that it incorporates graph structure directly \cite{tang2024graphgpt}. The knowledge-graph community has its own version of this convergence, joining the symbolic structure of knowledge graphs with the fluency of language models along a roadmap of mutual reinforcement \cite{pan2024roadmap}, and the area as a whole has been surveyed as it has grown \cite{jin2024llmongraphs}. Whether language models can solve graph problems posed purely in natural language has been studied directly, with mixed results that improve when the model can call graph tools or follow structured procedures \cite{wang2023nlgraph,zhang2023graphtoolformer}, and dedicated surveys now track both the graph-for-language and language-for-graph directions \cite{peng2024graphragsurvey,ren2024llmgraphsurvey}. A related strand uses graphs as structured memory for language-model agents, organizing what an agent has seen into a navigable graph \cite{yang2026graphmem}. The deeper point beneath the activity is that text-attributed graphs are where the combination is most powerful, because the text supplies exactly the shared vocabulary that graphs in general lack, so a language model can align the features of otherwise incomparable graphs through their text. This is the clearest current route around the feature-alignment obstacle, and its limitation is equally clear, that graphs without meaningful text, molecules described only by atoms or sensor networks described only by signals, do not benefit from it directly. A few further observations round out the picture. Encoding a graph as text for a language model forces a choice of serialization, an order in which to list nodes and edges, and the model's answer can depend on that arbitrary order, which is a structural mismatch between a permutation-invariant object and a sequential reader. Benchmarks that pose graph-reasoning tasks to language models have found that they handle small instances and simple structural questions but degrade as graphs grow, which is unsurprising given that a language model has no built-in notion of message passing. The complementary strengths are what make the combination attractive, since a language model brings broad semantic knowledge and flexible reasoning while a graph network brings faithful handling of structure, and the text-attributed setting lets each contribute what it does best. The costs are real as well, since running a large language model over the nodes of a sizeable graph is expensive, so the elegance of the text-bridging route is tempered by questions of efficiency the pure-graph methods do not face.

\subsection{An honest assessment}
\label{subsec:gfm-assessment}

The state of the field can be stated plainly. Foundation models for graphs are real within families, where ULTRA transfers across knowledge graphs, pretrained models transfer across molecules, and task-fixed models classify nodes on arbitrary graphs, and self-supervised pretraining genuinely learns representations that transfer within a domain. The language-model route is powerful for text-attributed graphs, where text bridges the feature gap that blocks the general case. What does not yet exist is a true general foundation model spanning arbitrary graph types, because the feature-alignment problem is unsolved in general and there is no graph analogue of the token that works across all domains. The analogy to language is imperfect in a way that matters, since text is one modality with one vocabulary while graphs are a family of objects with no shared substrate, and the data situation differs as well, since the web-scale corpora behind language models have no equal in a graph world where data is fragmented and domain-specific. What success would even look like is itself worth stating, since a general graph foundation model would be one that, presented with a graph from a domain it had never seen, could perform a useful task on it with little or no domain-specific training, and the honest measure of progress is distance from that capability rather than performance on any single benchmark. By that measure the field has made real progress within families and little toward the fully general goal, which is neither a failure nor a vindication of the ambition but an accurate statement of where it stands. The honest verdict is therefore that the term means something narrower for graphs than for language: foundation models for graphs are established within families and emerging across them, the general ambition confronts a structural obstacle the text and image cases never faced, and the most promising route to generality runs through text-attributed graphs, where language supplies the shared vocabulary that the graph itself does not. That route, and the broader question of whether the obstacle can be overcome or only circumvented, is among the central open problems the closing chapters consider.

\FloatBarrier
\section{Future research directions}
\label{sec:future}

\begin{figure}[t]
\centering
\resizebox{0.92\textwidth}{!}{%
\begin{tikzpicture}[
  hdr/.style={font=\small\bfseries, align=center},
  nbox/.style={rounded corners=3pt, draw=gnngreen!80!black, fill=tintgreen, align=center, font=\footnotesize, text=gnnink, inner sep=3.5pt, minimum height=7.5mm, text width=32mm},
  mbox/.style={rounded corners=3pt, draw=gnnamber!85!black, fill=tintamber, align=center, font=\footnotesize, text=gnnink, inner sep=3.5pt, minimum height=7.5mm, text width=32mm},
  lbox/.style={rounded corners=3pt, draw=gnnred!80!black, fill=tintred, align=center, font=\footnotesize, text=gnnink, inner sep=3.5pt, minimum height=7.5mm, text width=32mm},
  ar/.style={-{Stealth[length=3mm]}, draw=gnnslate, line width=1.3pt}]
\node[hdr, text=gnngreen!55!black] (h1) at (0,3.2)   {Near-term};
\node[hdr, text=gnnrust] (h2) at (4.3,3.2) {Mid-term};
\node[hdr, text=gnnred!85!black] (h3) at (8.6,3.2) {Longer-term};
\node[nbox] (n1) at (0,2.4) {realistic benchmarks};
\node[nbox, below=1.4mm of n1] (n2) {efficient scaling};
\node[nbox, below=1.4mm of n2] (n3) {language models for text graphs};
\node[mbox] (m1) at (4.3,2.4) {reliable extrapolation};
\node[mbox, below=1.4mm of m1] (m2) {dynamic, continual learning};
\node[mbox, below=1.4mm of m2] (m3) {robust, fair, explainable};
\node[lbox] (l1) at (8.6,2.4) {feature-space unification};
\node[lbox, below=1.4mm of l1] (l2) {general foundation models};
\node[lbox, below=1.4mm of l2] (l3) {neuro-symbolic integration};
\draw[ar] (-1.9,-0.55)--(10.5,-0.55) node[right,font=\footnotesize,text=gnngray]{maturity};
\end{tikzpicture}}
\caption[A roadmap of research directions.]{The research directions arranged by horizon, from near-term work that extends current methods to longer-term problems whose solution would change what graph learning can do. The grouping is indicative rather than a schedule, since progress on the harder problems may arrive in any order.}
\label{fig:roadmap}
\end{figure}
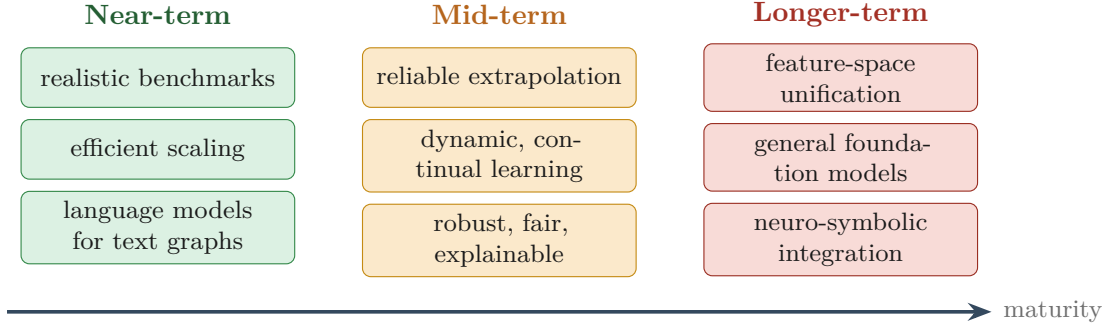

The challenges and foundation-model chapters identified the field's open problems, and this chapter organizes them into concrete research directions. The directions are not a list of independent wishes but the specific next steps the survey's analysis implies, each following from a difficulty the preceding chapters established: the feature-alignment obstacle that blocks general foundation models, the extrapolation problem that recurred across the scientific and shift-prone domains, the trust gap that widens with the stakes, and the questions of scale and integration that the methods themselves raise. \Cref{fig:roadmap} arranges the directions by horizon and \cref{tab:futuredirections} pairs each with the problem it addresses and the part of the survey it arises from. A theme runs through all of them. The field's first decade established that graph learning works and mapped where it helps, and its next phase is defined by a shift from capability to dependability and from the specific to the general, from models that solve particular problems on particular graphs toward models that can be trusted in consequential use and that transfer across the boundaries the survey found dividing the domains. The directions below are the concrete forms that shift takes.

\begin{table}[t]
\centering
\caption[Research directions and the problems they address.]{The research directions, each paired with the problem it addresses and the part of the survey from which it arises.}
\label{tab:futuredirections}
\small
\begin{tabularx}{\textwidth}{@{}lXl@{}}
\toprule
Direction & Problem it addresses & Arises from \\
\midrule
Feature-space unification     & no shared vocabulary across graphs   & foundation models \\
Reliable extrapolation        & failure outside the training distribution & science, shift \\
Calibrated uncertainty        & overconfidence in high-stakes use    & trust limits \\
Dynamic, continual learning   & graphs evolve over time              & many domains \\
Neuro-symbolic integration    & combining learning and reasoning     & knowledge graphs \\
Physics-informed models       & enforcing known constraints          & power, materials, climate \\
Efficient scaling             & cost on web-scale graphs             & scalability \\
Realistic benchmarks          & the benchmark--deployment gap        & cross-domain synthesis \\
\bottomrule
\end{tabularx}
\end{table}

\subsection{Unifying the feature space}
\label{subsec:fut-unify}

The highest-value open problem follows directly from the foundation-model chapter, where the obstacle to generality was that graphs share no common vocabulary. The direction it implies is the search for a universal encoder that maps a graph from any domain into a shared representation space,
\begin{equation}\label{eq:unify}
\Phi:\graph\longmapsto\mathcal{Z},
\end{equation}
on which downstream tasks across domains could then be built, so that the representation rather than the raw features becomes the common substrate. Three routes toward this are visible in current work. The first uses text as the bridge, since text-attributed graphs already carry a shared vocabulary in their node descriptions, and a language model can align otherwise incomparable graphs through it. The second learns the alignment directly, seeking encoders or graph tokenizers that place disparate graphs into one space without relying on text. The third manufactures an invariant where none was given, as the knowledge-graph foundation model did by representing relations through their interactions rather than by fixed identities. Whether any of these scales to graphs in full generality, or whether a universal encoder is achievable only within families of related graphs, is the question on which the prospect of general graph foundation models turns. The keystone status of this problem is worth making explicit. Almost every ambition the survey raised for generality, a single model serving many domains, transfer from data-rich to data-poor settings, and the few-shot and zero-shot capability that makes foundation models valuable, rests on having a representation that means the same thing across graphs. Without it, each domain remains an island, and progress in one does not carry to another except through the slow transfer of methods the cross-domain synthesis described. With it, the boundaries between domains would soften, and the accumulated data and pretraining of the whole field could in principle serve any new problem. This is why feature-space unification, abstract as it sounds, is the single advance that would most change what graph learning can do, and why the routes toward it, whether through text, through learned alignment, or through manufactured invariants, are worth pursuing even though none is yet known to succeed in full generality.

\subsection{Reliable extrapolation and trust}
\label{subsec:fut-trust}

A second cluster of directions addresses reliability, the recurring finding that models are dependable within their training distribution and uncertain beyond it. The scientific domains made the stakes vivid, since extrapolation to novel materials or a changing climate is exactly what those fields need and exactly where learned models are least sure, and the distribution-shift challenge showed the same pattern across healthcare, power, and fraud. The direction is twofold: methods that extrapolate more reliably, by building in physical constraints or invariances that hold beyond the data, and methods that know when they cannot, through calibrated uncertainty and the detection of inputs that fall outside the training distribution. Closely tied to this is the demand for trust, which the challenges chapter argued must be met on several fronts at once, since a deployable model in a high-stakes setting must be robust to manipulation, explainable to the people it affects, and fair across the groups it touches, and these properties interact rather than compose freely. The direction is models that are robust, interpretable, and fair by design rather than by post-hoc patching, together with the certified guarantees that bound worst-case behaviour rather than merely passing observed tests, since it is guarantees of this kind that high-stakes deployment ultimately requires. One technical idea worth singling out is invariant learning, which seeks a representation whose relationship to the label holds across environments rather than only on average, formalized as minimizing the worst-case loss over a set of environments,
\begin{equation}\label{eq:invariance}
\min_{\theta}\ \max_{e}\ \mathcal{L}_e(\theta),
\end{equation}
so that the learned predictor depends on features stable across conditions rather than on correlations particular to the training distribution. The appeal for graphs is direct, since the distribution shifts the survey catalogued, across hospitals, grids, and time, are exactly changes of environment, and a model that latched onto invariant rather than incidental structure would generalize across them. The harder half of reliability, knowing when extrapolation is unsafe, calls for uncertainty that is not merely produced but calibrated and validated, and for the field to treat the detection of out-of-distribution inputs as a first-class capability rather than an afterthought, since in a high-stakes setting a model that abstains when it should is more valuable than one that answers confidently and is wrong.

\subsection{Dynamics, scale, and integration}
\label{subsec:fut-integration}

A third cluster concerns capabilities the methods chapters showed to be underdeveloped. Real graphs change, with nodes and edges arriving and patterns drifting, and most methods still assume a static snapshot, so temporal and continual learning that updates a model as its graph evolves without forgetting what it knew is a direction many domains require. Scale raises a question the field has barely begun to answer, namely whether graph models improve predictably as data and parameters grow, in the way that the error of a language model falls as a power of its scale,
\begin{equation}\label{eq:scaling}
\mathrm{error}(N)\propto N^{-\alpha},
\end{equation}
or whether the fragmented, domain-specific nature of graph data prevents such scaling laws from holding, a question whose answer bears directly on whether the foundation-model strategy can work for graphs at all. Integration is the third strand, the incorporation into learned models of structure and knowledge that pure data-driven methods lack. Neuro-symbolic approaches combine learned representations with symbolic reasoning, training against an objective that adds a logical or relational constraint to the data loss,
\begin{equation}\label{eq:neurosym}
\mathcal{L}=\mathcal{L}_{\mathrm{data}}+\lambda\,\mathcal{L}_{\mathrm{logic}},
\end{equation} The unifying, invariance, scaling, and neuro-symbolic formulations, \cref{eq:unify,eq:invariance,eq:scaling,eq:neurosym}, frame the open directions discussed in this section.
which is natural in the knowledge-graph setting where rules and learning meet. Physics-informed models embed known physical law, a direction the power, materials, and climate chapters showed to be both possible and valuable. And the integration with language models, most effective on text-attributed graphs, is the route through which graph learning most directly joins the broader progress in artificial intelligence. Each integration imports something the data alone does not supply, and together they point away from purely data-driven graph learning toward models that combine learning with reasoning, physical knowledge, and language. Each strand has a concrete near-term form. Temporal graph learning has matured enough to have its own methods and benchmarks, and the open problem is less whether to model time than how to do so efficiently on graphs that change continuously rather than in discrete snapshots. The scaling question is being approached empirically, by training graph models across a range of sizes to see whether the smooth improvements that justified scaling language models appear, and the answer, still unsettled, will shape how much effort the field invests in ever-larger graph models. Neuro-symbolic integration is most developed where the symbolic side is already present, in knowledge graphs and in domains with known rules, and the broader question is whether the approach extends to settings where the symbols must themselves be discovered. The common thread across dynamics, scale, and integration is that each asks the field to move beyond the static, purely learned, single-graph model much of the methodology still assumes, toward models that are temporal, that grow predictably, and that draw on knowledge outside their training data.

\subsection{Toward a mature field}
\label{subsec:fut-mature}

A final direction is not a method but a change in practice, and it follows from the calibration theme the survey pressed in every chapter. The recurring gap between benchmark performance and deployed reliability is partly a problem of evaluation, since the benchmarks the field optimizes against often fail to reflect the conditions of use, and the direction is benchmarks built to mirror deployment, with splits that test out-of-distribution generalization, with the realistic class imbalances and adversarial pressures that real applications present, and with metrics that encode the costs a deployed system actually faces. Alongside better benchmarks the field needs standardized evaluation that makes results comparable and the honest reporting of the benchmark-deployment gap as a matter of course rather than as an occasional caveat. Reproducibility belongs to the same effort, since results that cannot be reproduced cannot be built upon, and the field has at times struggled with inconsistent evaluation protocols, tuned baselines stronger than reported, and comparisons that do not hold conditions fixed. A culture of careful, comparable, deployment-aware evaluation is less glamorous than a new architecture but arguably more valuable at the field's current stage, because it is what would let the community tell which of its many methods genuinely advance the state of the art and which only appear to. This maturation, from demonstrating that graph networks can work to understanding when and why they work and deploying them where they reliably do, is the connective direction beneath the others, and the directions together describe the path from the field's current state, capable and impactful but uneven and incompletely understood, toward a mature, trustworthy, and more general graph-learning capability.

\FloatBarrier
\section{Conclusion}
\label{sec:conclusion}

This survey set out to give a complete account of graph neural networks across the domains where they are used, and the path it took was deliberate. It began with the foundations, the idea that relational data is a graph and that learning on graphs means respecting permutation symmetry through message passing, and with the architectures, the convolutional, attention-based, and message-passing families and the spectral and spatial views that explain them. It then turned to twelve application domains in turn, from social networks and recommendation through knowledge graphs, molecules, healthcare, vision, traffic, power systems, wireless networks, fraud, industrial systems, materials, and climate, treating each with the same structure of graph construction, tasks, methods, strengths, and an honest reckoning of weaknesses. It drew those domains together in a synthesis, examined the technical challenges that constrain the field, assessed the prospect of graph foundation models, and laid out the research directions the analysis implies. 

Several findings recur with enough consistency to count as the survey's central conclusions. The first is that the graph construction is the most consequential decision in any application, more so than the choice of architecture, since a natural graph is a faithful match that the methods exploit, a constructed graph is a modelling choice whose quality silently determines the result, and a learned graph is the field's response to the discovery that even natural graphs are sometimes incomplete. The second is that the value of a graph network varies sharply and predictably by domain. It is largest where relational structure carries signal that no model examining records in isolation can reach, as in the coordination of a fraud ring, the geometry of a molecule, or the global structure of the atmosphere; it is a matter of speed rather than accuracy where the governing physics is already known, as in power-flow surrogates, interatomic potentials, and learned simulators, which earn their place by being fast and differentiable rather than more correct; and it is real but modest where a strong non-relational baseline already captures most of the signal, as in much of traffic forecasting and recommendation, where honest comparison is what separates a genuine contribution from a restatement of what simpler methods achieve.

Beneath the diversity of domains lay a small set of shared structures. A single spatio-temporal template, a graph signal evolving over time, recurred across traffic, energy, epidemics, climate, and industry, so that recognizing a problem as spatio-temporal forecasting on a graph was most of the work of solving it. A divide between transductive and inductive settings ran throughout, determining what generalization means and how readily a model reaches new graphs and new nodes. And the homophily assumption that connected nodes are alike, which makes the smoothing of ordinary message passing the right operation in social and citation networks, failed in the adversarial setting of fraud and forced those domains to rebuild the standard toolkit. The technical challenges proved to be connected rather than separate, the limits of depth and the failure with heterophily being two faces of the same smoothing bias, and robustness, explainability, and fairness being facets of a single demand for trust that interact rather than compose. Graph foundation models, finally, are real within families where a shared vocabulary exists or a single task is fixed, and are blocked in the general case by the absence of any vocabulary common to all graphs, with the most promising route to generality running through text-attributed graphs where language supplies the substrate the graph itself lacks.

The thread connecting all of this is a matter of calibration. The gap between performance on a benchmark and reliability in deployment was the rule rather than the exception across the survey, widest exactly where the stakes are highest, in healthcare, in safety-critical power systems, and in the adversarial, drifting world of fraud, and naming that gap honestly was a recurring obligation rather than an occasional caveat. The field's first phase demonstrated that graph networks work, often impressively, and its next phase is defined by a shift from capability to dependability and from the specific to the general, from models that solve particular problems on particular graphs toward models that can be trusted in consequential use and that transfer across the boundaries dividing the domains.

The honest verdict, then, is neither the unqualified enthusiasm that greets a new method nor the dismissal that follows inflated claims. Graph neural networks are a genuine and broadly useful approach, responsible for real advances that simpler methods could not have produced, in drug discovery, in weather forecasting, in the simulation of matter, and in the detection of coordinated fraud, and they are also uneven, sometimes an elaborate way to match what a strong baseline already does, and constrained by limits in expressiveness, scale, robustness, and generalization that the field is still learning to overcome. The relational structure of the world is genuine, and learning to exploit it is a worthwhile and lasting goal; the work that remains is to do so reliably, generally, and with a clear understanding of when the structure carries the signal and when it does not.

% <<< BODY:END >>>

%================================ BIBLIOGRAPHY ================================
\FloatBarrier
\bibliographystyle{IEEEtran}   % fallback: \bibliographystyle{unsrtnat}
%\bibliography{refs}

\end{document}